\documentclass[twoside,phd,etd]{Variables_submit}

\usepackage[T1]{fontenc}
\usepackage{longtable}
\usepackage{standalone}
\usepackage{pdfpages}
\usepackage{listings}
\usepackage{courier}
\usepackage{fancyhdr}
\usepackage[margin=0.3in,labelfont=bf,labelsep=none]{caption}
\usepackage{xspace}
\usepackage{multirow}
\usepackage{colortbl}
\usepackage{booktabs}
\usepackage{rotating}
\usepackage{hhline}
\usepackage{graphicx}
\usepackage{amssymb}
\usepackage{algorithmic}
\usepackage{caption}
\usepackage{blindtext}
\usepackage{etoolbox}
\usepackage{url}
\usepackage{balance}
\usepackage{mdwlist}
\usepackage{subcaption}
\usepackage{pifont}
\usepackage{dsfont}
\usepackage{amsmath,epsfig}
\usepackage[ruled,vlined]{algorithm2e}
\usepackage{verbatim}
\usepackage{makeidx}
\usepackage{bbm}
\usepackage{enumerate}
\usepackage{color}
\usepackage{amssymb}

\usepackage{amsthm}
\theoremstyle{plain}

\usepackage{caption}
\DeclareCaptionLabelFormat{mylabel}{#1 #2.\hspace{1.0ex}}
\DeclareCaptionFont{white}{\color{white}}
\DeclareCaptionFormat{listing}{\colorbox[cmyk]{0.43, 0.35, 0.35,0.01}{\parbox{\textwidth}{\hspace{15pt}#1 #2 \hspace{1ex} #3}}}
\usepackage[ruled,vlined]{algorithm2e}
\usepackage[Bjornstrup]{fncychap}

\DeclareCaptionType{copyrightbox}

\newtheorem{definition}{Definition}[chapter]

\newtheorem*{entity_lag*}{Entity Lag}
\newtheorem*{emerging_entities*}{Emerging Entities}
\newtheorem*{ee_ep*}{Emerging Entities in Event Pages}
\newtheorem*{news_density*}{News Reference Density}
\newtheorem*{event*}{Event}
\newtheorem*{entity*}{Entity}

\makeindex

\usepackage[bookmarksnumbered,pdfpagelabels=true,plainpages=false,colorlinks=true,bookmarks=true,linkcolor=blue,citecolor=blue,urlcolor=blue]{hyperref}

 \Year{2017}
 \Month{}
 \Author{Besnik Fetahu}

\TitleTop{Approaches for Enriching and Improving Textual Knowledge Bases}

 \Advisor{Prof. Dr. techn. Wolfgang Nejdl}

\Abstract{
\doublespace
\begin{footnotesize}

\emph{Verifiability} is one of the core editing principles in Wikipedia, where editors are encouraged to provide \emph{citations} for the added statements. Statements can be any arbitrary piece of text, ranging from a sentence up to a paragraph. However, in many cases, citations are either outdated, missing, or link to non-existing references (e.g. dead URL, moved content etc.). In total, 20\% of the cases such citations refer to \emph{news} articles and represent the second most cited source. Even in cases where citations are provided, there are no explicit indicators for the span of a citation for a given piece of text. In addition to issues related with the verifiability principle, many Wikipedia entity pages are incomplete, with relevant information that is already available in online news sources missing. Even for the already existing citations, there is often a delay between the news publication time and the reference time.

In this thesis, we address the aforementioned issues and propose automated approaches that enforce the \emph{verifiability} principle in Wikipedia, and suggest relevant and missing news references for further enriching Wikipedia entity pages. To this end we make the following contributions as part of this thesis:

\begin{itemize}
	\item \emph{Citation recommendation} -- we address the problem of finding and updating news citations for statements in Wikipedia entity pages. We propose a two-stage approach for this problem. First, we classify each statement whether it requires a news citation or citations from other categories (e.g. web, book, journal, etc.). Second, for statements that require a news citation, we formalize three properties of what makes a good citation, namely: (i) the citation should entail the Wikipedia statement, (ii) the statement should be central to the citation, and (iii) the citation should be from an authoritative source. We combine standard information retrieval techniques, where we use the statement to query a news collection, and build classification models based on the three properties to determine the most appropriate citation.
	\item \emph{Citation span} -- from the already existing citations in Wikipedia entity pages and the ones we recommend in our first problem, we propose an automated approach which determines the span of such citations.  We approach this problem by classifying which textual fragments in a paragraph are covered or hold true given a citation. We propose a sequence classification approach where for a paragraph and a citation, we determine the citation span at a fine-grained level.
	\item \emph{News suggestion} -- to account for the ever evolving nature of Wikipedia entities, with relevant information published on a daily basis in news articles, we propose a two-stage supervised approach for this problem. First, we suggest news articles to Wikipedia entities (article-entity placement) relying on a rich set of features which take into account the \emph{salience} and \emph{relative authority} of entities, and the \emph{novelty} of news articles to entity pages. Second, we determine the exact section in the entity page for the input article (article-section placement) guided by class-based section templates.
\end{itemize}

We perform extensive evaluation with real-world datasets, on news collections with more than 20 million news articles, and on the entire set of english Wikipedia entity pages. Our approaches perform with high accuracy on the three problems we address and show superior performance when compared to existing baselines and state of the art approaches.

 
\end{footnotesize}
}

\Zusammenfassung {
\doublespace
\begin{footnotesize}
\begin{sloppypar}

\emph{Nachweisbarkeit} ist eines der zentralen Editierungs-Prinzipien in Wikipedia. Editoren werden dazu angehalten, ihre hinzugef{\"u}gten Aussagen mittels \emph{Zitierungen} zu belegen. Aussagen k{\"o}nnen dabei beliebige Textst{\"u}cke sein, von S{\"a}tzen bis hin zu einem Absatz. In vielen F{\"a}llen sind Zitierungen jedoch veraltet, fehlen oder verweisen auf nicht existierende Referenzen (z.B. tote URLs oder verschobene Inhalte). In 20\% der F{\"a}lle verweisen Zitierungen auf \emph{News}-Artikel, die zweith{\"a}ufigste Art der Zitierung in Wikipedia. Selbst in den F{\"a}llen, in denen Zitierungen vorhanden sind, fehlen Angaben {\"u}ber die Spanne des Textes, die durch die Zitierung abgedeckt wird. Unabh{\"a}ngig von Problemen in Zusammenhang mit dem Nachweisbarkeits-Prinzip sind viele Wikipedia-Artikel unvollst{\"a}ndig, wobei oft relevante Information fehlt, die bereits in Online News-Quellen verf{\"u}gbar ist. Auch f{\"u}r die bereits hinzugef{\"u}gten Zitierungen existiert oft eine Verz{\"o}gerung zwischen der Ver{\"o}ffentlichungszeit des News-Artikels und der Zeit, zu der der Artikel in Wikipedia referenziert wurde.

In dieser Arbeit besch{\"a}ftigen wir uns mit den aufgezeigten Problemen und stellen automatisierte Ans{\"a}tze vor, die das \emph{Nachweisbarkeits}-Prinzip in Wikipedia durchsetzen und relevante und fehlende News-Referenzen vorschlagen mit dem Ziel, die Qualit{\"a}t von Wikipedia-Artikeln zu erh{\"o}hen. Diese Arbeit enth{\"a}lt die folgenden Beitr{\"a}ge:

\begin{itemize}
    \item \emph{Zitierungsempfehlungen} -- Wir besch{\"a}ftigen uns mit dem Problem des Findens und Erneuerns von News-Zitierungen f{\"u}r Aussagen in Wikipedia-Artikeln. F{\"u}r dieses Problem stellen wir einen zweiteiligen Ansatz vor. Zun{\"a}chst bestimmen wir mittels Klassifizierung, ob eine Aussage eine News-Zitierung oder eine Zitierung aus einer anderen Kategorie (z.B. Web, B{\"u}cher, Zeitschriften) ben{\"o}tigt. Im zweiten Schritt bestimmen wir drei Eigenschaften, die eine gute Zitierung ausmachen: (i) die Zitierung sollte die Wikipedia-Aussage enthalten, (ii) die Aussage sollte eine zentrale Rolle im zitierten Artikel einnehmen und (iii) der zitierte Artikel sollte aus einer verl{\"a}sslichen Quelle stammen. Wir kombinieren Standardtechniken des Information Retrieval und verwenden die gegebene Aussage, um eine passende News-Sammlung zusammenzustellen. Weiterhin entwickeln wir Klassifizierungsmodelle basierend auf den drei genannten Eigenschaften, um die passendste Zitierung zu ermitteln.
 \item \emph{Zitierungsspanne} -- Aus den bereits existierenden Zitierungen in Wikipedia Artikeln und den Zitierungen, die wir in unserem ersten Problem vorschlagen, entwickeln wir einen automatisierten Ansatz zur Bestimmung der Spanne einer Zitierung. Dazu klassifizieren wir, welche Textbausteine eines Absatzes durch eine Zitierung abgedeckt werden, bzw. als wahr angesehen werden k{\"o}nnen. Wir stellen einen Sequenzklassifizierungsansatz vor, der die Spanne einer Zitierung bezogen auf einen Absatz im Detail bestimmen kann.
 \item \emph{News-Vorschl{\"a}ge} -- Unter Ber{\"u}cksichtigung der st{\"a}ndigen Ver{\"a}nderung von Wikipedia-Artikeln sowie der Tatsache, dass t{\"a}glich neue relevante Informationen in Online News-Artikeln ver{\"o}ffentlicht werden, stellen wir einen zweiteiligen {\"u}berwachten Ansatz vor. Zun{\"a}chst schlagen wir News-Artikel f{\"u}r Wikipedia-Artikel vor (article-entity placement), basierend auf einer ergiebigen Menge an Features, die die \emph{relative Bedeutung} eines Wikipedia-Artikels und die \emph{Neuheit} von News-Artikeln ber{\"u}cksichtigen. Im zweiten Schritt bestimmten wir die Sektion des Wikipedia-Artikels, f{\"u}r die der News-Artikel vorgeschlagen werden soll (article-section placement).
\end{itemize}

Wir f{\"u}hren umfassende Evaluationen mit real-world Datens{\"a}tzen von News-Sammlungen mit mehr als 20 Millionen News-Artikeln und der gesammten Menge englischer Wikipedia-Artikel durch. Unsere Ans{\"a}tze erzielen hohe Genauigkeit und {\"u}bertreffen die Leistung von existierenden Baselines und State-of-the-Art Ans{\"a}tzen.

\noindent\textbf{Schlagw\"orter:}~\emph{Zitierungsempfehlungen, Zitierungsspanne, News-Vorschl{\"a}ge, Wikipedia Anreicherung}

\end{sloppypar}
\end{footnotesize}
}

\Foreword{
\singlespace

The work in this thesis and generally throughout the entire course of the Ph.D studies I have published and co-authored several papers and journals in the fields of Text Mining, Natural Language Process, Semantic Web, Information Retrieval and Human-Computer Interaction.

The core contributions of this thesis in the individual chapters are published in the following venues:

\begin{enumerate}
	\item The contributions in Chapter~\ref{ch1:citation_recommendation}, which deal with the problem of news suggestion for Wikipedia articles are published in: 
	\begin{itemize}
	\item \cite{DBLP:conf/cikm/FetahuMNA16} Besnik Fetahu, Katja Markert, Wolfgang Nejdl, Avishek Anand: Finding News Citations for Wikipedia. In \emph{Proceedings of the 25th ACM International on Conference on Information
               and Knowledge Management}, CIKM 2016, Indianapolis, IN, USA, October 24-28, 2016, pages 337--346.

	\end{itemize}
	
	\item The contributions in Chapter~\ref{ch5:cite_span}, which deal with the problem of determining the citation span for references in Wikipedia are published in:
	\begin{itemize}
	\item \cite{DBLP:conf/emnlp17/FetahuMA17} Besnik Fetahu, Katja Markert, Avishek Anand: Fine Grained Citation Span for References in Wikipedia.  In \emph{Proceedings of the Conference on Empirical Methods in Natural Language Processing}, EMNLP 2017, Copenhagen, Denmark, September 7-11, 2017.
	\end{itemize}

	\item The contributions in Chapter~\ref{ch3:news_suggestion}, which deals with the problem of automatically suggesting news articles to Wikipedia is published in:
	\begin{itemize}
		\item \cite{DBLP:conf/cikm/FetahuMA15} Besnik Fetahu, Katja Markert, Avishek Anand: Automated News Suggestions for Populating Wikipedia Entity Pages.  In \emph{Proceedings of the 25th ACM International on Conference on Information and Knowledge Management}, CIKM 2015, Melbourne, Australia, October 19 - 23, 2015, pages 323--332.
	\end{itemize}
	
	\item Finally, in Chapter~\ref{ch:news_wiki_lag} we present a study of the lag between Wikipedia and news, and in Chapter~\ref{ch:entity_search} we show an application use case of entity search in structured datasets, which are published in:
\begin{itemize}
		\item \cite{DBLP:conf/websci/FetahuAA15} Besnik Fetahu, Abhijit Anand, and Avishek Anand. How much is wikipedia lagging behind news. In \emph{Proceedings of the ACM Web Science Conference, WebSci 2015}, Oxford,
               United Kingdom, June 28 - July 1, 2015, pages 28:1--28:9.
		\item \cite{DBLP:conf/semweb/FetahuGD15} Besnik Fetahu, Ujwal Gadiraju, and Stefan Dietze. Improving Entity Retrieval on Structured Data. In \emph{The Semantic Web - ISWC 2015 - 14th International Semantic Web Conference}, Bethlehem, PA, USA, October 11-15, 2015, Proceedings, Part I, pages 474--491.

	\end{itemize}
\end{enumerate}

The complete list of publications during my PhD is shown below:

\begin{enumerate}
    \item \cite{DBLP:conf/emnlp17/FetahuMA17} Besnik Fetahu, Katja Markert, Avishek Anand: Fine Grained Citation Span for References in Wikipedia.  In \emph{Proceedings of the Conference on Empirical Methods in Natural Language Processing}, EMNLP 2017, Copenhagen, Denmark, September 7-11, 2017.
	\item \cite{DBLP:conf/icde/YuGFD17} Ran Yu, Ujwal Gadiraju, Besnik Fetahu, and Stefan Dietze: FuseM: Query-Centric Data Fusion on Structured Web Markup. In \emph{Proceedings of the 33rd IEEE International Conference on Data Engineering, ICDE},
               San Diego, CA, USA, April 19-22, 2017, pages 179--182.
    \item \cite{DBLP:journals/tochi/GadirajuFKSD17} Ujwal Gadiraju, Besnik Fetahu, Rickardo Kawase, Patrick Siehndel, and Stefan Dietze: Is the Crowd Smarter Than a 5th Grader? Using Worker Self-Assessments for Competence-based Pre-Selection. In \emph{ACM Transactions of Computer-Human Interaction}, 2017.
    
	\item \cite{DBLP:conf/cikm/FetahuMNA16} Besnik Fetahu, Katja Markert, Wolfgang Nejdl, Avishek Anand: Finding News Citations for Wikipedia. In \emph{Proceedings of the 25th {ACM} International on Conference on Information and Knowledge Management, {CIKM} 2016, Indianapolis, IN, USA, October 24-28, 2016}, pages 337--346.
	\item \cite{DBLP:conf/esws/YuGZFD16} Ran Yu, Ujwal Gadiraju, Xiaofei Zhu, Besnik Fetahu, Stefan Dietze: Towards Entity Summarisation on Structured Web Markup. In \emph{Proceedings of the Semantic Web - ESWC 2016 Satellite Events}, Heraklion, Crete, Greece, May 29 - June 2, 2016, Revised Selected Papers, pages 69--73.
	\item \cite{DBLP:conf/semweb/YuFGD16} Ran Yu, Besnik Fetahu, Ujwal Gadiraju, Stefan Dietze: A Survey on Challenges in Web Markup Data for Entity Retrieval. In \emph{Proceedings of the ISWC 2016 Posters \& Demonstrations Track co-located with 15th International Semantic Web Conference (ISWC 2016), Kobe, Japan, October 19, 2016.}
	\item \cite{DBLP:series/lncs/TaibiFDF16} Davide Taibi, Giovanni Fulantelli, Stefan Dietze, Besnik Fetahu: Educational Linked Data on the Web - Exploring and Analysing the Scope and Coverage. In \emph{Open Data for Education - Linked, Shared, and Reusable Data for Teaching and Learning}, pages 16--37. 
	\item \cite{DBLP:series/lncs/BeetzBDFGHKLTWY16} Jakob Beetz, Ina Bl{\"u}mel, Stefan Dietze, Besnik Fetahu, Ujwal Gadiraju, Martin Hecher, Thomas Krijnen, Michelle Lindlar, Martin Tamke, Raoul Wessel, Ran Yu: In \emph{3D Research Challenges in Cultural Heritage II - How to Manage Data and Knowledge Related to Interpretative Digital 3D Reconstructions of Cultural Heritage}, pages 231--255.

	\item \cite{DBLP:journals/bjet/TaibiCDMF15} Davide Taibi, Saniya Chawla, Stefan Dietze, Ivana Marenzi, Besnik Fetahu: Exploring TED talks as linked data for education. BJET 46(5): 1092-1096 (2015)
	\item \cite{DBLP:conf/cikm/FetahuMA15} Besnik Fetahu, Katja Markert, Avishek Anand: Automated News Suggestions for Populating Wikipedia Entity Pages. In \emph{Proceedings of the 24th ACM International Conference on Information and Knowledge Management, CIKM 2015}, Melbourne, VIC, Australia, October 19 - 23, 2015, pages 323--332.
	\item \cite{DBLP:conf/ectel/GadirajuFK15} Ujwal Gadiraju, Besnik Fetahu, Ricardo Kawase: Training Workers for Improving Performance in Crowdsourcing Microtasks. In \emph{Design for Teaching and Learning in a Networked World - 10th European
               Conference on Technology Enhanced Learning, {EC-TEL} 2015}, Toledo,
               Spain, September 15-18, 2015, Proceedings, pages 100--114.
	\item \cite{DBLP:conf/ht/GadirajuSFK15} Ujwal Gadiraju, Patrick Siehndel, Besnik Fetahu, Ricardo Kawase: Breaking Bad: Understanding Behavior of Crowd Workers in Categorization Microtasks. In \emph{Proceedings of the 26th {ACM} Conference on Hypertext {\&} Social
               Media, {HT} 2015}, Guzelyurt, TRNC, Cyprus, September 1-4, 2015, pages 33--38.
	\item \cite{DBLP:conf/semweb/FetahuGD15} Besnik Fetahu, Ujwal Gadiraju, Stefan Dietze: Improving Entity Retrieval on Structured Data. In \emph{The Semantic Web - {ISWC} 2015 - 14th International Semantic Web Conference}, Bethlehem, PA, USA, October 11-15, 2015, Proceedings, Part {I}, pages 474--491.
	\item \cite{DBLP:conf/websci/FetahuAA15} Besnik Fetahu, Abhijit Anand, Avishek Anand: How much is Wikipedia Lagging Behind News? In \emph{Proceedings of the {ACM} Web Science Conference, WebSci 2015}, Oxford, United Kingdom, June 28 - July 1, 2015, pages 28:1--28:9.
	\item \cite{DBLP:conf/wise/YuGFD15} Ran Yu, Ujwal Gadiraju, Besnik Fetahu, Stefan Dietze: Adaptive Focused Crawling of Linked Data. In \emph{Web Information Systems Engineering - {WISE} 2015 - 16th International Conference}, Miami, FL, USA, November 1-3, 2015, Proceedings, Part {I}, pages 554--569.
	\item \cite{DBLP:conf/www/TaibiFDF15} Davide Taibi, Giovanni Fulantelli, Stefan Dietze, Besnik Fetahu: Towards Analysing the Scope and Coverage of Educational Linked Data on the Web. In \emph{Proceedings of the 24th International Conference on World Wide Web
               Companion, {WWW} 2015}, Florence, Italy, May 18-22, 2015 - Companion Volume, pages 705--710.
               
	\item \cite{DBLP:conf/esws/FetahuDNCTN14} Besnik Fetahu, Stefan Dietze, Bernardo Pereira Nunes, Marco Antonio Casanova, Davide Taibi, Wolfgang Nejdl: A Scalable Approach for Efficiently Generating Structured Dataset Topic Profiles. In \emph{The Semantic Web: Trends and Challenges - 11th International Conference, {ESWC} 2014}, Anissaras, Crete, Greece, May 25-29, 2014. Proceedings, pages 519--534.
               
	\item \cite{DBLP:conf/icalt/NunesCKFCC14} Bernardo Pereira Nunes, Alexander Arturo Mera Caraballo, Ricardo Kawase, Besnik Fetahu, Marco A. Casanova, Gilda Helena Bernardino de Campos: A Topic Extraction Process for Online Forums. In \emph{{IEEE} 14th International Conference on Advanced Learning Technologies, {ICALT} 2014}, Athens, Greece, July 7-10, 2014, pages 541--543.
	\item \cite{DBLP:conf/semweb/TaibiDFF14} Davide Taibi, Stefan Dietze, Besnik Fetahu, Giovanni Fulantelli: Exploring type-specific topic profiles of datasets: a demo for educational linked data. In \emph{Proceedings of the {ISWC} 2014 Posters {\&} Demonstrations Track a track within the 13th International Semantic Web Conference, {ISWC} 2014}, Riva del Garda, Italy, October 21, 2014., pages 353--356.
	\item \cite{DBLP:conf/semweb/FetahuGD14} Besnik Fetahu, Ujwal Gadiraju, Stefan Dietze: Crawl Me Maybe: Iterative Linked Dataset Preservation. In \emph{Proceedings of the {ISWC} 2014 Posters {\&} Demonstrations Track a track within the 13th International Semantic Web Conference, {ISWC}} 2014, Riva del Garda, Italy, October 21, 2014., pages 433--436.
	\item \cite{DBLP:conf/wise/NunesKFCC14} Bernardo Pereira Nunes, Ricardo Kawase, Besnik Fetahu, Marco A. Casanova, Gilda Helena Bernardino de Campos: Educational Forums at a Glance: Topic Extraction and Selection. In \emph{Web Information Systems Engineering - {WISE} 2014 - 15th International Conference}, Thessaloniki, Greece, October 12-14, 2014, Proceedings, Part {II}, pages 351--364.
	\item \cite{DBLP:conf/www/FetahuDNCTN14} Besnik Fetahu, Stefan Dietze, Bernardo Pereira Nunes, Marco Antonio Casanova, Davide Taibi, Wolfgang Nejdl: What's all the data about?: creating structured profiles of linked data on the web. In \emph{23rd International World Wide Web Conference, {WWW} '14}, Seoul, Republic of Korea, April 7-11, 2014, Companion Volume, pages 261--262.

	\item \cite{DBLP:conf/dexa/NunesCCFLD13} Bernardo Pereira Nunes, Alexander Arturo Mera Caraballo, Marco Antonio Casanova, Besnik Fetahu, Luiz André P. Paes Leme, Stefan Dietze: Complex Matching of RDF Datatype Properties. In \emph{Database and Expert Systems Applications - 24th International Conference, {DEXA} 2013}, Prague, Czech Republic, August 26-29, 2013. Proceedings, Part {I}, pages 195--208.
	\item \cite{DBLP:conf/ectel/TaibiFDF13} Davide Taibi, Giovanni Fulantelli, Stefan Dietze, Besnik Fetahu: Evaluating Relevance of Educational Resources of Social and Semantic Web. In \emph{Scaling up Learning for Sustained Impact - 8th European Conference, on Technology Enhanced Learning, {EC-TEL} 2013}, Paphos, Cyprus, September 17-21, 2013. Proceedings, pages 637--638.
	\item \cite{DBLP:conf/esws/NunesDCKFN13} Bernardo Pereira Nunes, Stefan Dietze, Marco Antonio Casanova, Ricardo Kawase, Besnik Fetahu, Wolfgang Nejdl: Combining a Co-occurrence-Based and a Semantic Measure for Entity Linking. In \emph{The Semantic Web: Semantics and Big Data, 10th International Conference, {ESWC} 2013}, Montpellier, France, May 26-30, 2013. Proceedings, pages 548--562.
	\item \cite{DBLP:conf/icwe/FetahuND13} Besnik Fetahu, Bernardo Pereira Nunes, Stefan Dietze: Summaries on the Fly: Query-Based Extraction of Structured Knowledge from Web Documents. In \emph{Web Engineering - 13th International Conference, {ICWE} 2013}, Aalborg, Denmark, July 8-12, 2013. Proceedings, pages 249--264.
	\item \cite{DBLP:conf/kes/NunesKFDCM13} Bernardo Pereira Nunes, Ricardo Kawase, Besnik Fetahu, Stefan Dietze, Marco A. Casanova, Diana Maynard: Interlinking Documents based on Semantic Graphs. In \emph{17th International Conference in Knowledge Based and Intelligent Information and Engineering Systems, {KES} 2013}, Kitakyushu, Japan, 9-11 September 2013, pages 231--240.
	\item \cite{DBLP:conf/lak/NunesFC13} Bernardo Pereira Nunes, Besnik Fetahu, Marco Antonio Casanova: Cite4Me: Semantic Retrieval and Analysis of Scientific Publications. In \emph{Proceedings of the {LAK} Data Challenge}, Leuven, Belgium, April 9,
               2013.
	\item \cite{DBLP:conf/semweb/NunesFDC13} Bernardo Pereira Nunes, Besnik Fetahu, Stefan Dietze, Marco A. Casanova: Cite4Me: A Semantic Search and Retrieval Web Application for Scientific Publications. In \emph{Proceedings of the {ISWC} 2013 Posters {\&} Demonstrations Track}, Sydney, Australia, October 23, 2013, pages 25--28.
	\item \cite{DBLP:conf/semweb/FetahuDNTC13} Besnik Fetahu, Stefan Dietze, Bernardo Pereira Nunes, Davide Taibi, Marco Antonio Casanova: Generating structured Profiles of Linked Data Graphs. In \emph{Proceedings of the {ISWC} 2013 Posters {\&} Demonstrations Track}, Sydney, Australia, October 23, 2013, pages 113--116
	\item \cite{DBLP:conf/www/FetahuND13} Besnik Fetahu, Bernardo Pereira Nunes, Stefan Dietze: Towards focused knowledge extraction: query-based extraction of structured summaries. WWW (Companion Volume) 2013: 77-78
	\item \cite{DBLP:conf/www/TaibiFD13} Davide Taibi, Besnik Fetahu, Stefan Dietze: Towards integration of web data into a coherent educational data graph. In \emph{22nd International World Wide Web Conference, {WWW} '13}, Rio de Janeiro, Brazil, May 13-17, 2013, Companion Volume, pages 419--424.
	\item \cite{DBLP:conf/sigir/FetahuS12} Besnik Fetahu, Ralf Schenkel: Retrieval evaluation on focused tasks. In \emph{The 35th International {ACM} {SIGIR} conference on research and development in Information Retrieval, {SIGIR} '12}, Portland, OR, USA, August 12-16, 2012, pages 1135--1136.
\end{enumerate}

}

 \MemberA{Member 1}
 \MemberB{Member 2}
 \MemberC{Member 3}

\makeatletter
\newcommand\footnoteref[1]{\protected@xdef\@thefnmark{\ref{#1}}\@footnotemark}
\makeatother
\newcolumntype{P}[1]{>{\centering\arraybackslash}p{#1}}
\newcommand{\argmax}{\operatornamewithlimits{argmax}}
\newcommand {\N}{\mathcal{N}}

\begin{document}

\parskip 3pt
\frontmatter

\makepreliminarypages

\tableofcontents
\clearemptydoublepage

\listoffigures
\clearemptydoublepage
 
\listoftables
\clearemptydoublepage

\doublespace

\mainmatter


\chapter{Introduction}\label{introduction}

\section{Motivation}

The advent of Internet and Web 2.0 platforms, has led to major societal shifts. Nowadays, Web users have the means to consume and create content as part of various sharing platforms like Social Media, Blogs and other platforms. The increasing number of users in the Web~\cite{internet_stats} has shifted the focus of many organizations towards providing content online. This has led to many successful digitization projects like the Internet Archive~\cite{internet_archive}, the Million Books Project~\cite{books}, The New York Times~\cite{sandhaus2008new} etc. The impact of such digitization has societal benefits, in which information is easier accessible thus leading to a more informed society. For instance, in US alone, nearly 40\% of users consume their daily news through online news media platforms~\cite{pewnews}.

Apart from organizations that follow a certain discourse on providing information like news media, there is the other spectrum of the Web, where information is a direct result of the interaction between users and Web applications. The engagement and the number of users directly correlates to the value of an application. Examples include Social Media platforms like Twitter, Facebook etc., where user engagement is of importance to the success of these applications~\cite{DBLP:conf/www/Ribeiro14}. Such a phenomena is described by Simon~\cite{simon1971designing}, where user attention and engagement are some of the key available resources for organizations. This similarly applies to Web applications.

In this respect, one of the most known examples of such synergies between Web application and users on the Web is Wikipedia~\cite{wiki}. It represents an open, collaborative effort of creating encyclopedic content by Web users. At its core are Wikipedia editors, who provide content according to established principles and editing policies~\cite{wiki_policy}. There are approximately 284 different language versions of Wikipedia~\cite{wiki_languages}. Only in the english Wikipedia there are roughly 5 million articles, and a total of 30 million registered editors across all localized Wikipedias. 
The dynamics of content creation in Wikipedia, and the organizational structure and collaboration between editors has been subject to extensive research~\cite{keegan_editors_2012,kittur_power_2007,DBLP:conf/wikis/ZhangPL12,DBLP:conf/group/PancieraHT09,DBLP:conf/icwsm/Warncke-WangRTH15}.

Wikipedia, is one of the top visited websites overall\footnote{\small{In 2015 it was in the top 10 most visited Internet sites according to the Alexa Internet ranking \url{www.alexa.com}).}}. Due to the large number of editors, and its openness in terms of added content, to provide quality assurances, there are guidelines and policies~\cite{wiki_policy}, editor categorizations (e.g. \emph{admins} or \emph{novice} editors), and finally each revision of an article can be edited or deleted by other peer-editors. Despite the fact that the nature of these policies are guidelines and are not enforced, studies~\cite{ASI:ASI23172} show that Wikipedia in specific domains achieves comparable quality to expert curated encyclopedia like Britannica~\cite{britannica}.

The value of Wikipedia has been widely acknowledged. It serves as the backbone for a wide range of applications. It is used to construct knowledge graphs like DBpedia~\cite{DBLP:journals/ws/BizerLKABCH09}, YAGO~\cite{suchanek_yago:_2007}, which  are included in major search engines like Google KnowledgeGraph or Apple's Siri system. Furthermore, it has been widely used in fields such as text categorization~\cite{wang_building_2008}, entity disambiguation~\cite{Hoffart:2011:RDN:2145432.2145521} etc. Therefore, apart from its direct visitors, its content is used and accessed implicitly through other sources that are built upon Wikipedia.

The core role and popularity of Wikipedia on the Web and the large variety of applications can be traced to two main factors. First, articles in Wikipedia are constantly evolving and new articles are added by its community of editors. This is mostly influenced by emerging information from the Web. For example, for an existing Wikipedia article like \texttt{United States presidential election, 2016}\footnote{\url{https://en.wikipedia.org/wiki/United_States_presidential_election,_2016}}, there are news reports, blogs, and other sources reporting about this particular event. In many cases such emerging information is directly reflected in the corresponding Wikipedia articles, thus, keeping Wikipedia up to date. In some cases, real-world events are immediately reflected in Wikipedia within few minutes~\cite{keegan_hot_2011}. Second, due to the Wikipedia policies~\cite{wiki_policy}, especially the \emph{verifiability}\footnote{\url{https://en.wikipedia.org/wiki/Wikipedia:Verifiability}} policy, it recommends Wikipedia contributors to support  their additions with  references from authoritative external sources. In particular, this policy states that \emph{``articles should be based on reliable, third-party, published  with a reputation for fact-checking and accuracy.''}\footnote{\url{https://en.wikipedia.rg/wiki/Wikipedia:Identifying_reliable_sources}}. This policy, on the one hand, guides contributors towards both neutrality and the importance of authoritative assessment and, on the other hand, allows Wikipedia core editors to identify unreliable articles more easily via a lack of such citations. Citations therefore play a crucial role in ensuring and upholding Wikipedia reliability, leading to high quality and important information which is harnessed by the Web users in general and the above mentioned applications.

Despite the established policies and the speed with which editors provide content for Wikipedia articles, these articles vary heavily in terms of quality.  First, articles vary in their popularity, hence, their affinity to attract editors and thus provide content. For instance, 51\% of Wikipedia articles are categorized as \emph{Stub} or articles in a need for expansion~\cite{wiki_stats}. Second, given that new information regarding articles in Wikipedia constantly emerges, it is hard to keep all articles up to date. Apart from such information overload, the number of active editors\footnote{The definition of an \emph{active editor} based on Wikipedia refers to registered users that have contributed in Wikipedia in the last 30 days (for any time point of measurement).} at a given time point varies, and is far smaller than the total of 30 million registered editors~\cite{wiki_stats}. Naturally, the amount of active editors and other editors demographics and editor's interests will impact the \emph{coverage} of articles that are kept up to date. Furthermore, there is an inherent delay between the time a real-world event happens which is relevant to a Wikipedia article and the time it is reflected in Wikipedia~\cite{DBLP:conf/websci/FetahuAA15}. In addition, changes in a specific Wikipedia article may cause information on other articles to be inconsistent. Finally, for any provided citation in Wikipedia and the text it is cited in, it is not possible to determine the span of text it covers. This has implications in enforcing the \emph{verifiability} policy, where situation may arise in which a paragraph in a Wikipedia article containing a citation may be only partly covered by a reference.

\section{Scope of the Thesis}\label{sec:thesis_scope}

Motivated from the importance and wide use of Wikipedia as a resource for a wide range of tasks and its high popularity amongst Web users, we address three core issues which deal with consistency, keeping up to date, and providing trustworthy information for Wikipedia: (i) finding news citations for Wikipedia statements, (ii) citation span determination, and (iii) enrichment of Wikipedia entity pages with novel and important news articles. 

Before delving into the details of the problems we address, we clarify some notions we will use throughout this thesis. We will use \emph{Wikipedia article}, with which we refer to Wikipedia \emph{entity} and \emph{event} pages, whereas with \emph{Wikipedia entity page} we refer to only entities. Whereas, with \emph{Wikipedia statement} we will refer to the \emph{piece of text}, ranging from a sentence up to a paragraph, that \emph{has} or \emph{needs} a citation.

\paragraph{(I)} Despite the growing trends in terms of the number of entity pages in Wikipedia, and those that adhere to the editing policies in Wikipedia, there is a large set of entity pages whose already existing citations are either outdated or not accessible. Furthermore, as new information is constantly added by Wikipedia editors, it is of great importance to automate or at the very least aid the editors in finding the appropriate citations. Additionally, there are cases in which statements are explicitly marked with \emph{citation needed}; the trust and truthfulness of such statement is into question. This indeed may be the case where the statement is simply not true, however, more often the citation is simply missing and it can be found from sources like \emph{news collections}.

For this problem, where we are given a \emph{piece of text} (we will define later the granularity of the textual fragments that we consider) from a Wikipedia article, we lay out two fundamental research questions. 
\\\\
\emph{
\noindent\textbf{RQ1.1.} For a \emph{statement} from a Wikipedia entity page, how can we determine the required type of citation (e.g. \emph{news}, \emph{web}, \emph{book} etc.)? }
\\\\
\noindent The outcome from \textbf{RQ1.1} is of great importance in finding appropriate citations for any given statement which adhere to the Wikipedia policies. There are many scenarios where specific citation categories are preferred. For instance, for entity pages in the medical domain, a more authoritative reference would be a citation coming from a medical \emph{journal}. In other cases, the availability of specific sources may restrict the statements for which we can suggest a citation.

Next, after knowing the desired citation category of a statement, the problem is how to find such references to cite. This brings us to the second research question which we postulate as following.
\\\\\emph{
\noindent\textbf{RQ1.2.} For a Wikipedia statement that requires a news citation, how can we find news citations which provide evidence for the statement under consideration?}
\\\\
\noindent Automating the process of providing citations as postulated in \textbf{RQ1.2} has several advantages. First, it addresses the problem of \emph{long-tail} entity pages, which suffer due to the lack of interest by  Wikipedia editors. Second, because Wikipedia is at a constantly evolving state, providing citations in an automated manner will ease the process of editing and serve as a complementary mechanism for Wikipedia editors. Finally, through automation it is possible to enforce in an objective manner the Wikipedia policies without falling into issues that in many cases lead to edit wars and disputability in Wikipedia.

\paragraph{(II)} It is evident from the problem in \textbf{(I)} that in Wikipedia, determining the granularity for which a citation is valid is somewhat ill-defined. The reasons for this is that there are no explicit requirements and furthermore no means on specifying for what part of text a citation is valid.  There are several consequences as a result of this. For instance, if we take a paragraph from a Wikipedia article and a reference cited from a paragraph, we are not able to tell for which part of the paragraph the citation provides evidence for. We formalize the research question addressing this issue as following:
\\\\\emph{
\noindent\textbf{RQ2.} For a paragraph which we extract from a Wikipedia article and a reference cited within the paragraph, how can we accurately determine the span of the citation?}
\\\\
\noindent Determining the span of a citation, that is, singling out at a fine-grained level what a citation covers in a paragraph extracted from a Wikipedia article has important implications. By accurately knowing the span it is possible to have a closed cycle where for uncovered parts in a paragraph we find citation as postulated in \textbf{(I)}. Otherwise, for statements which do not have a citation from the appropriate source type, it may be an important signal on the truthfulness and validity of a statement.

\paragraph{(III)} Finally, apart from recommending citations for already existing content in Wikipedia article, and finding their corresponding span, a highly important issue remains with emerging information from Web sources regarding a specific Wikipedia article or missing information. Specifically, for a given news collection, we deal with the problem of suggesting news articles to Wikipedia articles, which in turn can be processed by Wikipedia editors in order to add the encoded information within these sources. To this end we postulate the following two research questions.
\\\\\emph{
\noindent\textbf{RQ3.1.} For a Wikipedia entity and a news collection, how can we find news articles in which the entity is a \emph{salient} concept and at the same time the news article provides important and novel information for the entity?}

\noindent After addressing the question \textbf{RQ3.1}, which we refer to as the \emph{article-entity} placement task, we proceed and answer the second question which deals with finding the appropriate \emph{section} in the Wikipedia article. 
\\\\\emph{
\noindent\textbf{RQ3.2.} For a Wikipedia entity and a suggested news article, how can we find the \emph{appropriate} section within the entity, and in case such a section is \emph{missing} how can we automatically add the appropriate section in the Wikipedia entity page and \emph{suggest} the news article for?}
\\\\
\noindent The second research question addresses an important issue on  suggesting novel information to Wikipedia articles and to specific sections. Due to the fact that information regarding Wikipedia articles constantly evolves, such sections in many cases might be missing. Therefore, to address fully \textbf{RQ3.2} one needs to be able to add missing sections for which the news is relevant. For example, for a Wikipedia article \texttt{Barack Obama} before his \emph{US presidential election}, the corresponding article did not contain a section about \emph{US Presidency}, hence, in this case, a new section should be suggested automatically, in order to suggest news articles at the appropriate section and at a fine-grained level.

\newpage
\section{Contributions of the Thesis}\label{sec:thesis_contributions}
\begin{figure}[h!]
	\centering
	\includegraphics[width=1.0\textwidth]{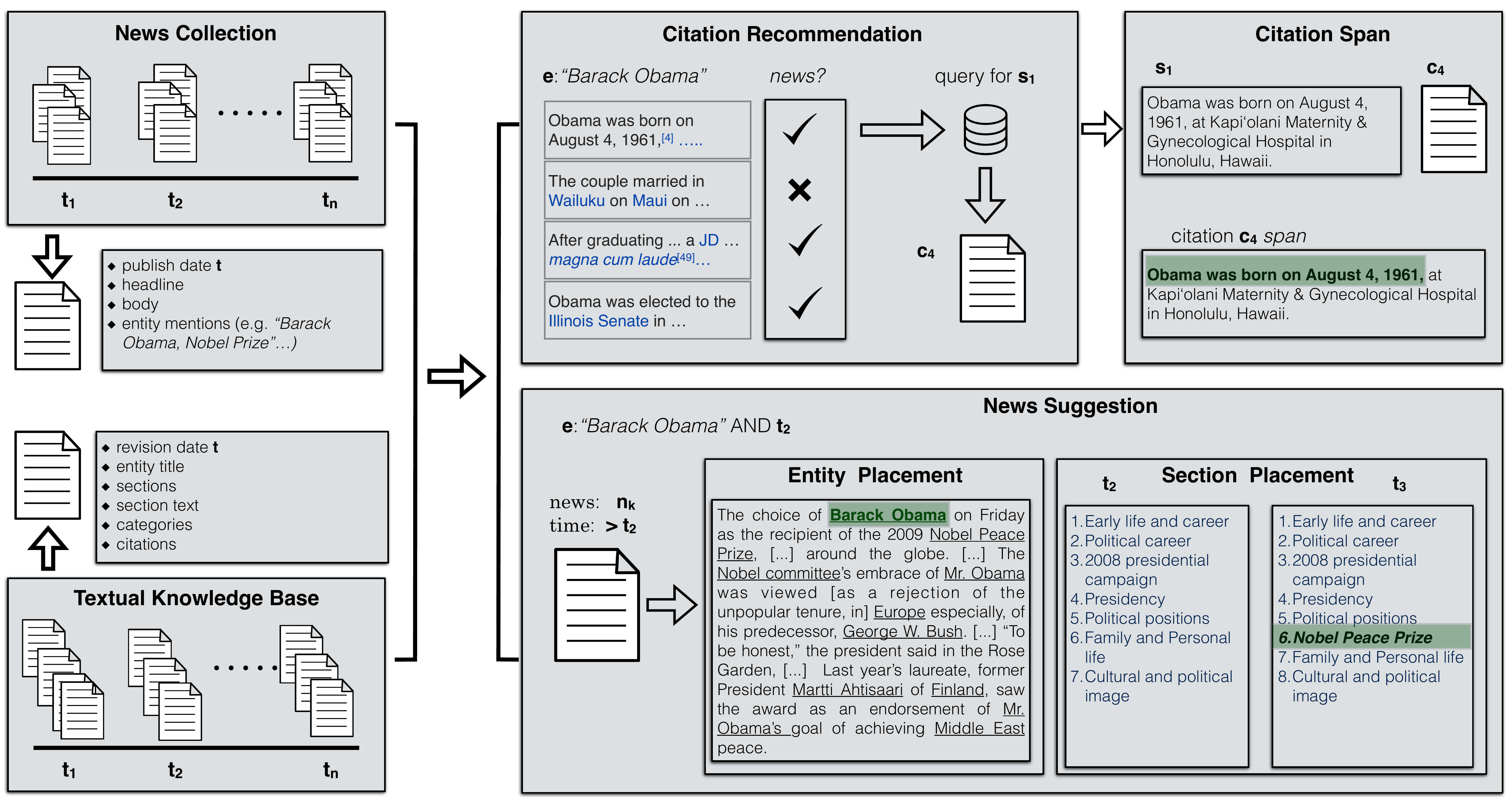}
	\caption{Overview of the proposed approach for enrichment and improvement of textual knowledge bases. The approach shows the three main steps: (i) \emph{Citation Recommendation}, (ii) \emph{Citation Span}, and (iii) \emph{News Suggestions}.}
	\label{fig:general_approach_outline}
\end{figure}

In this thesis, we answer the research questions formalized in the previous section. The contribution of this thesis is on enriching and improving the quality of Wikipedia as one of the most well known textual knowledge bases in the Web. Figure~\ref{fig:general_approach_outline} shows an outline of our contributions and the proposed solutions for the three core problems listed in Section~\ref{sec:thesis_scope}. 

\begin{enumerate}[\textbf{(I)}]
	\item \textbf{\emph{News Citation Recommendation for Wikipedia}:} In Chapter~\ref{ch1:citation_recommendation} we propose a novel approach for finding \emph{news citations} in Wikipedia. We address the two research questions in problem \textbf{(I)}. 

		\begin{itemize}
			\item \textbf{RQ1.1.}	Firstly, for a Wikipedia article and a specific \emph{statement}, we propose an approach which determines the \emph{type} of resources that are appropriate for citation. Dependent on the statement at hand, there is a range of 12 citation types (e.g. \emph{news, web, journal, book, report, $\ldots$}) which can be chosen from. This is a prerequisite for finding appropriate citations that enforce the Wikipedia editing policies, where \emph{authoritative} sources are suggested. Which source type is considered more authoritative is dependent on the article and the statement, however, if an article is about medicine a source from a journal is preferred over a news article.	To determine the citation category, we rely on \emph{language style} and other \emph{structural} attributes we extract from Wikipedia articles. This addresses question \textbf{RQ1.1}.
			\item \textbf{RQ1.2.} Second, we focus on only the cases where for any piece of text in Wikipedia the required and appropriate citation is of type news. The reason for focusing on this specific type of citation is motivated in Chapter~\ref{ch:news_wiki_lag}. We automatically construct a query to find news articles from a news corpus. We determine which news article to suggest as evidence for the given statement based on \emph{textual entailment}, \emph{news authority}  and other \emph{centrality} measures.
		\end{itemize}
		The contributions from this chapter are published in:
	\begin{itemize}
	\item \cite{DBLP:conf/cikm/FetahuMNA16} Besnik Fetahu, Katja Markert, Wolfgang Nejdl, Avishek Anand: Finding News Citations for Wikipedia. In \emph{Proceedings of the 25th ACM International on Conference on Information
               and Knowledge Management}, CIKM 2016, Indianapolis, IN, USA, October 24-28, 2016, pages 337--346.

	\end{itemize}

	\item[\textbf{(II)}] \textbf{\emph{Fine-grained Citation Span for References in Wikipedia}:} In Chapter~\ref{ch5:cite_span} we propose an approach, where for a paragraph containing a \emph{web} or \emph{news} citation we determine the coverage span of the citation. The work is of importance for automated approaches on enriching and expanding Wikipedia. In this way, we can determine to what extent an entity page adheres to the \emph{verifiability} policy. Furthermore, we provide explicit markings of the statements in Wikipedia that are covered by a citation. Hence, tying this together with the contribution in Chapter~\ref{ch1:citation_recommendation} we close the cycle in which we find citations for uncovered statements until for all statements we can provide citations, in case they exist in a given news collection.
	
	The contribution in this chapter has been published in:

	\begin{itemize}
		\item \cite{DBLP:conf/emnlp17/FetahuMA17} Besnik Fetahu, Katja Markert, Avishek Anand: Fine Grained Citation Span for References in Wikipedia.  In \emph{Proceedings of the Conference on Empirical Methods in Natural Language Processing}, EMNLP 2017, Copenhagen, Denmark, September 7-11, 2017.
	\end{itemize}

	\item[\textbf{(III)}] \textbf{\emph{Automated News Suggestion for Populating Wikipedia Pages}:} In Chapter~\ref{ch3:news_suggestion} we propose a novel approach for accounting for the ever evolving nature of Wikipedia entity pages, respectively, the emerging information in news and Web sources in general. Furthermore, due to the varying popularity of Wikipedia articles and respectively their affinity to attract editors to provide content for such articles, we can retrospectively suggest missing information for such articles. The proposed approach addresses the following research questions:
	
	\begin{itemize}
		\item \textbf{RQ3.1.} First, for a news article, we consider the dual problem of determining the \emph{salient} entities in the news article and consequentially if the news article is of importance for the Wikipedia entity page at hand. In this way, we ensure that only important information is suggested where the entity is a salient concept. This addresses the research question in \textbf{RQ3.1}.
		\item \textbf{RQ3.2.} Second, considering the section structure of Wikipedia articles, it is important to determine precisely for which section a suggested news article is relevant. Due to the fact that Wikipedia articles evolve with new information becoming available, and along with that the section structure will change. Therefore, in our proposed approach we account for the two attributes, by first determining the appropriate section for which we suggest the news article, and in case such a section is missing we suggest its addition into the section structure.
	\end{itemize}
	
	The contribution in this chapter has been published in:

	\begin{itemize}
		\item \cite{DBLP:conf/cikm/FetahuMA15} Besnik Fetahu, Katja Markert, Avishek Anand: Automated News Suggestions for Populating Wikipedia Entity Pages.  In \emph{Proceedings of the 25th ACM International on Conference on Information and Knowledge Management}, CIKM 2015, Melbourne, Australia, October 19 - 23, 2015, pages 323--332.
	\end{itemize}
\end{enumerate}

Apart from our holistic approach on dealing with enrichment and improvement of Wikipedia articles, we additionally make the following contributions which provide the context for the work carried in this thesis.

\begin{itemize}
	\item[\textbf{(A1)}] \textbf{\emph{How much is Wikipedia lagging behind News?}} The importance of news in Wikipedia is acknowledged by its editing policies~\cite{wiki_policy}, where authoritative and third-party sources like news articles are suggested for citation. To better understand the actual use of news as citations in Wikipedia we analyze how news are reflected in Wikipedia, respectively, the amount of time it takes for an entity to be reported in news and its occurrence in Wikipedia, and finally, how many of the citations from all citation categories are news citations. 
	
	The importance and interaction between news and Wikipedia is published in:
	\begin{itemize}
		\item \cite{DBLP:conf/websci/FetahuAA15} Besnik Fetahu, Abhijit Anand, and Avishek Anand. How much is wikipedia lagging behind news. In \emph{Proceedings of the ACM Web Science Conference, WebSci 2015}, Oxford,
               United Kingdom, June 28 - July 1, 2015, pages 28:1--28:9.
	\end{itemize}
	
	\item[\textbf{(A2)}] \textbf{\emph{Improving Entity Retrieval in Structured Data.}} Finally, we look into application use cases, where Wikipedia   drives several major industry projects on constructing knowledge graphs by Google~\cite{DBLP:conf/kdd/0001GHHLMSSZ14}, Yahoo!~\cite{Blanco:2011:EEE:2063016.2063023}, Microsoft~\cite{DBLP:conf/www/NieMSWM07} which drive the functionalities behind entity search. 
	
	In this case, we address several of the shortcomings that mostly deal with the nature of such structured datasets (e.g. DBpedia, Freebase etc.) and propose a new query similarity model for entity search. The contribution of this work has been published in:
	
	\begin{itemize}
		\item \cite{DBLP:conf/semweb/FetahuGD15} Besnik Fetahu, Ujwal Gadiraju, and Stefan Dietze. Improving Entity Retrieval on Structured Data. In \emph{The Semantic Web - ISWC 2015 - 14th International Semantic Web Conference}, Bethlehem, PA, USA, October 11-15, 2015, Proceedings, Part I, pages 474--491.
	\end{itemize}
\end{itemize}


\clearemptydoublepage
\chapter{Foundations and Technical Background}\label{ch2:foundations}

In this chapter, we introduce the technical background necessary to understand the work carried in this thesis. We first introduce the notion of knowledge bases, then continue on entity linking techniques. Next, we provide a thorough analysis of information retrieval techniques. We then describe clustering techniques, and finally conclude with supervised learning and feature selection algorithms.

\section{Knowledge Bases}\label{sec:knowledge_bases}

With the term \emph{knowledge base} (KB) we refer to RDF datasets published based on a set of \emph{linked data} principles introduced by Berners-Lee et al.~\cite{berners2001semantic}. Since then there has been a big push towards publishing data according to these principles. Nowadays, the number of datasets is in the range of thousands~\cite{lod_stats}. KBs are commonly referred to as  \emph{structured datasets, linked datasets} or \emph{RDF datasets}.

\subsection{Resource Description Framework -- RDF}
The term knowledge bases (KB) refers to RDF datasets published based on linked data principles introduced by Berners-Lee et al.~\cite{berners2001semantic}. This can be considered also as the inception of the field of Semantic Web. Since then, there has been a big push towards publishing data following the principles, known as the \emph{linked data} principles, introduced in \cite{berners2001semantic}. Nowadays, the number of datasets is in the range of thousands~\cite{lod_stats}. These datasets are represented in RDF format and  are interchangeably referred to as \emph{structured datasets, linked datasets} or \emph{RDF datasets}. 

Resource Description Framework, or RDF, is a graph data model proposed by W3C as a standard for knowledge representation~\cite{rdf}. The RDF data model consists of a set of \emph{resources}, \emph{predicates}, and \emph{literals}. 

A \emph{resource} refers to a real-world entity (e.g. person, organization etc.) or an abstract concept. We denote with $\mathcal{R}$ the set of resources in a KB. \emph{Literals}, $\mathcal{L}$, on the other hand, represent values like a string, date, number etc. Finally, \emph{predicates}, $\mathcal{P}$, represent a relation between resources or a resource and a literal. 

We define a KB to be the projection between these building blocks of the RDF data model. Hence, a knowledge base can be represented as following: $\mathcal{K} := \mathcal{R} \times \mathcal{P} \times (\mathcal{R} \cup \mathcal{L})$. Alternatively, a KB can be seen as simply a set of triples of the form $\langle s, p, o\rangle$, where $s\in \mathcal{R}$, $p \in \mathcal{P}$, and the object $o$ can represent a resource or a literal, thus, $o \in \mathcal{L}\cup \mathcal{R}$.

Listing~\ref{listing1} shows a set of triples describing a resource, which here represents  $s:=\text{\texttt{db:University\_of\_Hannover}}$.

\definecolor{mygray}{gray}{0.9}
\lstset{backgroundcolor=\color{mygray}}
\begin{lstlisting}[caption= RDF Resource Example for \emph{``University of Hannover''},label=listing1]
db:University_of_Hannover rdf:type 		owl:Thing .
db:University_of_Hannover rdf:type 		dbo:Agent .
db:University_of_Hannover rdf:type		dbo:EducationalInstitution .
db:University_of_Hannover rdf:type 		dbo:Organisation .
db:University_of_Hannover rdf:type		dbo:University .
db:University_of_Hannover dbo:city	 	db:Hannover .
db:University_of_Hannover dbo:country	db:Germany .
db:University_of_Hannover dbp:budget 	"€441.8 million"@en .
\end{lstlisting}

\subsection{RDF Schema -- RDFS}

A crucial construct in publishing RDF datasets is the organization of resources into \emph{classes}. This functionality is provided by RDFS~\cite{rdfs}. RDFS is an extension of the basic constructs provided by the RDF data model. A class in RDFS is defined by the set of triples in Listing~\ref{rdfs_class}. A resource $s$ is assigned to a class $c$ by the triple $\langle$ \emph{s}, \texttt{rdf:type}, \emph{c}$\rangle$, similar to Listing~\ref{listing1} where $\langle$ \texttt{db:University\_of\_Hannover}, \texttt{rdf:type}, \texttt{owl:Thing}$\rangle$.

\begin{lstlisting}[caption= RDFS Class Definition Example,label=rdfs_class]
dbo:University 		rdf:type					rdfs:Class .
dbo:University 		rdfs:subClassOf 	dbo:Organisation .
dbo:Organisation 	rdfs:subClassOf 	dbo:Agent .
dbo:Agent					rdfs:subClassOf		owl:Thing .
\end{lstlisting}

RDFS allows us to express hierarchy relations between different classes through \texttt{rdfs:subClassOf}. As such, the resource \texttt{db:University\_of\_Hannover} assigned to class \texttt{dbo:University}, by traversing the class hierarchy we assume that it is also of type \texttt{owl:Thing}. 

\subsection{Real-World Knowledge Bases}

The functionalities of RDF and RDFS, and similar Semantic Web initiatives on modeling and representing knowledge have resulted in many initiatives on representing data according to such standard and principles. Arguably, some of the most well known examples, include knowledge bases like DBpedia~\cite{DBLP:journals/ws/BizerLKABCH09}, YAGO~\cite{suchanek_yago:_2007}. Such KBs represent a subset of the information contained in Wikipedia articles, and represented according to linked data principles~\cite{berners2001semantic}.

These KBs are particularly interesting for this thesis. They allow to construct homogeneous groups of Wikipedia articles based on a type taxonomy constructed for Wikipedia articles based on the category structure in Wikipedia. Such categorization of articles according to a type taxonomy is useful for the approach in Figure~\ref{fig:approach_overview}. In many cases articles that are of different types, e.g., articles about \emph{Places} and \emph{Politicians}, have completely different structure and the problems we tackle behave differently for the different  types. Therefore, treating such articles separately accounts for accurate and reliable models.

\section{Entity Linking and Disambiguation}\label{sec:entity_linking}

In this section, we present an overview of state-of-the-art approaches on entity linking (EL) and disambiguation (NED) techniques. The task here corresponds to canonicalizing \emph{surface forms} or \emph{text phrases} to entities on a given database of entities. In majority of the cases~\cite{Hoffart:2011:RDN:2145432.2145521,DBLP:conf/i-semantics/MendesJGB11,DBLP:journals/software/FerraginaS12} Wikipedia is used as the target to link such surface forms to entities.

\begin{figure}[h!]
	\centering
	\includegraphics[width=0.8\textwidth]{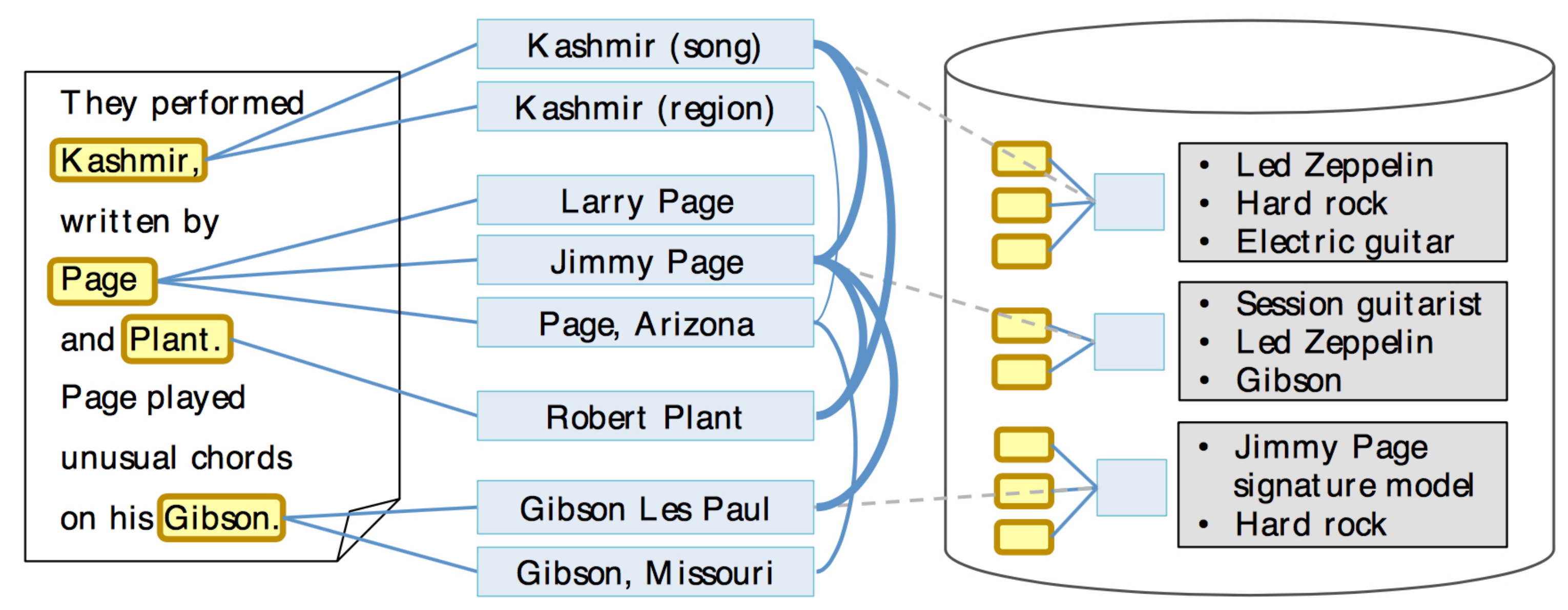}
	\caption{Mention-Entity graph for an example text snippet with ambiguous entity mentions~\cite{Hoffart:2011:RDN:2145432.2145521}.}
	\label{fig:ned}
\end{figure}

The core problem in this task is to resolve \emph{ambiguous} mentions of entities in free text. Figure~\ref{fig:ned} highlights the problem of resolving entity mentions from a text snippet into entities in Wikipedia.

We highlight two main differences between state of the art approaches on NED and EL. In NED systems, the candidate mentions from a text snippet are limited to only those that resolve to a named entity of type $\{$\emph{Person, Location, Organization, Time}$\}$. This is usually done through named entity recognition (NER) approaches like~\cite{DBLP:conf/acl/FinkelGM05}. Contrary to NED approaches, in EL there is no restriction in the mentions that can be resolved to entities on a target knowledge base. For instance, through EL systems one can resolve a mention referring to an event, e.g. \emph{U.S Presidential Elections 2016} to its corresponding Wikipedia article\footnote{\url{https://en.wikipedia.org/wiki/United_States_presidential_election,_2016}}, which through NED approaches is not possible. 

In this thesis, we opt for entity linking approaches due to their wider coverage of linking mentions from news and Web sources to Wikipedia articles. While the different entity linking and disambiguation approaches differ on their final result of disambiguated or linked entities, however, on common attribute of both applications is that they consider the \emph{coherence} and \emph{contextual} similarity between entity candidates in a textual snippet as part of their linking or disambiguation results.

\paragraph{NED.} AIDA~\cite{Hoffart:2011:RDN:2145432.2145521} is a state of the art approach on named entity  disambiguation. For a text snippet as shown in Figure~\ref{fig:ned}, AIDA performs the following operations to accurately link ambiguous entity mentions into a KB. Based on a pre-processing step where through NER are extracted mentions to named entities it generates a list of entity candidates from the target KB. Finally, the disambiguation is performed jointly by constructing a graph of mentions and entity candidates from the target KB. The goal is to find a dense sub-graph that fulfills the following properties: (i) high \emph{contextual similarity} between mentions in the text and the candidate entities in KB, (ii) weighted edges amongst the candidate entities, which measure their \emph{coherence}. The coherence in this case is a function of the number of incoming links two entities share in a KB. Common incoming links for any two entities $e_1$ and $e_2$ in a KB can be defined by the following triples $\langle$\emph{x}, \emph{p}, $e_1\rangle$ and $\langle$\emph{x}, \emph{p}, $e_2\rangle$.

\paragraph{EL.} The work by Milne and Witten~\cite{DBLP:conf/cikm/MilneW08} is one of the most notable works and serves as the basis for many EL approaches. It relies on \emph{anchor text} from Wikipedia articles which are used to learn models, and are later applied on textual resources to link specific phrases to Wikipedia articles. A commonality between AIDA and this work is the coherence between entities~\cite{witten2008effective} and is computed occurring in a textual snippet, and contextual similarity. The coherence score is computed as in Equation~\ref{eq:milne_witten}. 

TagMe~\cite{DBLP:journals/software/FerraginaS12} is a state of the art in entity linking. The advantage of TagMe compared to other existing approaches, is that it optimizes for short textual snippets as well. This provides an advantage considering that a significant proportion of news and web resources in general are not very lengthy. TagMe follows a similar scheme on performing the entity linking, however, with improvements on disambiguating surface forms that map to anchor texts in Wikipedia and their corresponding articles based on a voting scheme. In the voting scheme, for each anchor text and a candidate article it is computed an average relatedness score (see Equation~\ref{eq:milne_witten}) w.r.t the other articles that are candidates from other existing anchors in a textual snippet. Finally, to account for the efficiency and accuracy of the approach, the candidate articles are pruned based on the link probability (the score that an anchor text links to a Wikipedia article) and the voting score for each anchor and Wikipedia article pair.

\begin{equation}\label{eq:milne_witten}
rel(e_1, e_2) = \frac{\log \left( max(|E_1|, |E_2|) - \log(|E_1 \cap E_2) \right)}{\log(|E|) - \log(min(|E_1|,|E_2|)}
\end{equation}
where $E_1$ and $E_2$ represent the set of incoming links to entities $e_1$ and $e_2$, respectively. $E$ represent the set of all entities in a KB.

\section{Information Retrieval}\label{sec:information_retrieval}

Information Retrieval (IR) deals with the means on accessing and satisfying user information needs through querying of large collections, mostly of unstructured documents. Despite its foundations being on unstructured documents, IR has become a multi-modal field, providing techniques for access of multimedia objects and other structured datasets like KBs.

In this thesis, we will discuss relevant \emph{query models} within the scope of this thesis, respectively for textual collections and structured datasets. The main idea behind any query similarity model is the following.

\emph{For a document collection $\mathcal{D}$ which is projected into a vocabulary space of terms $\mathcal{V}$, and a query $q\in \mathcal{V}$, the task is to find relevant documents from $\mathcal{D}$ such that they satisfy the information need in $q$.}

From the task above we highlight two key points: (i) representation of documents and queries, and (ii) document relevance for a given query.

\subsection{Document and Query Representation}\label{sec:doc_representation}

\begin{figure}
	\centering
	\includegraphics[width=0.5\textwidth]{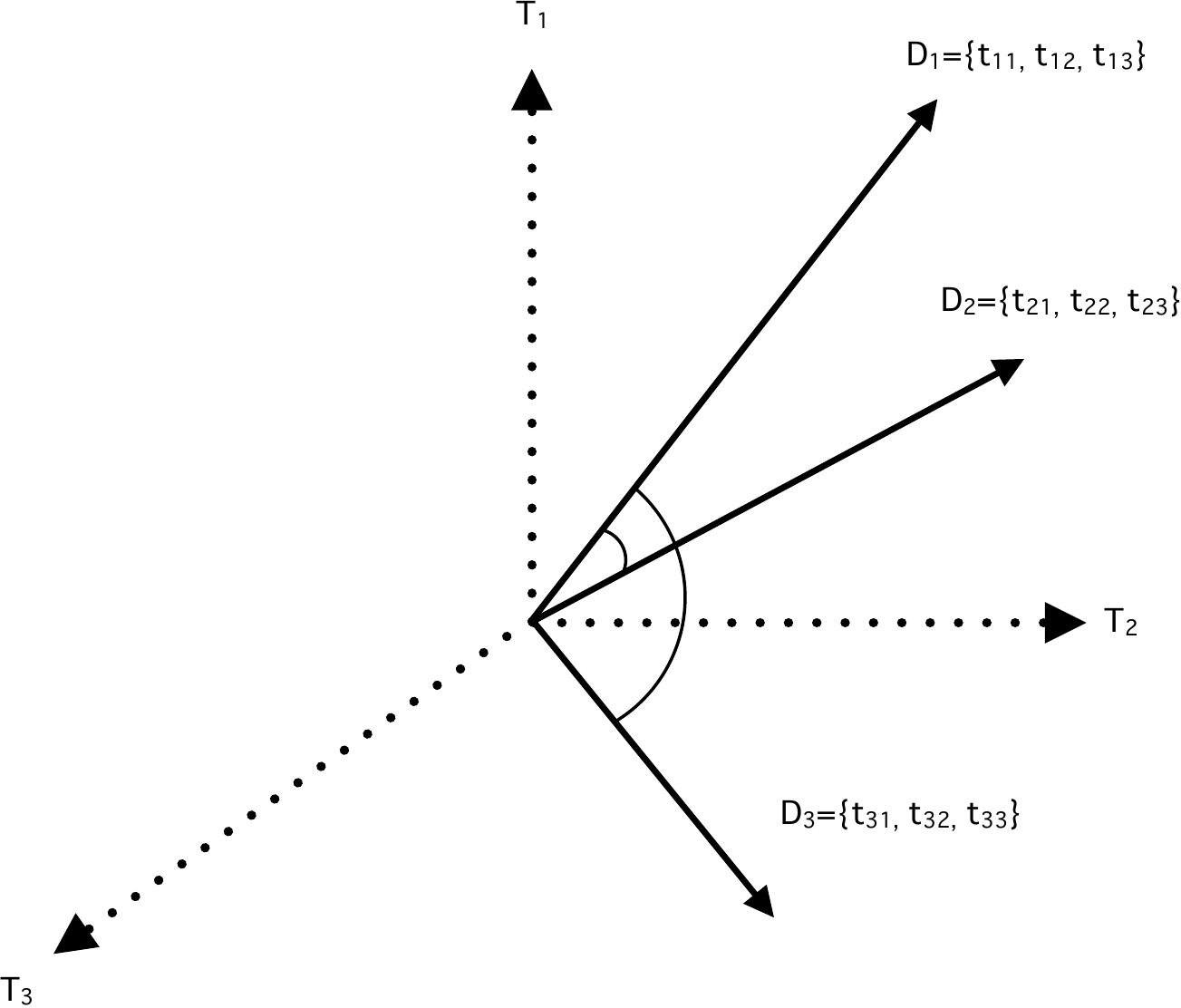}
	\caption{Vector representation of a document collection. Each term is drawn from a vocabulary and it represent a dimension in the document representation.}
	\label{fig:vector_space}
\end{figure}

The de-facto representation of documents and queries is based on the proposed vector space model by Salton et al.~\cite{DBLP:journals/cacm/SaltonWY75}. A document is represented by the terms occurring in the document, and for each term we can assign boolean indicator values or some form of weight reflecting the importance in the document. Similarly, queries are represented into the vector space model. Through this representation it is easy to compute the relevance of a document for a given query. This leads to ranked retrieval, where the documents are ranked according to their relevance to the query. Figure~\ref{fig:vector_space} shows a three-dimensional~\cite{DBLP:journals/cacm/SaltonWY75} representation of a document collection and the similarity between documents based on the dot product of the vector representations.

Here, it is assumed that the vector space representation is drawn from a vocabulary of terms $\mathcal{V}$ which corresponds to all the terms (words, or stemmed words) in the given document collection $\mathcal{D}$.

Hence, a crucial part in computing the similarity between a query and a document based on their vector space representations is the assignment of the weights for the term occurrences. The most widely used weighting scheme is based on the \emph{tf-idf}~\cite{manning2008introduction}. That is, the \emph{term frequency} or $tf$ measures the frequency of a term $v\in \mathcal{V}$ in a document $d\in \mathcal{D}$, whereas the $idf$ or \emph{inverse document frequency} counts the number of documents in which the term $v$ occurs. Through $tf$ it is measured the importance of term for a document, whereas with $idf$ it is measured how well such a term distinguishes a document from others. Their combination yields the trade-off between the two, and its simplest variation is computed as in Equation~\ref{eq:tf_idf}.

\begin{equation}\label{eq:tf_idf}
	tfidf = tf(v,d) \cdot \underbrace{\frac{|\mathcal{D}|}{df(v)}}_{\text{idf}}
\end{equation}
where, $df(v)=|d\in\mathcal{D}: v\in d|$, representing the number of document in $\mathcal{D}$ containing term $v$.

\noindent Through the $tfidf$ term weighting, document retrieval becomes a function of ranking documents based on the sum of the query terms present in a document $d$. For example, a basic ranked retrieval of documents is shown in Equation~\ref{eq:tf_idf_rank}.
\begin{equation}\label{eq:tf_idf_rank}
	r(q, d) = \sum_{v\in q}tfidf(v,d)
\end{equation}

\subsection{Information Retrieval Models}\label{subsec:retrieval_models}
\paragraph{Okapi BM25. } One of the most widely used retrieval models is BM25~\cite{robertson1995okapi}. Contrary to the model based solely on the \emph{tfidf} scores, BM25 requires parameter tuning that are dependent on the given document collection $\mathcal{D}$. Furthermore, in its query-document scoring function it takes into account the document length. The BM25 scoring model is shown in Equation~\ref{eq:bm25}.

\begin{equation}\label{eq:bm25}
r(q,d) = \sum_{v\in q}w_{tf}(q,d)\cdot w_{idf}(v)
\end{equation}
where, the term frequency score $w_{tf}(q,d)$ is computed as following:
\begin{equation}\label{eq:bm25_tf}
w_{tf}(q,d) = \frac{(k_1 + 1)\cdot tf(q,d)}{k_1 \cdot \left((1-b)+b \cdot \frac{len(d)}{avglen}\right) + tf(q,d)}
\end{equation}
where, the parameters $k_1$ ($k_1\geq 1$) and $b$ ($0\leq b \leq 1$) are tunable, and are usually set to values $k_1=1.2$ and $b=0.75$, respectively. Here, $b$ controls how much we normalize the term frequency scores according to the document length and its ratio to the average document length in $\mathcal{D}$. With $len(d)$ and $avglen$ we note the length of document $d$, and average document length in $\mathcal{D}$, respectively.

\noindent The inverse document frequency score $w_{idf}$ for a query term is computed as following:
\begin{equation}\label{eq:bm25_idf}
	w_{idf}(v) = \log \frac{N -df(v)+0.5}{df(v)+0.5}
\end{equation}
here, $N$ represents the number of documents in $\mathcal{D}$, and $df(v)$ as defined above represents the number of documents containing term $v$.

\paragraph{Divergence from Randomness -- \emph{DFR}.} Contrary to BM25, DFR is a non-parametric approach~\cite{Amati:2002:PMI:582415.582416}. The advantages of non-parametric models is that there is no need to rely on expensive parameter tuning approaches or have collection specific parameters. 

In \emph{DFR}, contrary to probabilistic models like BM25, where the documents are ranked based on their relevance, here the documents are ranked according to the \emph{gain} in retrieving a document given a query term. The basic idea behind \emph{DFR} is the weighting of the query terms for a given document collection. The weighting is done according to the scheme in Equation~\ref{eq:dfr_weights} which computes two probability distributions.

\begin{equation}\label{eq:dfr_weights}
	w(v) = -\log_2{P_1}^{(1-P_2)}
\end{equation}

\noindent The first probability distribution $P_1$ is a function of the within-document term frequency and reflects the chance of having a specific term frequency of a term $v$ in a document $d$. The intuition here is to measure the information gain of a specific term w.r.t a document collection. The higher the probability $P_1$ for a term the lower its gain. This is usually refers to the notions of \emph{``nonspeciality''} and \emph{``speciality''} words on a given document collection~\cite{DBLP:journals/jasis/Harter75}. Hence, for $P_1$ we have $-\log_2{P_1}$ which is a monotonically decreasing function and represents how informative a term is for a document w.r.t a collection.

\noindent In the second probability distribution $P_2$ it is measured the likelihood of the term occurrence only to a subset of documents in a collection called the \emph{elite documents}. This set can differ on how it is constructed. For instance, in Harter et al.~\cite{DBLP:journals/jasis/Harter75} this is defined as the set of documents which are important for a given term, while in \cite{Amati:2002:PMI:582415.582416} this is simply represented by the documents that contain a given term. Finally, the intuition here captures how \emph{informative} a term is, that is, the lower the expected score of $P_2$ the higher the gain of the term is as it can be seen from Equation~\ref{eq:dfr_weights}.

Finally, the two probability distributions are subject to the term frequency, the size of the document collection, the size of the elite set, and the frequency of a term in the elite set. Whereas, the probability models or the randomness models usually range from the \emph{Binomial distribution}, \emph{Poisson distribution}, etc.~\cite{Amati:2002:PMI:582415.582416}.

\section{Clustering Approaches}\label{sec:sampling_clustering}

In this section, we describe clustering approaches which are used in this thesis. Clustering techniques are often used in the context of unsupervised learning, where data items are grouped together based on a pre-defined notion of similarity. In some cases, clustering is used to improve efficiency of algorithms, where pair-wise similarity is required and as such by limiting the comparisons only to a limited set of data items provides drastic gains in terms of efficiency.

Here we will discuss three main clustering techniques. First, we describe the most fundamental clustering algorithm \emph{k--means}~\cite{hartigan1979algorithm} and a more sophisticated version of it, namely \emph{X--means}~\cite{pelleg2000x}. Next, we discuss \emph{Locality Sensitive Hashing} (LSH)~\cite{leskovec2014mining} which clusters data items based on generated min-hash signatures. Finally, we describe \emph{spectral clustering}~\cite{DBLP:journals/sac/Luxburg07} a clustering algorithm which relies on matrix factorization approaches.

\subsection{k--means Clustering}\label{subsec:kmeans}

\emph{k--means} or Lloyd's algorithm~\cite{hartigan1979algorithm} is one of the first and most widely used clustering algorithms. In this algorithm, data items with a multidimensional representation, where this representation can be arbitrary (e.g. document terms, feature values), are grouped into $k$ clusters. In the first step, the algorithm chooses $k$ random \emph{cluster centroids}, which correspond to random data items, and data items are associated with one of the clusters based on the minimal Euclidean distance between the centroid and the data item. The process is iterative, where after each iteration the cluster centroids are updated with the new representation from all its data items assigned to it, and the data items are re-assigned to their closest centroids. The process stops when the data items do not change their cluster assignment.

The quality of clustering is measured by the \emph{residual sum of squares} (RSS) which captures the distance from a data item to its cluster centroid. 

\begin{equation}\label{eq:rss}
	RSS_k=\sum_{\vec{x}\in \omega_k}|\vec{x}-\vec{\mu}({\omega_k})|^2
\end{equation}
where $\vec{x}$ represents an item assigned to cluster $\omega_k$, whereas $\vec{\mu}({\omega_k})$ represents the cluster centroid.

The main disadvantage of the \emph{k--means} algorithm is the choice of the $k$ for the number of clusters, and the initial assignment of the cluster centroids. For the latter issue, there are approaches that consider the inter-centroid distance in the initial step, trying multiple initial cluster assignments, or initial centroids which are based on prior information~\cite{DBLP:conf/icml/BradleyF98}.

\subsection{X--means Clustering}\label{subsec:xmeans}
 Pelleg and Moore~\cite{pelleg2000x} proposed \emph{X--means}, a clustering technique which overcomes three main drawbacks of the standard \emph{k--means} algorithm. First, it provides a more efficient algorithm for performing the clustering process. Second, it automatically finds the right number of clusters, which optimize the \emph{Bayesian Information Criterion} (BIC). Finally, it ensures that the clustering results do not fall into \emph{local minima}, as is the case for \emph{k--means}.
 
To find the right $k$ for clustering, as input it is required the \emph{lower} and \emph{upper} bound of the search space for which the $k$ can take values. The algorithm consists of three steps: (i) run the conventional \emph{k--means} for a specific $k$, (ii) improve the structure of the clusters by estimating whether new clusters should appear within the already existing $k$ clusters, that is, increase the number of $k$, and (iii) if $k > k_{max}$ report the best scoring model for BIC.

\noindent \emph{X--means} for a given $k_{max}$, in the worst case tries all $2^{k_{max}}$ configurations. That is, for each cluster, it considers if the BIC score is improved by further splitting into multiple clusters. The rate of splitting is determined by how close the current setting is to the true distribution of the data. 

In \emph{X--means}, BIC is computed as in Equation~\ref{eq:bic_xmeans}. $M_j$ corresponds to a clustering solution for a specific $k$, $D$ represents the data points, $|D|$ is the number of data points, $p_j$ is the number of parameters in the model $M_j$, finally $\hat{l}_j(D)$ represents the log-likelihood of the data according to the model $M_j$.

\noindent The estimation of the log-likelihood of the data is done under the assumption that the models are \emph{spherical Gaussians}, which is the assumption made by the \emph{k--means} algorithm. The log-likelihood $l(D)$ is shown in Equation~\ref{eq:xmeans_loglikelihood}, whereas by taking the derivative $\hat{l}$ we obtain the \emph{maximum log-likelihood}.
\begin{equation}\label{eq:bic_xmeans}
BIC(M_j) = \hat{l}_j(D) - \frac{p_j}{2} \cdot \log |D|
\end{equation}

\begin{equation}\label{eq:xmeans_loglikelihood}
	l(D) = \log \prod_i P(x_i) = \sum_i\left(\log \frac{1}{\sqrt{2\pi}\sigma^M} - \frac{1}{2\sigma^2}\lvert\lvert x_i -\mu_{(i)}\rvert\rvert^2 + \log \frac{|D_{(i)}|}{|D|}\right)
\end{equation}
the estimation of the variance $\sigma$ for a model $M_j$ with the assumption that the data is distributed under the spherical Gaussian model.

\subsection{Minhashing and Locality Sensitive Hashing}\label{subsec:lsh_clustering}

When dealing with a large set of documents or data items, for different clustering approaches a fundamental issue remains efficiency. For instance, sophisticated clustering approaches like \emph{spectral clustering}, computing \emph{adjacency matrices} or the pair-wise similarity between documents is highly expensive. This directly correlates to the number of documents, e.g. for 1000 documents the number of pairwise comparisons is already ${1000}\choose{2}$, with nearly 500k comparisons.

To improve on efficiency, there are approaches that construct signatures of documents and project them into a space where documents that may have slight similarity have similar representation. This gives rise to the computation of \emph{minhash signatures}, and respectively \emph{locality-sensitive hashing} clustering which uses the generated \emph{mihash} signatures in order to group likely similar items together.

Below we describe two techniques which allow to improve on efficiency of clustering large collections~\cite{leskovec2014mining}.

\paragraph{Minhash Signatures.} To avoid computation of pair-wise similarities between documents in a large collection, we need to be able to have compact representations of such documents to leverage it for clustering. \emph{Minhash signatures} compute the \emph{characteristics matrix}, which as rows contains the document representation (i.e. terms) and as columns the documents in a collection. A toy example from~\cite{leskovec2014mining} is shown in Table~\ref{tbl:minhash_signature}. Next, the minhash signature of a document from the characteristic matrix is the number of the row from the permuted set of terms representing the document collection where we encounter a term contained in a given document. For example the minhash signature of $h(d_1)=v_3$. 

A highly useful property of the compute minhash signatures is that for a random permutation of the terms representing the document collection, the probability that the minhash function produces the same signature for two documents is equal to the \emph{Jaccard} similarity. 

Enumerating all possible permutations of the representation space is expensive. One way to remedy such a problem and pick a random permutation of the rows representing a document collection is to map the rows through a set of hash functions. The hash functions map the rows into as many buckets as there are rows. Here one precaution to take is that the number of collisions or mapping of different rows into the same bucket is not high.

\begin{table}
	\centering
	\begin{tabular}{l | c | c | c | c| }
	\emph{item} & $d_1$ & $d_2$ & $d_3$ & $d_4$\\
	\hline\hline
		$v_1$ & 0 & 0 & 1 & 0\\
		$v_2$ & 0 & 0 & 1 & 0\\
		$v_3$ & 1 & 0 & 0 & 1\\
		$v_4$ & 1 & 0 & 1 & 1\\
		$v_5$ & 0 & 1 & 0 & 1\\
	\end{tabular}
	\caption{A characteristic matrix representing a set of documents (in the columns) and for the terms drawn from the vocabulary of terms from all documents.}
	\label{tbl:minhash_signature}
\end{table}

Instead of computing random permutations, we map the rows through a set of hash functions into random buckets (where the number of buckets is ideally as many as the number of rows). The minhash signatures are computed as following: for each document the signature corresponds to the number of hash functions, and for each column where the document has a specific term, in the signature each value is replaced with the lowest hash value mapping the rows to the random buckets.

\noindent Considering the same example as in Table~\ref{tbl:minhash_signature}, and if we take two hash functions $h_1(x)=x+1 \text{\texttt{ mod }} 5$ and $h_2(x)=3x+1 \text{\texttt{ mod }} 5$, with which we map the rows into their corresponding hash buckets, we have the following minhash signatures in Table~\ref{tbl:minhash_sign_ii}.

\begin{table}
	\centering
	\begin{tabular}{l || c | c | c | c|}
	& $d_1$ & $d_2$ & $d_3$ & $d_4$\\
 	\hline\hline
	$h_1$ & 1 & 3 & 0 & 1\\
	$h_2$ & 0 & 2 & 0 & 0\\
	\end{tabular}
	\caption{Minhash signatures after mapping the rows into the hashing buckets based on $h_1$ and $h_2$ and after replacing each hash column with the lowest hash bucket for each document for all its non-zero entries.}
	\label{tbl:minhash_sign_ii}
\end{table}

\paragraph{Locality Sensitive Hashing.} Through \emph{minhash signatures} we  can compress the representation of large collection while preserving the similarities between documents. However, we are still left with a combinatorial problem of pair-wise comparisons between documents. 

\noindent One way to remedy this problem is through the \emph{Locality Sensitive Hashing} (LSH) approach. It groups together documents that are up to a certain degree similar. We can do this by splitting the minhash signatures into $b$ bands or buckets consisting of $k$ rows  each. For each band we associate a hash function which takes the $k$ rows from each band and for each item and generates a hash value corresponding to some bucket. Similar vectors will be bucketed together by the same hash function. Furthermore, since the minhash signatures are generated from random permutations of the characteristic matrix, in one of the bands if there is any similarity between any two document they will be bucketed together by one hash function.

For any given threshold of Jaccard similarity $s$ for which two documents are considered candidates for pair-wise comparison, there are probabilistic guarantees that they agree on all rows in a band $b$ is $s^r$, whereas that they disagree in at least one row is $1-s^r$. Finally, we can generalize this and provide the probability of the signatures agreeing on all rows in at least one band $b$ with $1-(1-s^r)^b$.

\subsection{Spectral Clustering}\label{subsec:spectral_clustering}

\emph{Spectral clustering} is a popular clustering algorithm due to its ability to group data points of any arbitrary shape. Other algorithms like \emph{k--means} make strong assumptions on the form of the clusters, hence, they are not flexible in detecting arbitrary shapes like \emph{spectral clustering}. Furthermore, through \emph{spectral clustering} we do not get stuck in local minima and there is no need to restart the clustering process with different cluster centroids to obtain optimal results~\cite{DBLP:journals/sac/Luxburg07}. 

We explain the intuition behind these favorable properties of \emph{spectral clustering}. In spectral clustering, for any given set of data items (documents or any other type of data) we perform the following steps to compute the clusters:

\paragraph{Similarity Graph. } Assume we have a collection of documents $\mathcal{D}$. First, we construct the \emph{similarity graph}, $G=(V,E)$ where the vertices correspond to the documents $d\in \mathcal{D}$ and the edges are between documents which have some notion of similarity $s(d_i,d_j)$. In spectral clustering, there are mainly three approaches on generating the graph $G$: (i) $\epsilon$--neighborhood graph consists of all connected vertices whose distance based on $s(d_i,d_j)$ is less than $\epsilon$, (ii) $k$--nearest neighbors, with the graph consisting of the $k$ nearest vertices for $d_i$, and (iii) \emph{fully connected graph} where all the vertices are connected if their similarity is above zero, and the edge weights correspond to the similarity score $s$.

\noindent In the case of the similarity graph it is important the choice of the similarity measure in order to get meaningful clusters, however, one pre-condition for applying spectral clustering is such that the similarity $s(d_i, d_j)$ should be always positive and should be a symmetric measure.

\paragraph{Graph Laplacians. } From the constructed similarity graph $G$, we compute the \emph{graph Laplacians}. There are two main graph Laplacians that are used in spectral clustering: (i) \emph{unnormalized} and (ii) \emph{normalized} graph Laplacians. 

The unnormalized graph Laplacian $L=D-W$, where $D$ is the diagonal matrix, where each entry in the diagonal matrix corresponds to the sum of weights or similarities between all the items in that row, $d_i=\sum_{j}s_{ij}$ and $W$ is the adjacency matrix computed for data under consideration. Whereas, the normalized graph Laplacian is computed as $L_{sym}= I - D^{-1/2}WD^{-1/2}$, where $I$ represents the \emph{identity matrix}.

\paragraph{Clustering. } From $L$ (independent of which one is used), compute the first $k$ \emph{eigenvectors} ($u_1, u_2, \ldots, u_k$), which results in the matrix $U\in \mathbb{R}^{n\times k}$. Now from the computed eigenvectors from all the data points, usually a simple \emph{k--means} clustering algorithm is employed to cluster the items based on their eigenvectors, where as a similarity measure between such vectors is used the Euclidean distance.

Similarly as for many other existing clustering algorithms, an issue here is to determine the right number of clusters. However, there are simple heuristics that leverage the eigenvector space to find the right number of clusters. One common technique is to consider the distribution of \emph{eigenvalues}, such that the number of clusters $k$ corresponds to the first $k$ eigenvalues, that is, $\lambda_1, \ldots, \lambda_k$ are very small, but $\lambda_{k+1}$ is relatively high.

\section{Supervised Learning and Feature Selection}\label{sec:supervised_learning}

In this section we describe supervised learning approaches, and additionally distinguish a class of the so called \emph{structured prediction} in supervised learning. Finally, we describe how we can quantify the importance of features which are used in the learning algorithms.

\subsection{Supervised Learning}

For learning, we assume the setting where we are given a set of data items represented in an $n$--dimensional feature space. That is $X\in \mathbb{R}^{k\times n}$, where $X=\langle X_1, X_2, \ldots, X_k\rangle$. The task is to predict a discrete output $Y$, hence, we can generalize the learning approaches as to learning the function $f: X \rightarrow Y$.

\paragraph{Logistic Regression -- LR. }  It is one of the most simplistic and widely used supervised learning algorithms~\cite{bishop2006pattern}. For a set of training data $X$ it learns $n$ feature weights based on the \emph{maximum likelihood principle} (MLP). The classification model is represented by a linear function which for an instance $X_i\in X$ chooses the $y\in Y$ that maximizes the probability $P(Y=y|X_i)$. 

LR estimates the likelihood of an instance $X_i$ belonging to a class $Y=y$ as shown in Equation~\ref{eq:logit}. The most probable class is chosen by simply taking the $y\in Y$ which maximizes the probability $P(Y=y|X_i)$.

\begin{equation}\label{eq:logit}
P(Y=y|X_i) = \frac{exp(\theta_y + \sum_{j=1}^{n}\theta_{y,j}X_{i,j})}{\sum_{i=1}^{k}exp(\theta_y + \sum_{j=1}^{n}\theta_{y,j}X_{i,j})}
\end{equation}
where $\theta_{y,j}$ ($j=1,\ldots,n$) represents feature weights that  are estimated based on MLP for a given set of training data. This is done by providing $\theta$ as a free parameter, namely $\theta \leftarrow \argmax_{\theta}\prod_{i=i}^{k}P(Y_i|X_i, \theta)$.

While logistic regression has found wide adaptation for many classification tasks, one main disadvantage is its linearity, that is, it can classify accurately instances that are only linearly separable.

\paragraph{Random Forests -- RF. } They belong to a widely technique of supervised learning, namely \emph{ensemble learning}~\cite{Breiman2001}. In Random Forests (RF), the input space from a set of training instances is split into $K$ classification trees resulting into a \emph{forest}. The generation of trees follows two main principles. First, the input feature space is split into $K$ random vectors, or feature subsets chosen randomly, resulting in classification trees $h_K$. In the second approach, in the case of low-dimensional input feature space, one can employ a \emph{linear combination} of features and consequentially generate classification trees based on the CART algorithm~\cite{DBLP:books/wa/BreimanFOS84}.

The classification in RFs is performed based on the \emph{majority voting} scheme. That is, for a given instance $X_i$ we generate $K$ labels from the $K$ trees in our random forest and finally pick the label $y$ which was predicted by the majority of $K$ trees.

RFs~\cite{Breiman2001} provide theoretical guarantees on the performance of the classifiers, which is measured through a \emph{margin function} shown in Equation~\ref{eq:margin_rf}.

\begin{equation}\label{eq:margin_rf}
	margin(X, Y) = avg_k\mathbb{I}(h_K(X)= Y) - \max\limits_{Y'\neq Y} \mathbb{I}(h_k(X)=Y')
\end{equation}

\noindent The margin measures the distance in number of votes from the correct label for an instance versus the incorrect label which has the highest votes on average.

\begin{figure}[h]
	\centering
	\includegraphics[width=0.5\textwidth]{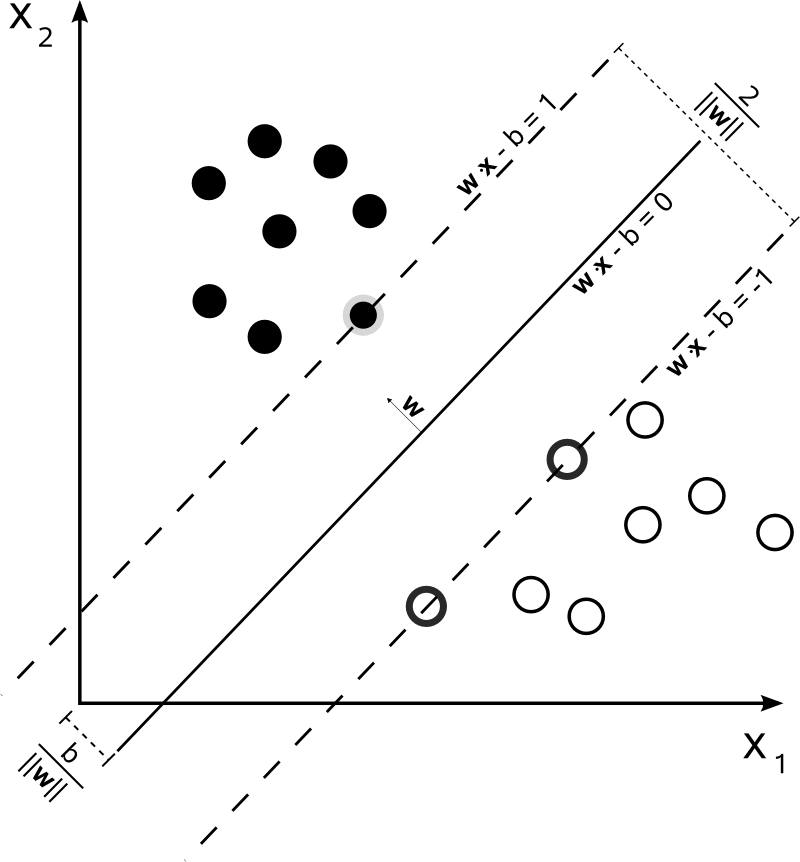}
	\caption{An example of linearly separable instances. The support vectors are the ones which are at the margins and they define the maximum distance between the support vectors between the two classes.}
	\label{fig:svm_support_vectors}
\end{figure}

\paragraph{Support Vector Machines -- SVM. } Are a widely used supervised learning approach when the input feature space for training instances it is high. Introduced first by Vapnik~\cite{DBLP:journals/ml/CortesV95}, the  task is to construct a \emph{hyperplane} which separates input instances linearly. An optimal hyperplane is constructed based on so called \emph{support vectors}, which determines the maximal margin between support vectors of different classes. Figure~\ref{fig:svm_support_vectors} shows an example of support vectors for an optimal hyperplane.

In more details, in SVMs for linear model $y(x) = \mathbf{w}^T\phi(X) + b$, where $\phi(X)$ denotes the multi-dimensional representation of instance $X$, we find the optimal weights $\mathbf{w}$ and parameter $b$ based solely on the support vectors. In this case, based on the linear model $y(x)$ the support vectors satisfy the equation $y_i(\mathbf{w}^T\phi(X) +b)=1$. In order to find the optimal weights, the task is to find the support vectors that maximize the distance between support vectors of two classes. Furthermore, in SVMs, through a \emph{kernel trick}, it allows for feature spaces that are larger than the number of instances to be classified efficiently. In cases where the instances are not linearly separable one can employ non-linear kernels which maps the feature space into a non-linear one where the classification can be performed accurately.

\noindent The optimal weights are estimated subject to the support vectors and are discussed in details in~\cite{DBLP:journals/ml/CortesV95,bishop2006pattern}. 

Finally, in the case of multi-class classification problems, SVMs consider two schemes which convert the multi-class classification into a binary classification problem. The first scheme refers to as \emph{one-versus-the-rest} (OvA), where the class of interest is considered as the positive class, whereas the others as negative classes. The class with the highest confidence based on the OvA scheme is assigned as the class of an instance. The second scheme, \emph{one-versus-one} (OvO) trains ${K \choose 2}$ classifiers for all pairs of classes, and classifies an instance based on the majority voting scheme from the resulting classifiers.

\subsection{Structured Prediction}\label{subsec:structured_prediction}

The main difference between discrete classification models (previously described approaches) and \emph{structured prediction} are the following. In the first case, we optimize for classifying an instance or data point to a discrete output, e.g. binary label or any multinomial variable. In structured prediction, the task is to predict jointly a sequence of instances or data points. That is, the instances in a sequence are dependent on each other and thus the name structured prediction aka sequence classification.

Examples of structured prediction are tasks such as gene segmentation~\cite{bernal2007global}, Named Entity Recognition~\cite{mccallum2003early} etc.

In structured prediction we have the following setup. For a set of instances $X=\langle X_0, X_1, \ldots, X_k\rangle$ which have the corresponding sequence of labels $\mathbf{y}=\langle y_0, y_1, \ldots y_k\rangle$, where $\mathbf{y}$ is a discrete set of values. 

\noindent The task is to predict for $X$ the vector of labels $\mathbf{y}$. To do this in an accurate manner, we need to be able to encode the dependencies between instances in the sequence $X$. Hence, this gives rise to \emph{graphical models} which can encode such inter-dependencies between instances in a sequence.

\paragraph{Conditional Random Fields -- CRF.} One of the most well known graphical models, for structured prediction, \emph{conditional random fields}, was proposed by Lafferty and McCallum~\cite{DBLP:conf/icml/LaffertyMP01}. CRFs belong to the so called \emph{discriminative models} where for a given input sequence $X$ and their labels $\mathbf{y}$, the goal is to model the conditional probability $p(\mathbf{y}|X)$. Here, similarly as in other classification approaches, the input instances are represented in a $k$--dimensional feature space. The advantage of CRFs and other structured prediction models is that it allows for arbitrary dependencies between instances and correspondingly the labels associated with them.

\begin{figure}[h]
	\centering
	\begin{subfigure}[t]{0.45\textwidth}
        \centering
        \includegraphics[width=0.7\textwidth]{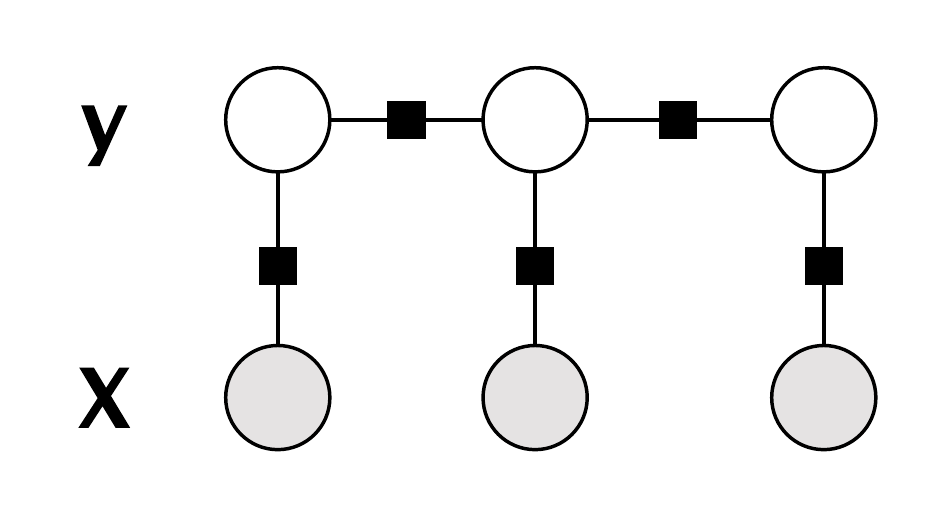}
        \caption{Linear-Chain CRF}
        \label{fig:linear_crf}
    \end{subfigure}
    ~
	\begin{subfigure}[t]{0.45\textwidth}
        \centering
        \includegraphics[width=0.7\textwidth]{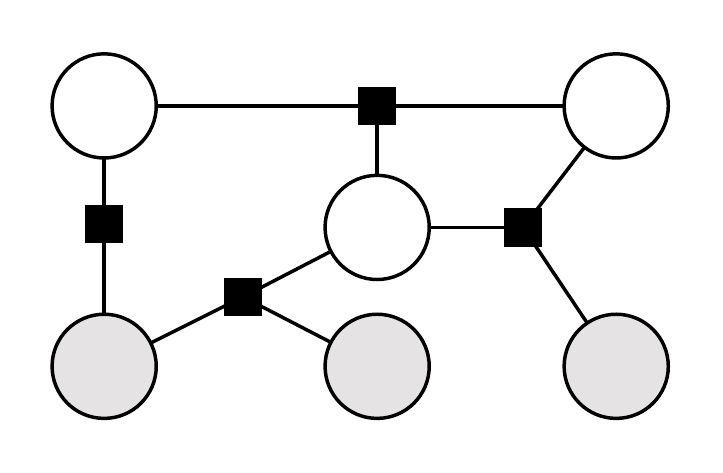}
        \caption{General CRF}
        \label{fig:general_crf}
    \end{subfigure}
    \caption{Modeling of dependencies between instances in a sequence according to linear-chain CRFs and general CRFs.}
    \label{fig:crf_models}
\end{figure}

\noindent For this purpose, CRFs represent a sequence as a graphical model, where we distinguish between the \emph{linear-chain} and \emph{general} CRFs. In \emph{linear-chain} CRFs the dependencies between instances in a sequence are limited to immediate neighbors as shown in Figure~\ref{fig:linear_crf}. In \emph{general} CRFs, the dependencies can be arbitrary and are subject to the problem which we try to model.

Figure~\ref{fig:crf_models} shows the two cases of CRFs. The models are represented as by the graph $G=(X,F,\mathbf{y})$, which is also referred as a \emph{factor graph}. This is because of the nodes $F$ in \emph{black squares} which represent the factors. Through factors in CRFs, we represent the dependencies between the instances in a sequence and their dependency w.r.t the labels. As it can be seen from Figure~\ref{fig:general_crf} such factors nodes can be arbitrary.

The factors in G are represented by a family of functions $F=\{\Psi_a\}$, which vary in the two models. We first show the estimation of $p(\mathbf{y}|X)$ for linear-chains and then discuss the difference with the general CRFs.

In linear-chains we estimate $p(\mathbf{y}|X)$ as shown in Equation~\ref{eq:linear_crf}. The factors $\Psi_a$ here are between two labels $y_i$ and $y_{i-1}$ and correspondingly between the observed sequence $X_i$ and $y_i$. Hence, the features $f_k(y_i, y_{i-1}, X_i)$ are computed only between the three nodes in $G$. Here we distinguish, between $f_{ij}(y,y',X)$ which represents features modeling the transition between two consecutive labels $y_i$ and $y_j$. Next, we distinguish $f_{io}(y,y',X)$ which models the dependency between the observed value $X$ w.r.t the label $y$.

\begin{equation}\label{eq:linear_crf}
 	p(\mathbf{y}|X) = \frac{1}{Z(X)}\prod\limits_{i=1}^{n}\exp\left\{\sum\limits_{k=1}^{K}\theta_k f_k(y_i, y_{i-1}, X_i)\right\}
\end{equation}
where $Z(X)$ is a normalization factor which allows us to turn the computed estimates into a probability distribution. $\theta_k$ are parameters which indicate the importance of certain features, and they are usually estimated through standard parameter estimation methods like the Maximum Likelihood or other more sophisticated approaches like Quasi-Newton Optimizers~\cite{bishop2006pattern}.

\begin{equation}\label{eq:linear_crf_normalization}
Z(X) = \sum_{y\in\mathbf{y}}\prod\limits_{i=1}^{n}\exp\left\{\sum\limits_{k=1}^{K}\theta_k f_k(y_i, y_{i-1}, X_i)\right\}
\end{equation}

Finally, the only difference between linear-chain and general CRFs is in the factor nodes. Since we can have arbitrary dependencies between the input sequences, labels and the factors, instead of $\prod_{i=1}^{n}$ which simply follows the linear-chain, here we have the set of factors $\Psi_a$ in $G$ as shown in Equation~\ref{eq:general_crf}. Figure~\ref{fig:general_crf} shows that the set of factors is arbitrary and can be between different sequences (in linear-chain a sequence was only factored w.r.t the label, and correspondingly the label was factor w.r.t the previous label) and label nodes in $G$.

\begin{equation}\label{eq:general_crf}
 	p(\mathbf{y}|X) = \frac{1}{Z(X)}\prod_{\Psi_a \in G}\exp\left\{\sum\limits_{k=1}^{K(a)}\theta_k f_k(y_i, y_{i-1}, X_i)\right\}
\end{equation}

\subsection{Feature Selection}\label{subsec:feature_selection}

An important aspect that is considered in in supervised learning is \emph{feature selection}. The importance of feature selection is manifold. Through feature selection we choose features that have higher discriminative power on distinguishing the instances of different classes, and further improve on efficiency of the generated classifiers. Furthermore, we can improve the generalizability of a trained model on unseen examples.

\paragraph{Information Gain. } We explain a standard feature selection algorithm that is based on the concept of mutual information gain~\cite{DBLP:conf/icml/YangP97}. 

\noindent The intuition behind IG feature selection is to assign higher weight to features, respectively to feature values assigned to instances, which enable us to distinguish them into different classes. Table~\ref{tbl:ig_examples} shows an example behind the intuition of IG feature selection. It is evident that the example features have high discriminatory power to distinguish between instances belonging to the different classes. 

\begin{table}[h]
	\centering 	
	\begin{tabular}{l l l}
	\toprule
	\emph{word frequency} & \emph{class=winter} & \emph{class=summer}\\
	\midrule
	$snow$ & 10 & 0\\
	$ski$ & 25 & 0\\
	$warm$ & 2 & 50\\
	$beach$ & 10 & 120\\
	\bottomrule
	\end{tabular}
	\caption{An example feature of word occurrence for classifying documents into \emph{winter} and \emph{summer} class.}
	\label{tbl:ig_examples}
\end{table}

To quantify the discriminative or classification power of a specific feature, respectively its values, we measure based on Equation~\ref{eq:ig_measure}.
\begin{equation}\label{eq:ig_measure}
	IG(f) = H(Y) - H(Y|f) = - \sum\limits_{y\in \mathbf{y}}P(y)\log P(y) + \sum\limits_{k=1}^{n}\sum\limits_{y\in \mathbf{y}} P(y|f^k)\log P(y|f^k)
\end{equation}

We see from $IG(f)$ that through the information gain we simply measure how well we can split based on the feature values instances belonging to the different classes (the right most part of the equation).


\clearemptydoublepage

\chapter{News in Wikipedia}\label{ch:news_wiki_lag}

In this chapter, we analyze how news media, respectively, news articles reporting about certain Wikipedia entity pages are reflected in Wikipedia. As we will show later on, news represent the second most cited source in Wikipedia. Furthermore, based on Wikipedia editing policies~\cite{wiki_policy}, news fall into the category of suggested source types as an authoritative source for citation. 


Before delving into details of the main contributions of this thesis, we first analyze the importance of news in Wikipedia. In a controlled study we see if there is a need for automated approaches on suggesting news articles for Wikipedia articles. 

The implications of this study are manifold and can be leveraged as following. For instance, automated knowledge base construction tasks can rely on news as a source or an indicator to add or update entities. First, news could be a primary source for addition of emerging entities~\cite{DBLP:conf/www/HoffartAW14}. Secondly, knowledge bases that build upon Wikipedia can periodically refresh their contents. They constantly deal with the natural trade-off between the maintenance costs of a fresh and consistent state with the loss of useful information. For newsworthy entities and events, understanding this delay in appearing in Wikipedia would suitably help knowledge bases improve their maintenance or characterize the information loss.

We study how fast Wikipedia reacts to real world events reported in news. We carry out this study on the Wikipedia revision history and the New York Times news corpus for the overlapping years between 2001 and 2007. We analyze and define \emph{lag} as the time difference between when an entity or event was reported in news and the first time it appeared in Wikipedia. Specifically, we answer the following questions:

 \begin{itemize}
 	\item What fraction of external references in entity pages are news articles?
 	\item How much does Wikipedia lag behind news articles and how does it evolve?
 	\item Which categories or classes of entities in news lead or lag Wikipedia?
 	\item How do events reported by news articles lag with the Wikipedia event pages?
 \end{itemize}

\section{Collection Alignment}\label{sec:alignment}
To carry out this study, we first align the two collections, Wikipedia and NYT corpus. The detailed descriptions of the datasets in our experimental setup are given below:

\begin{itemize}
\item \textbf{Wikipedia --} The \emph{English Wikipedia revision history}~\cite{wiki} contains the full edit history from January~2001 to December~2013. We consider all versions including versions that were marked as minor edits. 
    
\item \textbf{News -- }  The \emph{New York Times Annotated corpus}~\cite{nyt} comprises of more than 1.8 million articles from the New York Times published between 1987 and 2007. Every article has an associated publication time and we refer to this as the time of the article. Since Wikipedia was released in 2001 and the NYT corpus is valid until 2007, we consider the sub-collections from both corpora that are overlapping in time, between 2001 and 2007.

\end{itemize}

\subsection{Preliminaries and Setup}\label{subsec:setup}

\paragraph{Preliminaries.}

Before delving into detail in the lag analysis, it is necessary to introduce the entity and event notions.

\begin{entity*}	
An entity is something which has a canonical (i.e., uniquely identifiable) representation in Wikipedia. In other words, it represents a real world concept, e.g. \texttt{People, Organization, Location}. We refer to the Wikipedia page dedicated to a given entity as an Entity Page.
\end{entity*}

\begin{event*}
It is defined as a real-world event that has a Wikipedia article, e.g. \texttt{U.S Elections 2004}. The Wikipedia article dedicated to the event is referred to as the Event Page. 
\end{event*}

\begin{figure}[ht!]
\centering
\includegraphics[width=1.0\textwidth]{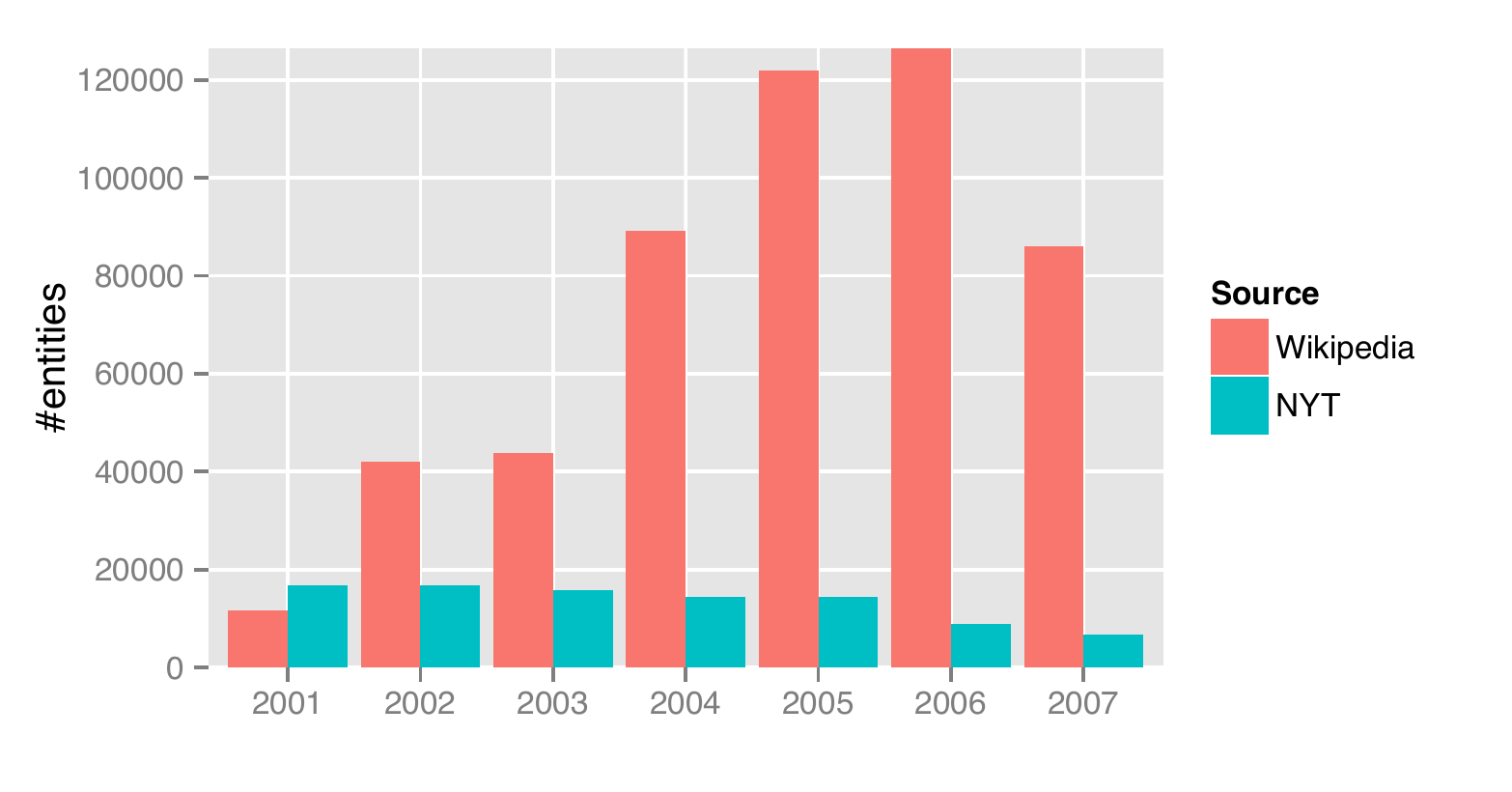}
\caption{Number of entities appearing in the corresponding years in Wikipedia, and those extracted from the entity linking process in the NYT corpus.}
\label{fig:entity_year_dist}
\end{figure}

\paragraph{Setup.}
The experimental setup is as following. We first link the free text mentions of entities from the news articles in NYT corpus to Wikipedia through \emph{entity linking} (see Section~\ref{sec:entity_linking}), for which we rely on TagMe!~\cite{DBLP:journals/software/FerraginaS12}. To maintain high accuracy of the disambiguated entities, we filter out entities with a low threshold\footnote{Entities with a disambiguation score lower than 0.3 are filtered out.}. In total, we analyze 1.8 million NYT articles, resulting in approximately 506,151 distinct entities (after filtering for the appropriate threshold). Figure~\ref{fig:entity_year_dist} shows the distribution of extracted entities for the years 2001--2007, alongside the number of entities appearing in Wikipedia at the respective years. 

The final set of entities for our experimental analysis comprises of a collection of 180,478 entities that appear only in the years 2001-2007.

\begin{figure*}[ht!]
\centering
 \includegraphics[width=0.85\textwidth]{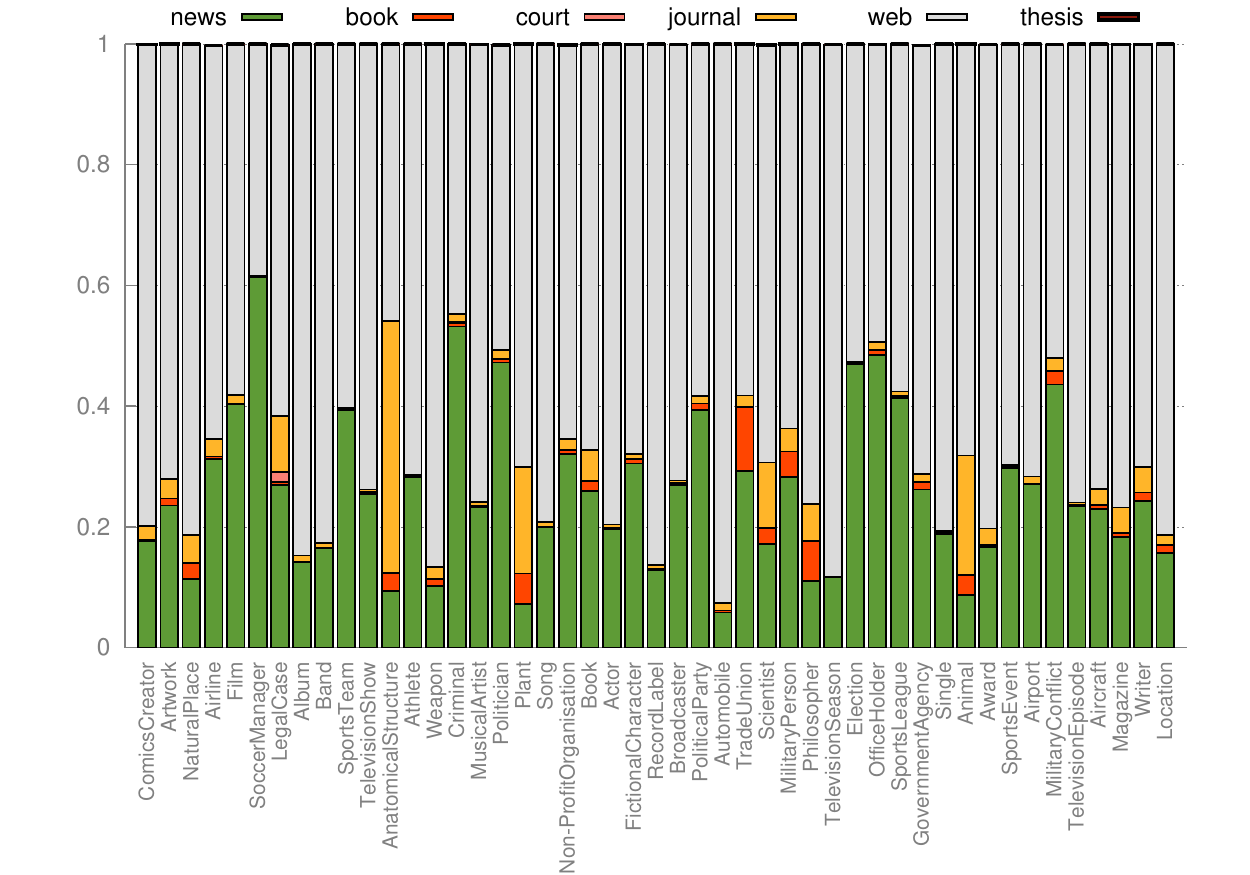}
 \caption{News Reference Density for the different entity types. The reference density of a given reference type is measured as the fraction of references of that type over all references for the entity page.}
 \label{fig:cite_density}
\end{figure*}

\section{News Reference Density in Wikipedia}\label{sec:news-dyn}

To start off we want to investigate how news impacts Wikipedia by studying such news references in \emph{entity pages}. We categorize entity pages based on their types that are associated in DBpedia\footnote{\url{http://wiki.dbpedia.org/Ontology}}, through the triples $\langle e \text{\texttt{ rdf:type }} type\rangle$. Entity pages, typically contain references to qualify the stated facts therein. These references are broadly classified into the following source types -- \texttt{web, news, book, report} and \texttt{journal}, etc., by Wikipedia\footnote{\url{http://en.wikipedia.org/wiki/Wikipedia:Citation_templates}}. We first study the distribution of news references(of type \texttt{news}) in entity pages across \emph{entity types} and define it below.

\begin{news_density*}\label{def:news_density}
News Reference Density (NRD) of an entity is the fraction of news references over all references of all types in the page. Similarly reference densities of other citation types are defined.
\end{news_density*}

\begin{figure*}[ht!]
 \includegraphics[width=1.0\textwidth]{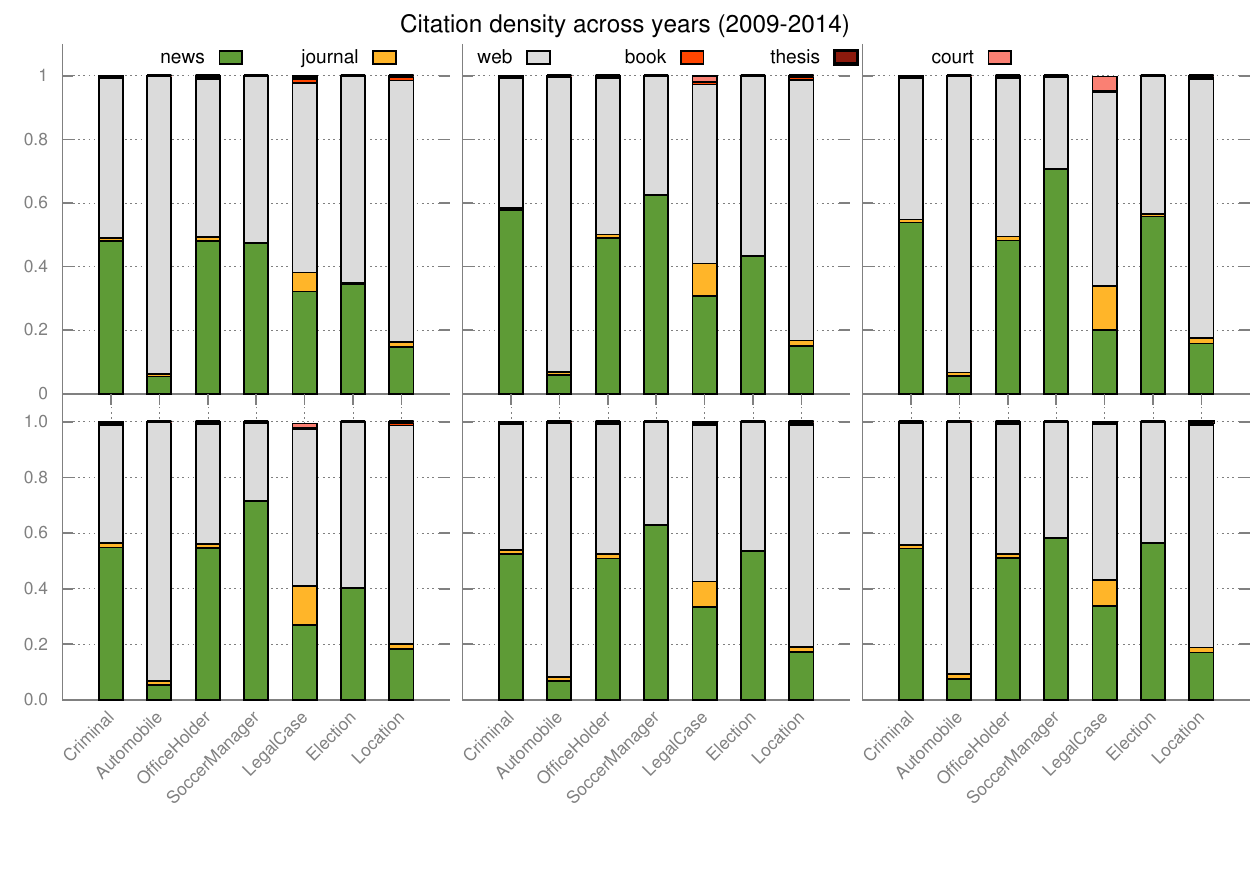}
 \caption{ Reference density for the different entity types. The plots show the reference density for years 2009-2014, in order from left to right.}
 \label{fig:domain_year_cite_density}
\end{figure*}

We observe that, as expected, most of the references are from the \texttt{web}. However, the second most dominant type of reference are news references constituting 20\% of overall references. The NRD varies across entity categories as shown in Figure~\ref{fig:cite_density}. While types \texttt{OfficeHolders} (mostly politicians) have a high news density, on the other hand \texttt{Bands} have high density for web references. The NRD in most cases is stable across years for the different entity types as shown in Figure~\ref{fig:domain_year_cite_density}. However, there are slight variations on the reference density for specialized types and the corresponding reference types, e.g. \texttt{LegalCase} and \texttt{Court} reference types.

Taking into account the organization of Wikipedia entity pages into section, we analyze the distribution of news densities across sections in an Wikipedia entities. We observe that sections in entity pages vary considerably across categories with only some of the sections being common among categories, e.g. `\emph{Early Life}' and `\emph{Career}'. When we look at the partial contribution of the sections to the page news reference density, we observe that while `\emph{Early Life and Career}' in \texttt{Politicians} have highest NRD contribution of 64\%, the section `\emph{Sports Team}' in \texttt{Athletes} has the highest contribution of 19\%. 

\section{Entity Lag}
\label{sec:entity-lag}

\begin{figure}[ht!]
\centering
  \includegraphics[width=0.7\textwidth]{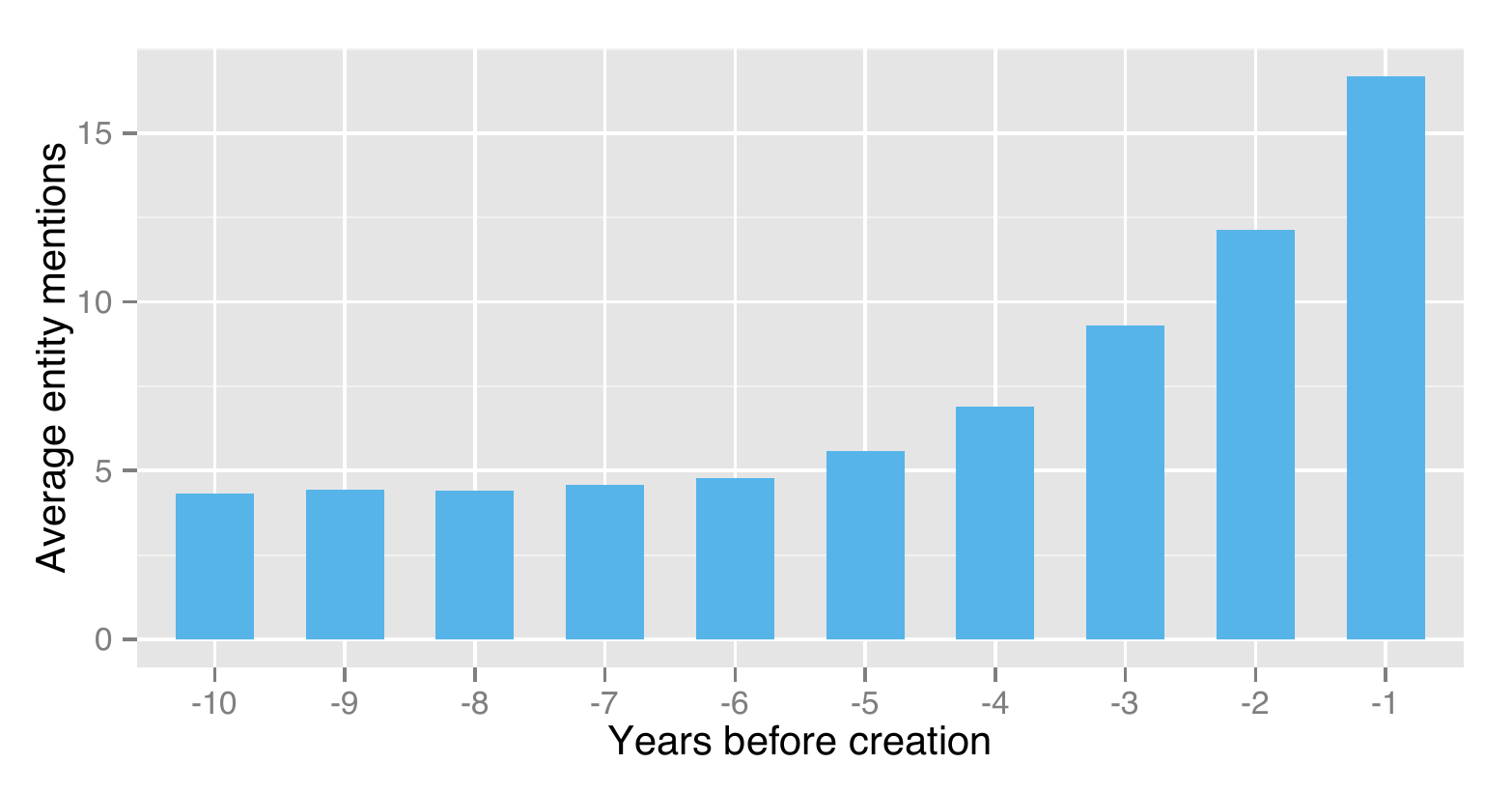}
  \caption{Entity mention counts in news articles before creation of Wikipedia entity page. Mention counts of entities peak a year before it is created in Wikipedia.}
  \label{fig:cumm_dist}
\end{figure}

\begin{figure*}[ht!]
\centering
  \includegraphics[width=1.0\textwidth]{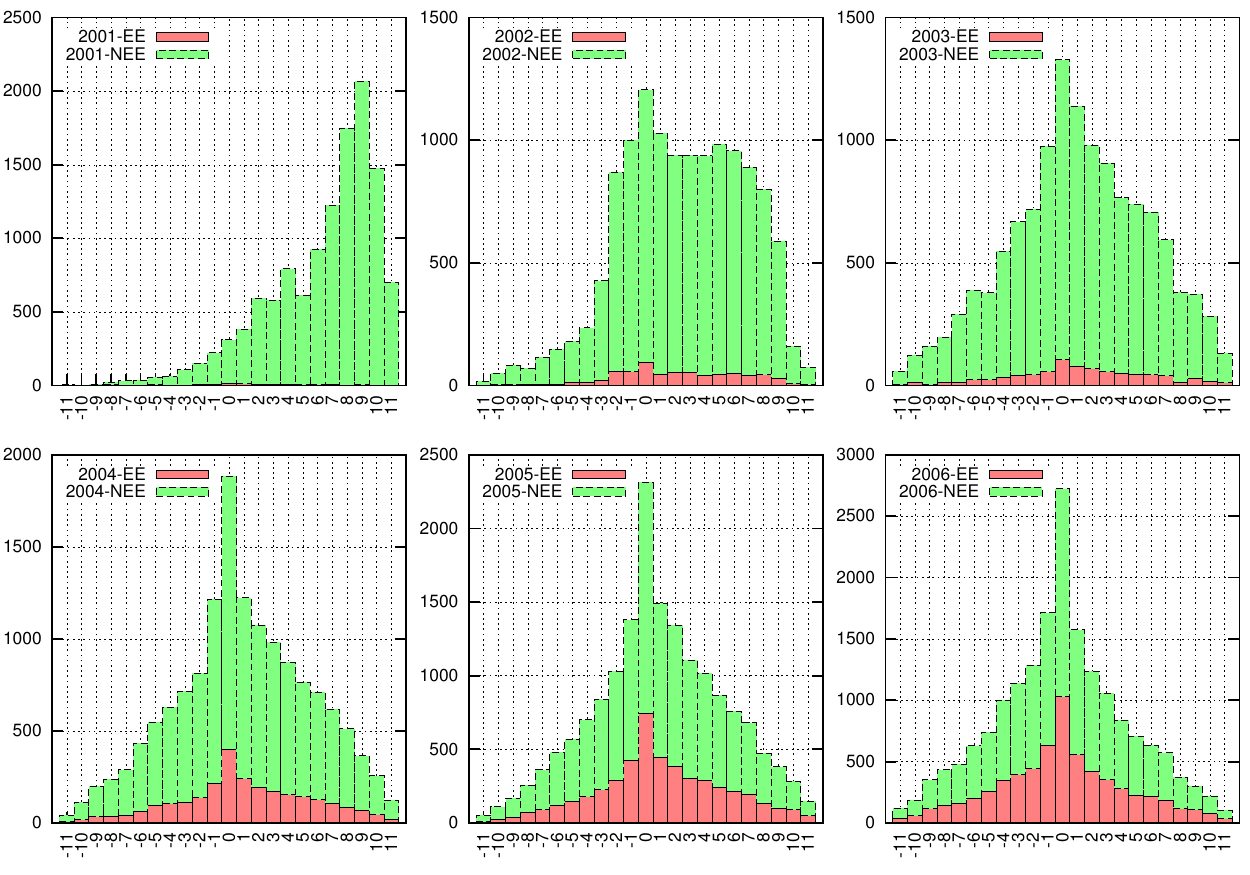}
  \caption{Entity lag in months. The emergent entities are shown in red, they are determined by filtering all entities from the subset of NYT that appear in earlier years before 2001. The y-axis is normalized using the \emph{sum} of entities having medium lag for the emerging and non-emerging entities, respectively. }
  \label{fig:hist-med-lag-dist-months}
\end{figure*}

From the aligned collection we analyze the behavior with which entities extracted from NYT are added in Wikipedia.

\paragraph{Entity Lag.}
This can be attributed to two factors: \emph{inherent popularity of the entity}, and \emph{evolution of authorship} of entity pages in Wikipedia. One explanation is that entities appearing in authoritative news sources like NYT reflect their popularity. Figure~\ref{fig:cumm_dist} shows the average entity mention distribution (in NYT) across years before the entity is first created in Wikipedia. This follows the assumption that an increase of entity mentions in news sources will eventually result in the creation of an entity in Wikipedia. Figure~\ref{fig:cumm_dist} shows that shortly before the entity creation in Wikipedia, the entity is mentioned most in news. The second factor, is that Wikipedia's authorship has increased with an ever growing number of editors, hence establishing itself as a independent source of information~\cite{keegan_hot_2011}, thus entities can be created from what is deemed as important by the editors in Wikipedia.

We measure the \emph{time span} between the entity mention and its creation time in Wikipedia, and define the \emph{entity lag} below. 

\begin{entity_lag*} 
The delay of the first appearance of an entity page relative to the first appearance of its mention in a news article is called \emph{entity lag} or simply \emph{lag} $lag(e_i)$.
$\mathbf{lag(e_i) = t_w(e_i) - t_n(e_i)}$, where $t_w(e_i)$ is the time of the first version of entity $e_i$ was authored and $t_n(e_i)$ is the publication time of its first mention in news.
\end{entity_lag*}

First, we analyze how the creation of entities in Wikipedia lag their mentions in news. We denote the entities with an absolute lag of less than a month as \emph{low lag entities}, less than a year as \emph{medium-lag entities} and the rest with more than a year as \emph{high-lag entities}. Figure~\ref{fig:hist-med-lag-dist-months} shows the lag distribution in months. 

We see that in the first year of Wikipedia the average lag was high with a majority of entities in Wikipedia lagging behind news. However, quite distinctly, the lag re-distributes towards a means of zero in the course of time into a Gaussian or normal distribution. We also see that the absolute number of entities with a lag of zeros go up, and the standard deviation reduces. The lag distribution through the years shifts to a normal distribution, with most of the entities centered around the mean, which in our case is zero. 

Since Wikipedia only started after 2001, we also consider the entities which were \emph{emergent} in news after 2001 (denoted by the red histogram). 

\begin{emerging_entities*}
An entity is considered as an \emph{emergent entity} (EE) if its first mention in NYT is after the time when Wikipedia was released, i.e., January 2001.
\end{emerging_entities*}

Emergent entities have a similar distribution like the existing entities. Since news articles are rich in political news and their coverage, we observe that emergent political topics and entities show low lag. An example is \texttt{Freedom Fries} which came into prominence in 2003 as a political euphemism for the actual French fries. On the other hand works of fiction like \texttt{The lost City} typically exhibit high lag. Similar to the non-emergent entities the lag distribution for \emph{emergent entities} is normal.

Based on this distributions we can provide rough estimates of the fraction of \emph{`newsworthy'} entities, which could be missed given a maintenance period. Services that periodically update their entity repositories would lose around half of the entities if their update periods is greater than a month than if they update daily. 


\begin{figure}[h]
	\centering
	\begin{subfigure}[t]{0.3\textwidth}
        \centering
        \includegraphics[width=1\textwidth]{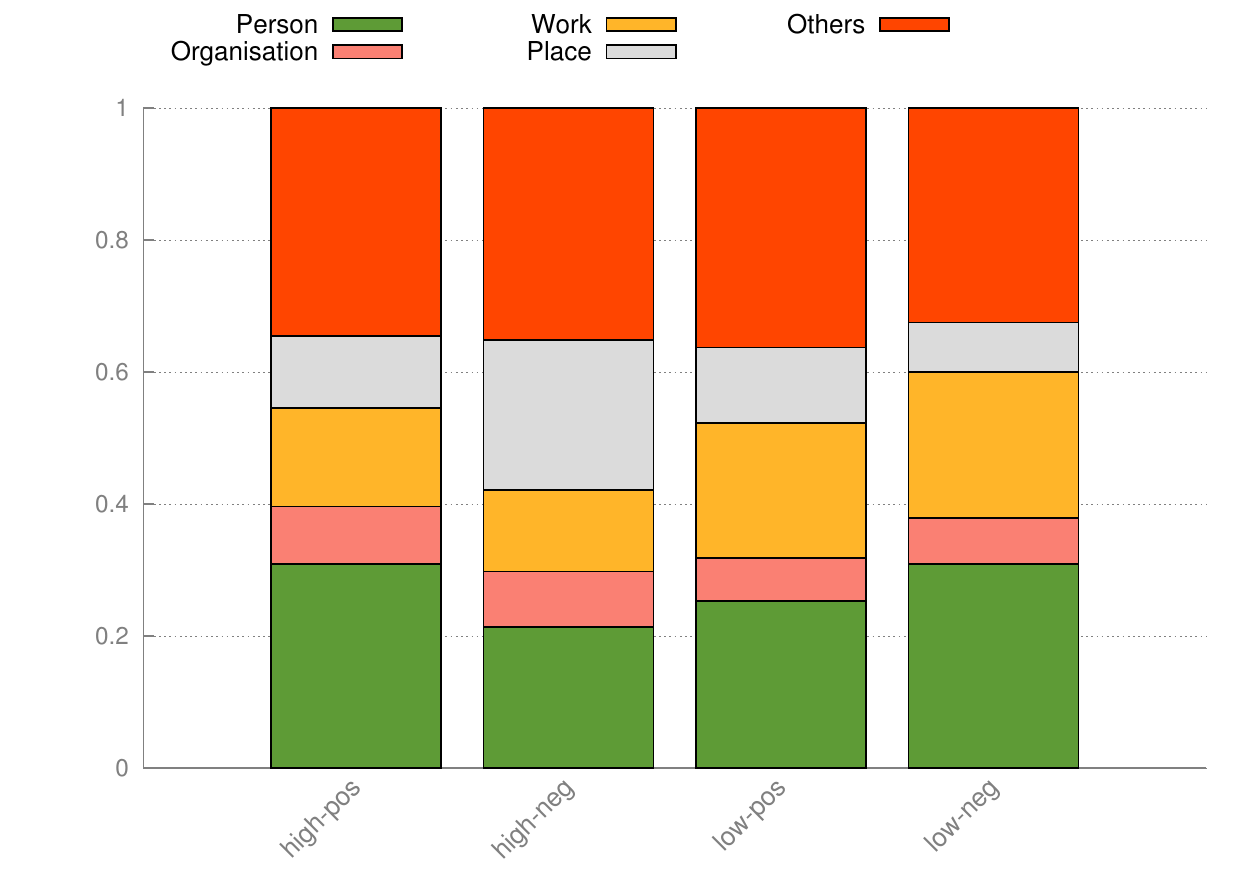}
        \caption{Overall}
        \label{fig:stacked_all}
    \end{subfigure}
    ~
	\begin{subfigure}[t]{0.3\textwidth}
        \centering
        \includegraphics[width=1\textwidth]{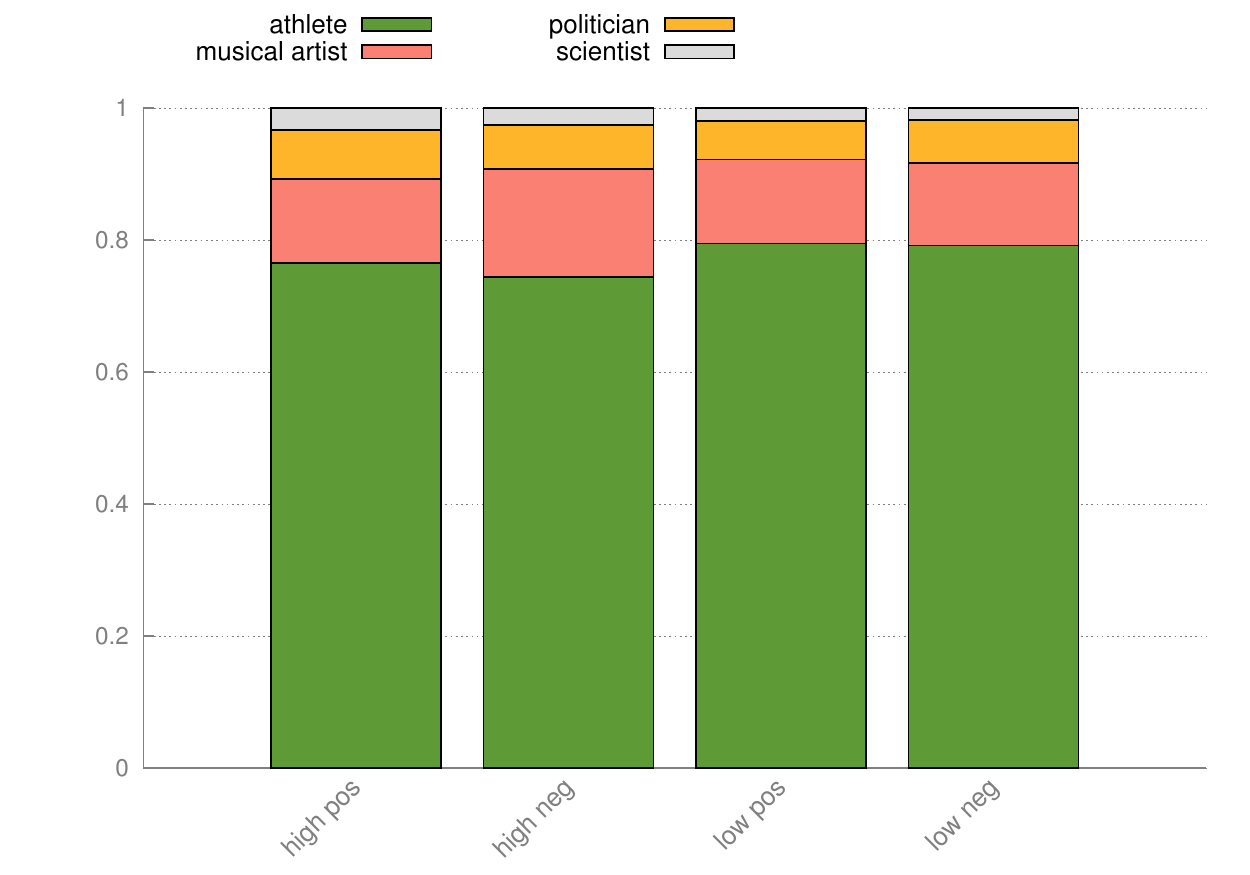}
        \caption{Person}
        \label{fig:stacked_person}
    \end{subfigure}
    ~
   	\begin{subfigure}[t]{0.3\textwidth}
        \centering
        \includegraphics[width=1\textwidth]{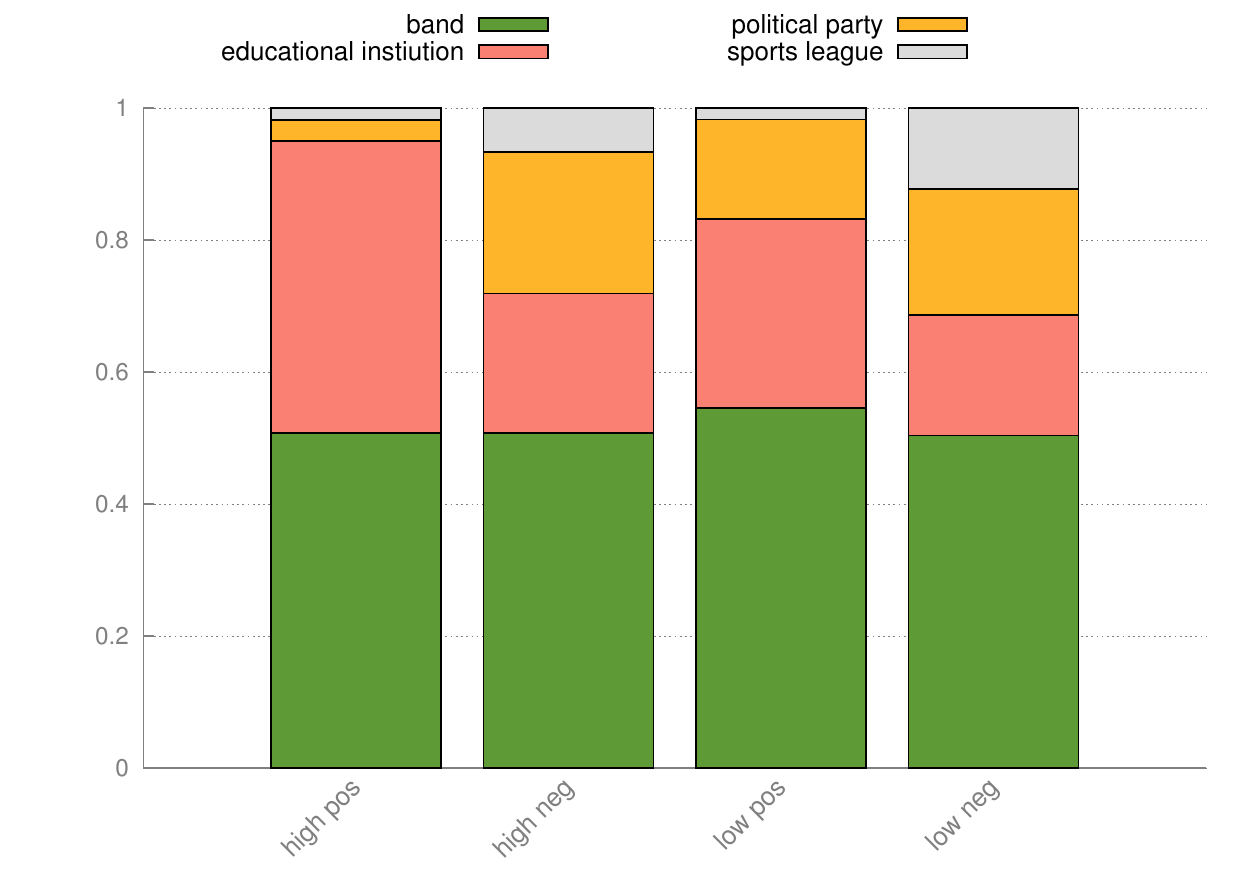}
        \caption{Organization}
        \label{fig:stacked_organisation}
    \end{subfigure}
    \caption{Lag distribution of different types. The y-axis values are normalized by the sum of the overall entities falling into the different \emph{lag classes}.}
    \label{fig:entity-types}
\end{figure}

\subsection{Lag for Entity Types}

To characterize which entity types show different lag behavior -- positive or negative, low or high -- we group entities based on their types. For example, \emph{Barack Obama} \texttt{isA US President isA  Politician isA Person}.

We take a coarse grained representation of entity types, that is \emph{Person, Work, Organization, Places, Other} and are presented in Figure~\ref{fig:stacked_all}. High-positive refers to high lag (Wikipedia lags news) whereas high-negative implies a high lead (Wikipedia leads news). It is natural to see that \emph{Places} have the highest negative lag since entity pages for many geographic locations were introduced during the early days of Wikipedia which we refer to as its \emph{bootstrapping period}.

We see that Wikipedia has a high positive lag for \emph{Persons} (almost 37\%) in comparison to other types. This means that most of the emergent entities are people rather than other entity types. A closer look into the four major subcategories of \emph{People} in Figure~\ref{fig:stacked_person} reveals that \emph{musicians} tend to be mentioned in Wikipedia earlier than news and we confirm that most of them, like the locations, were also created during the bootstrapping period. In the case of \emph{Organizations} in Figure~\ref{fig:stacked_organisation}, we make two observations. First, all educational institutions have a high lag and secondly political parties either have a high lead or a small lag. This suggests that political parties are quite popular entities in Wikipedia while educational institutes are not. 

The entity class \emph{Work} encompasses all types of \emph{books, musical composition and movies}. In general \emph{Work} is reported under low lag (around 21\%-22\%) as compared to its higher lag instances which is around around 12\%-14\%. In sum, artistic works and locations get reflected in Wikipedia sooner than other types while Wikipedia lags news for emerging personalities. The overall distribution of entity lag is shown in Table~\ref{tbl:overal_lag}.

\begin{table}[ht!]
\centering
\begin{tabular}{l l l}
\toprule
\texttt{lag type} & \texttt{negative (lag)} & \texttt{positive (lead)}\\
\midrule
high & 57.1\% & 8\%\\
medium & 22.2\% & 11\%\\
low & 0.2\% & 1.1\%\\
\bottomrule
\end{tabular}
\caption{Absolute entity lag distributions for all lag types. The numbers are aggregated over the years 2001-2006.}
\label{tbl:overal_lag}
\end{table}

\section{Event Lag}
\label{sec:event-lag}

We define \emph{event lag}, similarly to \emph{entity lag}, as the publication time difference between the first news article which reports the event and the Wikipedia event page. Events reported in the news can be as a reaction to an event in the past, or a build up to an upcoming event. We do not make a difference in both these cases and treat the first news article reporting the event as the inception of the event.

\begin{figure}[h!]
\centering
\includegraphics[width=0.7\columnwidth]{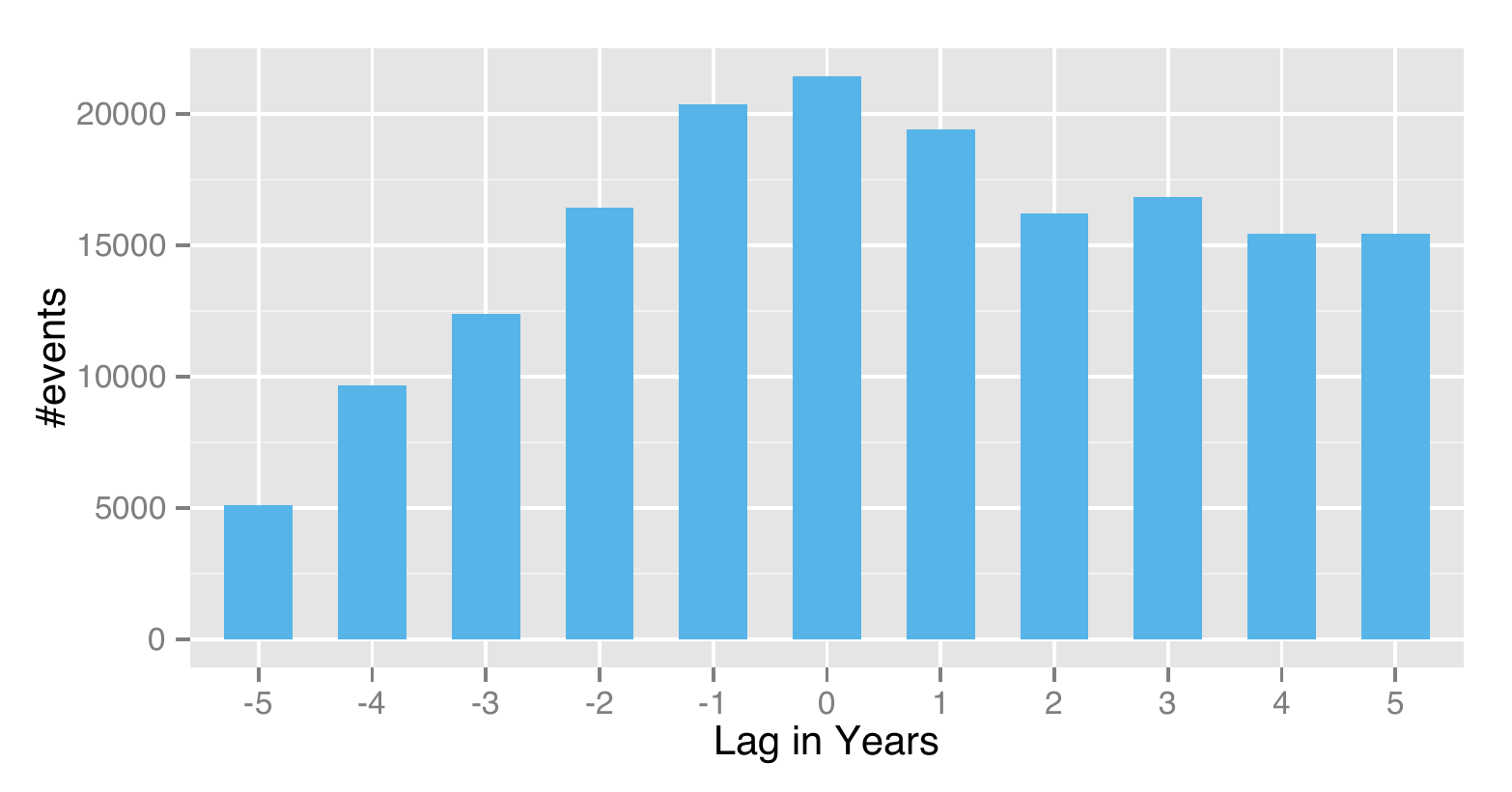}
\caption{Event news reference lag (in years) in Wikipedia. Most of Wikipedia events fall into \emph{low-lag} class, showing high dynamics of reporting real news events in Wikipedia.}
\label{fig:event_lag}
\end{figure}

\subsection{Emerging Entities in Event Pages}\label{sec:emerging-entities-events}

Finally, we study how events influence the creation of entities in Wikipedia. For this experiment we consider all events in DBpedia with their publication time (resource of type \texttt{dbpedia-owl:Event}). Unlike the previous experiments we do not rely on the NYT corpora and hence can consider the entire Wikipedia revision history.

The notion of the publication time corresponds to the first time the event page was introduced in Wikipedia. Next, we extract the explicitly linked entities in the event page and compare the publication times of the mentioned entities and the event publication time. To this effect, we make a simplistic assumption about the mentioned entities in the event page: \emph{entities created after the event page are created because of this event}.

\begin{ee_ep*}
Emerging entity density of an event page is the fraction of entities created after the event page. We refer to them  as emerging entities (note that this is different from the emergent entities in the previous section).
\end{ee_ep*}

For example, for the event ``\emph{Charlie Hebdo Shootings}''\footnote{\url{http://en.wikipedia.org/wiki/Charlie_Hebdo_shooting}}, created on 7th January, 2015, the mentioned entities therein, ``\emph{Corinne Rey}'' or ``\emph{Coco}''\footnote{\url{http://en.wikipedia.org/wiki/Coco_(cartoonist)}} were created five days later on 12th January.

\begin{figure}[ht!]
	\begin{subfigure}[t]{0.45\textwidth}
        \centering
        \includegraphics[width=1\textwidth]{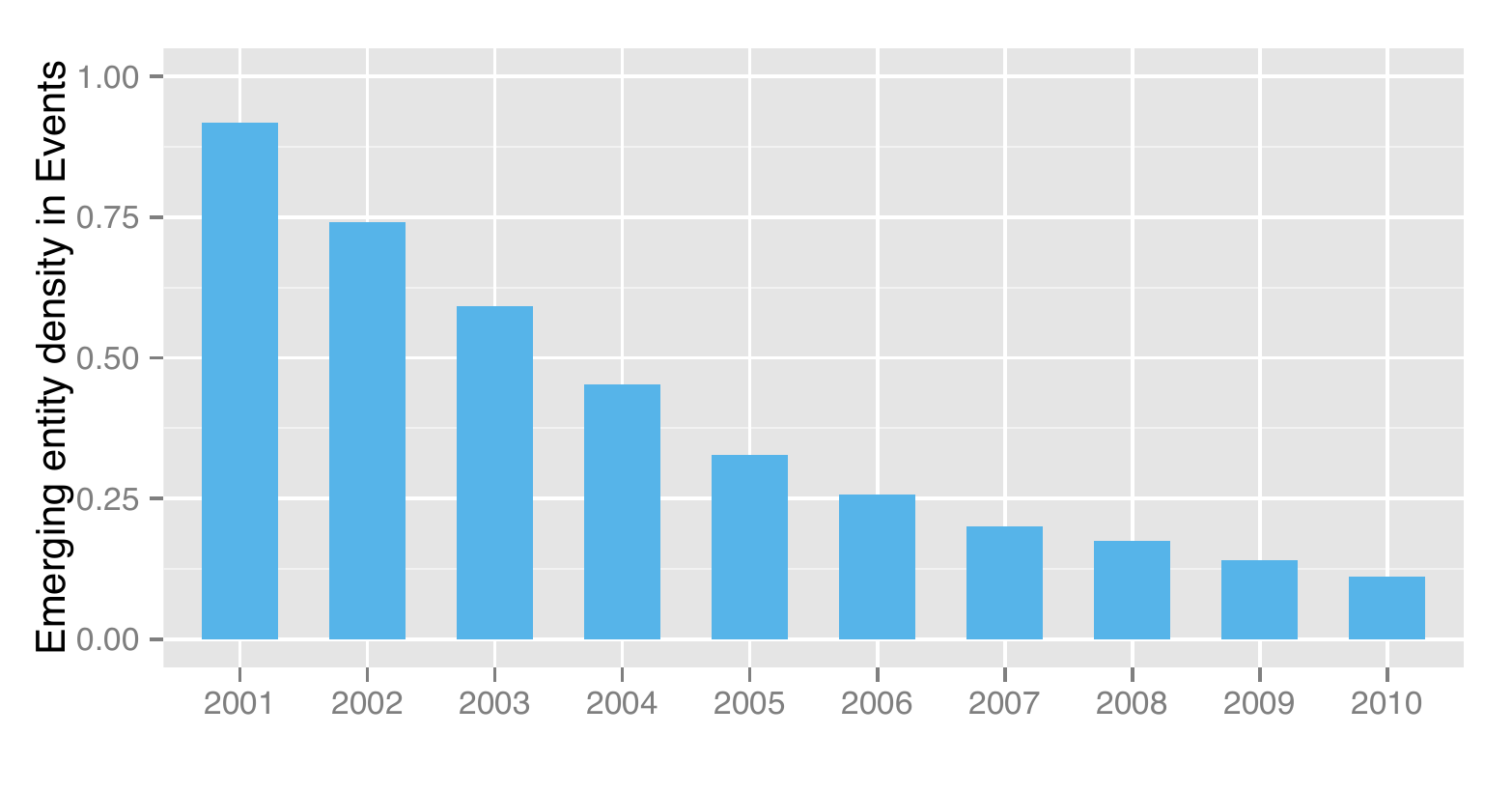}
        \caption{Emerging Entity Density}
        \label{fig:EED-yearly}
    \end{subfigure}
	~
 	\begin{subfigure}[t]{0.45\textwidth}
        \centering
        \includegraphics[width=1\textwidth,height=100pt]{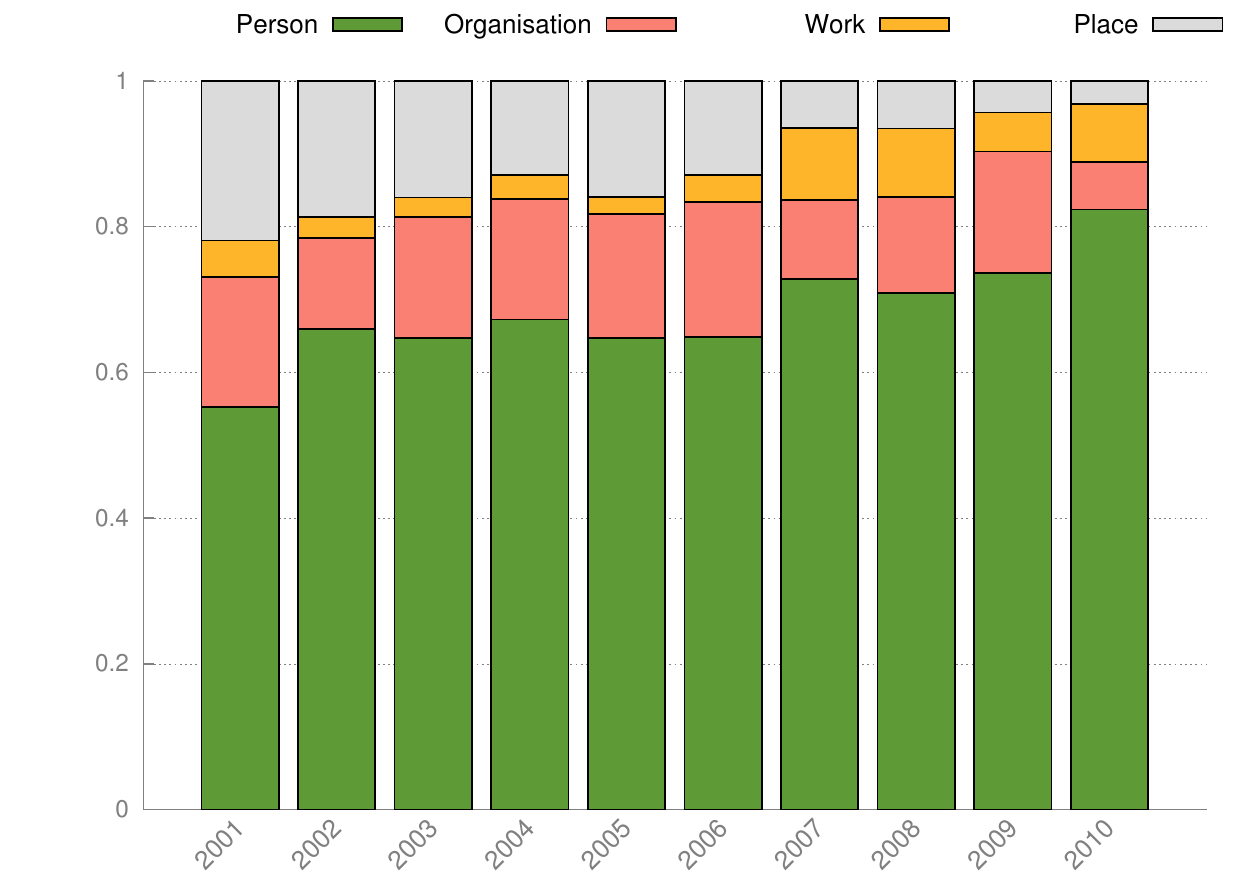}
        \caption{Emerging entity categories}
        \label{fig:stacked_organisation}
    \end{subfigure}
    \caption{Emerging entity density in Wikipedia event pages.}
    \label{fig:event-entity-density}
\end{figure}

The emerging entity density (EED) evolution from 2001-2010 is presented in Figure~\ref{fig:EED-yearly} where the y-axis represents the average emerging entity density of event pages in a given year. We have a total of 14,604 events with 179,981 entities with the exception of events from the last few years owing to the lack of event data in DBPedia for this period. We see that in the early years the EED of event pages was very high, sometimes above 80\%, meaning most of the entities mentioned in the event pages were emerging. Understandably, this declines every year resembling the phenomena of diminishing returns. However, we still see a high percentage of emerging entities in the recent event pages which point to the fact that event pages are great repositories of upcoming and emerging entities missing in the knowledge bases. We also observe that the curve, although decreasing, tends to stabilize in the recent years around 13\%. Finally, we look at the categories of emerging entities in Figure~\ref{fig:stacked_person} to find that people comprise the majority of the emergent entities consistently over the years. On the other hand, organizations were emergent between 2001-2005 but their EED contribution to event pages has been decreasing from 2006 onwards.

\section{Conclusions and Implications}\label{sec:conclusion}

The implications of this study are manifold. First, it shows that news collections are an important resource for mining emerging entities. The normal distribution of the entity lag shows that almost 50\% of the entities before occurring in Wikipedia are already mentioned in news.  Our experiments on news reference density show that a high proportion of facts about entities are qualified with a news reference. 

Secondly, entity and event repositories relying on Wikipedia can quantify the degree of loss or re-calibrate their update frequency based on the lag distribution. It is also possible to optimize emergent entity coverage by focusing on event pages. Interestingly, the lag for events is far lower than entity pages. 

Thirdly, event pages are containers of emergent entities with around 12\% of the entities linked to an entity being emergent.

Finally, this study provides useful insights and highlights the need for automated approaches to enrich Wikipedia entities and events with news citations. This is one of the key motivations for the contributions in this thesis in the upcoming chapters.


\clearemptydoublepage
\chapter{Related Work}\label{ch:related_work}

In this chapter, we review related literature, which focuses towards problems that we highlighted in Chapter~\ref{introduction} and compares the contributions of this thesis. 

In more details, in Section~\ref{sec:editor_dynamics} we review related work which analyzes the dynamics of Wikipedia editors, and the community structures within. This is related to our study in Chapter~\ref{ch:news_wiki_lag}, where we measure the entity and event lag in news and Wikipedia. Next, in Section~\ref{sec:page_gen}, we provide an overview of existing work and state of the art approaches on generating automatically Wikipedia articles, thus, related to our contributions in Chapter~\ref{ch1:citation_recommendation} and \ref{ch3:news_suggestion}. Similarly, in Section~\ref{sec:kba} and \ref{sec:salience} we list the shortcomings of existing knowledge base acceleration approaches and works on determining entity salience in news articles, as such these represent highly related works relevant for Chapter~\ref{ch1:citation_recommendation} and \ref{ch3:news_suggestion}. Finally, in Section~\ref{sec:cspan}, we present an overview of existing work in determining the citation span in scientific articles, which is related to our contribution in Chapter~\ref{ch5:cite_span}.

\section{Wikipedia Editor Dynamics}\label{sec:editor_dynamics}

One of the core parts of Wikipedia are its editors. Therefore, analyzing the behavior of editors and how they collaborate has many implications. First, since Wikipedia is collaboratively created, and thus the guidelines on high-quality edits and other policies are agreed collectively. Second, the analysis on the responsiveness of editors to real-world events and correspondingly how they diffuse such information into the Wikipedia pages is highly important. Our study in Chapter~\ref{ch:news_wiki_lag} answers one of the questions on how fast are Wikipedia editors to add information reported in news about Wikipedia articles. Below we review work that focuses solely on the Wikipedia editors and the structure of editors therein.

Kittur et al.~\cite{kittur_power_2007} analyses the structure of Wikipedia editors on how they collaborate. They further classify the \emph{collaborators} into five different classes based on the number of revisions. Next, they measure the population growth of the collaborators falling into the five different classes. They find that the shift on how content is provided mostly by collaborators with lower number of edits is due to the increased fraction of such users in the Wikipedia community structure. This, however, does not correlate with any decline of the content provided by collaborators with high number of edits, hence, is accounted to the higher fraction of low edit users. In contrast to the work from Kittur et al., we have a different focus in our analysis, namely that of entity and event lag in Wikipedia, without any distinction of the Wikipedia community structure. In \cite{Suh:2009:SNS:1641309.1641322} the authors analyze several aspects of Wikipedia's editors. They conclude that the number of edits is decreasing. Another slightly related work~\cite{Bar-Ilan:2014:TYW:2615569.2615643} analyzes the number of research papers about Wikipedia, here too they conclude that the number has been decreasing, however, papers that use Wikpedia's data has seen an increase.

Keegan et al.~\cite{keegan_hot_2011,keegan_editors_2012,hu_measuring_2007} focus on the dynamics of Wikipedia's coverage of real world entities. In~\cite{keegan_hot_2011}, the authors consider emerging events like the T\={o}hoku catastrophe\footnote{\url{http://en.wikipedia.org/wiki/2011_T\%C5\%8Dhoku_earthquake_and_tsunami}}. In the case of such high dynamic events, it is found out that for localized Wikipedias (e.g. Japanese), the corresponding event appears only six minutes after the event, whereas in the English Wikipedia, it appears in less than an hour. Furthermore, they analyze the co-authorship, concluding that within Wikipedia there are sub-communities that edit articles of the same topic. 

As a continuation of their work, in~\cite{keegan_editors_2012} the social network structure of Wikipedia collaborators is analyzed. The analysis is based on four main hypotheses that are based on two main set of attributes, article and editor attributes, respectively. The first hypothesis validates the fact that for breaking news articles attract more editors. The second hypothesis validates the co-authorship of articles in Wikipedia from collaborators that are categorized into three main classes: \emph{Experienced, Apprentice, Non-Expert}. Significant collaborations between the three classes of collaborators is found only on \emph{contemporary} articles (articles are divided into \emph{breaking, contemporary, historical}) between \emph{apprentice} and \emph{experienced} collaborators. The third hypothesis, analyzes the editor attributes and implies that experienced editors will edit more articles than others. The third hypothesis leads to the fourth and last hypothesis. It analyzes the fact that experienced editors are more likely to contribute to similar types of articles rather than to dissimilar. Strong correlation is found for editors belonging to the \emph{apprentice} class and for most of the article types. 

The works~\cite{keegan_hot_2011,keegan_editors_2012,hu_measuring_2007,kittur_power_2007} are related only to our study shown in Chapter~\ref{ch:news_wiki_lag}. Contrary to our analysis the work by Keegan et al. has as a main focus modeling the network structure of editors and how this reflects on the dynamics of Wikipedia and contemporary and emergent entities and events. On the other hand, in our analysis we focus on larger real world news corpus which inherently represents emerging entities and events. In addition, we also distinguished the lag of different entity types, and as a last diverging point  we analyzed how entities are co-created and their impact on the entity lag.

\section{Wikipedia Page Generation}\label{sec:page_gen}

\paragraph{Citation Sources.} Ford et al.~\cite{DBLP:conf/wikis/FordSMM13} analyze the citation behavior of Wikipedia editors with respect to their adherence to the citation guidelines. They investigate what types of sources are most often cited, i.e. \emph{primary, secondary} and \emph{tertiary} as defined in Wikipedia~\cite{wiki_policy}. Similar as to our study in Chapter~\ref{ch:news_wiki_lag}, they conclude that news are one of the top cited source in the \emph{secondary} type, while they see a growing trend of \emph{primary} sources due to their persistence on the web, contrary to the Wikipedia policies of preferring secondary sources. 

Luyt and Tan~\cite{DBLP:journals/jasis/LuytT10}, for a subset of Wikipedia entity pages from the domain of \emph{History}, they  analyze how citations are biased towards a specific group of sources. \cite{DBLP:conf/wikis/FordSMM13,DBLP:journals/jasis/LuytT10} emphasize the importance of citations in Wikipedia as a means to ensure the quality of entity pages.

Hunag et al.~\cite{DBLP:conf/jcdl/HuangWMG14,DBLP:conf/cikm/HuangKCMGR12} have considered the problem recommending citations in scientific publications. Despite the fact that this is significantly different from the approaches we propose in Chapter~\ref{ch1:citation_recommendation}, it reveals the importance on the \emph{verifiability} of statement be it in scholarly articles or Wikipedia.

The works regarding citation sources reveal that citation policies in Wikipedia, specifically, the use of third-party sources is important to ensure the high quality of Wikipedia articles, and at the same time, finding citations is an important problem that we encounter in other domains such as scientific literature. Therefore, through our contributions in this thesis, are highly important in other domains as well, apart from our work that focuses only in Wikipedia articles.

\paragraph{Wikipedia Enrichment.}
Sauper and Barzilay \cite{DBLP:conf/acl/SauperB09} propose an approach for automatically generating entire Wikipedia entity pages for specific entity types. The approach is trained on already-populated entity pages of a given type (e.g. `\emph{Diseases}') by learning templates about the entity page structure. For example, an entity of type  \texttt{Disease}, in majority of the cases they contain a \emph{``Treatment''} section. The approach works as following. For a new entity page, first, they extract documents via Web search using the \emph{entity title} and the \emph{section title} as a query, for example `\emph{Lung Cancer}'+`\emph{Treatment}'. Next, the task is to  identify the best paragraphs extracted from the resulting documents from the Web search. They rank the paragraphs via an optimized supervised \emph{perceptron model} for finding the most representative paragraph that is the least similar to paragraphs in other sections. That is, in case an entity has more than one section in its section template, then the paragraph ranking is optimized across all sections such that there is minimal content redundancy from the suggested paragraphs across sections. The top-1 paragraph for each section is added into the entity page.

A similar line of work was proposed by Taneva and Weikum~\cite{Taneva:2013:GEM:2505515.2505715}. They propose an approach that constructs short summaries for the long tail entities in Wikipedia. The summaries are called `\emph{gems}' and the size of a `\emph{gem}' can be user defined. The size of a summary is measured in terms of characters or words. They focus on generating summaries that are novel and diverse. However, they do not consider any structure of entities, which is present in Wikipedia.

Contrary to \cite{DBLP:conf/acl/SauperB09} and \cite{Taneva:2013:GEM:2505515.2505715}, in this thesis, specifically in Chapter~\ref{ch3:news_suggestion}, we focus on suggesting entire documents to Wikipedia entity pages, respectively to the appropriate sections. Furthermore, following the Wikipedia editing policies, we focus on news collections as authoritative sources and recommended third-party sources for citation in Wikipedia. The news articles that we suggest for entity pages is where the entity is a \emph{salient} concept in the news article, and that the information contained within the article is novel and is \emph{authoritative} for the entity of interest. The notion of \emph{authority} in this case we explain it subject to the entity of interest. That is, a news article provides authoritative information for an entity of interest if it co-occurs (within the news article content) with entities of higher authority (as measured through prior probability or any centrality measures). This has the advantage as it allows to adjust accordingly on what passes the \emph{suggestion threshold} for an entity page. 

The notion of relevance in \cite{DBLP:conf/acl/SauperB09} is computed implicitly through the ranking of documents from the Web search. Furthermore, they do not deal with the novelty of a suggested piece of information is proposed to \emph{stub} Wikipedia entity pages.

Finally, through our contribution in Chapter~\ref{ch3:news_suggestion} and \ref{ch1:citation_recommendation}, we address several limitations in this line of work. First of all, the approache \cite{DBLP:conf/acl/SauperB09} has the problem of reproducibility and maintainability. This is due to the fact that they rely on Web search which is an uncontrolled variable in this case, and the corresponding ranking is subject to proprietary methodological issues. Secondly, we aim at updating the already existing Wikipedia entity pages, while the focus of related work is on populating \emph{stub} Wikipedia pages. Finally, both the approaches in~\cite{DBLP:conf/acl/SauperB09} and \cite{Taneva:2013:GEM:2505515.2505715}  (finding paragraphs and summarization) could be used to process the news articles we suggest to Wikipedia pages. Our concentration on news is also novel.

\paragraph{Wikipedia Quality Measures} Anderka et al.~\cite{DBLP:conf/sigir/AnderkaSL12} propose a supervised approach to predict quality flaws in Wikipedia pages. A quality flaw in Wikipedia is usually annotated with specific \emph{cleanup} tags. They train a model to predict quality flaws, where among the top--10 quality flaws they identify \emph{unreferenced, refimprove, primary sources} as some of the most serious flaws. The work in \cite{DBLP:conf/sigir/AnderkaSL12} is complementary to the contributions of this thesis in Chapter~\ref{ch1:citation_recommendation} and \ref{ch3:news_suggestion}. Majority of the quality flaws are due to content in Wikipedia pages not following the editing policies, and one of the major editing policies deals with citations and the source of such citations. Hence, the approaches we propose can address many of the quality flaws. 

As part of future work, we foresee combining our approaches together with the one proposed by Anderka et al.~\cite{DBLP:conf/sigir/AnderkaSL12}, and quantify the improvements we gain in terms of quality through our approaches.

\section{Knowledge Base Acceleration}\label{sec:kba}

\paragraph{Cumulative Citation Recommendation (CCR).} TREC introduced the CCR track in the Knowledge base acceleration track in 2012. For a stream of news and social media content and a target entity from a knowledge base (Wikipedia), the goal of the task is to generate a score for each document based on how pertinent it is to the input entity. Balog et al.~\cite{DBLP:conf/riao/BalogRTN13,DBLP:conf/sigir/BalogR13} propose approaches that find entity mentions in the document collection and rank them according to how central the entity is in the respective documents. This however is a filtering task for documents towards checking if they are relevant for a pre-defined set of entities. In contrast, in our task we aim at finding news citations as evidence for Wikipedia statements.

In the other spectrum of KBA tasks is that of enriching structured knowledge bases.  For a specific information extraction template and given corpus, the task is to analyze the corpus and find worthwhile mentions of an entity or snippets that match the templates. West et al.~\cite{DBLP:conf/www/WestGMSGL14} consider the problem of knowledge base completion, through question answering and complete missing facts in Freebase based on templates, i.e. \emph{Frank\_Zappa} \texttt{bornIn} \emph{Baltymore, Maryland}. In contrast, in Chapter~\ref{ch3:news_suggestion}, we do not extract facts for pre-defined templates but rather suggest news articles based on their relevance to an entity. In cases of long-tail entities, we can suggest to add a novel section  through our abstraction and generation of section templates at entity class level.

\section{Entity Salience and Filtering}\label{sec:salience}

An important aspect in this thesis, is to determine salient entities on a given corpus. In our thesis, we focus solely on news articles, which have a distinct language style, referred to as the \emph{pyramid style}, where the important information is mentioned first. In this section, we review some of the most prominent works in this field, and compare to our proposed entity salience models in Chapter~\ref{ch3:news_suggestion}.

Determining which entities are prominent or salient in a given text has a long history in NLP, sparked by the linguistic theory of Centering~\cite{walker1998centering}. Salience has been used in pronoun and co-reference resolution \cite{DBLP:conf/acl/Ng10}, or to predict which entities will be included in an abstract of an article \cite{DBLP:conf/eacl/DunietzG14}. Frequent features to measure salience include the frequency of an entity in a document, positioning of an entity, grammatical function or internal entity structure (POS tags, head nouns etc.). These approaches are not currently aimed at knowledge base generation or Wikipedia coverage extension but we postulate that an entity's salience in a news article is a prerequisite to the news article being relevant enough to be included in an entity page. Therefore, in Chapter~\ref{ch3:news_suggestion}, as part of our entity salience model, we use the salience features in~\cite{DBLP:conf/eacl/DunietzG14}.  However, these features are document-internal --- we will show that they are not sufficient to predict news inclusion into an entity page and add features of entity authority, news authority and novelty that measure the relations between several entities, between entity and news article as well as between several competing news articles.

As we will show in Chapter~\ref{ch3:news_suggestion}, our entity salience models which relies on the language style of news significantly outperform existing approaches~\cite{DBLP:conf/eacl/DunietzG14}.

\section{Citation Span}\label{sec:cspan}

The related literature in this section focuses solely on the contributions in Chapter~\ref{ch5:cite_span}. The approaches we propose in determining the span of a citation in Wikipedia articles are novel and as such do not have directly related works in the same domain. However, there exist previous work that consider this problem in the scientific domain of, on determining the scope of a reference in a scientific article. As we will discuss in more details below and later show in our experimental evaluation, there are major differences on how citations are used in Wikipedia and in scientific publications.

\paragraph{Scientific Text.} One of the first attempts to determine the citation span in text was carried in the context of document retrieval~\cite{DBLP:journals/ipm/OConnor82}. Here we encounter the first mention of \emph{citing statements}, which refers to the sentences in a given document that are relevant or describe a reference to another document or scientific article. In this work, the \emph{citing statements} from a document were used as an index to retrieve the \emph{cited} document. The citing statements are extracted based on heuristics starting from the citing sentence and are expanded with sentences in a window of +/-2 sentences, depending if they contain cue words like \emph{`this', `these',$\ldots$ `above-mentioned'}. We consider the approach in~\cite{DBLP:journals/ipm/OConnor82} as a baseline.

Kaplan et al.~\cite{DBLP:journals/jip/KaplanTT16} proposed the task of determining the \emph{citation block} based on a set of \emph{textual coherence} features (e.g. grammatical or lexical coherence). The citation block \emph{starts} from the citing sentence, with succeeding sentences classified if they belong to the block. The classification of succeeding sentences after the citing sentence is performed through supervised models either relying on standard discrete output classifiers like SVMs or sequence prediction through CRFs. 

Abu-Jbara et al.~\cite{DBLP:conf/naacl/Abu-JbaraR12} determine the citation block by first segmenting the sentences and then classifying individual words as being \emph{inside}/\emph{outside} the citation. Finally, the segment is classified depending on the word labels (majority of words being inside, at least one, or all of them). This approach is not applicable in our case due to the fact that words in Wikipedia text are not domain or genre specific as one expects in scientific text, and as such their classification does not work.

\paragraph{Citations in IR.} The importance of determining the citation span has been acknowledged in the field of Information Retrieval (IR). The focus is on building citation indexes~\cite{Garfield108} and improve the retrieval of scientific articles~\cite{DBLP:conf/cikm/RitchieRT08,Ritchie:2006:FBI:1629808.1629813}. Indeed, citing statements, are highly valuable as they provide an abstraction or summary of the cited documents and as such can greatly improve the retrieval process of scientific articles. In majority of the cases, the citing statements consist of the citing sentence and sentences extracted from fixed window size. 

\paragraph{Citations for Summarization.} Citations have been successfully employed to generate summaries of scientific articles~\cite{DBLP:conf/coling/QazvinianR08,DBLP:journals/jasis/ElkissSFESR08}. In all cases, citing statements are either extracted manually or from heuristics such as extracting only the citing sentences. Similarly~\cite{DBLP:conf/ijcai/NanbaO99} expands the summaries in addition to the citing sentence based on cue words (e.g. \emph{`In this', `However'} etc.). The work in~\cite{DBLP:conf/acl/QazvinianR10} goes one step beyond and considers sentences which do not \emph{explicitly} cite another article. The task is to assign a binary label to a sentence, indicating whether it contains context for a cited paper. Since the work in ~\cite{DBLP:conf/acl/QazvinianR10} tackles a similar problem to the one we address in Chapter~\ref{ch5:cite_span}. Similarly, here too the premise is that citations are marked explicitly and additional citing sentences are found dependent from them. We compare against it and show that independent of the methodological differences, that there is a major difference between scientific articles and the language in web and news sources.

\paragraph{}
The language style and the composition of citations in Wikipedia and in scientific text differ significantly. Citations are \emph{explicit} in scientific text (e.g. \emph{author names}) and are usually the first word in a sentence~\cite{DBLP:conf/naacl/Abu-JbaraR12}. In Wikipedia, citations are \emph{implicit} and there are no cue words in text which link to the provided citations. For example, for a statement in Wikipedia articles a reference to a web or news source is supposed to provide evidence for the statement therein. Hence, the link between the two is only explicit when we analyze the content of the statement and the reference, that is, there are no explicit cues in the statement that point to the reference. Therefore, the proposed methodologies and features from the scientific domain do not perform optimally in the case of determining the span of a citation in Wikipedia. 

In the case of \cite{DBLP:conf/acl/QazvinianR10}, we show that despite the similarities in the objectives of their approach, however, due to the differences in domains and the sentence level granularity it leads to erroneous spans in Wikipedia. This similarly holds for~\cite{DBLP:journals/ipm/OConnor82} where as shown in Table~\ref{tbl:sentence_citation}, in Wikipedia, the citation span needs to be performed at the sub-sentence level.

Related to our problem is the work on addressing quotation attribution. Pareti et al.~\cite{DBLP:conf/emnlp/ParetiOKCK13} propose an approach for addressing the \emph{direct} and \emph{indirect} quotation attribution. The task is mostly based on lexical cues and specific \emph{reporting verbs} that are the signal for the majority of direct quotations. However, in the case of quotation attribution the task is to find the \emph{source, cue}, and \emph{content} of the quotation, whereas in our case, for a given citing paragraph and reference we simply assess which text fragment is covered by the reference.
\paragraph{}
The approach in Chapter~\ref{ch5:cite_span} is the first attempt to determine the span of citations for \emph{web} or \emph{news} references in Wikipedia. The approach therein can help further improve and enforce the editing policies by explicitly marking the span of a citation in a Wikipedia article.


\clearemptydoublepage
\chapter{Finding News Citations for Wikipedia Entities}\label{ch1:citation_recommendation}

Wikipedia has become the most used Internet encyclopedia, and indeed, one of the most popular websites overall.\footnote{\small{In 2017 it was in top--10 most visited sites according to Alexa \url{www.alexa.com}.}} In addition, due to Wikipedia's inclusion into widely used applications such as Google KnowledgeGraph or Apple's Siri system, its content will influence the knowledge, and potentially the behavior of millions of users, even if they do not visit the Wikipedia site directly. Therefore, it is essential that its content is accurate and reliable.

Contrary to traditional encyclopedias,  Wikipedia is not authored mainly by experts. Articles are authored collaboratively by more than just a small number of contributors and the identity and expertise of authors is hard to verify. This leaves Wikipedia articles open to addition of inaccurate content, spamming or vandalism, and calls into question its reliability. A substantial number of reliability studies have compared Wikipedia against other reference works (such as the \textit{Encyclopedia Britannica} or drug package information) or subjected them to expert review: The exhaustive survey in \cite{ASI:ASI23172} concludes that the results of these studies have overall been favorable to Wikipedia when it comes to accuracy of facts, although some works (especially on medical articles) found errors of omission.\footnote{\small{The standard for medical information should be higher for obvious reasons and omitted information for side effects or risks can be crucial.}}

These surprisingly favorable results on the reliability of Wikipedia can in all probability be traced to a small number of Wikipedia editorial policies~\cite{wiki_policy}, one of which we are concerned with in this chapter. The \textit{Verifiability} policy requires Wikipedia contributors to support their additions with citations from authoritative external sources. In particular, Wikipedia policy states that \emph{``articles should be based on reliable, third-party, published sources with a reputation for fact-checking and accuracy.''}\footnote{\label{reliability}\small{\url{https://en.wikipedia.org/wiki/Wikipedia:Identifying_reliable_sources}}} This policy, on the one hand, guides contributors towards both neutrality and the importance of authoritative assessment, and on the other hand, allows Wikipedia core editors to identify unreliable articles more easily by the lack of such citations. Citations therefore play a crucial role in ensuring and upholding Wikipedia reliability.

For current and recent events, as we showed in Chapter~\ref{ch:news_wiki_lag}, news citations are one of the most-used sources~\cite{DBLP:conf/websci/FetahuAA15}. Again, Wikipedia encourages the use of news outlets as citations: \emph{``news reporting from well-established news outlets is generally considered to be reliable for statements of fact''}\footnoteref{reliability}. Indeed, news are the second-most widely used citation category in Wikipedia (with 1.88 million citations in our English Wikipedia snapshot). However, 26\% of these are no longer available due to dead or redirected links.  In addition, new information is added all the time and will need verification. For both these purposes, an automatic way of finding an authoritative news citation for any fact(s) one might wish to update, locate again or add would greatly facilitate Wikipedia editing and improve its reliability. Moreover, if no such citation can be found, it can guide contributors or core editors towards questioning their edits.

In this chapter, we describe an approach, which deals with the problem of finding automatically news citation for Wikipedia entities. In particular, we make the following contributions: 

\begin{itemize}
	\item We analyze for which type of Wikipedia statements a news citation is appropriate (in contrast to, for example, a scientific journal citation), taking into account the type and structure of entity the statement is about, as well as the language the statement is written in. We provide a supervised learning algorithm for \emph{statement classification} into the different citation categories. 
	\item We develop a citation discovery algorithm which formalizes three properties of a good citation, namely that it \emph{entails} the statement it supports, that it is from an \emph{authoritative} source and that the statement it supports is \emph{central} to it.
	\item We establish a large-scale evaluation framework for citation discovery which uses crowdsourcing for measuring our approach's precision.
\end{itemize}

To the best of our knowledge, this is the first work that automatically discovers citations for fine-grained Wikipedia statements. We show that news citations can be discovered with high precision, in large contemporary news collections. In particular, we with high accuracy recover the same or very similar citations as the ones originally given by Wikipedia contributors in the presence of numerous strong distractors or even find citations which are preferable to the original ones (as established via crowdsourcing).

\section{Problem Definition and Approach Outline}\label{sec:problemdefinition}

In this section, we describe the terminology and problem definition for finding \emph{news citations} for Wikipedia entities.

\subsection{Terminology and Problem Definition}\label{subsec:problem_definition}

We operate on a specific snapshot of Wikipedia $\mathbf{W}$ where the text in each Wikipedia page $e \in \mathbf{W}$ is organized into \emph{sections} denoted by $\Psi(e)$. Additionally, entity pages are organized into a \emph{type} structure, which is a \emph{directed-acyclic-graph} (DAG) induced by the Wikipedia categories. 

This is routinely exploited by knowledge bases like YAGO (e.g. \emph{Barack\_Obama} \texttt{isA Person}) and we leverage this type structure where each page $e$ belongs to a set of types $T(e)$. We, however, modify the original YAGO type structure to make it \emph{depth consistent} as explained in Section~\ref{subsec:learning_framework}.

\subsubsection{Citations and Wikipedia Statements} \label{sec:statements_citations}

Here, we describe the necessary terminology to explain the proposed approach for finding news citation for Wikipedia entity pages.

\begin{itemize}
	\item \textbf{Citation:} In Wikipedia pages, any \emph{piece of text} can be supported by a \emph{citation}. The citation points to an external information source, such as a news article, blog, book or journal, that is considered as \emph{evidence} for the fact mentioned in the text. 

	Citations in Wikipedia are categorized into a predefined set of \emph{16 citation categories}, $c=\{$\texttt{web}, \texttt{news}, \texttt{books}, \texttt{journal}, \texttt{map}, \texttt{comic}, \texttt{court}, \texttt{press release}, $\ldots\}$. Figure~\ref{fig:cite_type_dist} shows the distribution of the different citation types.

\item \textbf{Statement:} We refer to the \emph{piece of text} from a Wikipedia page, which \emph{has} or \emph{needs} a citation as a \emph{Wikipedia statement} or simply a \emph{statement}. 

We restrict statements to a single sentence or a sequence of sentences that occur between two consecutive citation markers or a citation marker and paragraph beginning/end.  A \emph{citation marker} is either an actual citation or a placeholder \texttt{citation needed}\footnote{\small{\url{https://en.wikipedia.org/wiki/Template:Citation_needed}}}.

Each statement $s$ in a page $e$ belongs to a section $\psi \in \Psi (e)$, and the set of statements extracted from a section $\psi$ of $e$ is represented as $S(e,\psi)$.

\item \textbf{Anchors and Entities: } Typically words or phrases in statements link to other Wikipedia pages which represent entities through \emph{anchors}. 

We denote these links to other pages or entities starting from a statement $s$ as $\gamma(s)$, and $T(s) \,\,=\,\, \{T(e) \,|\, e \in \gamma(s) \}$ the corresponding entity types.
\end{itemize}

\subsubsection{Citation Finding Tasks}\label{subsubsec:citation_finding_tasks}
Here, we present an overview of our approach on finding news citations. We decompose our approach into two main steps, which we describe below.

\begin{itemize}
	\item \textbf{Statement Categorization.} In this task the aim is to determine for a statement $s$ the appropriate citation category $c$. This step is in particular important as it allows us to consider in the later steps only those statements for which we are in hold of an appropriate corpus, e.g. \emph{news collection}.
	
	 We define the following function, which for an entity page $e$ and a statement $s$ predicts the correct citation category $c$.

\begin{equation}
SC:  f(s,e) \rightarrow c, \text{where $c\in\{$\texttt{web, news,$\ldots$}}\}
\end{equation}

We want to categorize $s$ as a requiring a \emph{news} citation if a news citation is the most appropriate citation type. This is based on the hypothesis that each statement typically has a preferred citation category, which we need to determine before making a high precision citation recommendation. We call a statement $s$ that requires a news citation as a \emph{news statment}.

\item \textbf{Citation Discovery.} In this task we are given a statement $s$, which based on the previous step, we classify its citation category to be $c=$\emph{news}, and a news collection $\N$. 

We define the \emph{citation discovery} as the task of finding news articles $n\in \N'$, where $\N' \subseteq \N$, which can be suggested as a citation for statement $s$.

\begin{equation}
FC: f(s,e,\mathcal{N}) \rightarrow \langle s, n\rangle \in \{`correct', `incorrect'\}
\end{equation}
\end{itemize}

\subsection{Approach Overview}\label{subsec:approach_overview}

\begin{figure*}[ht!]
	\centering
	\includegraphics[width=1.0\textwidth]{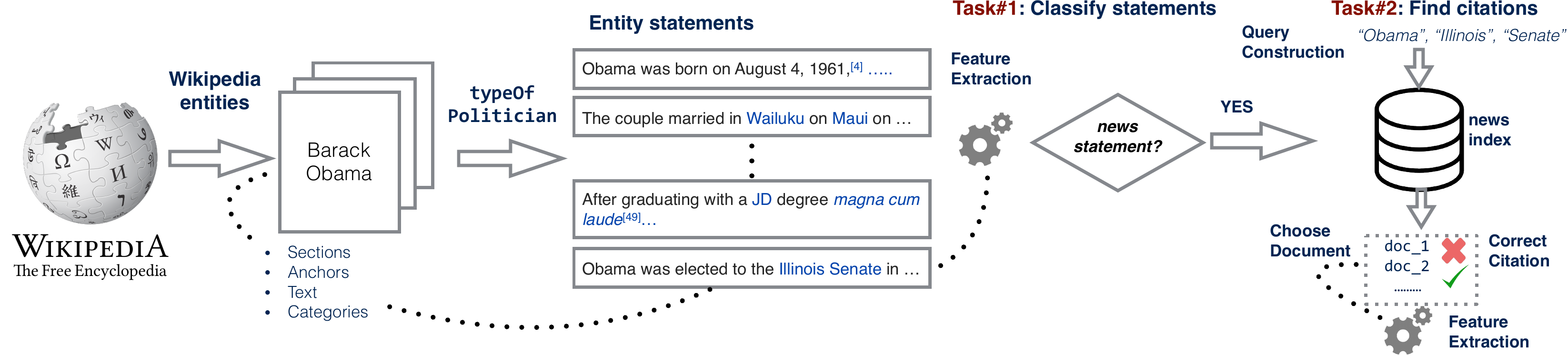}
	\caption{Approach overview. In the first task, we classify statements into one of the citation categories, and in the second we find the appropriate news article for citation for a news statement.}
	\label{fig:approach_overview}
\end{figure*}

Figure~\ref{fig:approach_overview} shows an overview of the proposed approach. For an entity, we extract the section and type structure. Additionally, from the individual section we extract its statements and for each statement we run both steps for statement categorization and citation discovery.

\begin{enumerate}
	\item \textbf{\emph{Statement Categorization--SC.}} In this step, we predict the citation category of a Wikipedia statement $s$ via a supervised machine learning model which we train on features we extract from the statement itself and the entity $e$. We train a multi-class classification model, where the classes correspond to the citation categories $c$.
	\item \textbf{\emph{Citation Discovery--FC.}} For all the \emph{news statements} as classified in the previous step, we find evidence via news articles from our news collection $\N$.  We retrieve \emph{candidate} news articles from a news collection $\N$ through standard \emph{information retrieval} methods with $s$ serving as our query, and classify each candidate as either an appropriate citation for $s$ or not.
\end{enumerate}

\section{Wikipedia Ground-Truth}\label{sec:datasets_groundtruth}

Before delving into the details of the proposed approach, we first describe the ground-truth dataset we use to evaluate our approach. We describe two main aspects: (i) the extract of statements from Wikipedia and their corresponding citation categories, and (ii) the news collection which consists of news articles already references from Wikipedia entity pages.

\subsection{Ground-Truth: Wikipedia News Statements}

We extract all statements as defined previously in Section~\ref{subsec:preliminaries} from a Wikipedia snapshot $\mathbf{W}$\footnote{The snapshot corresponds to the state of Wikipedia accessed at 2015-07-01}. Additionally, for each statement we extract all the citations to the external references. A statement can have more than one citation and in all cases we assume that a citation is valid for the entire statement\footnote{As a statement can have different clauses, sometimes extracted citations only serve as evidence for part of the statement.  We, however, do not distinguish at this level of granularity but assume that all associated citations support the whole statement.}

In total, we extract 6.9 million statements which point to 8.8 million citations. After analyzing the set of entity pages to which these statement are assigned, we are left with 1.65 million entities and a total of 668 distinct sections.

As described previously, each citation belongs to a category which is chosen by the Wikipedia editor from a citation category template. However, since there are no policies that enforce a strict categorization of citations, in some cases, Wikipedia editors assign the wrong categories. For instance, in cases where the citation refers to a news article, instead of \texttt{news} category, categories like \texttt{web} are assigned.

\paragraph{Ground-Truth Curation.}
For example, from the collection of citations in \textbf{W}, the top--3 news domains \emph{BBC}, \emph{NYTimes}, \emph{Guardian}, are often cited in categories other than \texttt{news}.\footnote{Thus, the citation \small{\url{http://news.bbc.co.uk/1/hi/uk_politics/7433479.stm}}  from the entity \texttt{Liam} \texttt{Byrne} has been categorized as \texttt{web}, although the more specific \texttt{news} category  would have been appropriate.} Most of such \emph{violations} by the editors occur when citing news under the category \texttt{web}. Indeed, the citation category \texttt{web} has a wide scope of its validity, however, in cases where a more fine-grained category can be chosen than the use of \texttt{web} category is inappropriate.

In most cases such violations can be accurately corrected by applying simple heuristics. We propose two simple, yet effective heuristics for this purpose:

\begin{itemize}
	\item \textbf{Majority Voting.} Citations from the same domain URL are tagged with different categories. We resolve such cases based on \emph{majority voting}. In case a domain is cited more often under the \texttt{news} category, then all citations to the same domain are changed to \texttt{news}.

	\item \textbf{URL Patterns.} In this heuristic we look for patterns in the URL, specifically for `\emph{/news/}' and `\emph{http://news.}'. This rule is applied to \texttt{web} statements, and in case the URL matches one of the patterns, we change its category to \texttt{news}.
\end{itemize}

Table~\ref{tbl:gt_cite_changes} shows the top--4 most frequent citation categories and the impact of our ground-truth curation rules. Rule application changes the citation category for 1.65 million citations, approximately 18\% of all citations in \textbf{W}. The cells in the table show the number of statements that are changed from the category in the \emph{row} to the category in the \emph{column} table.

\begin{table}[ht!]
\centering
\begin{tabular}{l l l l l }
\toprule
& book & journal & news & web\\
\cmidrule{2-5}
book & 0 & 2,650 & 1,155 & 71,801\\
journal & 14,905 & 0 & 13542 & 110,133\\
news & 5,698 & 2,770 & 0 & \textbf{391,634}\\
web & 16,549 & 25,109 & \textbf{944,977} & 0\\
\bottomrule
\end{tabular}
\caption{The cells show the number of statements that are changed from one category to another category after ground-truth curation.}
\label{tbl:gt_cite_changes}
\end{table}

\begin{figure}[ht!]
\centering
\includegraphics[width=0.8\columnwidth]{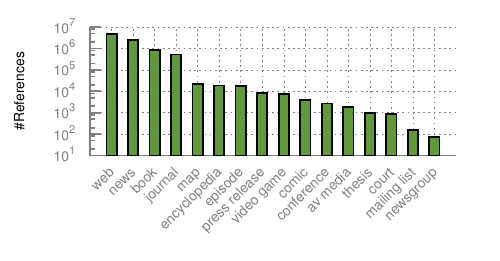}
\caption{Statement distribution by citation category.}
\label{fig:cite_type_dist}
\end{figure}

Finally, we assume that a statment $s$ is a \emph{news statement} if it contains at least one news citation (after the ground-truth curation). Figure~\ref{fig:cite_type_dist} shows the statement distribution across the citation categories. It is evident that \texttt{web} and \texttt{news} are the two most popular categories, with 5.3 and 1.88 million citations, coming from 1.2 million and 436k entities, respectively.

\subsection{Wikipedia News Collection}\label{subsec:data_wiki_news}

From the \emph{news statements}, we extract the cited news articles and construct the news collection, which we refer as the \emph{Wikipedia news collection}, $\mathcal{N}^{W}$. The collection $\N^W$ serves as our ground-truth for the \emph{citation discovery} task. 

We distinguish the following sub-collections from $\N^W$. With $N_s$ we denote the set of news articles cited by the news statement $s$, and with $N_t\subseteq\N^W$ we refer to the set of news articles cited from news statements $s$ whose entities are of type $t$.

\begin{figure}[ht!]
\centering
\includegraphics[width=0.8\columnwidth]{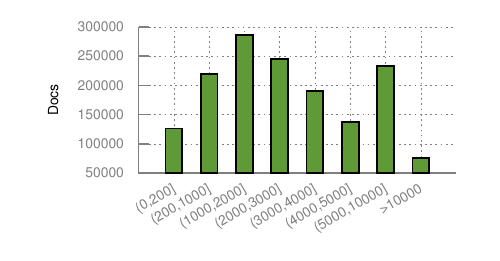}
\caption{News article length (in number of characters) distribution.}
\label{fig:news_length}
\end{figure}

The collection $\N^W$ consists of 1.88 million news articles, from which we successfully crawled 1.55 million articles. The remaining 19\% point to non-existent articles (dead links, moved content, etc.). Furthermore, some of the successfully crawled URLs point to the \emph{index pages}. This can be noticed  when we consider the article length (in terms of characters) in Figure~\ref{fig:news_length}. After filtering news articles whose length was below 200 characters, we are left with 1.39 million articles. This presents a decrease of 26\% from the original set of 1.88 million news articles.

An additional issue we notice in $\N^W$ are citations to non-English news articles. We find that 23\% of articles in $\N^W$ are in languages other than English. We use the Apache Tika\footnote{\small{\url{http://tika.apache.org}}} for detecting the language of the news articles.

\section{Statement Categorization}\label{sec:statement_classification}

In the statement categorization task, we are given a statement $s$ and the entity $e$ from which it is extracted. We compute features that exploit the language style of $s$ and the type and section structure of $e$ to categorize $s$ into one of the citation categories $c$. We learn a multi-class classifier (Section~\ref{subsec:learning_framework}) with classes corresponding to citation categories $c$ and optimize for predicting news statements.  

Table~\ref{tbl:sc_feature_list} shows an overview of the features that we compute for the function $SC$ to categorize a statement $s$.

\begin{table}[h!]
\centering
\begin{tabular}{p{3cm} p{10cm} p{2cm}}
\toprule
feature & description &\\
\midrule
$\# verbs\_attr$ & the number of verbs of attribution in statement $s$& \multirow{7}{2cm}{\emph{Language Style}} \\
$\# POS$ & the frequency of POS tags in $s$ & \\
$\lambda(s)$ & temporal proximity of $s$ to the time-point of  $\mathbf{W}$ &\\
$discourse$ & explicit discourse annotations of $s$ & \\
$\#quotations$ & the number of quotations in $s$ & \\
$\theta(s,N_t)$ & KL score between the LM of $s$ and that of $N_t$ & \\
$LDA(s, N_t)$ & similarity of $s$ to the topic model computed from  $N_t$ & \\[3ex]

$p(s= news | \psi)$ & \multirow{2}{10cm}{section and type priors, with \emph{min, max} and \emph{avg} scores}&  \multirow{4}{2cm}{\emph{Entity Structure}}\\
$p(s=news | t)$ & & \\
$p(s=news | t',t)$ & type co-occurrence probability between $t\in T(e)$ and all the possible types from entities in $s$, that is $t'\in T(s)$ & \\
$p(s=news|t,\psi)$ & type-section probability scores & \\
\bottomrule
\end{tabular}
\caption{Feature list for statement categorization.}
\label{tbl:sc_feature_list}
\end{table}

\subsection{Statement Language-Style}

We hypothesize that Wikipedia news statements are similar to the language style of news, as they often paraphrase cited news articles. Different genres (such as \textit{news, recipes, sermons, FAQs, fiction \ldots}) differ in their linguistic properties as the different functions they fulfill influence linguistic form \cite{biber1991variation}. For example, we expect news reports (which center mostly on past events) to contain more past tense verbs than a recipe which gives instructions via verbs in the imperative. We use features that were successful in automatic genre classification including structural features via parts-of-speech  as well as lexical surface cues \cite{petrenz2011stable}.

\paragraph{Part of Speech Density.} Frequency of part-of-speech (POS) tags, determined via the Stanford tagger, allows us to capture some of the structural properties of text. For example, news statements can be characterized by a high number of past tense verbs as well as proper nouns. We normalize the POS tag frequency w.r.t the sum of all tags in a statement, to account for varying statement length.

\paragraph{Verbs of Attribution and Quotation Marks.} News articles often report statements by persons of repute, witnesses or other sources. We approximate this kind of behavior in news articles through two features: Firstly, we count \emph{verbs of attribution} in $s$, via a list of 92 such verbs (\textit{claim, tell etc}) with POS tag \texttt{VB*} and normalize w.r.t the total number of \texttt{VB*}. Secondly, we use \textbf{quotation marks} as a potential indicator of \emph{paraphrasing}. The feature simply counts the number of quotation marks in $s$, normalized w.r.t the statement length.

\paragraph{Temporal Proximity $\lambda(s)$.} Most Wikipedia ness statements refer to relatively recent events, i.e. events close to the time of the Wikipedia snapshot. We use \emph{temporal expressions} such as dates and years as distinguishing features for news statements. We use a set of hand-crafted regular expression rules to extract temporal expressions.\footnote{This proved to be more scalable than state-of-the-art extractors like   HeidelTime~\cite{Strotgen:2010:HHQ:1859664.1859735} and Stanford's CoreNLP~\cite{DBLP:conf/acl/FinkelGM05} module}. We use the following rules: (1) \texttt{DD Month YYYY}, (2) \texttt{DD MM YYYY}, (3) \texttt{MM DD YY(YY)}, (4) \texttt{YYYY}, with different delimiters (whitespace, `-', `.'). We then compute $\lambda(s)=|Year(\mathbf{W}) - Year(s)|$.

In this case, for temporal proximity we could rely on more accurate temporal cues to measure the distance between the news statement and the revision date of an entity when we first encounter $s$. However, to do so, we would need to trace all the changes in an entity page to be able and determine the accurate revision date for $s$. Therefore, we opt for the more simplistic approach where the distance is in terms of years, and is with respect to the year of the Wikipedia snapshot.

\paragraph{Discourse Analysis.} We use discourse connectives to annotate the statements $s$ with \emph{explicit discourse relations} based on an approach proposed by Pitler and Church~\cite{DBLP:conf/acl/PitlerN09}. The annotations belong to the categories $\{$\emph{temporal, contingency, comparison, expansion}$\}$, following the Penn Discourse Treebank annotation \cite{prasad2008penn}. Some of the explicit discourse relations are particularly interesting (i.e., \emph{temporal}) as they represent a common language construct used in news articles that report event sequences. The features are boolean indicators on whether $s$ contains a specific explicit discourse relation.

\paragraph{Language Model and Topic Model Scoring.} As surface lexical features have been shown to be efficient in genre recognition \cite{sharoff2010web}, we compute n--gram (up to n=3) language models with \emph{Kneser-Ney} smoothing (LM) from news articles $N_t$ and compute the Kullback-Leibler divergence score between the corresponding language models, $\theta(s,N_t)$. The score shows how likely $s$ can be constructed from the LM from news articles cited from entities of type $t$. 

Similarly, we compute topic models using the LDA framework~\cite{DBLP:journals/jmlr/BleiNJ03}, where the score is the Jaccard similarity between $s$ and the topic terms from $N_t$.

\subsection{Entity-Structure Based Features}

Determining if a statement requires a news citation solely on language style is not always feasible. We exploit the entity structure of $e$ and compute the probability of statements having a news citation given its types $T(e)$ and sections $\Psi(e)$.

\paragraph{Section-Type Probability.} A good indicator of the likelihood that a statement $s$ requires a news citation is the entity type it belongs to and the section that it appears in. For instance, for type \texttt{Politician}, news statements have higher density in section \emph{`Early Life and Career'} as these tend to be more reflected in news. 

Since the number of combinations between entity types and sections can be large. One precaution we need to take into account is over-fitting. We avoid over-fitting by filtering out entity types with fewer than 10 statements. Similarly, we filter out sections with fewer than 10 statements, and in which case all the statements belong to the same citation category.

We compute the probability of a statement $s$ having a news citation given the entity type it appears in and the corresponding section in the entity page $e$.

\begin{equation}
p(s=news|t,\psi) = \frac{\sum_{e\in \mathbf{W} \wedge t\in T(e)}\sum_{s \in S(e,\psi)}1_{s \text{\texttt{ typeOf } news}}}{\sum_{e\in \mathbf{W}\wedge t\in T(e)}{|S(e,\psi)|}}
\end{equation}

The $p(s=news|t,\psi)$ probability is likely to be a sparse feature, so we compute type and section news-priors. We compute section $p(s=news|\psi)$ and type news-priors $p(s=news| t)$ as the ratio of news statements over the total number of statements (of any citation category) that belong to a section $\psi$ or type $t$, respectively.

Since $s$ is associated with an entity $e$, which consequentially may be associated with a set of types $T(e)$, we aggregate the computed type news-priors and the section-type probability into their \emph{min, max} and \emph{avg} scores.

\paragraph{Type Co-Occurrence.} From the entity types $T(s)$ and $T(e)$ we measure the likelihood of type co-occurrence in news. The probability simply counts the co-occurrence between $t$ and $t'$ in
news statements with respect to their total co-occurrence. Examples of highly co-occurring types in news are \texttt{Politician} and \texttt{Organization}.

\begin{equation}
p(s=news|t',t) = \frac{\sum_{e\in \mathbf{W}\wedge t\in T(e)}\sum_{s \in S(e)\wedge t'\in T(s)}1_{s \text{\texttt{ typeOf } news}}}{\sum_{e\in \mathbf{W}\wedge t\in T(e)}\sum_{s \in S(e)}1_{t'\in T(s)}}
\end{equation}

\subsection{Learning Framework}\label{subsec:learning_framework}

\paragraph{Learning Setup.} Wikipedia consists of a highly diverse set of entities. A model trained on all entities is unlikely to work. For example, the types \texttt{Location} and \texttt{Politician} represent two highly divergent groups with regard to their page structure, the statements they contain and the way they are reported in news.

Therefore, we learn $SC$ for individual types in the YAGO type taxonomy. The advantages of type specific functions $SC$ is that they are trained on homogeneous entities, which helps the models to predict with higher accuracy. We take only types that have more than 1000 entity instances, resulting in a total of 672 types from the YAGO type taxonomy. The types are organized from very broad types such as  (\texttt{owl:Thing}) to very specific types like \texttt{Serie\_A\_Players}.

To utilize the specialization and generalization in a principled manner we transform the YAGO type taxonomy (DAG) into a hierarchical DAG. This is utilized later on in order to find the right level of type granularity for learning $SC$. That is, if we start at the root of the type taxonomy, we will have all possible entities, and as we go down the type taxonomy the number of entities per type decreases, however, the entities are more homogeneous. 

\begin{figure}
	\centering 
	\includegraphics[width=0.5\textwidth]{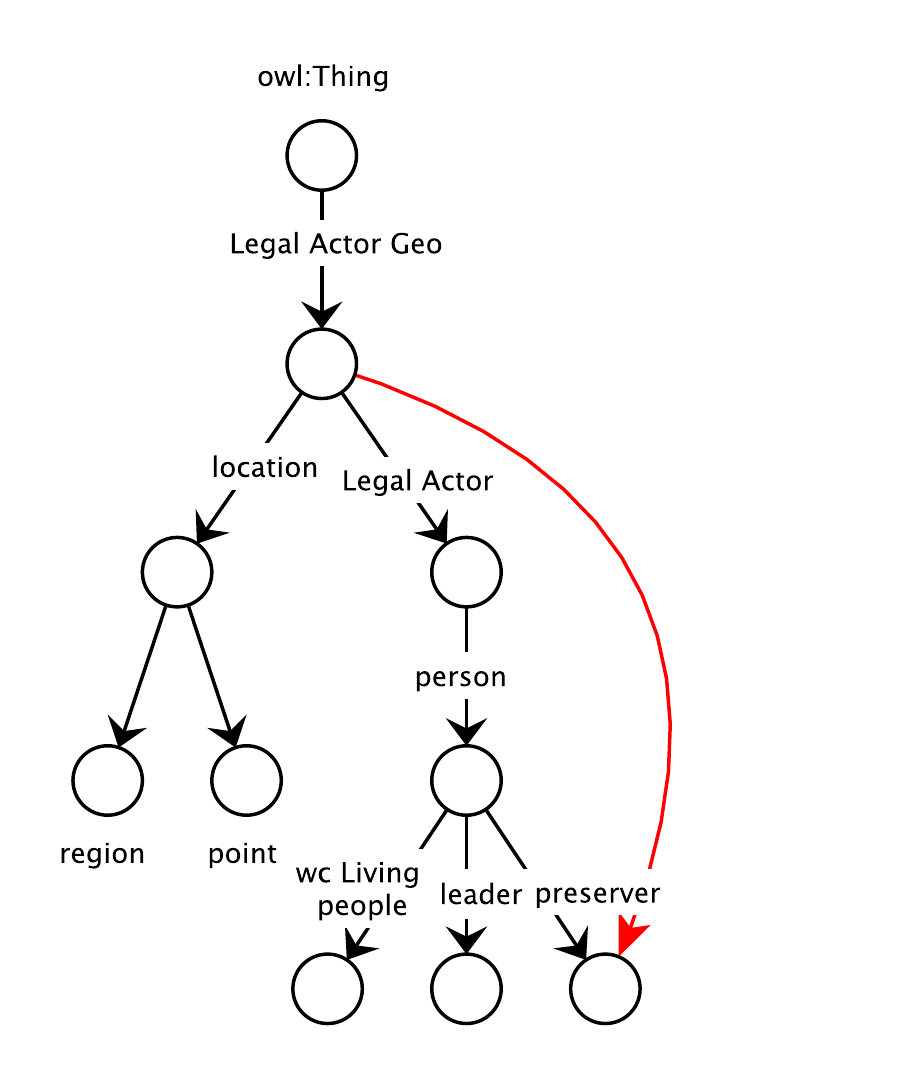}
	\caption{An example sub-tree from the YAGO type taxonomy. The edge in red represents the edge we need to delete since it is not depth-consistent.}
		\label{fig:yago_taxonomy}
\end{figure}

We assume that the hierarchy is rooted at \texttt{owl:Thing} and all internal nodes are \emph{depth-consistent}, i.e. all paths from the root to the node are of the same length. We obtain this by a simple heuristic whereby for every \emph{child type} $\rightarrow$ \emph{parent type} we remove edges where the parent's \emph{depth level} in the taxonomy is higher than the \emph{minimum level} from other parent nodes. Figure~\ref{fig:yago_taxonomy} shows an example of how we construct a depth consistent type taxonomy.

With this hierarchical type-taxonomy, we can determine the \emph{optimal level} of type granularity such that we have optimal performance in categorizing statements. For learning the type specific $SC$, we keep 10\% of entity instances for evaluation and the remainder for training. It is important to note that when we learn $SC$ for a given type, the training instances are sampled through \emph{stratified sampling} from all its \emph{children types}.

\paragraph{Learning Model.} The functions $SC$ represent \emph{multi-class} classifiers with classes corresponding to the citation categories. Since we want to predict the news category $c=$`\emph{news}' with high accuracy, one question is why we do not pose this as a \emph{binary} classification problem, where a statement is categorized as news or not. We used the \emph{multi-class} classifiers because they give us a more balanced distribution when compared to merging all non-news statements into a single category.

Finally, we opt for \emph{Random Forests} (RF)~\cite{Breiman2001} as our supervised machine learning model. The detailed explanation of RF is shown in Chapter~\ref{ch2:foundations}. We experimented with other models, but the differences in performance are marginal, and RF have superior learning time. We train the models on the full feature set in Table~\ref{tbl:sc_feature_list}.

\section{Citation Discovery}\label{sec:missing_citations}

For the citation discovery task, we follow the \emph{citation policy}\footnote{\small{\url{https://en.wikipedia.org/wiki/Wikipedia:Citing_sources}}} guidelines in Wikipedia and single out three key properties on what makes a \emph{good citation}.

\begin{enumerate}
 	\item the statement should be \emph{entailed} by the cited  news article
 	\item the statement should be \emph{central} in the cited news article
	\item the cited news article should be from an \emph{authoritative} source 
\end{enumerate}

We approach the citation discovery for news statements as follows. We use statement $s$ as a query (see Section~\ref{subsec:query_construction}) to retrieve the top--$k$ news articles from $\N$ as citation candidates for $s$. We then classify the candidate citations as either `\emph{correct}' or `\emph{incorrect}', depending on whether they meet the above criteria of a \emph{good citation}.

To do so, we compute features for each pair $\langle s, n_i\rangle$, w.r.t the individual sentences of a news article $n_i$. The feature vectors become the following: $\langle s, [\sigma_i^1, \sigma_i^2,\ldots, \sigma_i^j]\rangle$, where $\sigma_i^j$ represents the $j$-th sentence from $n_i$.

Since the number of sentences $\sigma_i$ varies across news articles, we aggregate the individually computed features at sentence level into the corresponding \emph{min, max}, \emph{average}, \emph{weighted average}, and \emph{exponential decay function} scores as shown below.

\begin{equation}
\langle s, \min_{j}F(\sigma_i^j), \max_{j}F(\sigma_i^j), Avg(F(\sigma_i)), \sum_{\sigma_i^j}\frac{1}{j}* F, \sum_{\sigma_i^j} F^{\frac{1}{j}},\ldots\rangle
\end{equation}
where $F$ is a feature from the complete feature list in Table~\ref{tbl:entailment_citations_feature_list}, \ref{tbl:centrality_citations_feature_list}, and Section~\ref{subsec:authority}.

\subsection{Query Construction}\label{subsec:query_construction}

We use the text from the statement $s$ to query our news collection $\N$. The length of a statement can be from a single sentence to an entire paragraph. For this reason, we need to employ efficient query construction approaches that extract the keywords from $s$ such that we increase the coverage and likelihood of retrieving relevant news article candidates for statement $s$.

It has been shown that in similar cases where the query corresponds to a sentence or paragraph, query construction (QC) approaches are necessary to increase the accuracy of IR models. Henzinger et al.~\cite{DBLP:conf/www/HenzingerCMB03} propose several QC approaches that weigh query terms based on the \emph{tf--idf} score, and other variations of weighting the individual terms.

We experimented with different QC approaches from \cite{DBLP:conf/www/HenzingerCMB03} and their impact on finding news articles in $\mathcal{N}^W$. From the many variations proposed in \cite{DBLP:conf/www/HenzingerCMB03}, we find one approach \emph{QCA1Base}, which had the best performance in terms of improving the coverage of relevant articles in top--$k$. In \emph{QCA1Base}, the terms extracted from the statements are weighted based on \emph{tf--idf}. The score of $tf$ simply represents the term frequency of a word in $s$, whereas $idf$ is computed with respect to other terms in our statement collection.

In principle, one should consider all retrieved articles from the result set. However, this is not only computationally expensive for our subsequent learning step but also unbalances our training set. The reason for this is that, news statements on average do not have more than one citation. Hence, if we go to high retrieval depths in order to have perfect recall, we have the vast majority of documents being irrelevant for $s$. 

To determine a reasonable retrieval depth, we experimented with 1000 randomly chosen statement queries with QC and determined the hit-rate at retrieval depth $k$ , i.e.  whether the cited article is retrieved in the top---$k$ articles.

\begin{figure}[ht!]
\centering
\includegraphics[width=0.7\columnwidth]{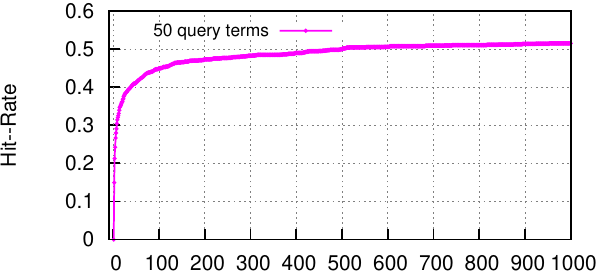}
\caption{{Hit-rate of articles in $\N^W$ up to rank 1000 (x--axis) for 1000 news statements, respectively \emph{QCA1Base} queries.}}
\label{fig:index_coverage}
\end{figure}

Figure~\ref{fig:index_coverage} shows the hit-rate in top--1000 with top 50 ranked query terms and with \emph{divergence from randomness} (see Chapter~\ref{ch2:foundations}) query similarity measure~\cite{Amati:2002:PMI:582415.582416} for our \emph{random sample} of 1000 news statements.

We focus on the top--100 retrieved news articles as potential citations for $s$, as the achieved hit-rate beyond the top--100 shows only minor improvement. In Figure~\ref{fig:index_coverage}, we also note that the hit-rate does not go beyond 50\%. We found that most of the news articles that are not retrieved are either missing or non-English articles in $\N^W$.

\subsection{Textual Entailment Features}\label{subsec:entailment}

\begin{table}[ht!]
\centering
\begin{tabular}{l p{10cm}}
\toprule
& \emph{entailment}\\
\midrule
$\mathbf{J}^{n}(s, \sigma_i^j)$ & n--gram overlap between $s$ and $\sigma_i^j$ and similarity headline of $n_i$ and $s$  \\

$\mathbf{J}^{P}(s, \sigma_i^j)$ &  NNP phrase overlap between $s$ and $\sigma_i^j$ \\

$\theta^1(s,\cdot)$ & $s$ unigram LM from news article $n_i$ and n--gram LM from news articles in $N_t$ \\

$K(s, \sigma_i^j)$ & tree kernel similarity between $s$ and $\sigma_i^j$  \\

$LDA(s,N_t)$ & term overlap between $s$ and topic terms from news in $N_t$  \\

$freq(e)$ & occurrence frequency of $e$ in the title and body of $n_i$ \\

\emph{baseline features} & retrieval score from the IR model for $n_i$ and its rank \\
\bottomrule
\end{tabular}
\caption{Entailment feature set for the citation discovery task.}
\label{tbl:entailment_citations_feature_list}
\end{table}

As the citation is supposed to give strong evidence for the statement's content, in the ideal case the cited news article should fully \textit{entail} the statement, i.e. the statement should be derivable from the news article. The recognition of textual entailment has been the study of extensive research in the last 10 years; cf \cite{dagan2013recognizing} for an overview. A full treatment of entailment needs extensive world knowledge and inference rules; we here restrict ourselves to much simpler lexical and syntactic similarity methods used in baseline entailment systems and leave the extensions to future work.\footnote{\small{Off-the-shelf entailment systems exist but are too slow to use at scale.}}

\paragraph{IR Baseline Features.} We use the retrieval model as a pre-filter to find candidate news articles as citations for $s$. The retrieval model also provides us with two possible features for the learning model: firstly, a \emph{matching score} of $n_i$ for query $s$, where the score corresponds to the \emph{divergence from randomness} query similarity measure~\cite{Amati:2002:PMI:582415.582416}. Secondly, the retrieval rank of $n_i$. We use the IR model as our \emph{baseline} and hence refer to them as \emph{baselines features}.

\paragraph{Tree Kernel Similarity.} Lexical similarity measures in many cases fail to capture the joint \emph{semantic} and \emph{syntactic} similarity. For this purpose, we consider the \emph{tree kernel} similarity measure proposed in~\cite{DBLP:conf/emnlp/Kate08}. We first compute the \emph{dependency parse trees} of $s$ and $\sigma_i^j$ using the Stanford tagger~\cite{toutanova2000enriching}, and then compute the tree kernel, $K(s, \sigma_i^j)$. Tree kernel similarity through the dependency parse tree measures the maximum matching subtrees between $s$ and $\sigma_i^j$, where the matching subtrees have the same syntactic and semantic meaning. We refer the reader to \cite{DBLP:conf/emnlp/Kate08} for details.

\paragraph{LM \& Topic Model Scoring.} From an article $n_i$ we compute a \emph{unigram LM} and compute $\theta(s, n_i)$ as the likelihood of $s$ being generated from the computed LM. In addition, we compute \emph{n--gram} LM (with $n$ up to 3) from articles in $N_t$, and compute the score $\theta^n(s, N_t)$ accordingly.  

Similarly, we compute \emph{LDA topic models}~\cite{DBLP:journals/jmlr/BleiNJ03} for entity types, specifically from articles in $N_t$. This follows the intuition that content usually is clustered around specific topics, i.e. for type \texttt{Politician} most discussions are centered around politics, career, etc. The topic score is the Jaccard similarity between $n_i$ and the topic terms.

\subsection{Centrality Features}\label{subsec:centrality} 

\begin{table}[ht!]
\centering
\begin{tabular}{l p{10cm}}
\toprule
& \emph{centrality}\\
\midrule
$\mathbf{J}(s, \sigma_i^c)$ & Jaccard similarity between $s$ and central sentence $\sigma_i^c$ \\

$\mathbf{J}^{P}(s, \sigma_i^c)$ &  NNP phrase overlap between $s$ and $\sigma_i^c$ \\

$\mathbf{J}^{n}(s, \sigma_i^c)$ & n--gram overlap between $s$ and $\sigma_i^c$ \\

$K(s, \sigma_i^c)$ & tree kernel similarity between $s$ and $\sigma_i^c$\\

$\phi(e,n_i)$ & the relative entity frequency of $e$ in $n_i$ \\

$\phi(\gamma(s),n_i)$ & relative entity frequency of $e\in \gamma(s)$ in $n_i$ \\
\bottomrule
\end{tabular}
\caption{Sentence centrality feature set for the citation discovery task.}
\label{tbl:centrality_citations_feature_list}
\end{table}

\paragraph{Similarity to most central news sentence.} As described above we compute similarity features between $s$ and sentences in $n_i$. However, some sentences in $n_i$ are more central than others. Hence, the computed features between the pairs $\langle s, [\sigma_i^1, \sigma_i^2,\ldots, \sigma_i^j]\rangle$, do not have uniform weight. Therefore, we find the most \emph{central sentence} $\sigma_i^c$ in $n_i$ and distinguish the computed entailment/similarity features between $s$ and $\sigma_i^c$.

We compute centrality of a sentence in $n_i$ through the TextRank approach introduced in~\cite{DBLP:conf/emnlp/MihalceaT04}. We first construct a graph $G=(V,E)$ from $n_i$, where $V$ corresponds to the sentences of $n_i$, with edges in $E$ weighted with the Jaccard similarity between any two sentences, in this case $\sigma_i^j \in V$. Computation of centrality for any vertex $\sigma_i^j$ is similar to that of PageRank, with slight changes accounting for the weighted edges between vertices.

\begin{equation}
\Gamma(\sigma_i) = (1-d) + d * \sum\limits_{\sigma_j \in In(\sigma_i)}\frac{\mathbf{J}(\sigma_i, \sigma_j)}{\sum\limits_{\sigma_k\in Out(\sigma_j)}{\mathbf{J}(\sigma_j, \sigma_k)}} \Gamma(\sigma_j)
\end{equation}
where $d$ is the damping factor ($d=0.85$), a common value in PageRank computation. The computation converges if the difference in the score of $\Gamma(\sigma_j)$ in two consecutive iterations is small.

\paragraph{Relative Entity Frequency.} The importance of $e$ in $n_i$ is crucial when finding citations for $s$. This importance is partially mirrored simply in how often $e$ is mentioned in $n_i$. However, another genre-typical property of news is its inverted pyramid structure, i.e. the most important information is mentioned at the beginning of the article. We therefore  measure relative entity frequency of $e$ in $n_i$ based on an approach described in Chapter~\ref{ch3:news_suggestion}. It attributes higher weight to entities appearing in the top paragraphs of $n_i$, where the weight follows an exponential decay function.

\begin{equation}\label{eq:rel_freq}
\phi(e,n) =  \frac{|\rho(e,n)|}{|\rho(n)|}\sum\limits_{\rho\in \rho(n)}\left(\frac{tf(e,\rho)}{\sum\limits_{e}tf(e',\rho)}\right)^{\frac{1}{\rho}}
\end{equation}
where $\rho$ represents a news paragraph from $n$ and $\rho(n)$ indicates the set of all paragraphs. $tf(e,\rho)$ indicates the frequency of $e$ in $\rho$. With $|\rho(e,n)|$ and $|\rho(n)|$ we indicate the number of paragraphs in which entity $e$ occurs and the total number of paragraphs.

Additionally we consider the relative entity frequency for entities in $e\in\gamma(s)$ and measure the \emph{minimum, maximum} and \emph{average} relative entity frequency scores.

\subsection{News-Domain Authority Features}\label{subsec:authority}

Wikipedia's editing policy distinguishes clearly between more and less-established news outlets and prefers the former.  We therefore compute the authority of news domains w.r.t entity types and sections.

We will denote the \emph{domain} of the news article referred from $s$ as $D[s]$, and with $D$ any arbitrary domain. 

\paragraph{Type-Domain Authority.} Authority of news domains is non-uniformly distributed across types. For types such as  \texttt{Politician} the authority of domains like \emph{BBC} is higher than for types such as  \texttt{Athletes}, where a domain specialized in sports news is more likely to be  authoritative. We capture the \emph{type-domain   authority} as follows:
\begin{equation*}
p(D|t) = \frac{\sum_{e\in\mathbf{W}\wedge t\in T(e)}\sum_{s\in S(e)}1_{D=D[s]}}{\sum_{e\in\mathbf{W}\wedge t\in T(e)}\sum_{s \in S(e)}D[s]}
\end{equation*}

\paragraph{Section-Domain Authority.} We measure the authority of domains associated to certain entity sections. The density of news references across sections varies heavily. Therefore, it is natural to consider the authority of news domains for a given section.
\begin{equation*}
p(D|\psi) = \frac{\sum_{e\in\mathbf{W}}\sum_{s \in S(e,\psi)}1_{D=D[s]}}{\sum_{e\in\mathbf{W}}\sum_{s \in S(e,\psi)}D[s]}
\end{equation*}

Note that these features compute news outlet authority with regard to current Wikipedia usage, which we seek to re-create. An alternative we intend to look at in future work is to measure authoritativeness via Wikipedia-external measures of news outlets, such as page visits or interlinkage.

\section{Statement Categorization Evaluation}\label{sec:t1_results}

Here we describe the evaluation of our approach for the statement categorization task. Since we consider the type taxonomy from YAGO, we have a hierarchy of models. Each statement belongs to an entity, which in turn is a child to a type (node) in the hierarchy. Consequently, we construct each model from training instances (statements) that are its children. We focus on two aspects (i) performance of models at varying depths, and (ii) performance of various feature classes.


\subsection{Experimental Setup}\label{subsec:t1_exp_setup}

\textbf{Setup.} We consider 672 entity types from our YAGO taxonomy, for which we learn individual $SC$ models. We consider types that have more than 1000 entity instances. The level of granularity in the YAGO taxonomy has a maximum depth of 20, while the root type is \texttt{owl:Thing} containing all possible entities.

\textbf{Train/Test.} We learn the $SC$ models using up to 90\% of the entity instances of a type $t$ as training set, and the remainder of 10\% for evaluation. We use \emph{stratified sampling} to pick  entities of type $t$ and its subtypes for the train and test set. We train and test $SC$ models over 6 million statements coming from 1.3 million entities.

\textbf{Metrics.} We evaluate the performance of $SC$ with \emph{precision} $P$, \emph{recall} $R$ and $F1$. A statement is considered to be classified correctly, if the predicted citation category matches to the citation category in our ground-truth.

\subsection{Results and Discussion}\label{subsec:t1_results}

The following discussion focuses on the results for the statement categorization task for the \texttt{news} citation category. We report the classification results for our function $SC$ only for the first three type levels in the YAGO taxonomy, specifically the immediate child \texttt{Legal Actor Geo} of \texttt{owl:Thing}.\footnote{For readability we remove the \emph{wordnet} prefix from the types and their numerical ID values.} 

Table~\ref{tbl:yagoLegalActorGeo_t1} shows the results for $SC$ models evaluated over 61k entities and trained with up to 550k entities, depending on the training sample size, which we vary in between the ranges $\tau \in [1\%,90\%]$.  The results for the type \texttt{yagoLegalActorGeo} represent more than 47\% of the total set of entities in our evaluation dataset.

The overall performance of $SC$ for all types for $\tau=90\%$ measured through \emph{micro-average} precision is 0.57. Since a statement belongs to multiple types $T(s)$, we decide the category of $s$  based on majority as categorized from the individual $SC$ models.

\begin{table}[ht!]
\centering
\begin{tabular}{l p{3cm} P{0.6cm} P{0.6cm} P{0.6cm}  P{0.5cm} P{0.6cm} P{0.6cm}  P{0.6cm} P{0.6cm} P{0.6cm} }
\toprule
\multicolumn{11}{c}{\texttt{yagoLegalActorGeo}} \\
\midrule
\texttt{Parent Type} & \texttt{Child Type} & \multicolumn{3}{c}{$1 \leq \tau \leq 10$} & \multicolumn{3}{c}{$10 < \tau \leq 50$} & \multicolumn{3}{c}{$50 < \tau \leq 90$}\\
\midrule
& & \textbf{P} & \textbf{R} & \textbf{F1} & \textbf{P} & \textbf{R} & \textbf{F1} & \textbf{P} & \textbf{R} & \textbf{F1}\\
\cmidrule{3-11}

\texttt{owl:Thing} & \texttt{Legal Actor Geo} & 0.48 & 0.36 & 0.41 & 0.51 & 0.43 & 0.47 & 0.53 & 0.47 & 0.50\\[2ex]

\hline

\multirow{2}{2.4cm}{\texttt{Legal Actor Geo}} & \texttt{Legal Actor} & 0.51 & 0.34 & 0.41 & 0.54 & 0.41 & 0.47 & \textbf{0.56} & \textbf{0.45} & \textbf{0.50}\\[1.5ex]
& \texttt{location} & 0.30 & 0.29 & 0.29 & 0.34 & 0.40 & 0.37 & 0.36 & 0.45 & 0.40\\

\hline

\multirow{2}*{\texttt{location}} & \texttt{region} & 0.30 & 0.28 & 0.29 & 0.35 & 0.40 & 0.37 & 0.37 & 0.44 & 0.40\\[1ex]
& \texttt{point} & 0.30 & 0.1 & 0.14 & 0.38 & 0.22 & 0.28 & 0.39 & 0.26 & 0.32\\[2ex]

\texttt{Legal Actor} & \texttt{person} & 0.53 & 0.36 & 0.43 & 0.56 & 0.43 & 0.49 & \textbf{0.58} & \textbf{0.46} & \textbf{0.51}\\

\hline

\multirow{5}*{\texttt{person}} & \texttt{preserver} & 0.63 & 0.31 & 0.42 & 0.67 & 0.46 & 0.54 & \textbf{0.67} & \textbf{0.49} & \textbf{0.57}\\[1ex]
&  \texttt{authority} & 0.53 & 0.20 & 0.29 & 0.62 & 0.24 & 0.35 & 0.65 & 0.33 & 0.44\\[1ex]
&  \texttt{contestant} & 0.59 & 0.43 & 0.50 & 0.62 & 0.52 & 0.57 & 0.64 & 0.56 & 0.60\\[1ex]
&  \texttt{leader} & 0.53 & 0.26 & 0.34 & 0.59 & 0.34 & 0.43 & 0.61 & 0.37 & 0.46\\[1ex]
&  \texttt{wc Living people} & 0.55 & 0.37 & 0.44 & 0.58 & 0.44 & 0.50 & 0.59 & 0.47 & 0.52\\[1ex]
\bottomrule
\end{tabular}
\caption{Results for the \emph{statement classification} for entities of type \texttt{yagoLegalActorGeo}. Results are for the different sample ranges $\tau$ and shown different levels of entity types in the YAGO type hierarchy.}
\label{tbl:yagoLegalActorGeo_t1}
\end{table}

\subsubsection{Level of Type Granularity}

As expected, we observe that model performance depends on the type level (cf. Table~\ref{tbl:yagoLegalActorGeo_t1}). A unified model from heterogeneous training instances performs poorly:  the $SC$ model for the main type \texttt{Legal Actor Geo} achieves a precision P=0.527 with high variance across its subtypes. Comparing the types at depth level 3 (last group of rows), the difference in terms of precision can go as high as 15\% between \texttt{Legal Actor Geo} and the best performing subtype \texttt{preserver}.

At higher depths, the performance often improves significantly as the instances belonging to a given type become more homogeneous. For example, the fine grained type \texttt{wcat Italian footballers} has a precision of P=0.87 and recall of R=0.58, which constitutes a 50\% precision and a 26\% recall improvement over its parent type \texttt{Person}. However, the performance improvement is not monotonically increasing. In some fine-grained types, there is in fact a performance reduction which can be attributed to over-fitting. This suggests that there is indeed a sweet spot in terms of choice of the best performing model for an instance. We observed that the instances that are children of \texttt{person} showed best performances between levels 5 and 8.

Our models perform poorly for types such as \texttt{location} since  location pages have a lower news density. We again observe that news articles are usually centered around people and its instances benefit the most from our approach. We also observe that the performance of our approach is sensitive to the type hierarchy. The choice of YAGO as a taxonomy is due its fine-grained types. However, there exist many long-tail entities that are direct descendants from the higher levels and fail to leverage the homogeneity of fine-grained types. We also perform poorly on such instances. 

In the YAGO taxonomy, the entities are distributed normally with a mean at depth level 8, which contains around 36\% of entities. The long tail with types lower than depth level 8 accounts for 28\% of entities in the YAGO taxonomy.

We focussed on the category \texttt{news} in our discussion and in Table~\ref{tbl:yagoLegalActorGeo_t1}.  Performance of $SC$ models for the categories $c=\{$\texttt{web}, \texttt{book}, \texttt{journal}$\}$ and type \texttt{person} is P=0.62 and R=0.59, P=0.29 and R=0.69, and P=0.25 and R=0.26, respectively. The relatively high score for the \texttt{web} category can be attributed to the high density of statements of category \texttt{web}, accounting for more than 54\% of the total statements. Hence, by always choosing \texttt{web} as the category of a statement we get an average precision of 0.54.

\subsubsection{Convergence and Feature Ablation}

\begin{figure}[h!]
\centering
\includegraphics[width=0.8\columnwidth]{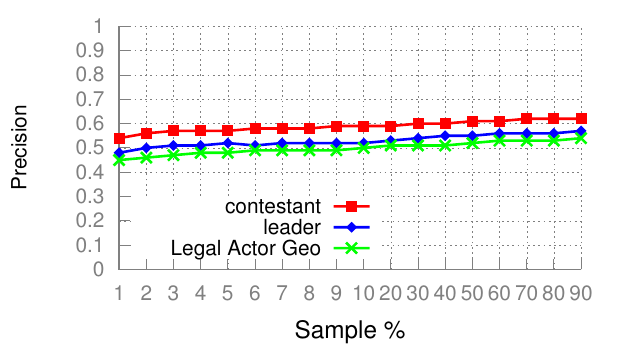}
\caption{{Learning curve for $SC$ measured for different sample sizes.}}
\label{fig:t1_learning_curve}
\end{figure}

\paragraph{Convergence.} We measure the amount of training data required for the models to converge to optimal performance. Figure~\ref{fig:t1_learning_curve} shows the learning curve for some of the types reported in Table~\ref{tbl:yagoLegalActorGeo_t1}. We see that $SC$ models converge and achieve optimal performance early on with a sample around 7\% to 10\%. 

\begin{figure}[h!]
\centering
\includegraphics[width=0.8\columnwidth]{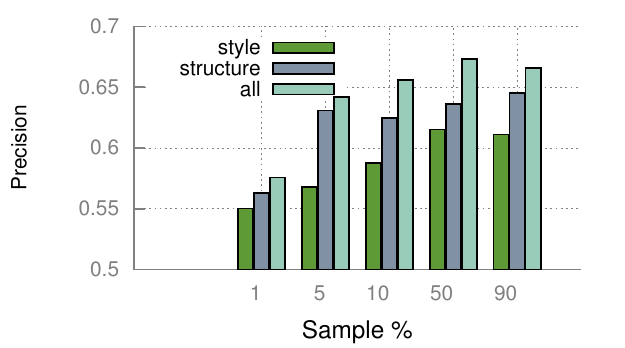}
\caption{{Feature ablation for features in $SC$ for type \texttt{preserver}}}
\label{fig:feature_ablation_yagoLegalActorGeo}
\end{figure}

\paragraph{Ablation.}  We apply a feature ablation test for the different different features groups from Section~\ref{sec:missing_citations}. Figure~\ref{fig:feature_ablation_yagoLegalActorGeo} shows the results for the feature groups \emph{language style}, and \emph{entity structure}. The highest gain is achieved with the feature group \emph{entity structure}, which reveals the challenging nature of the task where language style features cannot be applied alone.

\section{Citation Discovery Evaluation}\label{sec:t2_evaluation}

In this section, we evaluate the \emph{citation discovery} task for \emph{news statements}. We perform an extensive evaluation for approximately 22k news statements and discover citations from a real-word news collection with more than 20 million articles for a timespan of two years.

\subsection{Statement and News Collection}\label{subsec:data_gdelt_news}

To establish a realistic evaluation of the citation discovery step, we consider only a subset of our initial Wikipedia news collection $\N^W$ between the year range of 2013 and 2015. The resulting subset contains 22k news statements with 27k news article citations in $\N^W$\footnote{A statement can have more than one citation.}. We denote this temporal slice of news articles in $\N^W$ by $\N^W_{13-15}$.

The reason for considering only the subset $\N^W_{13-15}$ is because we are in hold of a large news collection, with up to 40k daily news articles, and with more than 20 million news articles between the years 2013 and 2015. The news collection we construct from the GDelt project\footnote{\url{http://gdeltproject.org/}} and denote it with $\N^G$. The merged news collection of $\N^W_{13-15}$ and $\N^G$ we denote with $\N$.

In $\N$, the collection $\N^W_{13-15}$ accounts for less than 1\% of the total news articles, hence, making the task of citation discovery difficult. This represents a realistic scenario, where we are given a large collection of news articles and from the many irrelevant articles we are to find an appropriate citation for a news statement $s$.

\subsection{Evaluation Strategies}\label{subsec:eval_strategy}

Here we describe the two evaluation strategies we employ for evaluating the performance of our models. In this task, we could simply count an news article as relevant for a news statement $s$ if it exists in our ground-truth, that is $n\in N_s$. However, given that we are in hold of a high coverage real-world news collection, there are news articles that still may be relevant for $s$, however, not exist in the ground-truth we extract from Wikipedia. Therefore, we outline two evaluation strategies below.

\paragraph{Evaluation Strategy \texttt{E1}:} In this scenario, we, for each news statement $s$, only consider the pairs $\langle s, n\rangle$, where $n\in N_s$ as correct and all other possible citations as incorrect. This allows for fully automatic evaluation but is only a \emph{lower bound} for $FC$, as there can be additional articles that are relevant for $s$ but do not exist in $N_s$.  We therefore also consider a variant \texttt{E1+FP}, where we consider $n'\notin N_s$ as additional correct citations if the similarity (based on the \emph{Jaccard} similarity) to one of the articles in $N_s$ is above 0.8.

\paragraph{Evaluation Strategy  \texttt{E2}:} \texttt{E2} assesses the true performance of $FC$. In this case, apart from already existing citations for $s$ from $N_s$, we assess through \emph{crowd-sourcing} the appropriateness as citations of articles $n\in \N \wedge n \notin N_s$.

We set up the crowd-sourcing experiment for \texttt{E2} as follows. For a statement $s$ and an article $n_i\notin N_s$ marked as correct by $FC$, we ask the crowd to compare $n_i$ with the ground-truth article $n \in N_s$ and answer the question \emph{`Which   of the two shown news articles is an appropriate citation for the   statement?'}. The workers are shown $s$ as well as $n_i$ and the  ground truth article in random order without an indication which one is the ground truth.  We provide the following response options: (i) \emph{first}, (ii) \emph{second}, (iii) \emph{both}, (iv) \emph{none}, and (v) \emph{insufficient info}. We deployed the experiment in CrowdFlower\footnote{\small{\url{https://www.crowdflower.com}}} and chose only high quality workers to ensure the reliability of our experiments\footnote{We select workers with the highest quality as   provided by the CrowdFlower platform.}. Furthermore, we removed workers who did not spend the minimum amount of two minutes to assess the appropriateness of a citation\footnote{The amount of two minutes was decided based on the number of citations the workers had to assess per page (consisting of 5 citations to assess).}.

\begin{figure}[h!]
	\centering
	\includegraphics[width=1.0\textwidth]{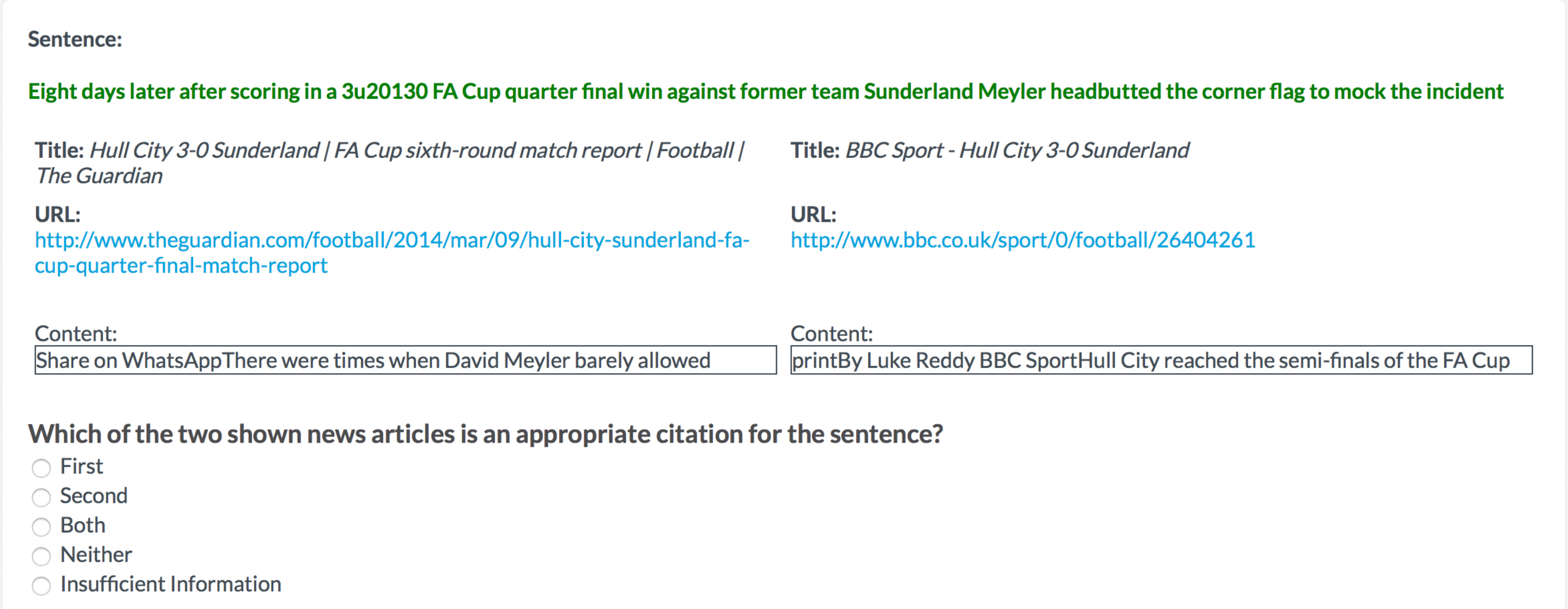}
	\caption{Evaluation interface in CrowdFlower shown to crowd-workers for assessing the relevance of news articles for a news statement $s$.}
	\label{fig:e2_job}
\end{figure}

We collect three judgments per question. We count citations as correct which are ground-truth articles or articles which the majority of workers judge as appropriate citations. Figure~\ref{fig:e2_job} shows the visual interface which we display to crowd-workers to assess the relevance of news citation for statement $s$.

\subsection{Experimental Setup}\label{subsec:t2_exp_setup}

\paragraph{Retrieval model.} We use the retrieval model in \cite{Amati:2002:PMI:582415.582416}, specifically the implementation provided   by Solr\footnote{\small{\url{http://lucene.apache.org/solr/}}}. We   use the top-100 retrieved news articles for a statement as candidate citations, from which we perform feature extraction and learn our SC models.

\paragraph{Learning Setup.} We learn classifiers specific to entity types for a total of 83 types. We limit ourselves to types that have news statements in the date range 2013-2015 and with at least 100 entity instances. From our set of 22k statements, we randomly sample statements from each entity type if they have more than 1000 instances, otherwise we take all statements. Training and testing data consist of the pairs $\langle s, n_i\rangle$, where $s$ is a news statement, and $n_i$ is one of the top--100 citation candidates which we retrieve from $\N$. We split training and testing data per statement $s$, where each $s$ and all its candidates are included \emph{completely} either in the training or test set.

\paragraph{Learning Approach.}  We learn the $FC$ models as supervised binary classification models using random forests RF\cite{Breiman2001}.  We predict $\langle s, n\rangle\in$\emph{`correct', `incorrect'}, i.e. if a candidate news
article is an appropriate citation for $s$ or not. We optimize for the \emph{`correct}' class. The correct labels in training and automatic evaluation $E1$ are all part of $\N^W_{13-15}$, which makes up less than 1\% of our news collection $N$. Therefore, we learn $FC$ as a \emph{cost-sensitive} classifier.

\paragraph{Metrics.} We evaluate the performance of $FC$ models through standard evaluation metrics like precision $P$, recall $R$, and $F1$ score. A candidate news article $n$ is considered to be a relevant or appropriate citation for the news statement $s$, if it exists in our ground-truth, that is $N_s$ or is marked by our human evaluators as relevant.

\paragraph{Baselines.} We consider two baselines (\textbf{B1} and \textbf{B2}) for this task. For $\mathbf{B1}$, we use the \emph{divergence from randomness} model~\cite{Amati:2002:PMI:582415.582416} to retrieve news articles from $\N$ for $s$ and simply  suggest the top--1 article as citation. In $\mathbf{B2}$ we learn a supervised model based on the \emph{IR baseline features} (see Table~\ref{tbl:t2_baseline_results_e1}).

\subsection{Results and Discussion}\label{subsec:t2_results}

Table~\ref{tbl:t2_results_e1} and \ref{tbl:t2_baseline_results_e1} show the results for all evaluation strategies for the citation discovery task.  We only show the detailed results for the top--10 best performing entity types out of the 83 types in our evaluation. Please note here that we have only 83 types, the reason for this difference in contrast to the first task of statement categorization is as a results from the constructed news sub-collection $\N^W_{13-15}$. That is, the statements that have citations to $\N^W_{13-15}$ belong to only 83 types.

The results in each row in Table~\ref{tbl:t2_results_e1} show the best performance we achieve for the individual types, while varying the variables such as the \emph{training sample size} and the  number of features, which we select based on Information-Gain feature selection algorithms. We show results with a maximum of 60\% training sample size.

We report additionally the overall performance of $FC$ models across all 83 types through \emph{micro-average} in the last row in Table~\ref{tbl:t2_results_e1}. 

\subsubsection{E1: Automated Evaluation}\label{subsec:e1_t2_results}

In Table~\ref{tbl:t2_results_e1}, in the third column, we show the evaluation results for the strategy \texttt{E1}. 

Results for \texttt{E1} are encouraging given the fact that in top--100 news candidates retrieved from $\mathcal{N}$ only 1\% of the news are `\emph{correct}' (on average one relevant citation in $\N^W_{13-15}$ per statement). Furthermore, as shown in Figure~\ref{fig:index_coverage} the highest recall we get at top--100 is on average around 45\%.

We achieve the best performance in terms of precision for the entity type \texttt{football} \texttt{player}, with precision P=0.80 and a recall of R=0.30. For F1 the best performing type in this setup is the entity type \texttt{player} with F1=0.57.

Using the evaluation strategy \texttt{E1+FP}, we consider as relevant all \emph{false positive} (FP) articles which are highly similar to the ground-truth articles $N_s$ (above 0.8 similarity). Even though the FP articles do not exist in our ground-truth, the high similarity to the ground-truth article is a  strong indicator for them being relevant citations. Using this strategy, the results improve for some of the types with up to 8\% in terms of precision. For type \texttt{entertainer} we have an increase of 11\%. In absolute numbers, by considering the highly-similar FP articles as relevant we gain an additional 757 news articles out of 12,877, i.e.  an additional 6\% news citations.

\begin{table}[h!]
\centering
\begin{tabular}{p{3cm}  p{0.5cm} p{0.5cm} p{1cm}  l l p{0.3cm} p{0.3cm} }

\toprule
& \multicolumn{3}{c}{\texttt{E1}} & \texttt{E1--FP}  & \texttt{E2}  & & \\
\cmidrule{2-8}
\texttt{type} & \textbf{P} & \textbf{R} & \textbf{F1} & \multicolumn{1}{c}{\textbf{P}} & \multicolumn{1}{c}{\textbf{P}}  & \#f. & \%\\
\cmidrule{1-8}
\texttt{player} & 0.67 & 0.46 & 0.55 & 0.71 $\blacktriangle$ (5.63\%) & \textbf{0.85} $\blacktriangle$ (21.18\%) & 20 & 60\\
\texttt{entertainer} & 0.70 & 0.33 & 0.45 & 0.78 $\blacktriangle$ (10.26\%) & \textbf{0.90} $\blacktriangle$ (22.22\%) & 40 & 60\\
\texttt{causal agent} & 0.73 & 0.28 & 0.41 & 0.77 $\blacktriangle$ (5.19\%) & \textbf{0.88} $\blacktriangle$ (17.05\%) & 40 & 60\\
\texttt{location} & 0.55 & 0.26 & 0.35 & 0.62 $\blacktriangle$ (11.29\%) & \textbf{0.83} $\blacktriangle$ (33.73\%) & 30 & 60\\
\texttt{artist}  & 0.67 & 0.21 & 0.32 & 0.67 & \textbf{0.85} $\blacktriangle$ (21.18\%) & 50 & 60\\
\texttt{football player} & 0.80 & 0.30 & 0.43 & 0.80 & \textbf{0.90} $\blacktriangle$ (11.11\%) & 50 & 60\\
\texttt{wcat Living people} & 0.67 & 0.23 & 0.34 & 0.70 $\blacktriangle$ (4.29\%) & \textbf{0.85} $\blacktriangle$ (21.18\%) & 50 & 50\\
\texttt{creator} & 0.74 & 0.25 & 0.38 & 0.74 & \textbf{0.91} $\blacktriangle$ (18.68\%) & 50 & 50\\
\texttt{organism} & 0.69 & 0.30 & 0.41 & 0.70 $\blacktriangle$ (1.43\%) & \textbf{0.83} $\blacktriangle$ (16.87\%) & 40 & 60\\
\texttt{person} & 0.64 & 0.35 & 0.46 & 0.66 $\blacktriangle$ (3.03\%) & \textbf{0.85} $\blacktriangle$ (24.71\%) & 20 & 60\\
\midrule
\emph{micro-average} & 0.67 & & \multicolumn{1}{c}{} & \multicolumn{1}{l}{0.71 $\blacktriangle$ (5.6\%)} & \multicolumn{1}{l}{\textbf{0.86 $\blacktriangle$ (22.00\%)}}\\
\bottomrule
\end{tabular}
\caption{Top--10 best performing entity types for the $FC$ task. \texttt{E1+FP} and \texttt{E2} columns show the improvement for P over \texttt{E1}. Right most column shows the configuration (\#f represents the number of top--$k$ most important features, and \% is the percentage of training data) for the $FC$ models. The last row shows the \emph{micro-average} precision across all $FC$ models.} 
\label{tbl:t2_results_e1}
\end{table}

\begin{table}[h!]
\centering
\begin{tabular}{l l l l l l l}

\toprule
& \multicolumn{3}{c}{\texttt{B1}} &  \multicolumn{3}{c}{\texttt{B2}}\\
\cmidrule{2-7}
\texttt{type} & \textbf{P} & \textbf{R} & \textbf{F1} & \textbf{P} & \textbf{R} & \textbf{F1}\\
\cmidrule{1-7}
\texttt{player} & 0.37 & 0.36 & 0.37 & 0.31 & 0.28 & 0.29\\
\texttt{entertainer} & 0.32 & 0.31 & 0.31 & 0.16 & 0.18 & 0.17\\
\texttt{causal agent} & 0.26 & 0.26 & 0.26 & 0.17 & 0.21 & 0.19\\
\texttt{location} & 0.21 & 0.19 & 0.20 & 0.21 & 0.23 & 0.22\\
\texttt{artist} & 0.31 & 0.31 & 0.31 & 0.24 & 0.27 & 0.25\\
\texttt{football player} & 0.31 & 0.31 & 0.31 & 0.29 & 0.38 & 0.33\\
\texttt{wcat Living people} & 0.27 & 0.26 & 0.26 & 0.21 & 0.18 & 0.2\\
\texttt{creator} & 0.34 & 0.32 & 0.33 & 0.25 & 0.24 & 0.24\\
\texttt{organism} & 0.29 & 0.28 & 0.28 & 0.22 & 0.19 & 0.2\\
\texttt{person} & 0.26 & 0.24 & 0.25 & 0.21 & 0.23 & 0.22\\
\midrule
\emph{micro-average} & 0.25 & & & 0.21 & & \\
\bottomrule
\end{tabular}
\caption{The results for the two baselines $\mathbf{B1}$ and $\mathbf{B2}$ for the top--10 best performing entity types for the $FC$ task. The last row shows the \emph{micro-average} precision across all $FC$ models.} 
\label{tbl:t2_baseline_results_e1}
\end{table}

Baselines \textbf{B1} and \textbf{B2} in Table~\ref{tbl:t2_baseline_results_e1} show the difficulty of the citation discovery task. In particular, we show that standard IR models struggle with this task. Choosing only the top--1 article for citation (\textbf{B1}) achieves only up to P=0.37. On the other hand, for \textbf{B2}, we see that we cannot learn well using only the IR baseline features, and perform even worse than using \textbf{B1}.

\subsubsection{E2: Automated+Crowdsourced Evaluation}\label{subsec:e2_t2_results}

For \texttt{E2}, we report results after re-evaluating performance of $FC$ models via gathering judgements for false positive (FP) news articles suggested as citations for $s$. We evaluate 11,803 false positive news article citation candidates for the top--10 entity types in Table~\ref{tbl:t2_results_e1}, from 6.9k news statements. As reported above, crowd-workers could choose between both ground truth and our suggestion being correct, one of them or neither. The inter-rater agreement between workers was 64\%. Table~\ref{tbl:e2_eval} shows how these false positives were assessed.

\begin{table}[h]
  \centering
  \begin{tabular}{l|r}
  \toprule
    both & 4,506 (38.2\%) \\
    ground truth only & 3,768 (31.9\%)\\
    our suggestion only & 2,287 (19.4\%)\\
    neither & 1,242 (10.5\%)\\
    \midrule
    all & 11,803 (100\%)\\
    \bottomrule
  \end{tabular}
  \caption{Relevant citation distribution for  \texttt{E2}.}
  \label{tbl:e2_eval}
\end{table}

We see that in many cases our suggestion was equal to (38.2\%) or even preferred (19.4\%) over the ground-truth suggestion. Hence, our method can even improve citation quality in Wikipedia.  

In the \texttt{E2} column in Table~\ref{tbl:t2_results_e1} we show the updated results for $FC$ after collecting judgments for false positive news articles. We see that for most of the types we have an average gain of 18\% in terms of precision. We achieve the biggest gain of 28\% for the entity type \texttt{location}. For the types \texttt{football player, creator, entertainer}, we can suggest news citations with 90-91\% precision. Please note that we do not report the recall score for \texttt{E2}, since assessing the appropriateness of  every article in $\N$ as a citation for $s$ is not feasible. The recall score is only reported w.r.t the ground-truth articles in $\N^W_{13-15}$.

\section{Pipeline Evaluation}\label{sec:evaluation_pipe}

For the evaluation of both tasks in a pipeline scenario, we randomly sample 1000 statements from all categories and ran the process of citation discovery through both steps. Each statement is associated with multiple entity types, as they are extracted from $e$ where $T(e)$ is a set of types. For the \emph{statement categorization} task we perform the evaluation based on our ground-truth; for the \emph{citation discovery} we evaluate the suggested citations as in evaluation strategy \texttt{E2}. Note, that here in the evaluation pair we have a news article (that we suggest) and a resource that can be of any type including \emph{book, web, journal}.

\paragraph{Statement Categorization.}  We set up \emph{statement categorization} as a \emph{majority voting} categorization. For each statement and the type specific classifiers $SC$ we predict the category and pick the category that has the majority of votes. In contrast to the \emph{statement categorization} in Section~\ref{sec:statement_classification}, where the original task aimed at showing for which types this task can be performed accurately, we now aim to set up citation discovery in an automated manner.

Based on the ground-truth, 340 out of the 1000 statements were \emph{news statements}. We categorize 368 as news statements, out of which 263 are correct, i.e. P=0.72 and R=0.77. It is interesting to see that we can leverage additional information through \emph{majority   voting}, where for the same statement and its associated types we can predict with high accuracy the citation category label of $s$.

\paragraph{Citation Discovery.} For the \emph{citation discovery} task we ran it based on the generic $FC$ model trained on statements belonging to all types, namely \texttt{owl:Thing}. We could use the type
specific $FC$, with additional costs for computing type specific features.

In the second task, from the 368 statements classified as news statements, we ran the citation discovery model $FC$. We are able to suggest 549 news citations for 78 statements. Based on crowd-sourcing evaluation, we suggest 346 relevant citations, i.e.  a precision of P=0.63, out of which 200 citations are citations that were preferred over existing ones in the ground-truth.  For 146 cases the citations we suggest are considered to be equally appropriate as the existing ones in the ground-truth, for 116 citations the ground-truth ones were preferred over the ones we suggested. Note that our $FC$ models suggest citations for $s$ only in case they fulfill the criteria in Section~\ref{sec:missing_citations}, thus, enforcing high accuracy.

\section{Conclusion} \label{sec:conclusion}

In this chapter, we define and attempt to solve the automatic news citation discovery problem for Wikipedia. We define two tasks -- \emph{sentence categorization} and the \emph{citation discovery} -- towards finding the correct news citation for a given Wikipedia statement. For the sentence categorization task, we learn a multi-class classifier to predict if a statement requires a news statement. For the news citation discovery problem, we first find the likely candidates by a retrieval model over a real-world news collection followed by a binary classification for the top-ranked candidates.

We find that statement categorization is a hard problem due to lack of context for the NLP-based features to perform well. However, the Wikipedia page and its type structure provide important cues towards accurate classification. On the other hand, we perform well on the citation discovery task with 67\% precision (for top-categories) using the automated evaluation, which further improves to over 80\% when crowd-sourced. This shows that we not only identify the correct ground truth articles present in Wikipedia, but in some cases our suggestions are a better fit compared to the sources in Wikipedia.

Finally, the contributions in this chapter, are highly important in upholding the core principles of Wikipedia editing policies. Through our proposed approach, we are able to provide citations to statements and as shown in our experimental evaluation, even improve the existing citations further with more authoritative and up-to-date information.

\clearemptydoublepage
\chapter{Fine Grained Citation Span for References in Wikipedia}\label{ch5:cite_span}

Citations uphold the c
Citations uphold the crucial policy of \emph{verifiability} in Wikipedia. This policy requires Wikipedia contributors to support their additions with citations from authoritative external sources (web, news, journal etc.). 
In particular, it states that \emph{``articles should be based on reliable, third-party, published sources with a reputation for fact-checking and accuracy''\footnote{\url{https://en.wikipedia.org/wiki/Wikipedia:Identifying_reliable_sources}}}. 

Not only are citations essential in maintaining reliability, neutrality and authoritative assessment of content in such a collaboratively edited platform; but lack of citations are essential signals for core editors for unreliability checks.

\begin{figure}[ht!]
	\centering
\includegraphics[width=0.8\columnwidth]{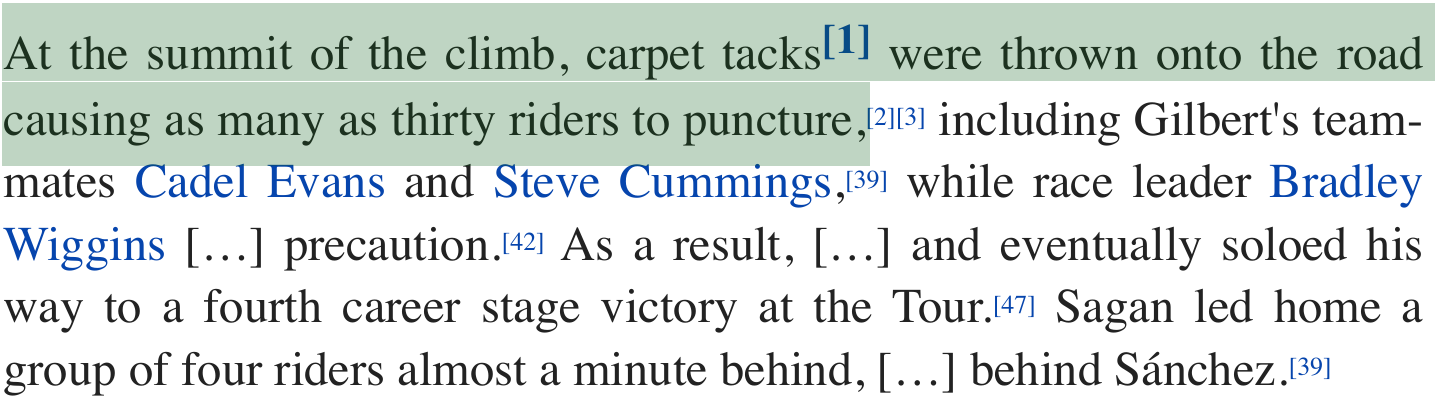}
	\caption{Sub-sentence level span for citation $^{\mathbf{[1]}}$ in a citing paragraph in a Wikipedia article.}
	\label{fig:span_example}
\end{figure}

However, there are two problems when it comes to citing
facts in Wikipedia.  First, there is a long tail of Wikipedia pages
where citations are missing and hence facts might be unverified.
Second, citations might have different span \emph{granularities},
i.e., the text encoding the fact(s), for which a citation is intended,
might span less than a sentence (see Figure~\ref{fig:span_example}) to
multiple sentences. We denote the different \emph{pieces of text}
which contain a citation marker as \emph{fact statements} or simply
\emph{statements}.  For example, Table~\ref{tbl:sentence_citation}
shows different \emph{statements} for several citations.  The
aim of this work is to automatically and accurately determine
\emph{citation spans} in order to improve
coverage~\cite{DBLP:conf/cikm/FetahuMA15,DBLP:conf/cikm/FetahuMNA16}
and to assist editors in  verifying citation quality at a
fine-grained level.

Earlier work on span determination is mostly concerned with scientific
texts~\cite{DBLP:journals/ipm/OConnor82,DBLP:journals/jip/KaplanTT16},
operates at sentence level and exploits explicit
authoring cues specific to scientific text. Although Wikipedia has
well formed text, it does not follow explicit scientific guidelines
for placing citations. Moreover, most statements can only be inferred
from the citation text.

In this chapter, we propose a novel approach for determining the citation span for references in Wikipedia. We start from a paragraph that contains either a citation, and operate at the sub-sentence level, that is, we determine the span of citation at the sub-sentence level. 

We loosely refer to the sub-sentences as text fragments, and consider a sequence prediction approach using a \emph{linear-chain CRF}~\cite{DBLP:conf/icml/LaffertyMP01}.  We limit to citations referring to \emph{web} and \emph{news} sources, as they are accessible online and present the most prominent sources in Wikipedia~\cite{DBLP:conf/websci/FetahuAA15}. By using recent work on moving window language models~\cite{Taneva:2013:GEM:2505515.2505715} and other aspects that take into account the paragraph structure which has a citation, we incrementally attempt to classify continuous text fragments as text that belong to a given citation. We ensure the coherence and accuracy of the selected fragments at all times and for all citation span cases as shown in Table~\ref{tbl:sentence_citation}.

\begin{table}[ht!]
\begin{tabular}{l p{10cm}}
\toprule
\texttt{sub sentence} & {Obama was born on August 4, $1961^{[c_1]}$, at Kapi'olani Maternity $\cdots$ Honolulu$^{[c_2]}$; he is the first $\cdots$ been born in Hawaii.$^{[c_3]}$}. \\[1.5ex]

\texttt{sentence} & {He was reelected to the Illinois Senate in 1998, $\cdots$ in 2002.$^{[c_1]}$} \\[1.5ex]

\texttt{multi sentence} & {On May 25, 2011, Obama $\cdots$ to address $\cdots$ UK Parliament in Westminster Hall, London. This was $\cdots$ Charles de Gaulle $\cdots$ and Pope Benedict XVI.$^{[c_1]}$} \\
\bottomrule
\end{tabular}
\caption{Varying degrees of citation span granularity in Wikipedia text.}
\label{tbl:sentence_citation}
\end{table}

By determining the citation span, thus, coupling together with the approach on citation recommendation in the previous chapter, we close a cycle of providing citations for Wikipedia, and consequentially determining at a fine-grained level the validity of such a citation for a given paragraph in Wikipedia.

\section{Problem Definition and Terminology}\label{sec:problem_definition}

In this section, we describe the terminology and define the problem of determining the \emph{citation span} in text in Wikipedia articles.

\paragraph{Terminology.} We consider Wikipedia articles $W=\{e_1,\ldots, e_n\}$ from a Wikipedia snapshot. We distinguish \emph{citations} to \emph{external references} in text and denote them with $\langle p_k, c_i\rangle$, where $c_i$ represents a citation which occurs in paragraph $p_k$ with positional index $k$ in an entity $e\in W$. We will refer to $p_k$ as the \emph{citing paragraph}. Furthermore, with \emph{citing sentence} we refer to the sentence in $s \in p_k$, which contains $c_i$.  Note that $p_k$ can have more than one citation as shown in Table~\ref{tbl:sentence_citation}.


\paragraph{Problem Definition.} The task of determining the \emph{citation span} for a citation $c$ and a paragraph $p$, respectively $\langle p, c\rangle$ (or simply $p_c$), is subject to the citing paragraph and the citation content. In particular, we refer with \emph{citation span} to the \emph{textual fragments} from $p$ which are covered by $c$. The fragments correspond to the sequence of \emph{sub-sentences} $\mathcal{S}(p)=\langle \delta_1^1, \delta_1^2,\ldots,\delta_1^k,\ldots,\delta_n^m\rangle$. We obtain the sequence of sub-sentences from $p$ by splitting the sentences into sub-sentences or text fragments based on the following punctuation delimiters ($\{,!;:?\}$). These delimitors do not always provide  a perfect semantic segmentation of sentences into facts. A more involved approach could be taken akin to work in text summarization, such as  Zhou and Hovy~\cite{zhou2006summarization} or \cite{DBLP:journals/tslp/NenkovaPM07} who  consider \emph{summary units} for a similar purpose.

Formally we define the \emph{citation span} in Equation~\ref{eq:cite_span} as the function of finding the subset $\mathcal{S}'\subseteq\mathcal{S}$ where the fragments in $\mathcal{S}'$ are covered by $c$.

\begin{equation}\label{eq:cite_span}
\varphi(p, c)\rightarrow \mathcal{S}'\subseteq\mathcal{S}, \;\; s.t.\;\; \delta \in \mathcal{S}' \wedge c \vdash \delta
\end{equation}
where $c \vdash \delta$ states that $\delta$ is covered in $c$.

We note here the subtle difference between a statement as defined in the previous chapter, where we defined it to be as the \emph{inter-citation} text. That is, a statement either started at the beginning of a paragraph or at the end of a sentence which has a citation. We use such definition as one of our baselines for comparison, but as we will see later in the experimental evaluation, this proves to be to coarse grained and accounts for large amount of erroneous spans for a citation.

\section{Citation Span Approach}\label{sec:approach}

We approach the problem of citation span detection in Wikipedia as a
\emph{sequence classification} problem. For a citation $c$ and a
citing paragraph $p$, we chunk the paragraph into textual fragments at
the \emph{sub-sentence} granularity, shown in
Equation~\ref{eq:cite_span}.

\begin{figure}[ht!]
\centering
\includegraphics[width=0.8\columnwidth]{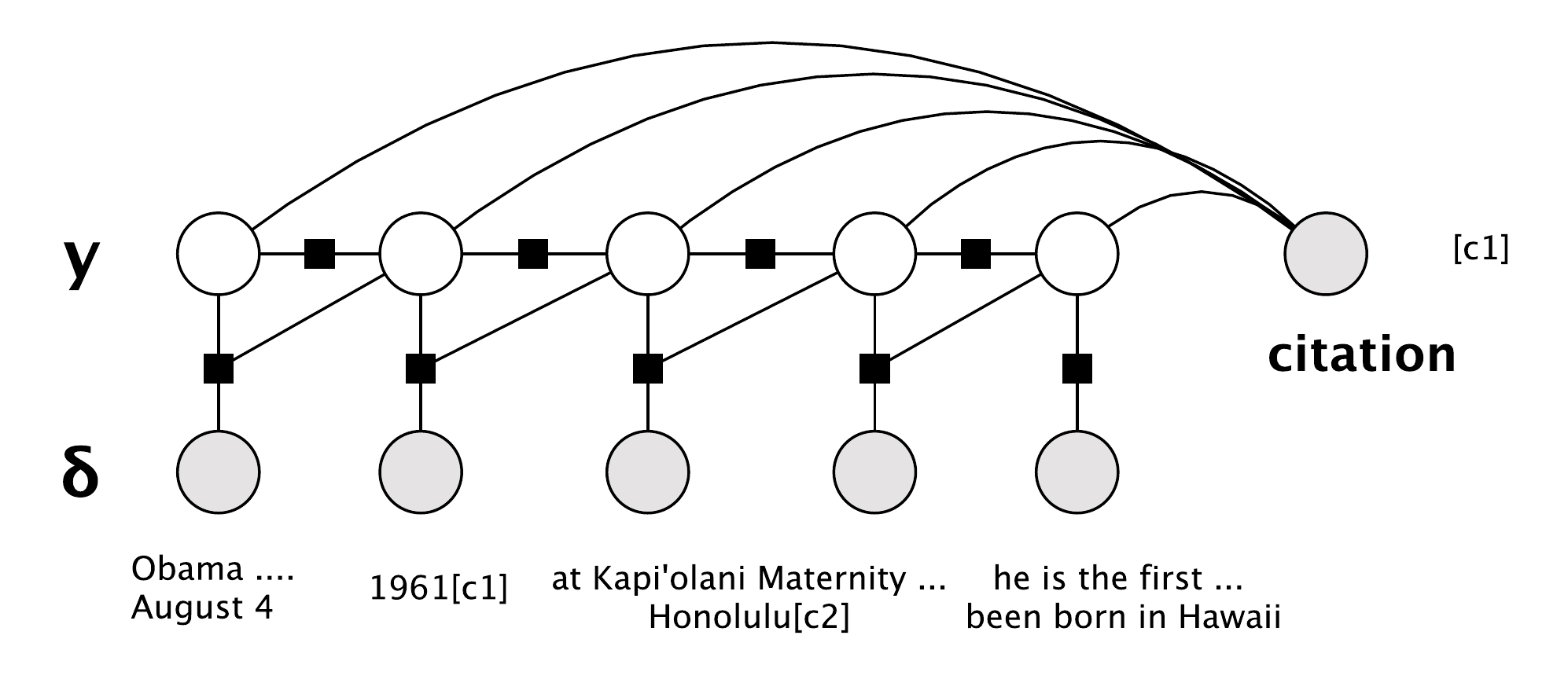}
\caption{Linear chain CRF representing the sequence of text fragments in a paragraph. In the factors we encode the fitness to the given citation.}
\label{fig:crf_chain}
\end{figure}

Figure~\ref{fig:crf_chain} shows an overview of the sequence
classification of textual fragments. We use a \emph{linear chain
  CRF}~\cite{DBLP:conf/icml/LaffertyMP01}, where for any fragment
$\delta$ we predict the label corresponding to a random variable
$\textbf{y}$ which is either \emph{`covered'} or
\emph{`not-covered'}. We opt for CRFs since we can encode global
dependencies between the text fragments and the actual citation, thus,
ensuring the coherence and accuracy of the predicted labels.

In the following we describe the features we compute for the factors $\Psi(y_i,y_{i-1}, \delta_i)$ for a sequence $\delta_i$ w.r.t the citation $c$. We determine the fitness of $\delta_i$ holding true or being covered by $c$. We denote with $f_k$ the features for the factors $\Psi_i(y_{i}, y_{i-1}, \delta_i)$ for sequence $\delta_i$ for the linear-chain CRF in Figure~\ref{fig:crf_chain}. 

\subsection{Structural Features}\label{subsec:struct_factors}

An important aspect to consider for the citation the span is the structure of the citing paragraph, and correspondingly its sentences. For a textual fragment $\delta$, we extract the following structural features shown in Table~\ref{tbl:struct_factors}. 

\begin{table}[ht!]
\centering
\begin{tabular}{l p{12.5cm}}
\toprule
factor & description\\
\midrule
$f_i^{c'}$ &  presence of other citations in $\delta_i$ where $c'\neq c$\\
$f^{\#s}$ & the number of sentences in $p$\\
$f_i^{|\delta_i|}$ & the length in terms of characters of the sub-sequence\\
$f_i^{s}$ &  check if $\delta_i$ is in the same sentence as the citation $c$\\
$f_i^{s\neq s'}$   & check if $\delta_i$ is in the same sentence as $\delta_{i-1}$\\
$f_i^{c}$ &  the distance of sequence $\delta_i$ to the sequence which contains citation $c$\\
\bottomrule
\end{tabular}
\caption{Structural features for a sequence $\delta_i$.}
\label{tbl:struct_factors}
\end{table}

From the features in Table~\ref{tbl:struct_factors} we highlight $f_i^{c}$ which specifies the distance of $\delta$ to the sequence that cites $c$. The closer a sequence is to the citation the higher the likelihood of it being covered in $c$. In Wikipedia, depending on the citation and the paragraph length, the validity of a citation is densely concentrated to its nearby sequences (preceding and succeeding). 

Furthermore, the features $f^{\#s}$ and $f_i^{s}$, respectively the number of sentences together with the feature considering if $\delta$ is in the same sentence as the sequence holding $c$ are strong indicators of predicting accurately the label of $\delta$. That is, it is more likely for a sequence $\delta$ to be covered by the citation if it appears in the same sentence or sentences nearby to the citation marker. 

However, as shown in Table~\ref{tbl:sentence_citation} there are three main citation span groups, and as such relying only on the structure of the citing paragraph does not yield optimal results. Hence, in the next group we consider features that tie the individual sequences in the citing paragraph with citation as shown in Figure~\ref{fig:crf_chain}.

\subsection{Citation Features}\label{subsec:cite_features}

A core indicator whether a textual fragment $\delta$ is covered in $c$ is based on the lexical similarity between $\delta$ and the content in $c$. We gather such evidence by computing two main similarity measures in this case. We compute the features $f_i^{LM}$ and $f_i^{J}$ between $\delta$ and paragraphs in the citation content $c$.

In details, $f_i^{LM}$ corresponds to a moving language window proposed in \cite{Taneva:2013:GEM:2505515.2505715}. In this case, for each word in either a paragraph in the citation $c$ or the sequence $\delta$, we associate a language model $M_{w_i}$ based on its context $\phi(w_i)=\{w_{i-3}$, $w_{i-2}$, $w_{i-1}$, $w_i$,  $w_{i+1}$,  $w_{i+2}$,  $w_{i+3}\}$ with a window of +/- 3 words. The parameters for the model $M_{w_i}$ are estimated as in Equation~\ref{eq:lm_model} for all the words in the context $\phi(w_i)$ and their frequencies denoted with $tf$. With $M_\delta$ and $M_p$ we denote the overall models as estimated in Equation~\ref{eq:lm_model} for the words in the respective fragments.

\begin{equation}\label{eq:lm_model}
P(w|M_{w_i})= \frac{tf_{w,\phi(w_i)}}{\sum_{w'\in \phi(w_i)}tf_{w',\phi(w_i)}}
\end{equation}

Finally, we compute the similarity of each word in $w \in \delta$ against the language model of paragraph $p \in c$ in Equation~\ref{eq:lm_sim_feature}, which corresponds to the Kullback-Leibler divergence score.

\begin{equation}\label{eq:lm_sim_feature}\small
f_i^{LM} = \min\limits_{p \in c}\left[-\sum_{w \in \delta} P(w|M_{\delta}) \log{\frac{P(w|M_{\delta})}{P(w|M_p)}}\right]
\end{equation}

The intuition behind $f_i^{LM}$ is that for the fragments $\delta$ we take into account the word similarity and the similarity in the context they appear in w.r.t a paragraph in a citation. In this way we ensure that the similarity is not by chance but is supported by the context in which the word appears. Finally, another advantage of this model is that we localize the paragraphs in $c$ which provide evidence for $\delta$.

Aa an additional feature we compute $f_i^{J}$ which corresponds to the maximal \emph{jaccard} similarity between $\delta_i$ and paragraphs $p\in c$. 

Finally, as we will show in our experimental evaluation in Section~\ref{sec:setup}, there is a high correlation between the citation span length and the length of a citation content in terms of sentences. Hence, we add as an additional feature $f^c$ the length in terms of the number of sentences for $c$.

\subsection{Discourse Features}\label{subsec:disc_factors}
An indicator which helps on tying sequences in a sentence is the presence of discourse senses. 

The discourse senses and the sense type establish if fragments in a sentence are semantically related. We annotate a sentence with explicit discourse senses based on an approach proposed in \cite{DBLP:conf/acl/PitlerN09}. It relies on the occurrence of discourse cues. The explicit discourse senses belong to one of the following: \emph{temporal, contingency, expansion, comparison}. 

After extracting the discourse sense for a sentence, specifically the discourse cue, based on its position we determine to which fragment it belongs and mark the sequence accordingly with the discourse sense. We denote with $f_i^{disc}$ the discourse feature for the sequence $\delta_i$.

\subsection{Temporal Features}\label{subsec:temp_factors}

An important aspect that we consider here is the temporal difference between two consecutive fragments $\delta_i$ and $\delta_{i-1}$. 
 If there exists a temporal date expression in $\delta_i$ and $\delta_{i-1}$ and they point to different time-points, this presents an indicator on the transitioning between the states $y_i$ and $y_{i-1}$. That is, there is a higher likelihood of changing the state in the sequence $\mathcal{S}$ for the labels $y_i$ and $y_{i-1}$.
 

We compute the temporal feature here $f_i^{\lambda(i,i-1)}$ indicating the difference in \emph{days} between any two temporal expression extracted from $\delta_i$ and $\delta_{i-1}$. We extract the temporal expression through a set of hand-crafted regular expressions. We use the following expressions: (1) \texttt{DD Month YYYY}, (2) \texttt{DD MM YYYY}, (3) \texttt{MM DD YY(YY)}, (4) \texttt{YYYY}, with delimiters (whitespace, `-', `.').

\section{Experimental Setup}\label{sec:setup}

We outline the experimental setup for evaluating the citation span approach and the competitors for this task. The data and the proposed approaches are provided as an appendix of this thesis, and can be accessed at the URL\footnote{\url{http://l3s.de/~fetahu/emnlp17/}}.

\subsection{Dataset}\label{subsec:dataset}

We evaluate the citation span approaches on a random sample of Wikipedia entities (snapshot of 20/11/2016). For the sampling process, we first group entities based on the number of citations (\emph{web} or \emph{news}\footnote{Wikipedia has an internal categorization of citation based on the reference they point to.}) and sample from the specific groups. This is due to the inherent differences in citation span for entities with varying number of citations. 

\begin{figure}[ht!]
\centering
\includegraphics[width=0.8\linewidth]{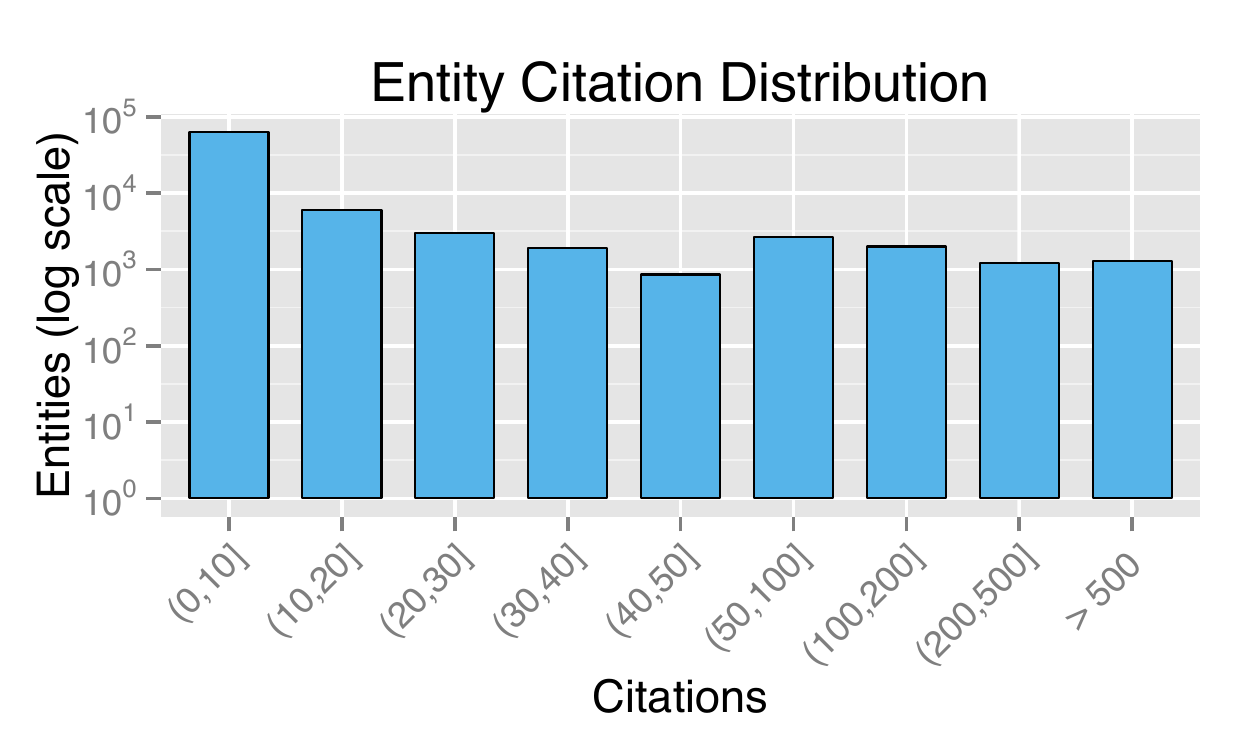}
\caption{Entity distribution based on the number of news citations.}
\label{fig:entity_news_cite_dist}
\end{figure}

For instance, entities with high number of citations tend to have shorter span per citation. Figure~\ref{fig:entity_news_cite_dist} shows the distribution of entities from the different groups. From each sampled entity we extract  the \emph{citing paragraphs} that contain either a \emph{web} or \emph{news} citation. Our sample consists of 500 citing paragraphs from 134 entities.

\subsection{Ground-Truth}\label{subsec:ground_truth}

\paragraph{Setup.} The authors of this work manually determined the citation span of $c$ in paragraph $p$. The reason for this is due to high evaluation efforts which cannot be enforced in crowdsourcing frameworks, thus raising quality concerns.

We set strict guidelines that help us generate reliable ground-truth annotations. We follow two main guidelines: (i) requirement to read and comprehend the content in $c$, and (ii) matching of the textual fragments from $p$ as either being supported \emph{explicitly} or \emph{implied/inferred} in $c$. In the case where the citation is not appropriate for the paragraph we simply skip such cases\footnote{This is the cases when the language of $c$ is not english.}.

Since the task requires reading and comprehending the entire content in $c$ and $p$, it takes on average up to 2.4 minutes to perform the evaluation for a single item. It is worth noting that for a textual fragment in $p$ and for citation $c$ we extract \emph{binary} judgements (`covered`,`not-covered'). 

\paragraph{Citation Span Stats.} Following the definition in Equation~\ref{eq:cite_span} we determine the citation span at the sub-sentence granularity level. Table~\ref{tbl:gt_stats} shows the distribution of citations falling into the specific spans for the citing paragraphs. We note that the majority of citations have a span between half a sentence\footnote{The sub-sentence span is the ratio of sub-sentences covered over the total number of sub-sentences in a sentence.} and up to a sentence, yet, the remainder of more than 20\% of citation span across multiple sentences in such paragraphs.

We define the citation span as the ratio of sub-sentences which are
covered by a given citation over the total number of sub-sentences in
the sentence, consequentially in the citing paragraph. That is, a citation
is considered to have a span of one sentence if it covers all its
sub-sentences.

\begin{equation}\label{eq:cite_span}
span(c,p) = \sum\limits_{s \in p}\frac{\#\delta^{s} \in \mathcal{S'}}{\#\delta^{s}}	
\end{equation}
where $\delta^s$ represents a sequence in sentence $s\in p$, which are part of the the ground-truth.

\begin{table}[h!]
\centering
\begin{tabular}{l l l l l l l}
\toprule
 & total & $\leq .5$ & $(.5,1]$ & $(1,2]$ & $(2,5]$ & $>5$\\
 \midrule
news & 316 & 0.11 & 0.64 & 0.17 & 0.07 & 0.02\\
web & 190 & 0.07 & 0.64 & 0.14 & 0.13 & 0.03\\
\bottomrule
\end{tabular}
\caption{Citation span distribution based on the number of sub-sentences in the citing paragraph. }
\label{tbl:gt_stats}
\end{table}

\begin{figure}[ht!]
\centering
\includegraphics[width=0.8\columnwidth]{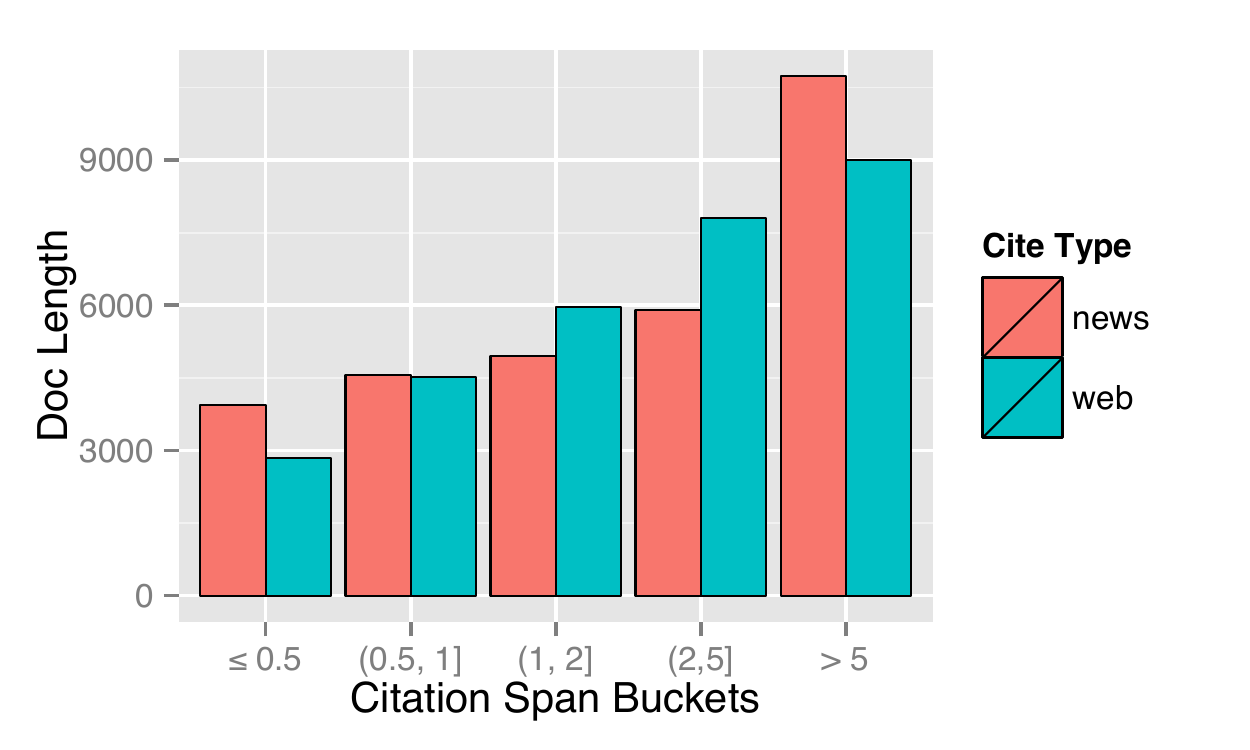}
\caption{Average document length for the different span buckets for citation types \emph{web} and \emph{news}.}
\label{fig:doc_length_cite_span}
\end{figure}

In Figure~\ref{fig:doc_length_cite_span} we analyze a possible factor to the variance in the citation span. It is evident that for longer documents the span increases. This is intuitive since such documents carry more information and consequentially their span in the citing paragraphs can be larger. An example is the Wikipedia article \texttt{2008 US Open (tennis)} which has a citing paragraph with a citation span of 7 sentences for an article of 30k characters long\footnote{\url{http://news.bbc.co.uk/sport1/hi/tennis/7601195.stm}}. We encoded this in the \emph{citation} features $f^c$.

Additionally, within the different citation spans we analyze how many of them contain \emph{skips} for two cases: (i) skip a sequence within a sentence, and (ii) skip sentences in $p$. The results for both cases are presented in Table~\ref{tbl:gt_skips}.
\begin{table}[h!]
\centering
\begin{tabular}{r l l l l}
\toprule
\emph{span} & \multicolumn{2}{c}{\emph{news}} & \multicolumn{2}{c}{\emph{web}}\\
 & \emph{skip} $\delta$ & \emph{skip} $s$ & \emph{skip} $\delta$ & \emph{skip} $s$\\
 \midrule
$\leq 0.5$ & 6\% & - & - & -\\
$(0.5,1]$ & - & - & - & 1\%\\
$(1,2]$ & - & 8\% & - & 19\%\\
$(2,5]$ & 5\% & 18\% & - & 21\%\\
$>5$ & - & 20\% & - & 67\%\\
\bottomrule
\end{tabular}
\caption{The percentage of citations in a span with \emph{sequence skips} and \emph{sentence skips}.}
\label{tbl:gt_skips}
\end{table}
From the results in Table~\ref{tbl:gt_stats} and \ref{tbl:gt_skips} we see that simple heuristics on selecting complete sentences or selecting consecutive sequences do not account for the different citation span cases and skips at the sentence and paragraph level. This leads to suboptimal results and introduces erroneous spans.

\subsection{Baselines}\label{subsec:baselines}

We consider the following baselines as competitors for our citation span approach. 

\paragraph{Inter-Citation Text -- IC.} This baseline was employed in the citation recommendation task in Chapter~\ref{ch1:citation_recommendation}. The span in this case consists of sentences which start either at the beginning of the paragraph or at the end of a sentence that contains a citation. The citation span granularity of this baseline is at the sentence level. 

\paragraph{Citation-Sentence-Window -- CSW.} The span consists of sentences in a window of +/- 2 sentences from the citing sentence~\cite{DBLP:journals/ipm/OConnor82}. The other sentences are included if they contain specific cue words in fixed positions. The cue words are drawn from a fixed vocabulary of 12 words that are specific for scientific literature (e.g. \emph{`above-mentioned', `these', 'therefore', $\ldots$}).

\paragraph{Citing Sentence -- CS.} The span consists of only the \emph{citing sentence}. This presents a strong baseline, which based on the distribution of citation span in Table~\ref{tbl:gt_stats}, around 64\% of citations have a span of up to one sentence. However, as we will see later in our evaluation, the amount of erroneous span is large in the other cases where the span is below or more than a sentence.

\paragraph{Markov Random Fields - MRF.} MRFs~\cite{DBLP:conf/acl/QazvinianR10} model two functions. First, \emph{compatibility}, which measures the similarity of sentences in $p$, and as such allows to extract non-citing sentences. Second, the \emph{potential}, which measures the similarity between sentences in $c$ with sentences in $p$. We use the provided implementation by the authors~\cite{DBLP:conf/acl/QazvinianR10}.

\textbf{Citation Span Plain -- CSPC.} A plain classification setup using the features in Section~\ref{sec:approach}, where the sequences are classified in isolation. We use Random Forests~\cite{Breiman2001} and evaluate them with 5-fold cross validation.

\subsection{Citation Span Approach Setup -- CSPS} For our approach $CSPS$ as mentioned in Section~\ref{sec:approach}, we opt for linear-chain CRFs and use the implementation in \cite{okazaki2007crfsuite}. We evaluate our models using 5-fold cross validation, and learn the optimal parameters for the CRF model through the L-BFGS approach~\cite{liu1989limited}.

\subsection{Evaluation Metrics}\label{subsec:metrics}

We measure the performance of the citation span approaches through the following metrics. We will denote with $W'$ the sampled entities, with $\mathbf{p} =\{p_c, \ldots\}$ ($p_c$ refers to $\langle p, c\rangle$) the set of sampled paragraphs from $e$, and with $|\mathbf{p}|$ the total items from $e$.

\paragraph{Mean Average Precision -- $MAP$.} First, we define \emph{precision} for $p_c$ as the ratio $P(p_c) = {|\mathcal{S}'\cap\mathcal{S}^{t}|}/{|\mathcal{S}'|}$ of fragments present in $\mathcal{S}'\cap \mathcal{S}^t$ over $\mathcal{S}'$. We measure MAP as in Equation~\ref{eq:map}. 

\begin{equation}\label{eq:map}
MAP = \frac{1}{|W'|}\sum_{e\in W'}\frac{\sum_{p_c \in \mathbf{p}}P(p_c)}{|\mathbf{p}|}
\end{equation}

\paragraph{Recall -- $R$.} We measure the recall for $p_c$ as the ratio $\mathcal{S}'\cap\mathcal{S}^t$ over the all fragments in $\mathcal{S}^t$, $R(p_c)=|\mathcal{S}'\cap\mathcal{S}^t|/|\mathcal{S}^t|$. We average the individual recall scores for $e\in W'$ for the corresponding $\mathbf{p}$.

\begin{equation}\label{eq:recall}
R = \frac{1}{|W'|}\sum_{e\in W'}\frac{\sum_{p_c \in \mathbf{p}}R(p_c)}{|\mathbf{p}|}
\end{equation}

\paragraph{Erroneous Span -- $\Delta$.} We measure the amount of extra \emph{words} or extra \emph{sub-sentences} (denoted with $\Delta_w$ and $\Delta_\delta$) added by text fragments that are not part of the ground-truth $\mathcal{S}^t$. The ratio is relative to the number of words or sub-sentences in the ground-truth for $p_c$. We compute $\Delta_w$ and $\Delta_\delta$ in Equation~\ref{eq:extra_words} and~\ref{eq:extra_subs}, respectively.

\begin{equation}\label{eq:extra_words}
\Delta_w = \frac{1}{|W'|}\sum_{e\in W'}\frac{1}{|\mathbf{p}|}\sum_{p_c \in \mathbf{p}}\frac{\sum_{\delta\in \mathcal{S}'\setminus \mathcal{S}^t}words(\delta)}{\sum_{\delta\in \mathcal{S}^t}words(\delta)}
\end{equation}

\begin{equation}\label{eq:extra_subs}
\Delta_\delta = \frac{1}{|W'|}\sum_{e\in W'}\frac{1}{|\mathbf{p}|}\sum_{p_c \in \mathbf{p}}\frac{|\mathcal{S}'\setminus \mathcal{S}^t|}{|\mathcal{S}^t|}
\end{equation}
\section{Results and Discussion}\label{sec:results}

In the section, we describe in details the evaluation results. We focus mainly in two main outcomes. The robustness of our approach on determining the citation span at a fine-grained level, and the amount of error introduced while determining the citation span, that is, the amount of \emph{erroneous span}.

\subsection{Citation Span Robustness}\label{subsec:robustness}
Table~\ref{tbl:cspan_results} shows the results for the different approaches on determining the citation span for all  span cases shown in Table~\ref{tbl:gt_stats}. 

\paragraph{Accuracy.} Not surprisingly, the baseline approaches perform reasonably well. $CS$ which selects only the citing sentence achieves a reasonable $MAP=0.86$ and similar recall. A slightly different baseline $CSW$ achieves comparable scores with $MAP=0.85$. This is due to the inherent span structure in Wikipedia, where a large portion of citations span up to a sentence (see Table~\ref{tbl:gt_stats}). Therefore, in approximately 64\% of the cases the baselines will select the correct span. For the cases where the span is more than a sentence, the drawback of these baselines is in coverage. We show in the next section a detailed decomposition of the results and highlight why even in the simpler cases, a sentence level granularity has its shortcomings due to sequence skips as shown in Table~\ref{tbl:gt_skips}.

Overall, when comparing $CS$ as the best performing baseline against our approach $CSPS$, we achieve an overall score of $MAP=0.83$ (a
slight decrease of 3.6\%), whereas in term of F1 score, we have a decrease of 9\%. The plain-classification approach $CSPC$ achieves similar  score with $MAP=0.86$, whereas in terms of F1 score, we have a decrease of 8\%.  As described above and as we will see later on in
Table~\ref{tbl:cspan_bucket_results}, the overall good performance of the baseline approaches can be  attributed to the citation span distribution in our ground-truth.

On the other hand, an interesting observation is that sophisticated approaches, geared towards scientific domains like $MRF$
perform poorly. We attribute this to  \emph{language style}, i.e., in Wikipedia there are no explicit citation hooks that are present in
scientific articles. Comparing to $CSPS$, we outperform $MRF$ by a large margin with an increase in $MAP$ by $84\%$.

When comparing the sequence classifier $CSPS$ to the plain classifier $CSPC$, we see a marginal difference of 1.3\% for \emph{F1}. However, it will become more evident later that classifying jointly the text fragments for the different span buckets, outperforms the plain classification model.

\begin{table}[h!]
\centering\small
\begin{tabular}{r l l l r r}
\toprule
  & \multicolumn{1}{c}{MAP} & \multicolumn{1}{c}{R} & \multicolumn{1}{c}{F1} & \multicolumn{1}{c}{$\Delta_{w}$} & \multicolumn{1}{c}{$\Delta_{\delta}$}\\
 \midrule
  {MRF}  & 0.45 & 0.78 & 0.56 &  308\% & 278\%\\
  {IC}   & 0.72 & \textbf{0.94} & 0.77 & 113\% & 115\%\\
  {CSW}  & 0.85 & 0.84 & \textbf{0.82} & 38\% & 31\%\\
  {CS}   & \textbf{0.86} & 0.84 & \textbf{0.82} & 35\% & 27\%\\
  {CSPC} & \textbf{0.86} & 0.68 & 0.76 &  \textbf{26}\% & \textbf{23}\%\\
  {CSPS} & 0.83 & 0.69 & 0.75 &  32\% & 24\%\\

 \bottomrule
\end{tabular}
\caption{Evaluation results for the different citation span approaches.}
\label{tbl:cspan_results}
\end{table}

\paragraph{Erroneous Span.} One of the major drawbacks of competing approaches is the granularity at which the span is determined. This leads to erroneous spans. From Table~\ref{tbl:gt_stats} we see that approximately in $\sim$10\% of the cases the span is at sub-sentence level, and in 28\% the span is more than a sentence. 

The best performing baseline $CS$ has an erroneous span of $\Delta_w=35\%$ and $\Delta_\delta=27\%$, in terms of extra words and
sub-sentences, respectively. That is, nearly half of the determined span is erroneous, or in other words it is not covered in the provided
citation. The $MRF$ approach due to its poor $MAP$ score provides the largest erroneous spans with $\Delta_w=308\%$ and $\Delta_\delta=278\%$. The amount of erroneous span is unevenly distributed, that is, in cases where the span is not at the sentence level granularity the amount of erroneous span increases. A detailed analysis is provided in the next section.

Contrary to the baselines, for $CSPS$ and similarly for $CSPC$, we achieve the lowest erroneous spans with $\Delta_w=32\%$ and
$\Delta_\delta=26\%$, and $\Delta_w=24\%$ and $\Delta_w=23\%$, respectively.

Compared to the remaining baselines, we achieve an overall relative decrease of $9\%$ for $\Delta_w(CSPS)$, and $34\%$ for $\Delta_w(CSPC)$, when compared to the best performing baseline $CS$.

From the \emph{skips} in sequences in Table~\ref{tbl:gt_skips} and the unsuitability of sentence granularity for citation spans, we analyze
the locality of erroneous spans w.r.t to the sequence that contains $c$, specifically the distribution of erroneous spans \emph{preceding}
and \emph{succeeding} it. For the $CS$ baseline, $71\%$ of the total erroneous spans are added by sequences preceding the citing sequence,
contrary to $35\%$ which succeed it. In the case of $CSPS$, we have only $9\%$ of erroneous spans (for $\Delta_\delta$) preceding the citation.

\subsection{Citation Spans and Feature Analysis}\label{subsec:cite_struct_feature_ablation}

We now analyze how the approaches perform for the different citation spans in Table~\ref{tbl:gt_stats}\footnote{The models were retrained and tested for the different buckets with 5-fold cross validation.}. Additionally, we analyze how our approach $CSPS$ performs when determining the span without access to the content of $c$.

\begin{table*}[ht!]
	\centering\small
	\begin{tabular}{l p{1cm} p{0.6cm} p{.9cm} p{.6cm} p{.6cm} p{.6cm} p{.6cm} p{.6cm} p{.8cm} p{.6cm} p{.6cm} p{.6cm}}
		\toprule
		& \multicolumn{3}{c}{$\leq 0.5$} & \multicolumn{3}{c}{$(0.5,1]$} & \multicolumn{3}{c}{$(1,2]$} & \multicolumn{3}{c}{$> 2$}	\\[1.5ex]
		& MAP & R & F1 & MAP & R & F1  & MAP & R & F1 & MAP & R & F1 \\
		\midrule
		MRF 	& 0.15 & 0.88 & 0.27 & 0.44 & 0.80 & 0.61 & 0.59 & 0.74 & 0.57 & 0.59 & 0.63 & 0.55\\
		IC 	    & 0.32 & \textbf{1.00} & 0.45 & 0.77 & \textbf{0.99} & 0.83 & 0.73 & \textbf{0.84} & 0.74 & 0.72 & \textbf{0.81} & \textbf{0.73}\\
		CSW 	& 0.38 & \textbf{1.00} & 0.54 & 0.93 & 0.98 & 0.96 & 0.88 & 0.54 & 0.65 & 0.79 & 0.34 & 0.43\\
		CS 	    & 0.40 & \textbf{1.00} & 0.56 & 0.94 & 0.98 & 0.97 & 0.90 & 0.53 & 0.65 & \textbf{0.80} & 0.32 & 0.42 \\[1.5ex]
		
		\midrule 
		
		CSPC 	& 0.85 & 0.53 & 0.65 & \textbf{0.96} & 0.97 & 0.97 & \textbf{0.96} & 0.68 & 0.79 & 0.71 & 0.65 & 0.68\\
		CSPS 	& \textbf{0.87**} & 0.56 & \textbf{0.68**} & \textbf{0.96} & 0.98 & \textbf{0.98} & 0.88 & 0.73 & \textbf{0.80*} & 0.74 & 0.72 & 0.70\\[0.8ex]

		\midrule
		$\Delta_{F1}$ & 
		 & & $\blacktriangle 21\%$ & 
		 & & $0\%$ & 
		 & & $\blacktriangle 8\%$ & 
		 & & $\blacktriangledown 4\%$\\

		\bottomrule
	\end{tabular}
	\caption{Evaluation results for the citation span approaches for the different span cases. For the results of $CSPS$ we compute the relative increase/decrease of $F1$ score compared to the best result (based on $F1$) from the competitors. We mark in bold the best results for the evaluation metrics, and indicate with ** and * the results which are highly significant ($p<0.001$) and significant ($p<0.05$) based on \emph{t-test} statistics when compared to the best performing baselines (CS, IC, CSW, MRF) based on F1 score, respectively.}
	\label{tbl:cspan_bucket_results}
\end{table*}

\paragraph{Citation Spans.}  Table~\ref{tbl:cspan_bucket_results} shows the results for the approaches under comparison for all the citation span cases. In the case where the citation spans up to a sentence, that is $(0.5, 1]$, which presents the simplest citation span case, the baselines perform reasonably well. This is due to the heuristics they apply to determine the span, which in all cases includes the \emph{citing sentence}. In terms of $F1$ score, the baseline $CS$ achieves a highly competitive score of $F1=0.97$. Our approach $CSPS$ in this case has slight increase of 1\% for $F1$ and an increase of 3\% for $MAP$. $CSPC$ achieves a similar performance in this case.

However, for the cases where the span is at the sub-sentence level or
across multiple sentences, the performance of baselines drops
drastically. In the first bucket ($\leq 0.5$) which accounts for 9\%
of ground-truth data, we achieve the highest score with $MAP=0.87$,
though with lower recall than the competitors with $R=0.56$. The
reason for this is that the baselines take complete sentences, thus,
having perfect recall at the cost of accuracy. In terms of $F1$ score
we achieve 21\% better results than the best performing baseline  $CS$.

For the span of $(1,2]$ we maintain an overall high accuracy and recall, and have the highest $F1$ score. The improvement is 8\% in terms of $F1$ score.
Finally, for the last case where the span is more than 2 sentences, we achieve  $MAP=0.74$, a marginal increase of 3\%, however with lower recall, which results in an overall decrease of 4\% for $F1$. The statistical significance tests are indicated with ** and * in Table~\ref{tbl:cspan_bucket_results}.

\begin{figure}[h!]
	\centering
	\includegraphics[width=.8\columnwidth]{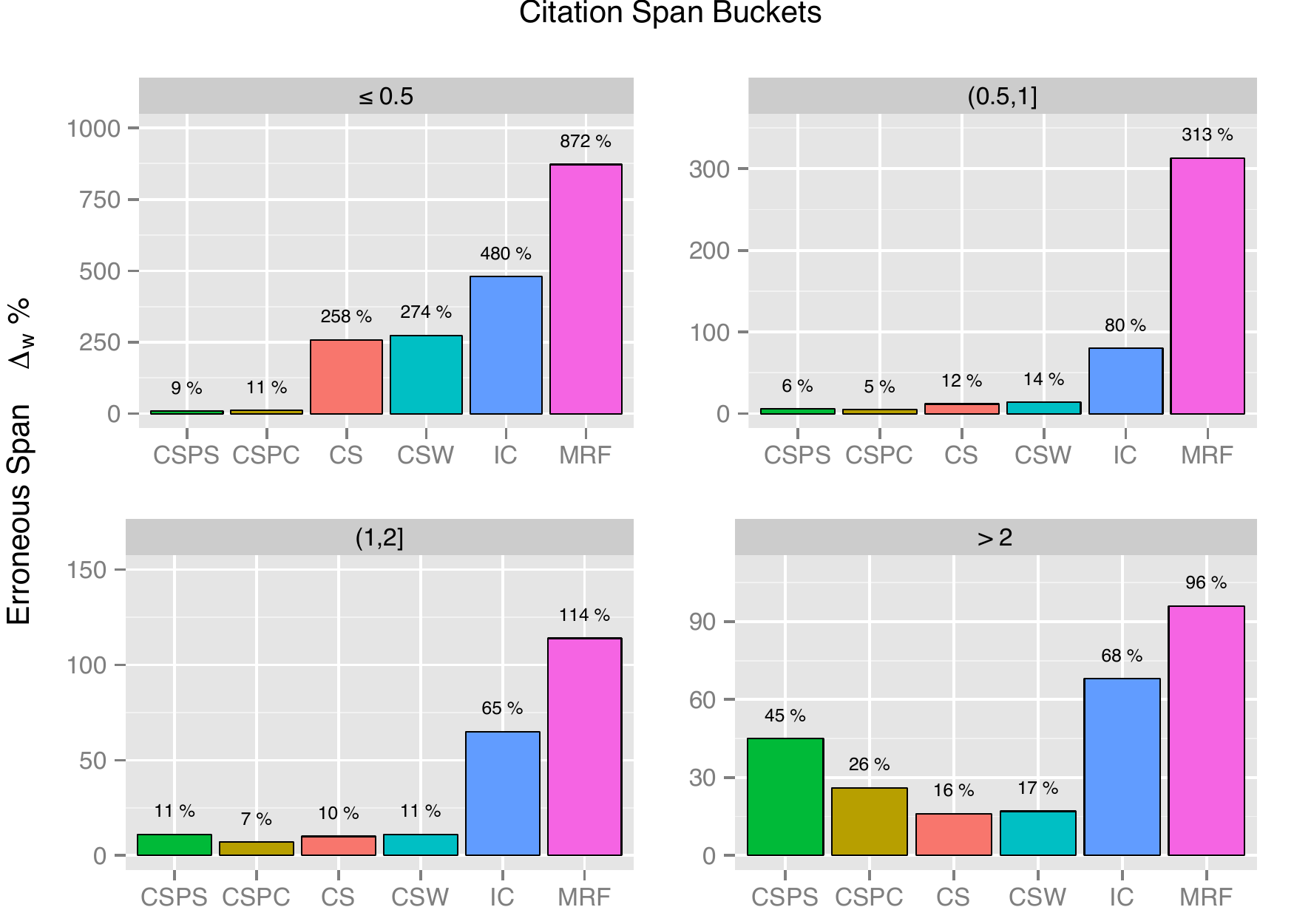}
	\caption{Erroneous span for the different citation span buckets. The y-axis presents the $\Delta_w$ whereas in the  x-axis are shown the different approaches.}
	\label{fig:spanbucket_results}
\end{figure}

\paragraph{Erroneous Span.} Figure~\ref{fig:spanbucket_results} shows the erroneous spans in terms of words for the metric $\Delta_w$ for all citation span cases. It is noteworthy that the amount of error can be well beyond 100\% due to the ratio of the suggested span and the actual span in our ground-truth, which can be higher.

In the first bucket (span of $\leq 0.5$) with granularity less than a sentence, all the competing approaches introduce large erroneous
spans. For $CSPS$ we have a $MAP=0.87$, and consequentially we have the lowest $\Delta_w=9\%$, while for $CSPC$ we have only $\Delta_w=11\%$. In contrast, the non-ML competitors introduce a minimum of $\Delta_w(CS)=182\%$, with MRFs having the highest error.  We also perform well in the bucket $(0.5,1]$. For larger spans, for instance, for $(1,2]$, we are still slightly better, with roughly 3\% less erroneous span when comparing $CSPC$ and $CS$. However, only in the case of spans with $>2$, we perform below the CS baseline. Despite, the smaller erroneous span, the $CS$ baseline never includes more than one sentence, and as such it does not include many erroneous spans for the larger buckets. However, it is by definition unable to recognize any longer spans.

\paragraph{Feature Analysis.} It is worthwhile to investigate the performance gains in determining the citation span without analyzing the content of the citation. The reason for this is that there are several citation categories for which access to the source cannot be easily automated. Models which can determine the span accurately without the actual content have the advantage of generalizing to other citation sources (e.g. \emph{books}) for which the evaluation is more challenging.\footnote{At worst, one needs to read and comprehend
  the entire book to determine if a fragment is covered by the citation.}

Here, we disregard the citation features from
Section~\ref{subsec:cite_features}. In terms of $MAP$, we have a
slight decrease with $MAP=0.82$ when compared to the model with the
citation features. For recall we have a drop of 3\%, resulting in
$R=0.67$. 

This shows that by solely relying on the structure of the citing paragraph and other structural and discourse features we can perform the task with reasonable accuracy.

\section{Citation Span Conclusion}\label{sec:conclusion}

In this work, we tackled the problem of determining the fine-grained citation span of references in Wikipedia. We started from the \emph{citing paragraph} and decomposed it into sequences consisting of sub-sentences. To determine accurately the span we proposed features that leverage the structure of the paragraph, discourse and temporal features, and finally analyzed the similarity between the citing paragraph and the citation content. We limited our approach only to \texttt{web} and \texttt{news} citations, due to the content of such references being accessible online.

We introduce both a standard classifier as well as a sequence classifier using
a linear-chain CRF model.
For evaluation we manually annotated a ground-truth dataset of
509 citing paragraphs. We reported standard evaluation metrics and
also introduced metrics that measure the amount of erroneous span.

We achieved a $MAP=0.86$, in the case of the plain classification model $CSPC$, and with a marginal difference for $CSPS$ with $MAP=0.83$, across all cases with an  erroneous span of $\Delta_w=26\%$ or $\Delta_w=32\%$, depending on the
model. Thus, we provide accurate means on determining the span and at
the same time decrease the erroneous span by 34\% compared to the best
performing baselines. Moreover, we excel at determining citation spans
at the sub-sentence level.

In conclusion, this presents an initial attempt on solving the
citation span for references in Wikipedia. As future work we foresee a
larger ground-truth and more robust approaches which take into account
factors such as a reference being irrelevant to a citing paragraph and
cases where the evidence for a paragraph is implied rather than
explicitly stated in the reference.

Finally, the contribution in this chapter, when coupled together with the approach on citation recommendation in Chapter~\ref{ch1:citation_recommendation} closes the cycle on finding citations for Wikipedia entities, and correspondingly determining accurately the span of such citations. This presents a major step on enforcing the Wikipedia editing policies, and enforcing the \emph{verifiability} principle.

\clearemptydoublepage
\chapter{Automated News Suggestion for Populating Wikipedia Entities}\label{ch3:news_suggestion}

In this chapter, we address the issue of suggesting novel and relevant content coming from news articles to Wikipedia entities. The work in this chapter is partly motivated by our initial study in Chapter~\ref{ch:news_wiki_lag}, where we analyzed the lag between the time facts or general information about an entity is reported in online news media, and the time it is added in Wikipedia. While popular entities and events in Wikipedia are updated instantly~\cite{keegan_hot_2011}, updating and maintaining in a timely manner long-tail entities in Wikipedia is subject to extensive research~\cite{DBLP:conf/acl/SauperB09,DBLP:conf/riao/BalogRTN13,DBLP:conf/sigir/BalogR13,DBLP:conf/eacl/DunietzG14}.

\begin{figure}[t!]
\centering
  \includegraphics[width=0.7\columnwidth]{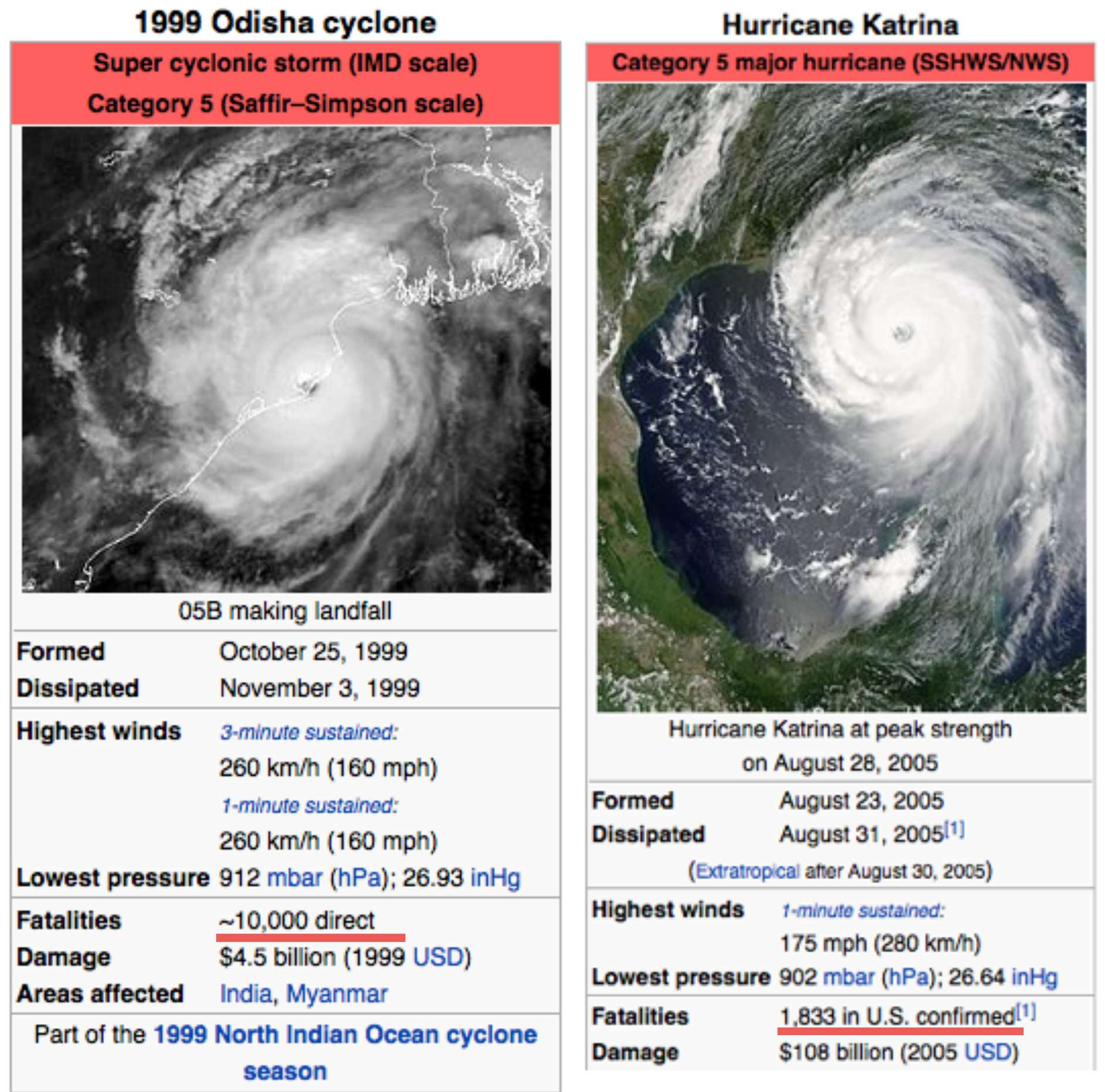}
  \caption{\small{Comparing how cyclones are reported in Wikipedia entity pages.}}
  \label{fig:cyclone}
\end{figure}

In many cases, Wikipedia entity pages are not comprehensive:  relevant information can either be \emph{missing} or added with a \emph{delay}. Consider the city of \emph{New Orleans} and the state of \emph{Odisha} which were severely affected by cyclones \emph{Hurricane Katrina} and \emph{Odisha Cyclone}, respectively. While \emph{Katrina} finds extensive mention in the entity page for \emph{New Orleans}, \emph{Odisha Cyclone} which has 5 times more human casualties (cf. Figure~\ref{fig:cyclone}) is not mentioned in the page for \emph{Odisha}. Arguably \emph{Katrina} and \emph{New Orleans} are more popular entities, but \emph{Odisha Cyclone} was also reported extensively in national and international news outlets. This highlights the lack of important facts in trunk and long-tail entity pages, even in the presence of relevant sources. In addition, previous studies have shown that there is an inherent delay or lag when facts are added to entity pages, Chapter~\ref{ch:news_wiki_lag}.

To remedy these problems, it is important to identify information sources that contain novel and salient facts to a given entity page. However, not all information sources are equal. The online presence of major news outlets is an authoritative source due to active editorial control and their articles are also a timely container of facts. In addition, their use is in line with current Wikipedia editing practice, as is shown in~\cite{DBLP:conf/websci/FetahuAA15} that almost 20\% of current citations in all entity pages are news articles. We therefore propose \emph{news suggestion} as a novel task that enhances entity pages and reduces delay while keeping its pages authoritative. 

Existing efforts to  populate Wikipedia~\cite{DBLP:conf/acl/SauperB09} start from an entity page and then generate candidate documents about this entity using an external search engine (and then post-process them). However, such an approach lacks in (a) reproducibility since rankings vary with time with obvious bias to recent news (b) maintainability since document acquisition for each entity has to be periodically performed. To this effect, our news suggestion considers a news article as input, and determines if it is valuable for Wikipedia. Specifically, given an input news article $n$ and a state of Wikipedia, the news suggestion problem identifies the entities mentioned in $n$ whose entity pages can improve upon suggesting $n$. Most of the works on knowledge base acceleration~\cite{DBLP:conf/riao/BalogRTN13,DBLP:conf/sigir/BalogR13,DBLP:conf/eacl/DunietzG14}, or Wikipedia page generation~\cite{DBLP:conf/acl/SauperB09} rely on high quality input sources which are then utilized to extract textual facts for Wikipedia page population. In this work, we do not suggest snippets or paraphrases but rather entire articles which have a high potential importance for entity pages. These suggested news articles could be consequently used for extraction, summarization or population either manually or automatically -- all of which rely on high quality and relevant input sources. 

We identify four properties of good news recommendations: \emph{salience}, \emph{relative authority}, \emph{novelty} and \emph{placement}. First, we need to identify the most salient entities in a news article. This is done to avoid pollution of entity pages with only marginally related news. Second, we need to determine whether the news is important to the entity as only the most relevant news should be added to a precise reference work. To do this, we compute the \emph{relative authority} of all entities in the news article: we call an entity more authoritative than another if it is more popular or noteworthy in the real world. Entities with very high authority have many news items associated with them and only the most relevant of these should be included in Wikipedia whereas for entities of lower authority the threshold for inclusion of a news article will be lower. Third, a good recommendation should be able to identify \emph{novel} news by minimizing redundancy coming from multiple news articles. Finally, addition of facts is facilitated if the recommendations are fine-grained, i.e., recommendations are made on the section level rather than the page level (\emph{placement}).

\paragraph{Approach and Contributions.} We propose a two-stage news suggestion approach to entity pages. In the first stage, we determine whether a news article should be suggested for an entity, based on the entity's \emph{salience} in the news article, its  \emph{relative authority} and the \emph{novelty} of the article to the entity page. The second stage takes into account the class of the entity for which the news is suggested and constructs \emph{section templates} from entities of the same class. The generation of such templates has the advantage of suggesting and expanding entity pages that do not have a complete
section structure in Wikipedia, explicitly addressing long-tail and trunk entities. Afterwards, based on the constructed template our method determines the best fit for the news article with one of the sections.

We evaluate the proposed approach on a news corpus consisting of 351,982 articles crawled from the \emph{news}  references in Wikipedia from 73,734 entity pages. Given the Wikipedia snapshot at a given year (in our case [2009-2014]), we suggest news articles that might be cited in the coming years. The existing news references in the entity pages along with their reference date act as our ground-truth to evaluate our approach. In summary, we make the following contributions.

\begin{itemize}
	\item  we propose a two-stage news suggestion approach for Wikipedia entity pages.
	\item we adopt and address the problem of determining whether a news article should be referenced to an entity considering the entity \emph{salience}, \emph{relative authority} and \emph{novelty} of the article for the entity page.
	\item we are able to place articles in a specific section of the entity page. Through \emph{section templates}, we address the problems of entities with a limited section structure by
class-based generalization i.e. we can expand entity pages with sections that come from entities of a similar class.
   \item an extensive evaluation on 351,982 news articles and 73,734 entity pages, using  their state for the years [2009-2013].
\end{itemize}

\begin{figure}[ht!]
\centering
  \includegraphics[width=1.0\textwidth]{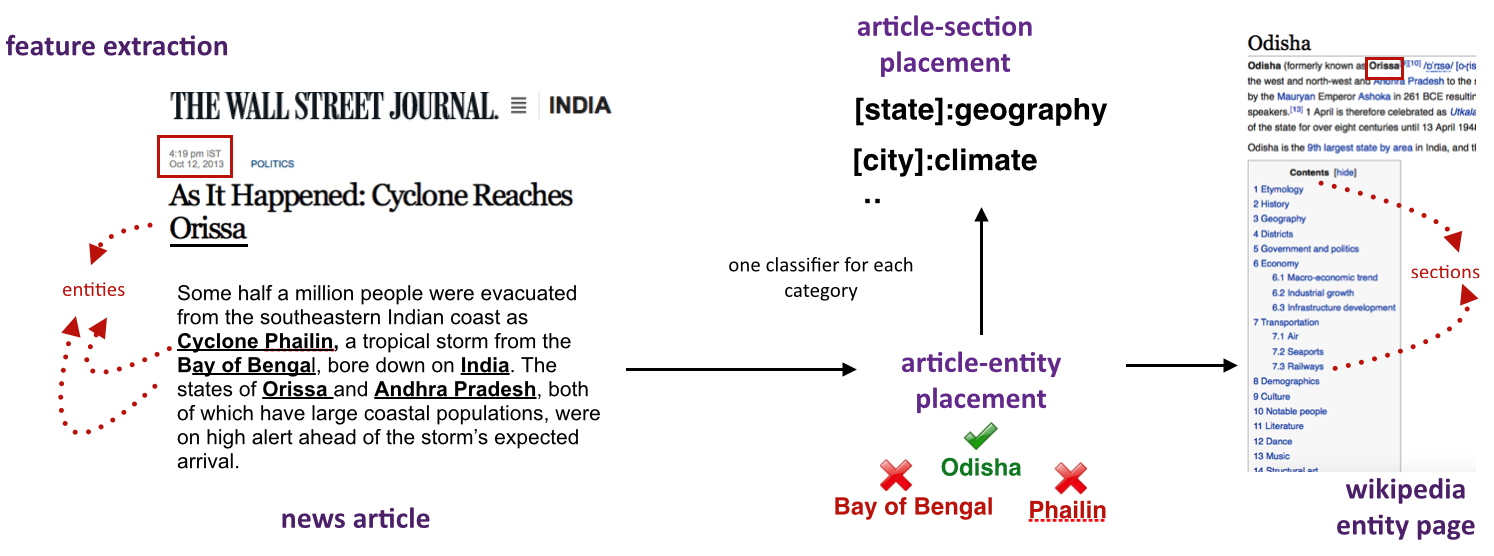}
  \caption{\small{News suggestion approach overview.}}
  \label{fig:approach}
\end{figure}

\section{Problem Definition and Approach Outline}\label{sec:problem_definition}

\subsection{Terminology and Problem Definition}\label{subsec:terminology}

We are interested in named entities $e$ mentioned in news articles. We canonicalize these mentions to entity pages in Wikipedia, a method typically known as \emph{entity linking}. We denote the set of entities extracted and linked from a news article $n$ as $\varphi(n)$. For example, in Figure~\ref{fig:approach}, entities are linked to Wikipedia entity pages (e.g. \emph{Odisha} is linked to the Wikipedia entity\footnote{\url{http://en.wikipedia.org/wiki/Odisha}}). For a collection of news articles $\mathbf{N}$, we further denote the resulting set of entities by $\mathbf{E}=\cup_{n \in \mathbf{N}}{\{e_i\}}$. 

Information in Wikipedia entities is organized into sections and evolves with time as more content is added. We refer to the state of Wikipedia at a time $t$ as $\mathcal{W}_t$ and the set of sections for an entity page $e$ as its \emph{entity profile} $S_e(t)$. Unlike news articles, text in Wikipedia could be explicitly linked to entity pages through anchors. The set of entities explicitly referred in text from section $s \in S_e(t)$ is defined as $\gamma(s)$. Furthermore, Wikipedia induces a category structure over its entities, which is exploited by knowledge bases like YAGO (e.g. \emph{Barack\_Obama} \texttt{isA Person}). Consequently, each entity page belongs to one or more entity categories or classes $c$. Now we can define our news suggestion problem below:

\begin{definition}[News Suggestion Problem] Given a set of news articles $\mathbf{N}=\{n_1,\ldots,n_k\}$ and set of Wikipedia entity pages $\mathbf{E}=\{e_1,\ldots, e_m\}$ (from $\mathcal{W}_t$) we intend to suggest a news article $n$ published at time $t_i > t$ to entity page $e$ and additionally to the most relevant section for the entity page $s \in S_e(t)$.
\end{definition}

\subsection{Approach Overview}

We approach the news suggestion problem by decomposing it into two tasks:

\begin{enumerate}
 \item \emph{AEP}: \emph{Article--Entity} placement
 \item \emph{ASP}: \emph{Article--Section} placement
 \end{enumerate}

In this first step, for a given entity-news pair $\langle n, e\rangle$, we determine whether the given news article $n\in\mathbf{N}$ should be suggested (we will refer to this as \emph{`relevant'}) to entity $e\in \mathbf{E}$. To generate such $\langle n, e\rangle$ pairs, we perform the \emph{entity linking} process, $\varphi(n)$, for $n$. 

The \emph{article--entity} placement task (described in detail in Section~\ref{subsec:article_linking}) for a pair $\langle n, e\rangle$ outputs a binary label (either \emph{`non-relevant'} or \emph{`relevant'}) and is formalized in Equation~\ref{eq:article_entity}.

\begin{equation}\label{eq:article_entity}
AEP: \langle e, n\rangle \rightarrow \{0,1\}, \,\,\,\; \forall e \in \varphi(n) \; \wedge\; n \in \mathbf{N}
\end{equation}

In the second step, we take into account all \emph{`relevant'} pairs $\langle n, e\rangle$ and find the correct \emph{section} for article $n$ in entity $e$, respectively its profile $S_e(t)$ (see Section~\ref{subsec:section_linking}). The \emph{article--section} placement task, determines the correct section for the triple $\langle n, e, S_e(t)\rangle$, and is formalized in Equation~\ref{eq:article_section}.

\begin{equation}\label{eq:article_section}
ASP: \langle e, n, S_e(t) \rangle \rightarrow \{s_1,\ldots, s_{k}\}, \; s \in S_e(t)
\end{equation}

In the subsequent sections we describe in details how we approach the two tasks for suggesting news articles to entity pages.

\section{News Article Suggestion}\label{sec:approach}

In this section, we provide an overview of the \emph{news suggestion} approach to Wikipedia entity pages (see Figure~\ref{fig:approach}). The approach is split into two tasks: (i) \emph{article-entity} (\emph{AEP}) and (ii) \emph{article-section} (\emph{ASP}) placement. For a Wikipedia snapshot $\mathcal{W}_t$ and a news corpus $\mathbf{N}$, we first determine which news articles should be suggested to an entity $e$.  We will denote our approach for \emph{AEP} by $\mathcal{F}_e$. Finally, we determine the most appropriate section for the \emph{ASP} task and we denote our approach with $\mathcal{F}_s$.

In the following, we describe the process of learning the functions $\mathcal{F}_e$ and $\mathcal{F}_s$. We introduce features for the learning process, which encode information regarding the entity \emph{salience}, \emph{relative authority} and \emph{novelty} in the case of AEP task. For the \emph{ASP} task, we measure the \emph{overall fit} of an article to the entity sections, with the entity being an input from \emph{AEP} task. Additionally, considering that the entity profiles $S_e(t)$ are incomplete, in the case of a missing section we suggest and expand the entity profiles based on \emph{section templates} generated from entities of the same class $c$ (see Section~\ref{subsubsec:as_sectiontemplates}).

\subsection{Article--Entity Placement}\label{subsec:article_linking}

In this step we learn the function $\mathcal{F}_e$ to correctly determine whether $n$ should be suggested for $e$, basically a binary classification model (0=\emph{`non-relevant'} and 1=\emph{`relevant'}). Note that we are mainly interested in finding the \emph{relevant} pairs in this task. For every news article, the number of disambiguated entities is around 30 (but $n$ is suggested for only two of them on average). Therefore, the distribution of \emph{`non-relevant'} and \emph{`relevant'} pairs is skewed towards the earlier, and by simply choosing the \emph{`non-relevant'} label we can achieve a high accuracy for $\mathcal{F}_e$.  
Finding the relevant pairs is therefore a considerable challenge. 

An article $n$ is suggested to $e$ by our function $\mathcal{F}_e$ if it fulfills the following properties. The entity $e$ is \emph{salient} in $n$ (a central concept), therefore ensuring that $n$ is about $e$ and that $e$ is important for $n$. Next, given the fact there might be many articles in which $e$ is \emph{salient}, we also look at the reverse property, namely whether $n$ is important for $e$. We do this by comparing the \emph{authority} of $e$ (which is a measure of popularity of an entity, such as its frequency of mention in a whole corpus) with the authority of its co-occurring entities in $\varphi(n)$, leading to a feature we call \emph{relative authority}. The intuition is that for an entity that has overall lower authority than its co-occurring entities, a news article is more easily of importance.\footnote{This is why people occurring infrequently in the news keep any press cutting mentioning them.} Finally, if the article we are about to suggest is already covered in the entity profile $S_e(t)$, we do not wish to suggest \emph{redundant} information, hence the \emph{novelty}.
Hence, the learning objective of $\mathcal{F}_e$ should fulfill the following properties. Table~\ref{tbl:importance_salience} shows a summary of the computed features for $\mathcal{F}_e$.

\begin{enumerate}
	\item \textbf{Salience:} entity $e$ should be a \emph{salient} entity in news article $n$ 
	\item \textbf{Relative Authority:} the set of entities $e' \in \varphi(n)$ with which $e$ co-occurs should have higher \emph{authority} than $e$, making $n$ important for $e$
	\item\textbf{Novelty:} news article $n$ should provide \emph{novel} information for entity $e$ taking into account its profile $S_{e}(t-1)$ 
\end{enumerate} 

\begin{table}[ht!]
\centering
\begin{tabular}{p{2cm} p{8cm} p{2cm}}
\toprule
feature & description\\
\midrule
$\Phi(e,n)$ & the relative frequency of $e$ in news article $n$. & \multirow{2}{2cm}{\emph{salience}}\\
\texttt{Baseline Features} & set of features as proposed by Dunietz and Gillick~\cite{DBLP:conf/eacl/DunietzG14} & \\[3ex]
$\widehat{\Gamma}(e|\varphi(n))$ & relative authority as the score of entities that have higher authority than $e$ and that co-occur in $n$. & \multirow{2}{2cm}{\emph{authority}}\\
$P(D)$ & measures the news domain authority. & \\
$\mathcal{N}(n|e)$ & measures the novelty of a news article $n$ for a given entity $e$ & \emph{novelty}\\
\bottomrule
\end{tabular}
\caption{\emph{Article--Entity} placement feature summary.}
\label{tbl:importance_salience}
\end{table}

\subsubsection{Salience-based features}

\paragraph{Baseline Features.} A variety of features that measure salience of an entity in text are available from the NLP community. We reimplemented the ones in Dunietz and Gillick~\cite{DBLP:conf/eacl/DunietzG14}. This includes a variety of features, e.g. positional features, occurrence frequency and the internal POS structure of the entity and the sentence it occurs in. Table 2 in \cite{DBLP:conf/eacl/DunietzG14} gives details. 

\paragraph{Relative Entity Frequency.} Although frequency of mention and positional features play some role in baseline features, their interaction is not modeled by a single feature nor do the positional features encode more than sentence position. We therefore suggest a novel feature called \emph{relative entity frequency}, $\Phi(e,n)$, that has three properties.: (i) It rewards entities for occurring throughout the text instead of only in some parts of the text, measured by the number of paragraphs it occurs in (ii) it rewards entities that occur more frequently in the opening paragraphs of an article as we model $\Phi(e,n)$ as an \emph{exponential decay} function. The decay corresponds to the positional index of the news paragraph. This is inspired by the news-specific discourse structure that tends to give short summaries of the most important facts and entities in the opening paragraphs. (iii) it compares entity frequency to the frequency of its co-occurring mentions as the weight of an entity appearing in a specific paragraph, normalized by the sum of the frequencies of other entities in $\varphi(n)$. 

\begin{equation}\label{eq:weighted_frequency}
\Phi(e,n) =  \frac{|p(e,n)|}{|p(n)|}\sum\limits_{p\in p(n)}\left(\frac{tf(e,p)}{\sum\limits_{e'}tf(e',p)}\right)^{p}
\end{equation}
where, $p$ represents a news paragraph from $n$, and with $p(n)$ we indicate the set of all paragraphs in $n$. The frequency of $e$ in a paragraph $p$ is denoted by $tf(e,p)$. With $|p(e,n)|$ and $|p(n)|$ we indicate the number of paragraphs in which entity $e$ occurs, and the total number of paragraphs, respectively. 

\subsubsection{Authority-based features}

\paragraph{Relative Authority.} In this case, we consider the comparative relevance of the news article to the different entities occurring in it. As an example, let us consider the meeting of the Sudanese bishop \emph{Elias Taban}\footnote{\url{http://en.wikipedia.org/wiki/Elias_Taban}} with \emph{Hillary Clinton}\footnote{\url{http://en.wikipedia.org/wiki/Hillary_Clinton}}. Both entities are salient for the meeting. However, in Taban's Wikipedia page, this meeting is discussed prominently with a corresponding news reference\footnote{\url{http://tinyurl.com/mshf7j2}}, whereas in Hillary Clinton's Wikipedia page it is not reported at all. We believe this is not just an omission in Clinton's page but mirrors the fact that for the lesser known Taban the meeting is big news whereas for the more famous Clinton these kind of meetings are a regular occurrence, not all of which can be reported in what is supposed to be a selection of the most important events for her. Therefore, if two entities co-occur, the news is more relevant for the entity with the lower a priori authority.

The \emph{a priori authority} of an entity (denoted by $\Gamma(e)$) can be measured in several ways. We opt for two approaches: (i) probability of entity $e$ occurring in the corpus $\mathbf{N}$, and (ii) authority assessed through centrality measures like PageRank~\cite{page1999pagerank}. For the second case we construct the graph $G=(V,E)$ consisting of entities in $\mathbf{E}$ and news articles in $\mathbf{N}$ as \emph{vertices}. The \emph{edges} are established between $n$ and entities in $\varphi(n)$, that is $\langle n \rightarrow \varphi(n)\rangle$, and the out-links from $e$, that is $\langle e \rightarrow \gamma(s_(t-1))\rangle$ (arrows present the \emph{edge} direction).

Starting from a priori authority, we proceed to \emph{relative authority} by comparing the a priori authority of co-occurring entities in  $\varphi(n)$. We define the \emph{relative authority} of $e$ as the proportion of co-occurring entities $e'\in \varphi(n)$  that have a higher a priori  authority than $e$ (see Equation~\ref{eq:avg_authority}.

\begin{equation}\label{eq:avg_authority}
\hat{\Gamma}(e|\varphi(n)) = \frac{1}{|\varphi(n)|}\sum\limits_{e'\in \varphi(n)}\mathbbm{1}_{\Gamma(e') > \Gamma(e)}
\end{equation}
As we might run the danger of not suggesting any news articles for entities with very high a priori authority, due to the strict inequality constraint, we can relax the constraint such that the authority of co-occurring entities is above a certain threshold.

\paragraph{News Domain Authority.} The news domain authority addresses two main aspects. Firstly, if bundled together with the \emph{relative authority} feature, we can ensure that dependent on the entity authority, we suggest news from authoritative sources, hence ensuring the quality of suggested articles. The second aspect is in a news streaming scenario where multiple news domains report the same event --- ideally only articles coming from authoritative sources would fulfill the conditions for the news suggestion task.

The \emph{news domain} authority is computed based on the number of news references in Wikipedia coming from a particular \emph{news domain} $D$. This represents a simple prior that a news article $n$ is from domain $D$ in corpus $\mathbf{N}$. We extract the domains by taking the base URLs from the news article URLs.

\subsubsection{Novelty-based features} An important feature when suggesting an article $n$ to an entity $e$ is the \emph{novelty} of $n$ w.r.t the already existing entity profile $S_{e}(t-1)$. Studies~\cite{Bernstein:2005:RDS:1099554.1099733} have shown that on comparable collections to ours (TREC GOV2) the number of duplicates can go up to $17\%$.  This figure is likely higher for major events concerning highly authoritative entities on which all news media will report.

Given an entity $e$ and the already added news references $N_{t-1}=\{n_1,\ldots, n_k\}$ up to year $t-1$, the \emph{novelty} of $n_{k+1}$ at year $t$ is measured by the KL divergence between the language model of $n_{k+1}$ and articles in $N_{t-1}$. We combine this measure with the \emph{entity} overlap of $n_{k+1}$ and $n'\in N_{t-1}$. The \emph{novelty} value of $n_{k+1}$ is given by the minimal divergence value. Low scores indicate low novelty for the entity profile $S_e(t)$. 

\begin{align}\label{eq:novelty} 
\mathcal{N}(n|e) = \min\limits_{n'\in N_{t-1}}\left\{\lambda \cdot D_{KL}\left(\theta(n') || \theta(n)\right) + (1-\lambda) \cdot J\left(\varphi(n'),\varphi(n)\right)\right\} 
\end{align} 
where $D_{KL}$ is the KL divergence of the language models ($\theta(n)$ and $\theta(n')$), whereas $\lambda$ is the mixing weight ($\lambda=\{0, \ldots, 1\}$) between the language models $D_{KL}$ and the entity overlap in $n$ and $n'$.

\subsection{Article--Section Placement}\label{subsec:section_linking}

\begin{table}[h!]
\centering
\begin{tabular}{p{1.5cm} p{6cm} p{7cm}}
\toprule
feature type & feature & description\\
\midrule

\multirow{2}{2cm}{\emph{Topic}} & $J(LDA(n), LDA(s_{t-1}))$ & \multirow{2}{7cm}{Topic similarity between an article $n$ and the section text, and with the already referenced news articles in $s$.}\\
& $J(LDA(n), N_{t-1})$ & \\[5ex]
  
\emph{Syntactic} & \emph{POS} & POS tag overlap (uni/bi/trigrams) between $n$ and $s$.\\ 
 
 \multirow{3}{2cm}{\emph{Lexical}} & $J(title(n), s_{t-1})$ &  \multirow{3}{7cm}{News title and top--$k$ paragraphs ($k=1 \ldots 5$) similarity with $s$.}\\
 & $D_{KL}(\theta(p(k) || \theta(s_{t-1}))$ & \\
 & $cos(p(n), s_{t-1})$ &  \\[3ex]

 \multirow{2}{2cm}{\emph{Entity-based}} & $J(\varphi(n), \gamma(s,t-1))$ & \multirow{2}{7cm}{Entity and entity type overlap between the news article and entities appearing in a section.}\\
 & $J(\text{\texttt{type}}(\varphi(n)), \text{\texttt{type}}(\gamma(s_{t-1})))$ & \\[3ex]
 
 \multirow{2}{2cm}{\emph{Freq.}} & \texttt{\#POS,|p(n)|,$|n|, |\varphi(n)|$} & \multirow{2}{7cm}{Frequency based features of POS tags, number of paragraphs, entities in a news article}\\
 & \texttt{top--$k(e)$, top--$k(\text{\texttt{type}}(e))$} & \\[5ex]
 \bottomrule
\end{tabular}
\caption{Feature types used in $\mathcal{F}_s$ for suggesting news articles into the entity sections. We compute the features  for all $s\in \widehat{S}_c(t-1)$ as well as $s_{t-1}$.}
\label{tbl:feature_list}
\end{table}

We model the \emph{ASP} placement task as a successor of the \emph{AEP} task. For all the \emph{`relevant'} news entity pairs, the task is to determine the correct entity section. Each section in a Wikipedia entity page represents a different topic. For example, \emph{Barack Obama} has the sections  \emph{`Early Life', `Presidency', `Family and Personal Life'} etc. However, many entity pages have an  incomplete section structure. Incomplete or missing sections are due to two Wikipedia properties. First, long-tail entities miss information and sections due to their lack of popularity. Second, for all entities whether popular or not, certain sections might occur for the first time due to real world developments. As an example, the entity \emph{Germanwings} did not have an \emph{`Accidents'} section before this year's disaster, which was the first in the history of the airline.

Even if sections are missing for certain entities, similar sections usually occur in other entities of the same class (e.g. other airlines had disasters and therefore their pages have an accidents section). We exploit such homogeneity of section structure and construct templates that we use to expand entity profiles. The learning objective for $\mathcal{F}_s$ takes into account the following properties:

\begin{enumerate}
	\item \textbf{Section-templates:} account for incomplete section structure for an entity profile $S_e(t)$ by constructing section templates $\widehat{S}_c$ from an entity class $c$
	\item \textbf{Overall fit:} measures the overall fit of a news article to sections in the section templates $\widehat{S}_c$
\end{enumerate}

\subsubsection{Section-Template Generation}\label{subsubsec:as_sectiontemplates}

Given the fact that \emph{entity profiles} are often incomplete, we  construct \emph{section templates} for every \emph{entity class}. We group entities based on their class $c$ and construct \emph{section templates} $\widehat{S}_c$. For different entity classes, e.g. \texttt{Person} and \texttt{Location}, the section structure and the information represented in those section varies heavily. Therefore, the section
templates are with respect to the individual classes in our experimental setup (see Figure~\ref{fig:entity_distribution}). 
\begin{equation}\label{eq:section_template}\small
\widehat{S}_c = \{s_1,\ldots, s_k\}, \forall S_e(t) \in \mathbf{E}
\wedge e \text{\texttt{ typeOf }} c \end{equation}

Generating \emph{section templates} has two main advantages. Firstly, by considering class-based profiles, we can overcome the problem of incomplete individual entity profiles and thereby are able to suggest news articles to sections that do not yet exist in a specific entity $S_e(t)$. The second advantage is that we are able to canonicalize the sections, i.e. \emph{`Early Life'} and \emph{`Early Life and Childhood'} would be treated similarly.

To generate the section template $\widehat{S}_c$, we extract all sections from entities of a given type $c$ at year $t$. Next, we cluster the entity sections, based on an extended version of \emph{k--means} clustering \cite{DBLP:journals/pami/KanungoMNPSW02}, namely \emph{x--means} clustering introduced in Pelleg et al. which estimates the number of clusters efficiently \cite{pelleg2000x}. As a similarity metric we use the cosine similarity computed based on the \emph{tf--idf} models of the sections. Using the \emph{x--means} algorithm we overcome the requirement to provide the number of clusters \emph{k} beforehand. \emph{x--means} extends the \emph{k--means} algorithm, such that a user only specifies a range [$K_{min}$, $K_{max}$] that the number of clusters may reasonably lie in.

\subsubsection{News-section fit}

The learning objective of $\mathcal{F}_s$ is to determine the overall fit of a news article $n$ to one of the sections in a given section template $\widehat{S}_c$. The template is pre-determined by the class of the entity for which the news is suggested as relevant by $\mathcal{F}_e$. In all cases, we measure how well $n$ fits each of the sections $s\in \widehat{S}_{c}(t-1)$ as well as the specific entity section $s' \in S_e(t-1)$. The section profiles in $\widehat{S}_c(t-1)$ represent the aggregated entity profiles from all entities of class $c$ at year $t-1$.

To learn $\mathcal{F}_s$ we rely on a variety of features that consider several similarity aspects as shown in Table~\ref{tbl:feature_list}. For the sake of simplicity we do not make the distinction in Table~\ref{tbl:feature_list} between the individual entity section and class-based section similarities, $s_{e}(t-1)$ and $s(t-1)$, respectively. Bear in mind that an entity section $s_e$ might be present at year $t$ but not at year $t-1$ (see for more details the discussion on entity profile expansion in Section~\ref{subsubsec:profile_expansion}).

\paragraph{Topic.} We use topic similarities to ensure (i) that the content of $n$ fits topic-wise with a specific section text and (ii) that it has a similar topic to previously referred news articles in
that section. In a pre-processing stage we compute the topic models for the news articles, entity sections $S_{e}(t-1)$ and the aggregated class-based sections in $\widehat{S}_{c}$. The topic models are computed using LDA~\cite{DBLP:journals/jmlr/BleiNJ03}. We only computed a single topic per article/section as we are only interested in topic term overlaps between article and sections. We distinguish two main features: the first feature measures the overlap of topic terms between $n$ and the entity section $s_{e}(t-1)$ and $s(t-1) \in \widehat{S}_c$, and the second feature measures the overlap of the topic model of $n$ against referred news articles in $N_{t-1}$ at time $t-1$.

\paragraph{Syntactic.} These features represent a mechanism for conveying the importance of a specific text snippet, solely based on the frequency of specific POS tags (i.e. \texttt{NNP, CD} etc.), as commonly used in text summarization tasks. Following the same intuition as in \cite{DBLP:conf/acl/SauperB09}, we weigh the importance of articles by the count of specific POS tags. We expect that for different sections, the importance of POS tags will vary. We measure the similarity of POS tags in a news article against the section text. Additionally, we consider \emph{bi-gram} and \emph{tri-gram} POS tag overlap. This exploits similarity in syntactical patterns between the news and section text.

\paragraph{Lexical.} As \emph{lexical} features, we measure the similarity of $n$ against the entity section text $s_{e}(t-1)$ and the aggregate section text $s(t-1)$. Further, we distinguish between the
overall similarity of $n$ and that of the different news paragraphs ($p(n)$ which denotes the paragraphs of $n$ up to the 5th paragraph). A higher similarity on the first paragraphs represents a more confident indicator that $n$ should be suggested to a specific section $s$. We measure the similarity based on two metrics: (i) the KL-divergence between the computed \emph{language models} and (ii) \emph{cosine} similarity of the corresponding paragraph text $p(n)$ and section text.

\paragraph{Entity-based.} Another feature set we consider is the overlap of \emph{named entities} and their corresponding \emph{entity classes}. For different entity sections, we expect to find a particular set of entity classes that will correlate with the section, e.g. `\emph{Early Life}' contains mostly entities related to family, school, universities etc.

\paragraph{Frequency.} Finally, we gather statistics about the number of entities, paragraphs, news article length, top--$k$ entities and entity classes, and the frequency of different POS tags. Here we try to capture patterns of articles that are usually cited in specific sections.

\section{Datasets and Pre-Processing}\label{sec:datasets}

\subsection{Evaluation Plan}
In this section we outline the evaluation plan to verify the effectiveness of our learning approaches. To evaluate the news suggestion problem we are faced with two challenges. 

\begin{itemize}
	\item \emph{What comprises the ground truth for such a task ?}
	\item \emph{How do we construct training and test splits given that entity pages consists of  text added at different points in time ?}
\end{itemize}

Consider the ground truth challenge. Evaluating if an arbitrary news article should be included in Wikipedia is both subjective and difficult for a human if she is not an expert. An invasive approach,
which was proposed by Barzilay and Sauper~\cite{DBLP:conf/acl/SauperB09}, adds content directly to Wikipedia and expects the editors or other users to redact irrelevant content over a period of time. The limitations of such an evaluation technique is that content added to long-tail entities might not be evaluated by informed users or editors in the experiment time frame. It is hard to estimate how much time the added content should be left on the entity page. A more non-invasive approach could involve crowdsourcing of entity and news article pairs in an IR style relevance assessment setup. The problem of such an approach is again  finding knowledgeable users or experts for long-tail entities. Thus the notion of \emph{relevance} of a news recommendation is challenging to evaluate in a crowd setup.

We take a slightly different approach by making an assumption that the news articles already present in Wikipedia entity pages are relevant. To this extent, we extract a dataset comprising of all news articles referenced in entity pages (details in Section~\ref{subsec:datasets}). At the expense of not evaluating the space comprising of  news articles absent in Wikipedia, we succeed in (i) avoiding restrictive assumptions about the quality of human judgments, (ii) being invasive and polluting Wikipedia, and (iii) deriving a reusable test bed for quicker experimentation.

The second challenge of construction of training and test set separation is  slightly easier and is addressed in Section~\ref{subsec:test_train}.

\subsection{Datasets}\label{subsec:datasets}

\begin{figure}[h!]
\includegraphics[width=1.0\columnwidth]{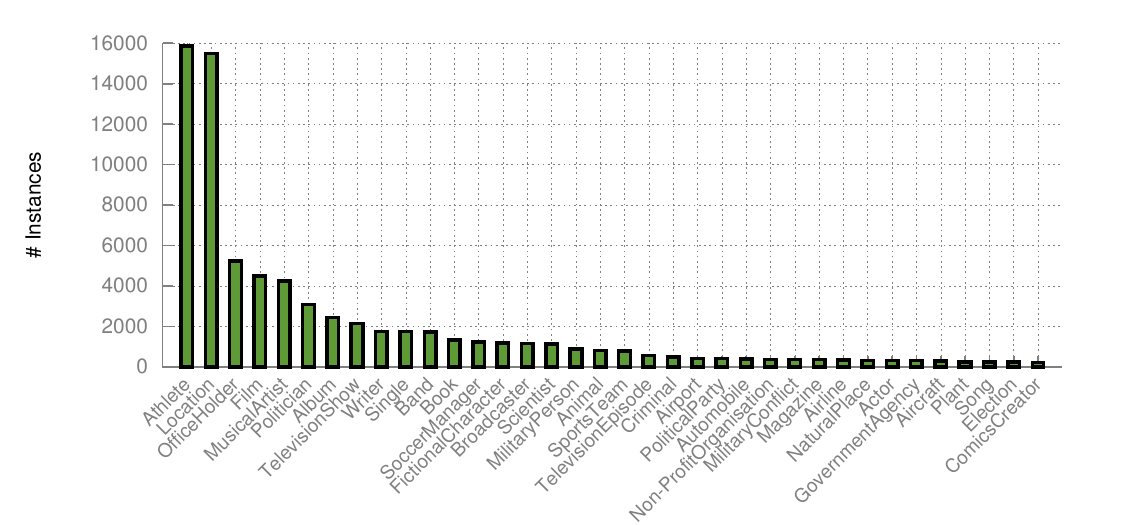}
\caption{Number of entities with at least one news reference for different entity classes.}
\label{fig:entity_distribution}
\end{figure}

The datasets we use for our experimental evaluation are directly extracted from the Wikipedia entity pages and their  revision history. The generated data represents one of the contributions of our work.\footnote{\url{http://l3s.de/~fetahu/cikm2015/data/}} The datasets are the following:

\paragraph{Entity Classes.} We focus on a manually predetermined set of \emph{entity classes} for which we expect to have news coverage. The number of analyzed \emph{entity classes} is $27$, including $73,734$
entities with at least one news reference. The \emph{entity classes} were selected from the DBpedia class ontology. Figure~\ref{fig:entity_distribution} shows the number of entities per class for the years (2009-2014).

\paragraph{News Articles.} We extract all news references from the collected Wikipedia entity pages.\footnote{A news reference in Wikipedia is denoted by the template \texttt{\{cite type=`news' | url=`'\}}} The extracted news references are associated with the sections in which they appear. In total there were $411,673$ news references, and after crawling we end up with $351,982$ successfully crawled news articles. The details of the news article distribution, and the number of entities and sections from which they are referred are shown in Table~\ref{tbl:news_dist}. 

\begin{table}[h!]
\centering
\begin{tabular}{c c c c }
\toprule
\texttt{year} & \texttt{\#news} & \texttt{\#entities} & \texttt{\#sections}\\
\midrule
\texttt{2009} & 42707 & 13550 & 3510 \\
\texttt{2010} & 78328 & 24953 & 8416 \\
\texttt{2011} & 73491 & 23144 & 6581 \\
\texttt{2012} & 81473 & 25980 & 8455 \\
\texttt{2013} & 69079 & 22121 & 8183 \\
\texttt{2014} & 29961 & 11088 & 4694 \\
\bottomrule
\end{tabular}
\caption{News articles, entities and sections distribution across years.}
\label{tbl:news_dist}
\end{table}

\paragraph{Article-Entity Ground-truth.} The dataset comprises of the news and entity pairs $\langle n, e\rangle \rightarrow \{0,1\}$. News-entity pairs are relevant if the news article is referenced in the entity page. Non-relevant pairs (i.e. negative training examples) consist of news articles that contain an entity but are not referenced in that entity's page. If a news article $n$ is referred from $e$ at year $t$, the features are computed taking into account the entity profiles at year $S_e(t-1)$.

\paragraph{Article-Section Ground-truth.} The dataset consists of the triple $\langle n, e, s\rangle$, where $s \in \widehat{S}_c$, where we assume that  $\langle n, e\rangle$ has already been determined as relevant.  We therefore have a multi-class classification problem where we need to determine the section of $e$ where $n$ is cited. Similar to the \emph{article-entity} ground truth, here too the features compute the similarity between $n$, $S_e(t-1)$ and $\widehat{S}_c(t-1)$.

\subsection{Data Pre-Processing}\label{subsec:pre_processing} 

We POS-tag the news articles and entity profiles $S_e(t)$ with the Stanford tagger~\cite{Toutanova:2003:FPT:1073445.1073478}.  For entity linking the news articles, we use TagMe!\cite{DBLP:journals/software/FerraginaS12} with a confidence score of 0.3. On a manual inspection of a random sample of 1000 disambiguated entities, the accuracy is above 0.9.  On average, the number of entities per news article is approximately 30. For entity linking the entity profiles, we simply follow the \emph{anchor} text that refers to Wikipedia entities.

\subsection{Train and Testing Evaluation Setup}\label{subsec:test_train} 

We evaluate the generated supervised models for the two tasks, \emph{AEP} and \emph{ASP}, by splitting the train and testing instances. It is important to note that for the pairs $\langle n, e\rangle$ and the triple $\langle n,e,\widehat{S}_{c}\rangle$, the news article $n$ is referenced at time $t$ by entity $e$, while the features take into account the entity profile at time $t-1$. This avoids any `overlapping' content between the news article and the entity page, which could affect the learning task of the functions $\mathcal{F}_e$ and $\mathcal{F}_s$. Table~\ref{tbl:train_test_instances} shows the statistics of train and test instances. We learn the functions at year $t$ and test on instances for the years greater than $t$. Please note that we do not show the performance for year 2014 as we do not have data for 2015 for evaluation.

\begin{table}[h!]
\centering
\begin{tabular}{p{1cm} l p{1.5cm} l l}
\toprule
 & \multicolumn{2}{c}{$\mathcal{F}_e$} & \multicolumn{2}{c}{$\mathcal{F}_s$}\\
 \toprule
 & \texttt{train} & \texttt{test} & \texttt{train} & \texttt{test} \\
\midrule
2009 & 74,005 & 469,386 & 19,399 & 218,757\\
2010 & 190,409 & 382,085 & 70,486 & 167,670\\
2011 & 286,588 & 292,398 & 115,286 & 122,870\\
2012 & 386,647 & 177,755 & 170,682 & 67,474\\
2013 & 471,209 & 59,172 & 218,538 & 19,618\\
\bottomrule
\end{tabular}
\caption{Number of instances for train and test in the \emph{AEP} and \emph{ASP} tasks.}
\label{tbl:train_test_instances}
\end{table}

\section{Results and Discussion}\label{sec:results}

\subsection{Article--Entity Placement}\label{subsec:article_entity_results} Here we introduce the evaluation setup and analyze the results for the \emph{article--entity (AEP)} placement task. We only report  the evaluation metrics for the \emph{`relevant'} news-entity pairs. A detailed explanation on why we focus on the \emph{`relevant'} pairs is provided in
Section~\ref{subsec:article_linking}.

\subsubsection{Evaluation Setup}\label{subsubsec:article_entity_setup}
\paragraph{Baselines.} We consider the following baselines for this task.

\begin{itemize}
\item \textbf{B1.} The first baseline uses only the salience-based features  by Dunietz and Gillick~\cite{DBLP:conf/eacl/DunietzG14}. 
\item \textbf{B2.} The second baseline assigns the value \emph{relevant} to a pair $\langle n, e\rangle$, if and only if  $e$ appears in the title of $n$.
\end{itemize}

\paragraph{Learning Models.} We use \emph{Random Forests} (RF)~\cite{Breiman2001}. We learn the RF on all computed features in Table~\ref{tbl:importance_salience}. The optimization on RF is done by splitting the feature space into multiple trees that are considered as ensemble classifiers. Consequently, for each classifier it computes the margin function as a measure of the average count of predicting the correct class in contrast to any other class. The higher the margin score the more robust the model.

\paragraph{Metrics.} We compute \emph{precision} P, \emph{recall} R and F1 score for the {\em relevant} class. For example, precision is the number of news-entity pairs we correctly labeled as relevant compared to our ground truth divided by the number of all news-entity pairs we labeled as relevant.

\subsubsection{Approach Effectiveness}\label{subsubsec:ae_results}
The following results measure the effectiveness of our approach in three main aspects: (i) overall \emph{performance} of $\mathcal{F}_e$ and comparison to baselines, (ii) \emph{robustness} across the years, and (iii) \emph{optimal} model for the \emph{AEP} placement task.

\begin{figure*}[ht!]
\centering
        \begin{subfigure}[b]{0.45\textwidth}
                \includegraphics[width=\textwidth]{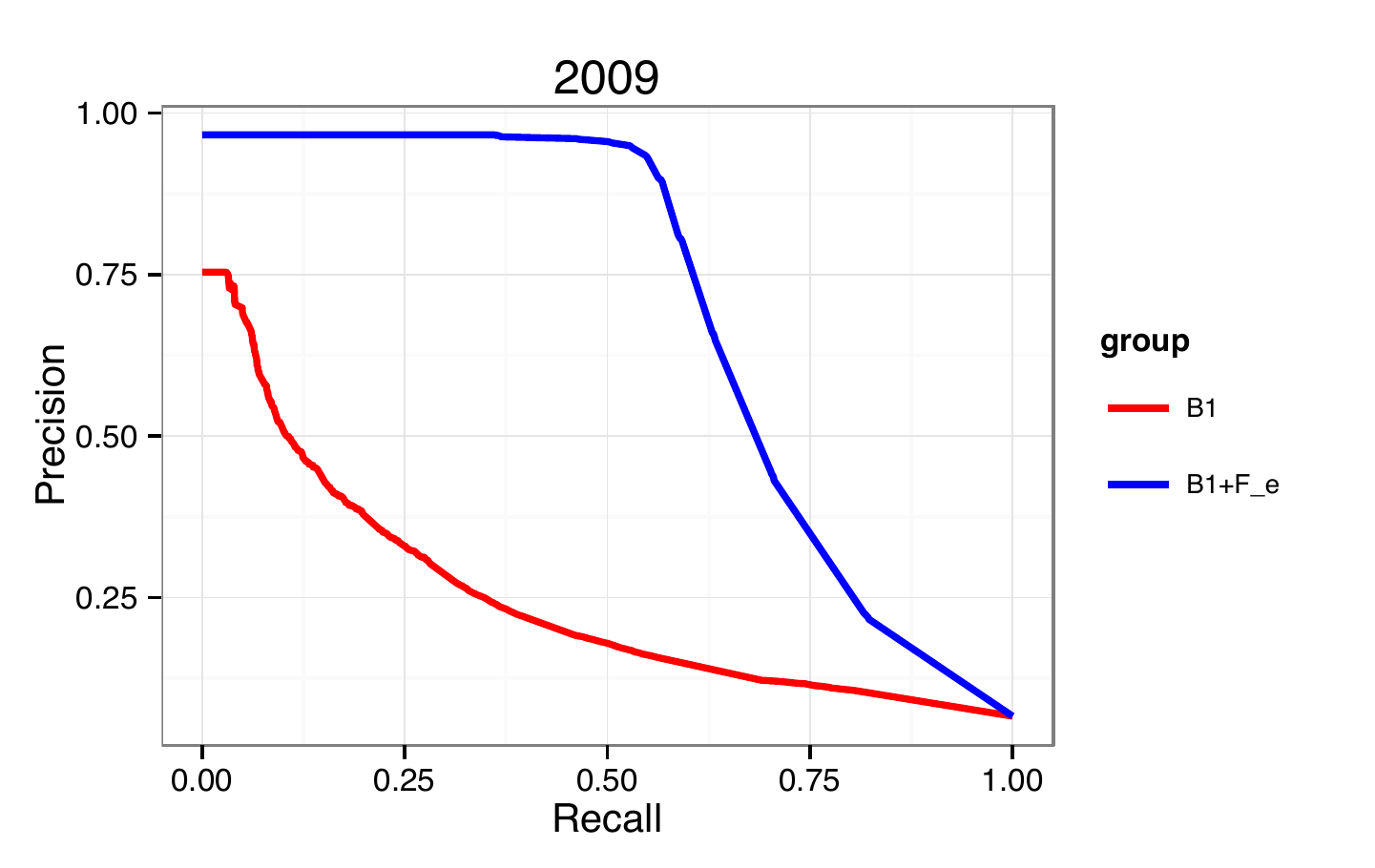}
                \caption{2009}
                \label{fig:salience_2009}
        \end{subfigure}
        \begin{subfigure}[b]{0.45\textwidth}
                \includegraphics[width=\textwidth]{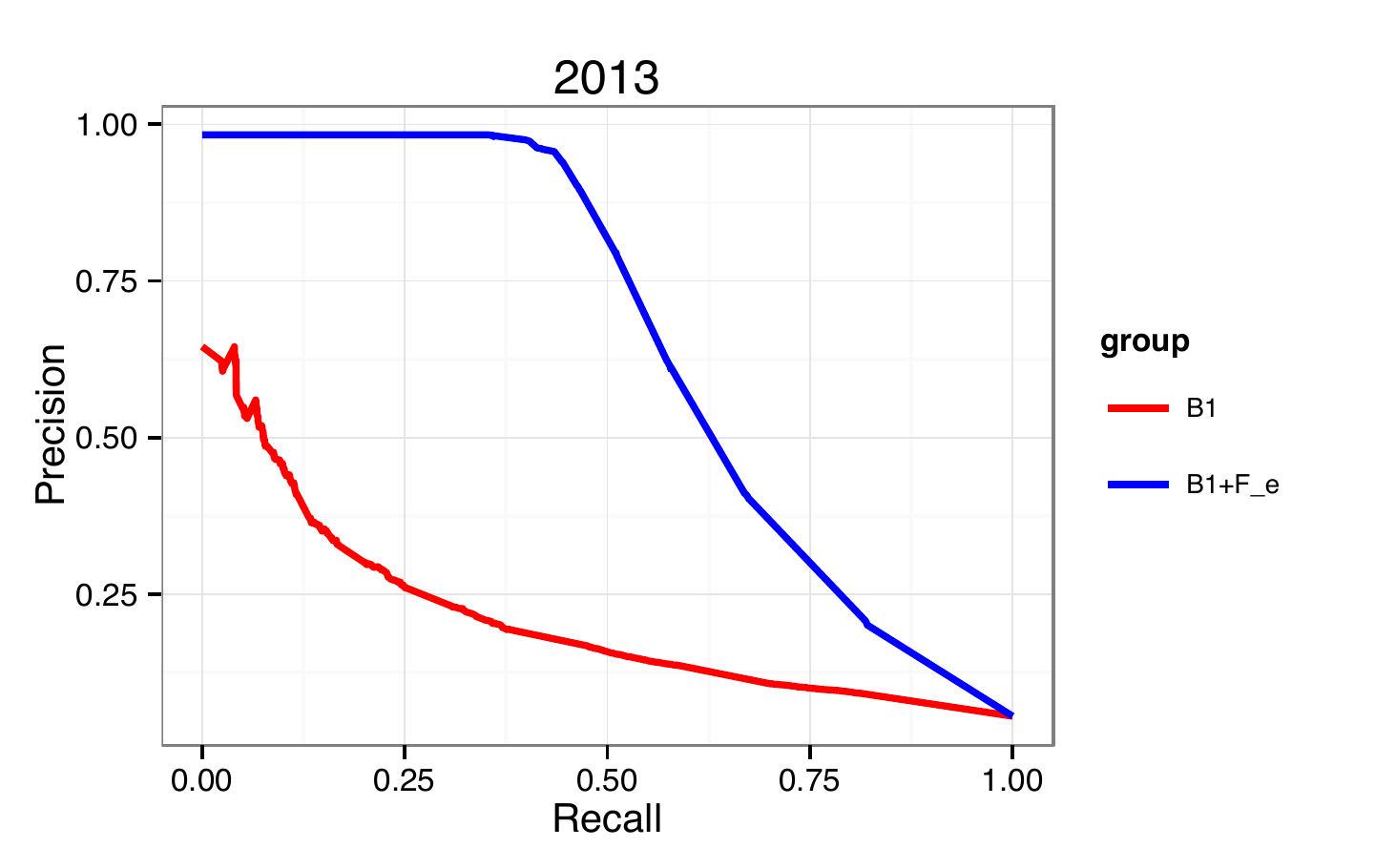}
                \caption{2013}
                \label{fig:salience_2013}
        \end{subfigure}
 	\caption{{Precision-Recall curve for the \emph{article--entity} placement task, in \emph{blue} is shown $\mathcal{F}_e$, and in \emph{red} is the baseline \textbf{B1}.}}
     \label{fig:salience_pr_curve}
 \end{figure*}

\paragraph{Performance.} Figure~\ref{fig:salience_pr_curve} shows the results for the years  2009 and 2013, where we optimized the learning objective with instances from year $t$ and evaluate on the years $t_i>t$ (see Section~\ref{subsec:test_train}).\footnote{We  only show  the first year 2009 and the last year 2013, since the difference to the other years is marginal.}
 The results show the \emph{precision--recall} curve. The \emph{red} curve shows baseline \textbf{B1}~\cite{DBLP:conf/eacl/DunietzG14}, and the \emph{blue} one shows the performance of $\mathcal{F}_e$. The curve shows for varying \emph{confidence scores} (high to low) the precision on labeling the pair $\langle e, n\rangle$ as \emph{`relevant'}. In addition, at each \emph{confidence score} we can compute the corresponding recall for the \emph{`relevant'} label. For high confidence scores on labeling the news-entity pairs, the baseline \textbf{B1} achieves on average a precision score of P=0.50, while $\mathcal{F}_e$ has P=0.93. We note that with the drop in the confidence score the corresponding precision and recall values drop
too, and the overall F1 score for \textbf{B1} is around F1=0.2, in contrast we achieve an average score of F1=0.67.

It is evident from Figure~\ref{fig:salience_pr_curve} that for the years 2009 and 2013, $\mathcal{F}_e$ significantly outperforms  the baseline \textbf{B1}. We measure the significance through the \emph{t-test} statistic and get a \emph{p-value} of $2.2e-16$. The improvement we achieve over \textbf{B1} in absolute numbers, $\Delta$P=+0.5 in terms of precision for the years between 2009 and 2014, and a similar improvement in terms of F1 score. The improvement for recall is $\Delta$ R=+0.4. The relative improvement over \textbf{B1} for P and F1 is almost 1.8 times better, while for recall we are 3.5 times better. In Table~\ref{tbl:article_entity_results} we show the overall scores for the evaluation metrics for \textbf{B1} and
$\mathcal{F}_e$. Finally, for \textbf{B2} we achieve much poorer performance, with average scores of P=0.21, R=0.20 and F1=0.21.

\paragraph{Robustness.} In Table~\ref{tbl:article_entity_results}, we show the overall performance for the years between 2009 and 2013. An interesting observation we make is that we have a very robust performance and the results are stable across the years. If we consider the experimental setup, where for year $t=2009$ we optimize the learning objective with only 74k training instances and evaluate on the rest of the instances, it achieves a very good performance. We predict with F1=0.68 the remaining 469k instances for the years $t\in (2009, 2014]$.

The results are particularly promising considering the fact that the distribution between our two classes is highly skewed.  On average the number of \emph{`relevant'} pairs account for only around $4-6\%$ of all pairs. A good indicator to support such a statement is the \emph{kappa} (denoted by $\kappa$) statistic. $\kappa$ measures agreement between the algorithm and the gold standard on both labels while correcting for chance agreement (often expected due to extreme distributions). The $\kappa$ scores for \textbf{B1} across the years is on average $0.19$, while for $\mathcal{F}_e$ we achieve a score of $0.65$ (the maximum score for $\kappa$ is 1).

\begin{table}[h!]
\centering
\begin{tabular}{p{1cm} l p{1cm} l p{1cm} l l}
\toprule
\texttt{year} & \multicolumn{2}{c}{P} & \multicolumn{2}{c}{R} & \multicolumn{2}{c}{F1}\\
\midrule
& \textbf{B1} & $\mathcal{F}_e$ & \textbf{B1} & $\mathcal{F}_e$ & \textbf{B1} & $\mathcal{F}_e$ \\
\midrule
2009 & 0.450 & 0.930 & 0.143 & 0.550 & 0.216 & 0.691\\
2010 & 0.503 & 0.939 & 0.128 & 0.540 & 0.204 & 0.685\\
2011 & 0.475 & 0.937 & 0.133 & 0.520 & 0.208 & 0.669\\
2012 & 0.476 & 0.935 & 0.110 & 0.515 & 0.177 & 0.664\\
2013 & 0.407 & 0.939 & 0.116 & 0.445 & 0.181 & 0.674\\
\bottomrule
\end{tabular}
\caption{\emph{Article--Entity} placement task performance.}
\label{tbl:article_entity_results}
\end{table}

\subsubsection{Feature Analysis}\label{subsubsec:ae_features} 

\begin{figure}[ht!]
\centering
    \includegraphics[width=0.8\columnwidth]{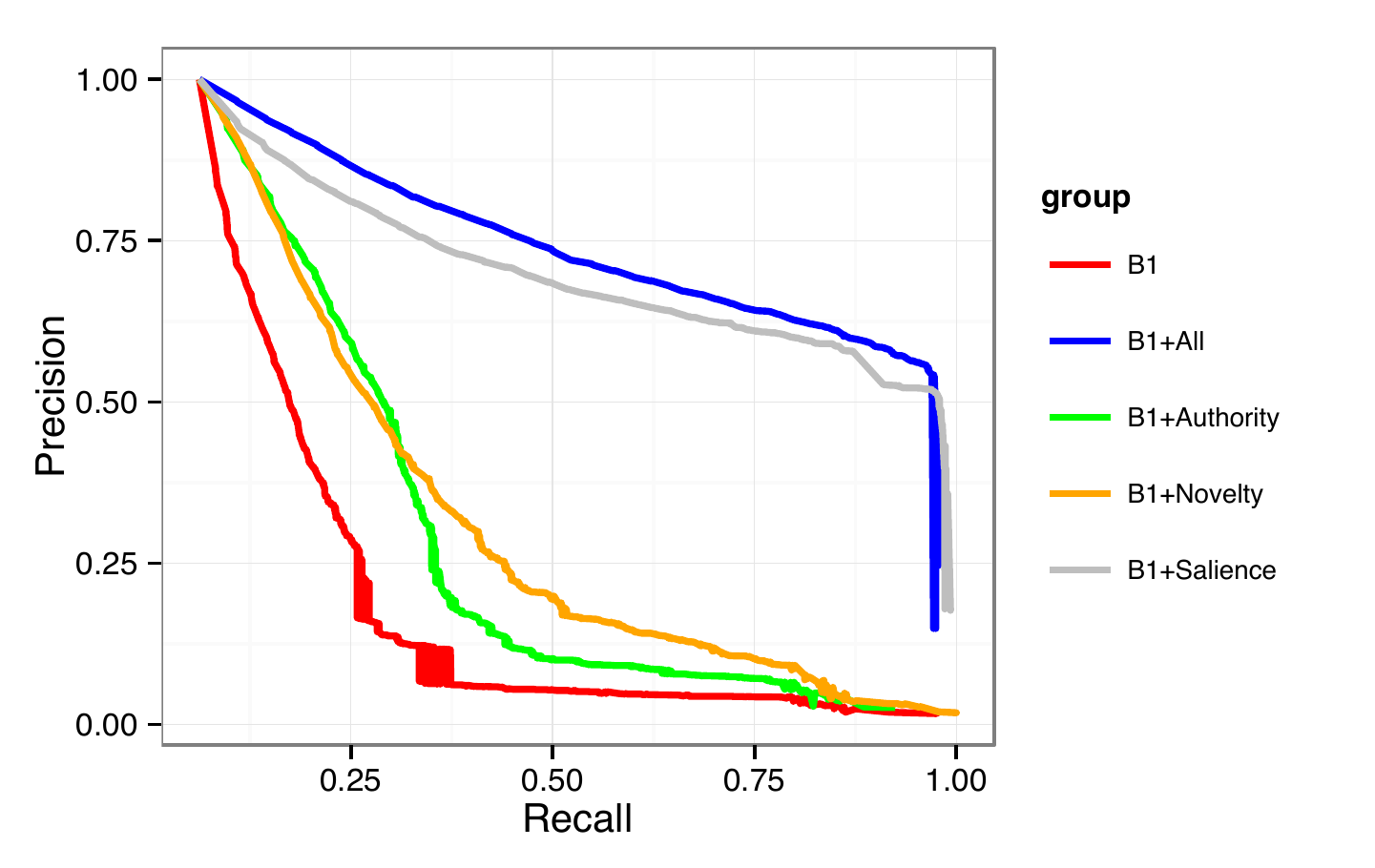}
    \caption{{Feature analysis for the \emph{AEP} placement task for $t=2009$.}}\
    \label{fig:salience_feature_pr_curve}
\end{figure}

In Figure~\ref{fig:salience_feature_pr_curve} we show the impact of the individual feature groups that contribute to the superior performance in comparison to the baselines. \emph{Relative entity frequency} from the \emph{salience} feature, models the entity salience as an exponentially decaying function based on the positional index of the paragraph where the entity appears. The performance of $\mathcal{F}_e$ with \emph{relative entity frequency} from the \emph{salience} feature group is close to that of all the features combined. The \emph{authority} and \emph{novelty} features account to a further improvement in terms of precision, by adding roughly a 7\%-10\% increase. However, if both feature groups are considered separately, they significantly outperform the baseline \textbf{B1}.

\subsection{Article-Section Placement}\label{subsec:article_section_results}
Here we show the evaluation setup for \emph{ASP} task and discuss the results with a focus on three main aspects, (i) the overall performance across the years, (ii) the \emph{entity class} specific performance, and (iii) the impact on \emph{entity profile} expansion by suggesting missing sections to entities based on the pre-computed templates.

\subsubsection{Evaluation Setup}\label{subsubsubsec:as_baselines}

\paragraph{Baselines.} To the best of our knowledge, we are not aware of any comparable approach for this task. Therefore, the baselines we consider are the following:
\begin{itemize}
    \item \textbf{S1}: Pick the section from template $\widehat{S}_c$ with the highest lexical similarity to $n$: \textbf{S1}$=\argmax_{s \in \widehat{S}_c(t-1)} \langle n, e, s\rangle$
    \item \textbf{S2}: Place the news into the most frequent section in $\widehat{S}_c$
\end{itemize}

\paragraph{Learning Models.} We use  \emph{Random Forests} (RF)~\cite{Breiman2001} and \emph{Support Vector Machines} (SVM)~\cite{chang2011libsvm}. The models are optimized taking into account the features in Table~\ref{tbl:feature_list}. In contrast to the \emph{AEP} task, here the scale of the number of instances allows us to learn the SVM models. The SVM model is optimized using the $\epsilon-SVR$ \emph{loss} function and uses the \emph{Gaussian} kernels. 

\paragraph{Metrics.} We compute \emph{precision} P as the ratio of news
for which we pick a section $s$ from $\widehat{S}_c$ and $s$ conforms
to the one in our ground-truth (see
Section~\ref{subsec:datasets}). The definition of \emph{recall} R and
F1 score follows from that of precision.

\subsubsection{Overall Article-Section Performance}\label{subsubsec:as_results}

\begin{figure}[h!]
    \centering
    \includegraphics[width=0.7\columnwidth]{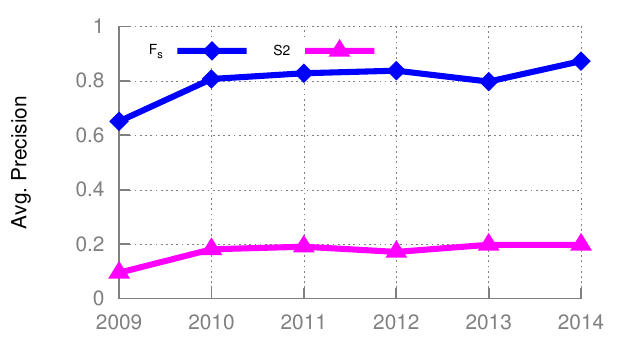}
    \caption{{\emph{Article-Section} performance averaged for all entity classes for $\mathcal{F}_s$ (using SVM) and \textbf{S2}.}}
     \label{fig:incremental_learning}
 \end{figure}

Figure~\ref{fig:incremental_learning} shows the overall performance and a comparison of our approach (when $\mathcal{F}_s$ is optimized using SVM) against the best performing baseline \textbf{S2}. With the
increase in the number of training instances for the \emph{ASP} task the performance is a monotonically non-decreasing function. For the year 2009, we optimize the learning objective of $\mathcal{F}_s$ with
around 8\% of the total instances, and evaluate on the rest. The performance on average is around P=0.66 across all classes. Even though for many classes the performance is already stable (as we will
see in the next section), for some classes we  improve further. If we take into account the years between 2010 and 2012, we have an increase of $\Delta$P=0.17, with around 70\% of instances used for training and
the remainder for evaluation. For the remaining years the total improvement is $\Delta$P=0.18 in contrast to the performance at year 2009.

On the other hand, the baseline \textbf{S1} has an average precision of P=0.12. The performance across the years varies slightly, with the year 2011 having the highest average precision of P=0.13. Always picking the most frequent section as in  \textbf{S2}, as shown in Figure~\ref{fig:incremental_learning}, results in an  average precision of P=0.17, with  a uniform distribution across the years.

\subsubsection{Article-Section Performance per Entity Class}

Here we show the performance of $\mathcal{F}_s$ decomposed for the different entity classes.  Specifically we analyze the 27 classes in Figure~\ref{fig:entity_distribution}. In Table~\ref{tbl:section_classifier}, we show the results for a range of years (we omit showing all years due to space constraints). For illustration purposes only, we group  them into four main classes ($\{$ \texttt{Person, Organization, Location, Event}$\}$) and into the specific sub-classes shown in the second column in Table~\ref{tbl:section_classifier}. For instance, the entity classes \texttt{OfficeHolder} and \texttt{Politician} are aggregated into \texttt{Person}--\texttt{Politics}.

It is evident that in the first year the performance is lower in contrast to the later years. This is due to the fact that as we proceed, we can better generalize and accurately determine the correct \emph{fit}
of an article $n$ into one of the sections from the pre-computed \emph{templates} $\widehat{S}_c$. The results are already stable for the year range $(2009, 2012]$. For a few \texttt{Person} sub-classes,
e.g. \texttt{Politics}, \texttt{Entertainment}, we achieve an F1 score above 0.9. These additionally represent classes with a sufficient number of training instances for the years $[2009, 2012]$. The lowest
F1 score is for the \texttt{Criminal} and \texttt{Television} classes.  However, this is directly correlated with the insufficient number of instances.

The baseline approaches for the \emph{ASP} task perform poorly. \textbf{S1}, based on  \emph{lexical similarity}, has a varying performance for different entity classes. The best performance is achieved for the class \texttt{Person -- Politics}, with P=0.43. This highlights the importance of our feature choice and that the \emph{ASP} cannot be considered as a \emph{linear function}, where the maximum similarity yields the best results. For different entity classes different features and combination of features is necessary. Considering that \textbf{S2} is the overall best performing baseline, through our approach $\mathcal{F}_s$ we have a significant improvement of over $\Delta$P=+0.64.

The models we learn are very robust and obtain high accuracy, fulfilling our pre-condition for accurate news suggestions into the entity sections. We measure the robustness of $\mathcal{F}_s$ through the $\kappa$ statistic. In this case, we have a model with roughly 10 labels (corresponding to the number of sections in a template $\widehat{S}_c$). The score we achieve shows that our model predicts with high confidence with $\kappa=0.64$.

\begin{table*}[ht!]
\centering
\begin{tabular}{p{2cm} p{2.5cm} l l l l l l l l l}
\toprule
\texttt{Entity class} & \texttt{Sub-Class} & \multicolumn{3}{c}{\texttt{2009}} & \multicolumn{3}{c}{\texttt{(2009,2012]}} & \multicolumn{3}{c}{\texttt{(2012,2014]}}\\
\midrule
& & P & R & F1 & P & R & F1 & P & R & F1 \\
\midrule

\multirow{6}{2cm}{\texttt{Person}} &  \texttt{Entert.} & 0.74 & 0.82 & 0.76 & 0.91 & 0.94 & 0.92 & 0.96 & 0.98 & 0.97\\
&  \texttt{Politics} & 0.92 & 0.94 & 0.93  & 0.92 & 0.95 & 0.93 & 0.94 & 0.96 & 0.95\\
&  \texttt{Scientists} & 0.48 & 0.68 & 0.55 & 0.89 & 0.94 & 0.91 & 0.93 & 0.95 & 0.94\\
&  \texttt{Sports} & 0.82 & 0.87 & 0.84 & 0.87 & 0.91 & 0.89 & 0.93 & 0.96 & 0.94\\
&  \texttt{Military} & 0.69 & 0.78 & 0.72 & 0.84 & 0.91 & 0.87 & 0.88 & 0.93 & 0.90\\
&  \texttt{Criminal} & 0.65 & 0.76 & 0.68 & 0.76 & 0.70 & 0.70 & 0.69 & 0.82 & 0.74\\[1.5ex]

\texttt{Organiz.} & \texttt{-} &  0.57 & 0.65 & 0.59  & 0.79 & 0.86 & 0.82 & 0.83 & 0.87 & 0.84\\[1.5ex]

\multirow{3}{2cm}{\texttt{Creative Work}} & \texttt{Television} & 0.53 & 0.65 & 0.56 & 0.75 & 0.73 & 0.71 & 0.73 & 0.77 & 0.75\\
& \texttt{Music} & 0.60 & 0.62 & 0.59 & 0.86 & 0.75 & 0.76 & 0.90 & 0.94 & 0.91\\
& \texttt{Written Work} & 0.66 & 0.77 & 0.70 & 0.73 & 0.83 & 0.77 & 0.72 & 0.79 & 0.74\\[1.5ex]

\texttt{Location} &\texttt{Location} & 0.78 & 0.76 & 0.72 & 0.86 & 0.90 & 0.87 & 0.92 & 0.96 & 0.94\\[1.5ex]
\texttt{Event} &\texttt{Event} & 0.56 & 0.68 & 0.61 & 0.86 & 0.87 & 0.85 & 0.69 & 0.72 & 0.69\\
\hline\hline
& \texttt{average} & 0.66 & 0.75 & 0.69 & 0.84 & 0.86 & 0.83 & 0.84 & 0.89 & 0.86\\
\bottomrule
\end{tabular}
\caption{\emph{Article-Section} placement performance (SVM based $\mathcal{F}_s$) for the different \emph{entity classes}. }
\label{tbl:section_classifier}
\end{table*}

\subsubsection{Entity Profile Expansion}\label{subsubsec:profile_expansion}

\begin{figure}[ht!]
\centering
\includegraphics[width=0.8\columnwidth]{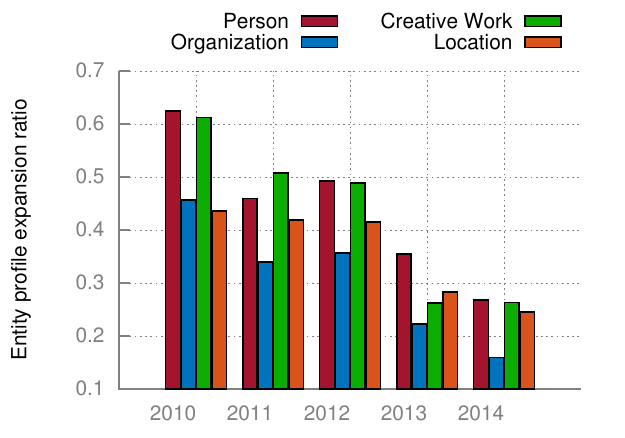}\vspace{-10pt}
\caption{{Correctly suggested news articles for $s\in {S}_e(t) \wedge s\notin {S}_e(t-1)$.}}
\label{fig:missing_sections}
\end{figure}

The last analysis is the impact we have on \emph{expanding} entity profiles $S_e(t)$ with new sections. Figure~\ref{fig:missing_sections} shows the ratio of sections for which we correctly suggest an article
$n$ to the right section in the section template $\widehat{S}_c(t)$. The ratio here corresponds to sections that are not present in the entity profile at year $t-1$, that is $s \notin S_{e}(t-1)$. However, given the generated templates $\widehat{S}_c(t-1)$, we can expand the entity profile $S_e(t-1)$ with a new section at time $t$. In details, in the absence of a section at time $t$, our model trains well on similar sections from the section template $\widehat{S}_c(t-1)$, hence we can predict accurately the section and in this case suggest its addition to the entity profile. With time, it is obvious that the expansion rate decreases at later years as the entity profiles become more `complete'. 

This is particularly interesting for expanding the entity profiles of long-tail entities as well as updating entities with real-world emerging events that are added constantly. In many cases such missing
sections are present at one of the entities of the respective entity class $c$. An obvious case is the example taken in Section~\ref{subsec:article_linking}, where the \emph{`Accidents'} is rather common for entities of type \texttt{Airline}. However, it is non-existent for some specific entity instances, i.e \emph{Germanwings} airline.

Through our \emph{ASP} approach $\mathcal{F}_s$, we are able to expand both \emph{long-tail} and \emph{trunk} entities. We distinguish between the two types of entities by simply measuring their section text length. The real distribution in the ground truth (see Section~\ref{subsec:datasets}) is 27\% and 73\% are \emph{long-tail} and \emph{trunk} entities, respectively. We are able to expand the entity profiles for both cases and all entity classes without a significant difference, with the only exception being the class \texttt{Creative Work}, where we expand significantly more \emph{trunk} entities.

\section{Conclusion}

In this chapter, we have proposed an automated approach for the novel task of suggesting news articles to Wikipedia entity pages to facilitate Wikipedia updating. The process consists of two stages. In the first stage, \emph{article--entity} placement, we suggest news articles to entity pages by considering three main factors, such as \emph{entity salience} in a news article, \emph{relative authority} and \emph{novelty} of news articles for an entity page. In the second stage, \emph{article--section} placement, we determine the best fitting section in an entity page. Here, we remedy the problem of incomplete entity section profiles by constructing section templates for specific entity classes. This allows us to add missing sections to entity pages. We carry out an extensive experimental evaluation on 351,983 news articles and 73,734 entities coming from 27 distinct entity classes. For the first stage, we achieve an overall performance with P=0.93, R=0.514 and F1=0.676, outperforming our baseline competitors significantly. For the second stage, we show that we can learn incrementally to determine the correct section for a news article based on section templates. The overall performance across different classes is P=0.844, R=0.885 and F1=0.860.

In the future, we will enhance our work by extracting facts from the suggested news articles. Results suggest that the news content cited in entity pages comes from the first paragraphs. However, challenging task such as the canonicalization and chronological ordering of facts, still remain.

Through this approach we are able to account for the constantly evolving nature of Wikipedia entities by suggesting important, novel, and authoritative information coming from news sources. The work in this chapter concludes the proposed approaches in this thesis, which account for a holistic framework of enriching and improving Wikipedia articles.

\clearemptydoublepage
\chapter{Entity Search as a Use Case of Wikipedia}\label{ch:entity_search}

In this chapter, we present the \emph{entity search} use case, which is one of the most well-known use cases of Wikipedia. We present work we carried in this field, specifically on improving entity search from structured datasets.

Wikipedia as one of the largest online encyclopedias is an important source for generating structured datasets. As such it has been successfully used in various projects on generating large knowledge graphs like DBpedia~\cite{DBLP:conf/semweb/AuerBKLCI07}, YAGO~\cite{suchanek_yago:_2007}. 

Similar datasets, like DBpedia, referred to as Web data form highly heterogeneous knowledge-graphs with more than 100 billion triples~\cite{DBLP:conf/www/PoundMZ10}. Web data consist of a wide variety of languages, schemas, domains and topics~\cite{DBLP:conf/esws/GueretGSL12}. Even though a large number of entities and concepts are highly overlapping, that is, they represent the same or related concepts, explicit links are still limited and often concentrated within large established knowledge graphs, like DBpedia.

The entity-centric nature of the Web of data has led to a shift towards tasks related to entity and object retrieval~\cite{DBLP:conf/semweb/BlancoCMT13,DBLP:conf/sigir/TononDC12} or entity-driven text summarization~\cite{DBLP:conf/sigir/DemartiniMBZ10}. Major search engine providers such as Google and Yahoo! already exploit such data to facilitate semantic search using knowledge graphs, or as part of similar efforts such as the \textit{EntityCube-Renlifang} project at Microsoft Research~\cite{DBLP:conf/www/NieMSWM07}. In such scenarios, data is aggregated from a range of sources calling for efficient means to search and retrieve entities in large data graphs. 

\emph{Entity search}~\cite{DBLP:conf/www/PoundMZ10,DBLP:conf/sigir/TononDC12} aims at retrieving relevant entities given a user query. The result is a ranked list of entities~\cite{DBLP:conf/semweb/BlancoCMT13}. By simply applying standard keyword search algorithms, like the BM25F~\cite{Blanco:2011:EEE:2063016.2063023,RobertsonBM25F}, promising results can be achieved. Other works rely on learning to rank approaches~\cite{Zhiltsov:2013:IES:2541154.2507868}, however, in the absence of training data, unsupervised retrieval models are preferred for ad-hoc queries.

In most cases, queries are entity centric. However, there are a large number of queries that are also topic-based, e.g. `\texttt{U.S. Presidents}'. Therefore, approaches like~\cite{DBLP:conf/sigir/TononDC12} have proposed retrieval techniques that make use of the explicit links between entities in the WoD for results or query expansion. For instance, following \texttt{owl:sameAs} or \texttt{rdfs:seeAlso} predicates from \texttt{dbp:Barack\_Obama}, one can retrieve co-references or highly related entities. However, considering the size of the WoD such statements are very sparse (see Figure~\ref{fig:related_types}). 

We propose a method for improving entity retrieval in two aspects. We improve the task by \emph{expanding} and \emph{re-ranking} the result set from a baseline retrieval model (BM25F), and address the sparsity of explicit links between entities through clustering of entities based on their similarity, using a combination of lexical and structural features. Finally, we re-rank the expanded result set, based on how likely it is an entity and its type to be relevant for a user query, defined as \emph{query type affinity}. 

We experimentally evaluate on  large structured dataset repository, namely the BTC12 dataset~\cite{btc-2012}, and use the SemSearch\footnote{\url{http://km.aifb.kit.edu/ws/semsearch10/}} query dataset. The individual steps in our approach are evaluated through a reliable crowdsourced evaluation approach. 

The main contributions of our work are as follows: (a) an entity retrieval model combining keyword search and entity clustering, and (b) an entity ranking model considering the query type affinity w.r.t the set of relevant entity types.

\section{Approach and Overview}\label{sec:motivationbackground}

\subsection{Preliminaries}\label{subsec:preliminaries}
The \textit{entity search (ES) task} concerns with retrieving a top--$k$ ranked set of entities from dataset for a given a user query $q$. User queries are typically entity centric. A \emph{dataset} in our case is a set of triples $\langle s, p, o \rangle$ (see Chapter~\ref{ch2:foundations}). An \emph{entity} profile of $e$ is the set of triples sharing the same subject URI $s$. The \emph{entity type} is determined by the triple $t_e=\langle s,$ \texttt{rdf:type}$, o \rangle$. 
Additionally, we define the \emph{query type} $t_q$, corresponding to the entity type in $q$, e.g. \emph{`Barack Obama'}, hence $t_q $\texttt{ typeOf Person}. 

\subsection{Motivation} \label{subsec:motivation_approach}

Recent studies \cite{DBLP:conf/sigir/TononDC12} have shown that \emph{explicit similarity statements}, which indicate some form of similarity or equivalence between entities, e.g. $\texttt{owl:sameAs}$, are useful for improving entity search results. However, explicit similarity statements usually are sparse and often focused towards a few well established datasets like DBpedia, Freebase etc. One reason for this is that these datasets represent known, and well structured graphs, which show a comparably high proportion of similarity statements linking similar entities within and beyond their original namespace.

\begin{figure}[ht!]
\centering
\begin{subfigure}[b]{0.5\textwidth}
                \includegraphics[width=\textwidth]{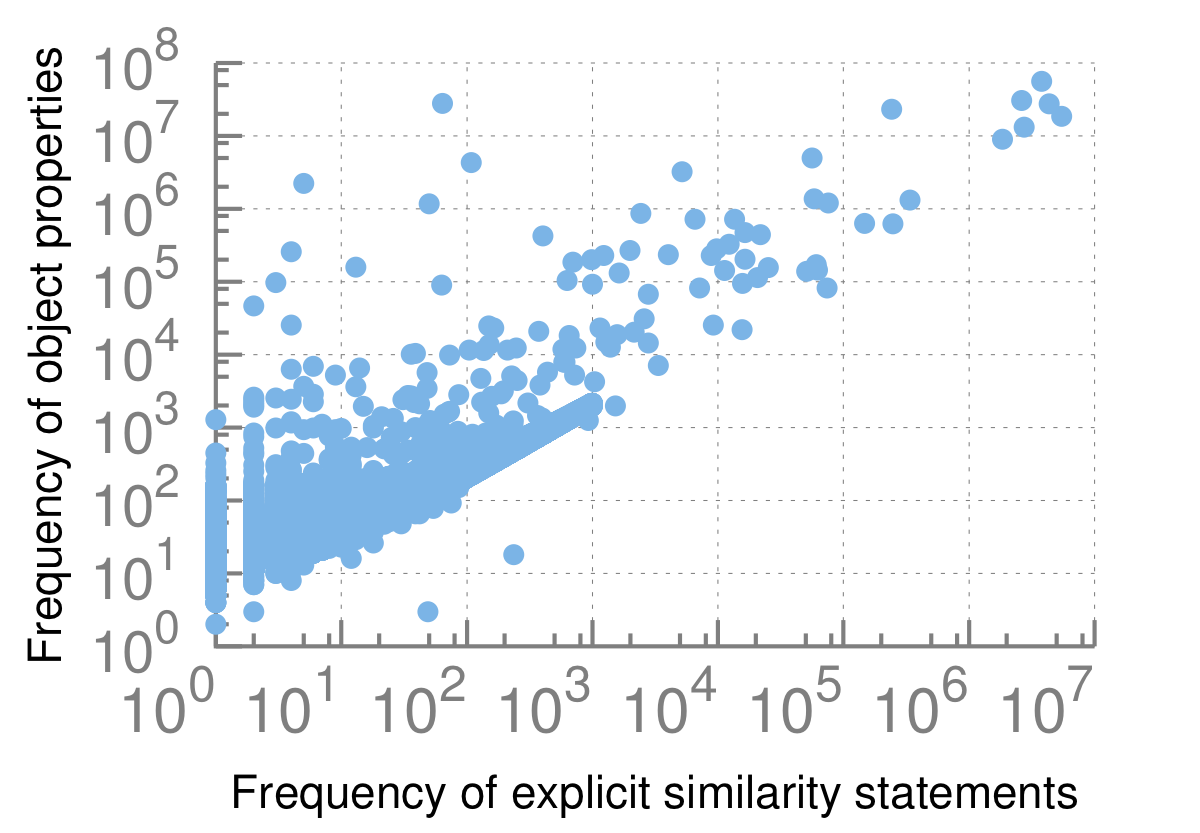}
                \caption{}
                \label{fig:related_types}
        \end{subfigure}%
        \begin{subfigure}[b]{0.5\textwidth}
                \includegraphics[width=\textwidth]{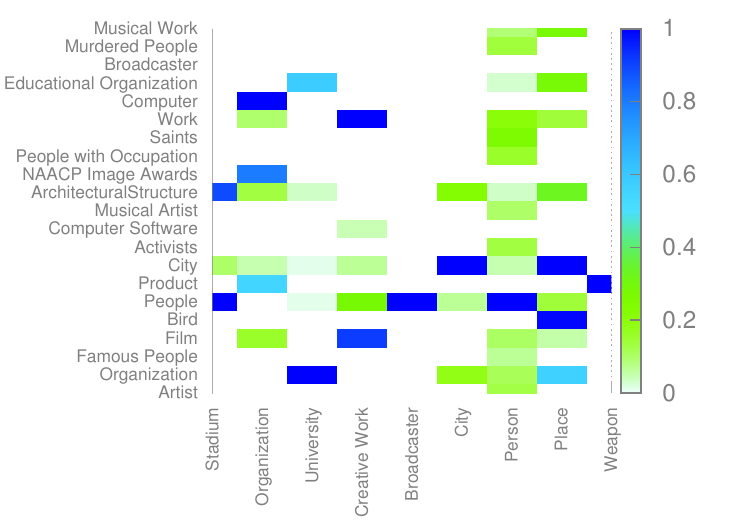}
                \caption{}
                \label{fig:query_affinity}
        \end{subfigure}
\caption{(a) Number of explicit similarity statements in contrast to the frequency of object property statements overall, shown for all data graphs. (b) Query type affinity shows the query type and the corresponding entity types from the retrieved and relevant entities.}
\end{figure}

Figure~\ref{fig:related_types} shows the total amount of explicit similarity statements (on the x--axis) that interlink entities in the BTC12 dataset. Referring to \cite{DBLP:conf/sigir/TononDC12}, here we specifically consider triples of the form $\langle e, p, e'\rangle$ where the predicate $p\in\{$\texttt{owl:sameAs}, \texttt{skos:related}, $\ldots$, \texttt{dbp:synonym}$\}$. These are plotted against the total number of \emph{object properties} (y--axis), where each point in the plot represents a graph in the BTC12 collection. From the figure, it is obvious that the number of explicit similarity statements is very sparse, considering the size of the dataset.

Nonetheless, missing links between entities can be partially remedied by computing their pair-wise similarity, thereby complementing statements like \texttt{owl:sameAs} or \texttt{skos:related}. Given the semi-structured nature of RDF data, graph-based and lexical features can be exploited for similarity computation. Particularly, lexical features derived from literals provided by predicates such as \textit{rdfs:label} or \textit{rdfs:description} are prevalent in LOD. Our analysis on the BTC12 dataset reveals that a large portion of entities (around 90\%) have an average literal length of 50 characters.

Furthermore, while the query type usually is not considered in state of the art ES methods, we investigated its correlation with the corresponding entity types from the query result set. We refer to a ground truth\footnote{\url{http://km.aifb.uni-karlsruhe.de/ws/semsearch10/Files/assess}} using the BTC10 dataset. We focus only on relevant entities for $q$. We analyze the \emph{query type affinity} of the result sets by assessing the likelihood of an entity in the results to be of the same type as the query type. Figure~\ref{fig:query_affinity} shows the query type affinity. On the x-axis we show the query type, whereas on the y-axis the corresponding relevant entity types are shown. Figure~\ref{fig:query_affinity} shows that most queries have high affinity with a specific entity type, with the difference being the query type \texttt{Person}, where relevant entities have a wider range of types.

We exploit the \emph{query type affinity} to improve the ranking of entities for a query $q$ (see Section~\ref{sec:retrieval}). Based on these observations, we argue that (a) \emph{entity clustering} can remedy the lack of existing linking statements and (b) entity \emph{re-ranking} considering the \emph{query type affinity} are likely to improve the entity retrieval task.

\subsection{Approach Overview} 
\begin{figure}[ht]
\centering
\includegraphics[width=1\textwidth]{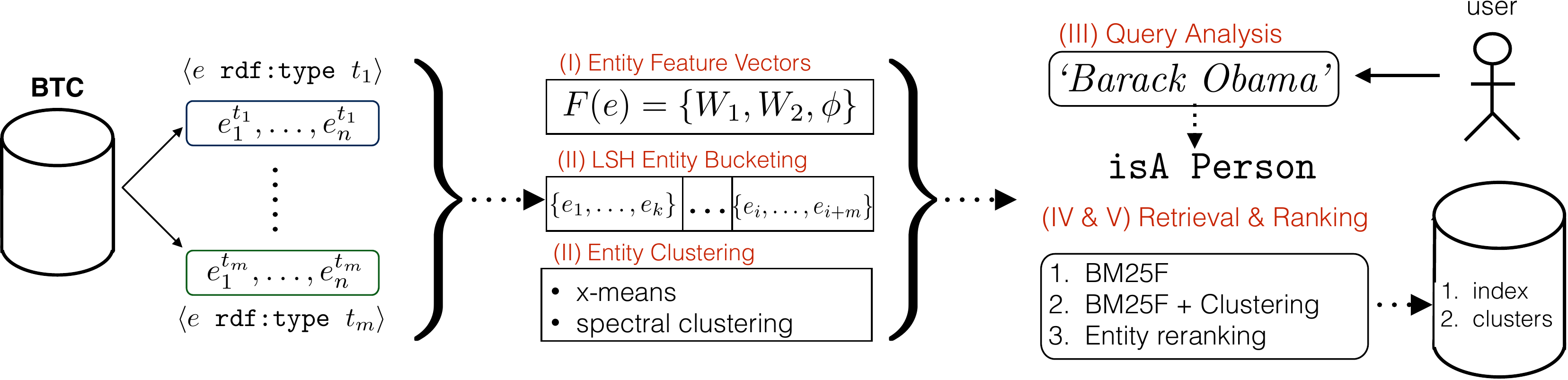}
\caption{Overview of the \emph{entity retrieval} approach.}
\label{fig:workflow}
\end{figure}

In this work we propose a novel approach for the \emph{entity search} task which builds on the observations described earlier. Figure~\ref{fig:workflow} shows an overview of the proposed approach. The individual steps are outlined below and described in detail in Section~\ref{sec:preprocessing} and \ref{sec:retrieval}. We distinguish between two main steps: (I) \emph{offline pre-processing}, including step I.a and I.b in the following overview, and (II) \emph{online entity retrieval}, covered by steps II.a to II.c.

\section{Data Pre-processing and Entity Clustering}\label{sec:preprocessing}

Here we describe the offline pre-processing to cluster entities and remedy the sparsity of explicit entity links.

\subsection{Entity Feature Vectors} 
Entity similarity is measured based on a set of structural and lexical features, denoted by the \emph{entity feature vector} $F(e)$. Finally, to avoid overfitting, we filter out items from $F(e)$ which have low frequency.

\paragraph{Lexical Features.} We consider a weighted set of \emph{unigrams} and \emph{bigrams} for an entity $e$, by extracting all textual literals used to describe $e$ denoted as $\mathbf{W}_1(e)$ and $\mathbf{W}_2(e)$. The weights are computed using the standard \emph{tf--idf} metric. High lexical similarity between an entity pair is a good indicator for expanding the result set from the corresponding cluster space.

\paragraph{Structural Features.} The feature set $\phi(e)$ considers the set of all object properties that describe $e$. The range of values for the structural features is $\phi(o,e) \rightarrow [0,1]$, i.e., to indicate if a object value is present in $e$.

\subsection{Entity Bucketing \& Clustering}\label{subsec:bucketing_clustering}

\paragraph{Entity Bucketing.} In this step we \emph{bucket} entities of a given \emph{entity type} by computing their \emph{MinHash} signature, which is used thereafter by the LSH algorithm~\cite{rajaraman2011mining}. This step is necessary as the number of entities is very large. In this way we reduce the number of pair-wise comparisons for the entity clustering, and limit it to only the set of entities within a bucket. Depending on the \emph{clustering algorithm}, the impact of bucketing on the clustering scalability varies. Since the LSH algorithm itself has linear complexity, bucketing entities presents a scalable approach considering the size of datasets in our experimental evaluation. 

\paragraph{Entity Clustering.} We cluster entities separately for the different entity types and the computed LSH buckets. We consider two clustering approaches: (i) \emph{X--means} and (ii) \emph{Spectral Clustering}. In both approaches we use Euclidean distance as the similarity metric. The dimensions of the Euclidean distance are the feature items in $F(\cdot)$. The similarity metric is defined in Equation~\ref{eq:entity_sim}, whereas the details of the two different clustering algorithms are explained in Chapter~\ref{ch2:foundations}.

\begin{equation}\label{eq:entity_sim}
	d(e, e') = \sqrt{\sum\left(\mathbf{F}(e) - \mathbf{F}(e')\right)^2}
\end{equation}
where the sum aggregates over the union of feature items from $\mathbf{F}(e), \mathbf{F}(e')$. The outcome of this process is a set of clusters $\mathbf{C}=\{C_1,\ldots,C_n\}$. The clustering process represents a core part of our approach from which we expand the entity results set for a given query, beyond the entities that are retrieved by a baseline as a starting point. The way the clusters are computed has an impact on the entity retrieval task, thus we present a thorough evaluation of cluster configurations in  Section~\ref{subsec:clustering_eval}.

\section{Entity Search: Expansion and Ranking}\label{sec:retrieval}

\paragraph{Result-set Expansion.}
From the initial result set $E_{b}=\{e_1,\ldots,e_k\}$ we obtain through BM25F, we expand the result set. From entities in $E_b$, we extract their corresponding set of clusters $\mathbf{C}$ as computed in the pre-processing stage. The result set is expanded with entities belonging to the clusters in $\mathbf{C}$. We denote the entities extracted from the clusters with $E_c$.

There are several precautions that we need to take into account in this step. We define two threshold parameters for expanding the result set. First, \emph{cluster size}, where for a number of entities above a threshold in a cluster we do not take the entities into account entities. The rationale is that clusters with a large number of entities tend to be less homogeneous, i.e. they tend to be a weak indicator of similarity. The second parameter deals with the \emph{number of entities} with which we expand the result set for a given entity cluster. The entities are considered based on their distance to the entity $e_b$. We experimentally validate the two parameters in Section~\ref{sec:evaluation}.

We measure the similarity of entities in  $E_c$ w.r.t query $q$ in Equation~\ref{eq:linkscore}, where the $\varphi$ measures \emph{string} distance of $e_c$ to $q$. Furthermore, this is done relative to entity $e_b$, such that if the $e_b$ is more similar to $q$, $\varphi(q,e_b) < \varphi(q,e_c)$ the similarity score will be increased, hence, the expanded entity $e_c$ will be penalized later on in the ranking. The second component represents the actual distance score $d(e_b, e_c)$.

\begin{equation}\label{eq:linkscore}
sim(q,e_c) = \lambda\frac{\varphi(q,e_c)}{\varphi(q,e_b)}  + (1-\lambda)d(e_{b}, e_{c})
\end{equation}
we set $\lambda=0.5$, such that entities are scored equally with respect to their match to query $q$ and the distance between entities. In this step we identify possibly relevant entities that have been missed by the scoring function of BM25F. Such entities could be suggested as relevant from the extensive clustering approaches that consider the structural and lexical similarity.

\paragraph{Query Analysis.}
Following the example in Figure~\ref{fig:query_affinity}, an important factor on the ranking of the result set is the \emph{query type affinity}. It models the relevance likelihood of a given entity type $t_e$ for a specific query type $t_q$. We give priority to entities that are likely to be relevant to the query type $t_q$ and are least likely to be relevant for other query types $t_q'$. The probability distribution is modeled empirically based on a previous dataset, BTC10. The score $\gamma$, we assign to any entity coming from the expanded result set is computed as in Equation~\ref{eq:type_affinity}. 

\begin{equation}\label{eq:type_affinity}
\gamma(t_e, t_q) = \frac{p(t_e | t_q)}{\sum\limits_{t_q'\neq t_q} \left(1 - p(t_e | t_q')\right)}
\end{equation}

An additional factor we use in the re-ranking process is the \emph{context score}. To better understand the query intent, we decompose a query \emph{q} into its \emph{named entities} and additional \emph{contextual terms}. An example is the query $q=\{\text{\emph{`harry potter movie'}}\}$ from our query set, in which case the contextual terms would be `\emph{movie}' and the named entity `\emph{Harry Potter}' respectively. In case of ambiguous queries, the contextual terms can further help to determine the query intent. The \textit{context score} (see Equation~\ref{eq:context}) indicates the relevance of entity $e$ to the contextual terms $Cx$ of the query $q$. For entities with a high number of textual literals, we focus on the main literals like \emph{labels, name} etc.  

\begin{equation}\label{eq:context}
context(q,e) = \frac{1}{|Cx|}\sum\limits_{c_x \in Cx}{\mathbbm{1}_{e\text{\texttt{ has }} c_x}}
\end{equation}

\paragraph{Ranking}
In the final step, we rank the expanded entity result set. The result set is the union of entities $\mathbf{E}=\mathbf{E}_b\cup\mathbf{E}_c$. In the case of entities retrieved through the baseline approach $e\in\mathbf{E}_b$, we simply re-use the original score, but normalize the values between $[0,1]$. For entities from $E_c$ we normalize the similarity score relative to the rank of entity $e_b$ (the position of $e_b$ in the result set) which was used to suggest $e_c$. This boosts entities which are the result of expanding top-ranked entities. 

\begin{equation}
rank\_score(e) = 
\begin{cases}
\frac{sim(q,e)}{rank(e_b)} & \text{ if } e \in E_c\\
bm25f(q,e) & \text{ otherwise }
\end{cases}
\end{equation}

The final ranking score $\alpha(e,t_q)$, for entity $e$ and query type $t_q$ assigns higher rank score in case the entity has high similarity with $q$ and its type has high relevance likelihood of being relevant for query type $t_q$. Finally, depending on the query set, in case $q$ contains contextual terms we can add $context(q,e)$ by controlling the weight of $\lambda$ (in this case $\lambda=0.5$).
\begin{align}\label{eq:rank}
\alpha(e,t_q) = \lambda \left(rank\_score(e) * \gamma(t_e, t_q)\right) + (1-\lambda) * context(q,e)
\end{align}

\section{Experimental Setup}\label{sec:experimentalsetup}

Here we describe our experimental setup, specifically the datasets, baselines and the ground truth. The setup and evaluation data are available for download\footnote{\url{http://l3s.de/~fetahu/iswc2015/}}.

\subsection{Evaluation Data}
\paragraph{Dataset.} We use the BTC12 dataset~\cite{btc-2012}. It represents one of the largest periodic crawls of Linked Data, also containing well-known knowledge bases like Freebase and DBpedia. The overall statistics of the data are: (i) 1.4 billion triples, (ii) 107,967 graphs, (iii) 3,321 entity types, and (iv) 454 million entities.

\paragraph{Entity Clusters.} The statistics for the generated clusters are as follows: the average number of entities fed into the \emph{LSH} bucketing algorithm is 77,485, whereas the average number of entities fed into \emph{x--means} and \emph{spectral} is 400. The number of generated entity buckets by LSH is 20,2009, while the number of clusters for \emph{x--means} and \emph{spectral} is 13 and 38, with an average of 10 and 20 entities per cluster respectively.

\paragraph{Query Dataset.} We use \emph{SemSearch}\footnote{\url{http://paragraph.aifb.kit.edu/ws/semsearch10/}} query set from 2010 with 92 queries. The SemSearch query set is a standard collection for evaluating entity retrieval tasks.

\subsection{Baseline and State of the Art}

\paragraph{Baseline.} We distinguish between two cases for the original BM25F baseline: (i) $\mathbf{B_{t}}$ and (ii) $\mathbf{B_{b}}$. In the first case, we use the \emph{title} or \emph{label} of an entity as a query field, whereas in the second case we use the full \emph{body} of an entity (consisting of all textual literals). The scoring of the fields is performed similar as in \cite{Blanco:2011:EEE:2063016.2063023}.

\paragraph{State of the art.} We consider the approach proposed in~\cite{DBLP:conf/sigir/TononDC12} as the state-of-the-art. Similar to their experimental setup, we analyze two cases: (i) $\mathbf{S1}$ and (ii) $\mathbf{S2}$. $\mathbf{S1}$ expands the entity set from the baseline approach with directly connected entities, and $\mathbf{S2}$ expands with entities up to the second hop. For further details we refer the reader to \cite{DBLP:conf/sigir/TononDC12}. In our experiments, we found that the $\mathbf{S2}$ did not result in any significant change in performance when compared to $\mathbf{S1}$, and we therefore do not report further on $\mathbf{S2}$.

\paragraph{Our approaches.} We analyze two entity retrieval techniques from our approach. The first is based on the \emph{x--means} clustering approach, which we denote by $\mathbf{XM}$. The second technique is based on \emph{spectral} clustering and is denoted by $\mathbf{SP}$. In both cases, we only expand the result set with entities coming from clusters with a total of ten entities associated with a cluster, and finally add only the most relevant entity based on the $sim(q,e_c)$ score. 

\paragraph{BTC indexes.} For the baseline, we generate a Lucene index, where we index entity profiles on two fields \texttt{title} and \texttt{body} (consisting of all the textual literals of an entity). The second index is an RDF index over the BTC dataset with support for SPARQL queries, for which we use the RDF3X tool~\cite{Neumann:2008:RRE:1453856.1453927}. The first index is used for the baseline approach, while the second for the state of the art approach.

\subsection{Ground Truth for Evaluation of Entity Retrieval}
\label{ss:q}

For each query, we establish the ground-truth of relevant entities from the result-set through crowdsourcing. Crowdsourced evaluation campaigns for the task of ad-hoc object retrieval have been shown to be reliable \cite{DBLP:conf/sigir/BlancoHHMPTT11,Halpin10evaluatingadhoc}. For each of the 92 queries, we pool the top 50 entities retrieved by the various methods, resulting in the top-k pooled entities corresponding to the query. By doing so we generate 4,600 query-entity pairs. 

We follow the key prescriptions for task design and deployment that emerged from the work of Blanco et al. \cite{DBLP:conf/sigir/BlancoHHMPTT11} to build a ground truth. Workers are asked to assess the relevance of each retrieved entity to the corresponding query on a 5-point Likert-type scale\footnote{\textit{1:Not Rel.}, \textit{2:Slightly Rel.}, \textit{3:Moderately Rel.}, \textit{4:Fairly Rel.} and \textit{5:Highly Rel.}}. 

We collect 5 judgements from different workers for each pair to ensure reliable relevance assessments and discernible agreement between workers. This results in a total of 23,000 judgements. The final relevance of an entity is considered to be the aggregated relevance score over the 5 judgements. We assess and compare the performance of the different methods by relying on the ground truth thus generated (see Section~\ref{sec:evaluation}).

\subsection{Evaluation Metrics}

Evaluation metrics assess the clustering accuracy and the retrieval performance. We indicate with $e_i^{rel}$ a retrieved relevant entity retrieved at rank $i$.

\paragraph{Cluster Accuracy.} As an initial evaluation, we assess the quality of our clusters. From a set of entities belonging to the same cluster, the accuracy is measured as the ratio of entities that \emph{belong together} over the total number of entities in a cluster, where assessments are obtained through crowdsourcing (see Section \ref{sec:evaluation}).

\paragraph{Precision.} P@k measures the precision at rank $k$, in our case $k=\{1,\ldots,10\}$. It is measured as the ratio of retrieved and relevant entities up to rank $k$ over the total number of entities retrieved up to rank $k$.

\begin{equation}\label{eq:p@k}
	P@k = \frac{1}{k}\sum_{i=1}^{k}e_i^{rel}
\end{equation}

\paragraph{Recall.} R@k is measured as the ratio of retrieved and relevant entities up to rank $k$ over the total number of relevant entities up to rank $k$, indicated by $|e_k^{rel}|$ in the equation below. 

\begin{equation}\label{eq:r@k}
	R@k = \frac{1}{|e_k^{rel}|}\sum_{i=1}^{k}e_i^{rel}
\end{equation}

\paragraph{Mean Average Precision}. MAP provides an overall precision of a retrieval approach across all queries under consideration.

\begin{equation}
	MAP = \frac{1}{|Q|}\sum_{q \in Q}P@k(q)
\end{equation}

\paragraph{Normalized Discounted Cumulative Gain}. It takes into account the ranking of entities generated using one of the retrieval approaches and compares it against the ideal ranking in the \emph{ground truth}.

\begin{align*}\label{eq:ndcg}
  nDCG@k = \frac{DCG@{k}}{iDCG@{k}}\;\;\;
  DCG@k = rel_1 + \sum\limits_{i=2}^{k}\frac{rel_i}{log_2{i}}
\end{align*}
where $DCG@k$ represents the discounted cumulative gain at rank $k$, and $iDCG@k$ is the ideal $DCG@k$ computed from the \emph{ground truth}.
\section{Evaluation and Discussion}\label{sec:evaluation}

In this section we report evaluation results of the two main steps in our approach. We first evaluate the quality of the pre-processing step, i.e., the clustering results for the \emph{x--means} and \emph{spectral} clustering algorithms. Next, we present the findings from our rigorous evaluation of the entity retrieval task.

\subsection{Cluster Accuracy Evaluation}\label{subsec:clustering_eval}
Considering the large number of clusters that are produced in the pre-processing step for a \textit{type} and \textit{bucket}, evaluating the accuracy and quality of all clusters is infeasible. Thus, we randomly pick 10 entity types and 10 buckets, resulting in 100 clusters for evaluation. For each cluster, we randomly select a maximum of 10 entities.

To evaluate the \emph{cluster accuracy}, we deploy atomic microtasks modeled such that a worker is presented with sets of 10 entities belonging to a cluster, along with a description of the entity in the form of the entity profile. The task of the worker is to pick the odd entities out (if any). We gather 5 judgments from different workers for each cluster. By enforcing restrictions available on the CrowdFlower platform, and following state of the art task design recommendations, we ensure that we receive judgments from the best workers (workers with high reputation as indicated by CrowdFlower). 

\begin{figure}[h!]
\begin{subfigure}[b]{0.5\textwidth}
                \includegraphics[width=\textwidth]{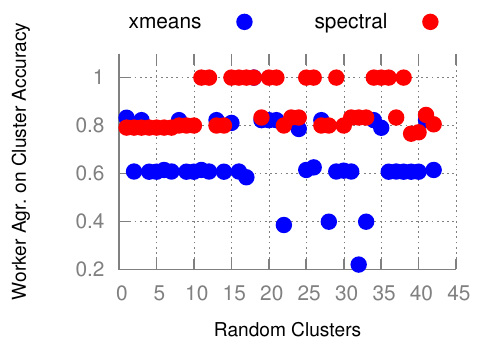}
                \caption{}
                \label{fig:cluster_agreement}
        \end{subfigure}%
        \begin{subfigure}[b]{0.5\textwidth}
                \includegraphics[width=\textwidth]{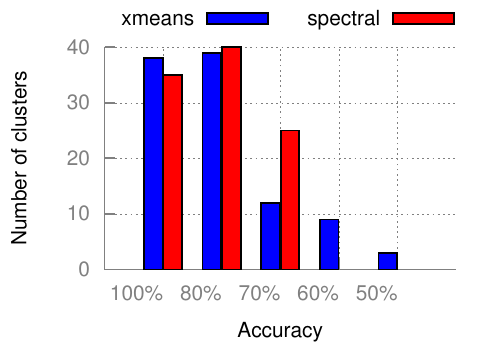}
                \caption{}
                \label{fig:cluster_accuracy_percentage}
        \end{subfigure}
\caption{(a) Worker agreement on cluster accuracy for \emph{spectral} and \emph{x--means} clustering. (b) Cluster accuracy for the \emph{spectral} and \emph{x--means} clustering approaches.}
\end{figure}

Figure~\ref{fig:cluster_accuracy_percentage} presents our findings for the evaluation of the clustering process. We note that for \emph{x--means} and \emph{spectral} clustering approaches, nearly 35\% and 38\% of the clusters are judged to be perfect respectively (i.e., the entities within the cluster were all found to belong together). 39\% of the clusters corresponding to \textit{spectral} clustering and 40\% of the clusters corresponding to \textit{x-means}, have an accuracy of 80\%. Considering its multidimensional representation of the entities, \emph{spectral} clustering has higher accuracy and it does not have clusters below 70\% accuracy. 
The lowest accuracy of 70\% for \emph{spectral} clustering implies that in each cluster there were only 3 entities that did not belong to the cluster. The implications of an accurate clustering process become clearer in the next section, where we assess the accuracy of finding relevant entities in the generated entity clusters.

Figure \ref{fig:cluster_agreement} presents the pairwise agreement between workers on the quality of each cluster. In case of the \emph{spectral} clustering, we observe a high inter-worker agreement of 0.75 as per Krippendorf's Alpha.
We observe a moderate inter-worker agreement of 0.6 as per Krippendorf's Alpha on the clusters resulting from \emph{x--means}.

\subsection{Entity Retrieval Evaluation}\label{subsec:retrieval_evaluation}

\begin{figure}[h!]
        \begin{subfigure}[b]{0.5\textwidth}
                \includegraphics[width=\textwidth]{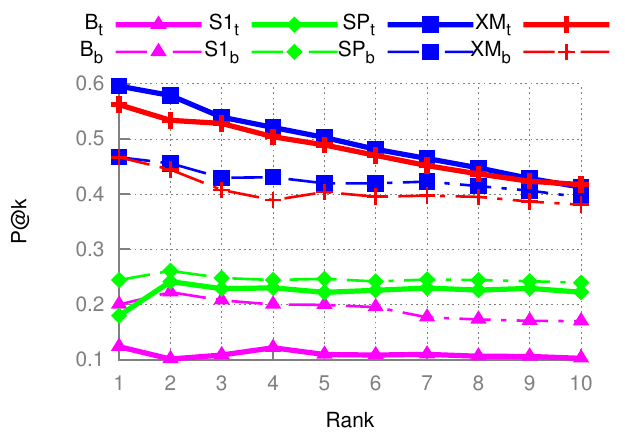}
                \caption{}
                \label{fig:precision_k}
        \end{subfigure}%
        \begin{subfigure}[b]{0.5\textwidth}
                \includegraphics[width=\textwidth]{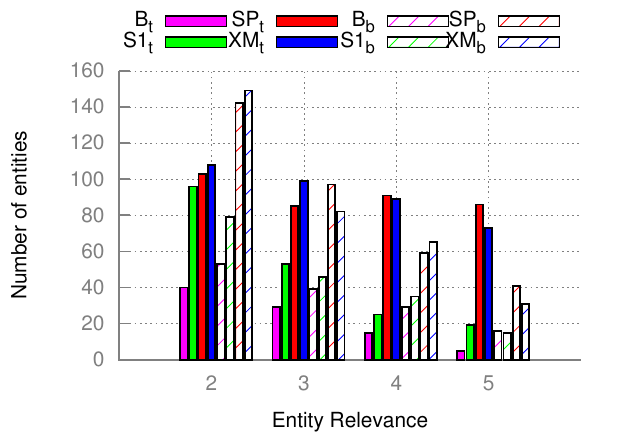}
                \caption{}
                \label{fig:relevance_histogram}
        \end{subfigure}
\caption{(a) P@k for the different entity retrieval approaches under comparison. (b) The relevant entity frequency based on their graded relevance (from \textit{2-Slightly Relevant} to \textit{5-Highly Relevant}) for the different methods.}
\end{figure}

Figure \ref{fig:precision_k} presents a detailed comparison between the $P@k$ for the different methods. The proposed approaches outperform the baseline and state of the art at all ranks. The precision is highest at $P@1=0.6$ whereas for the later ranks it stabilizes at 0.4. In contrast to our approach, the performance of the baseline and the state of the art is more uniform, and is around $P@k=0.25$. The best overall performing approach is the retrieval approach based on spectral clustering $SP$. Table~\ref{tbl:overal_results} shows the details about the performance of the respective approaches as measured for our evaluation metrics.

An interesting observation is that for our approaches the best performance is achieved when querying for the field \emph{title}. In the case of the baseline, the best performance is achieved when querying for the field \emph{body} ($B_b$) while the same is inconclusive in case of the state-of-the-art methods (\textbf{$S1_t$} and \textbf{$S1_b$}). We achieve a significantly higher retrieval performance when using the title field. This can be explained by the fact that entities that match a query on their \textit{title} field when compared to those that match a query on their \textit{body} field, have a higher likelihood of being an exact match. 

The high gain in performance through our methods (\textit{SP} and \textit{XM}) stems mainly from the two steps in our approach. The first step expands the result set with relevant entities as shown in Figure~\ref{fig:relevance_histogram}. The figure shows the number of relevant entities corresponding to the different grading scales as described in Section \ref{subsec:clustering_eval}. In all cases we note that our methods find more relevant entities. The second step which re-ranks the expanded result set helps in reducing the number of \emph{`non-relevant'} entities. We find that $S1_t$ has a 14\% decrease of non-relevant entities, whereas $SP_t$ and $XM_t$ depict a 35\% decrease, respectively. In second case where we query the \emph{body} field, the number of \emph{`non-relevant'} entities for $S1_b$ decreases by about 13\%, while $SP_b$ and $XM_b$ depict a 24\% decrease. 

\begin{table}[ht!]
\centering
\begin{tabular}{ p{1.2cm} p{1.2cm} p{1.2cm} p{1.2cm} p{1.2cm} p{1.2cm} p{1.2cm} p{1.2cm} p{1.2cm}}
\toprule
& $B_t$ & $B_b$ & $S1_t$ & $S1_b$ & $SP_t$ & $SP_b$ & $XM_t$ & $XM_b$\\
\midrule
P@10 & 0.103 & 0.170 & 0.222 & 0.240 & 0.413 & 0.394 & \textbf{0.417} & 0.381\\
R@10 & 0.052 & 0.089 & 0.112 & 0.118 & 0.206 & \textbf{0.219} & 0.216 & 0.215\\
MAP & 0.110 & 0.191 & 0.224 & 0.246 & \textbf{0.497} & 0.426 & 0.482 & 0.407\\
$Avg(R)$ & 0.031 & 0.058 & 0.063 & 0.074 & 0.132 & \textbf{0.133} & 0.131 & 0.130\\
\bottomrule
\end{tabular}
\caption{Performance of the different entity retrieval approaches. In all cases our approaches are significantly better in terms of P/R ($p<0.05$ measured for \emph{t-test}) compared to \emph{baseline} and \emph{state of the art}. There is no significant difference between $SP$ and $XM$ approaches.}
\label{tbl:overal_results}
\end{table}

\begin{figure}[ht!]
        \centering
        \includegraphics[width=0.6\textwidth,keepaspectratio]{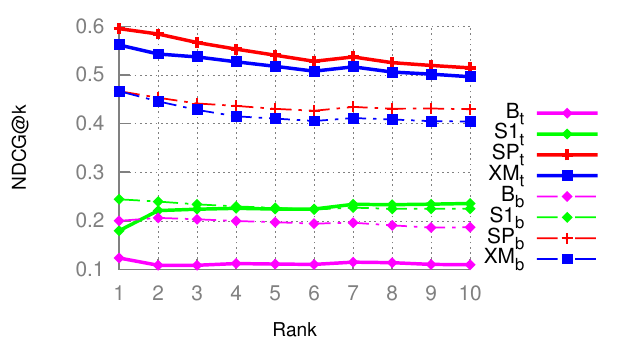}
        \caption{\small{NDCG$@k$ for $B1$, $S1$ and $SP$, $XM$}}
        \label{fig:ndcg}
\end{figure}

We additionally analyze the performance of the entity retrieval approaches through the $NDCG@k$ metric. Figure~\ref{fig:ndcg} shows the NDCG scores. Similar to our findings for $P@k$ presented in Table 1, our approaches perform best for the query field \emph{title} and significantly outperform the approaches under comparison.

\begin{figure}[ht!]
                \begin{subfigure}[b]{0.5\textwidth}
                \includegraphics[width=\textwidth]{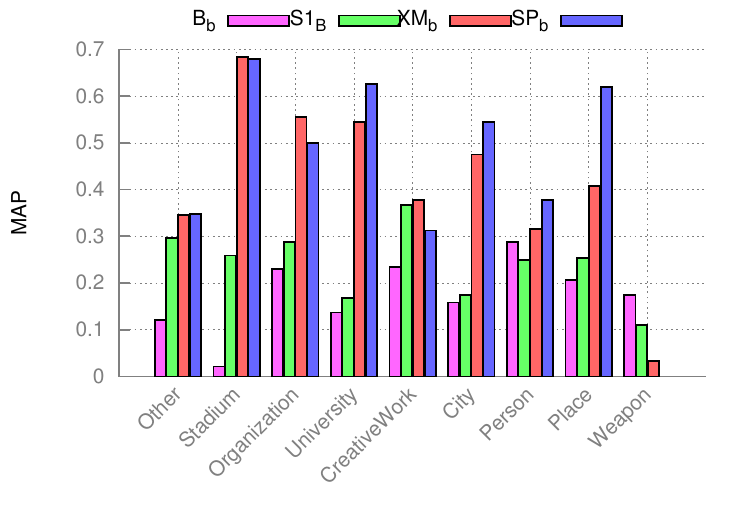}
                \caption{}
                \label{fig:query_improvement}
        \end{subfigure}%
        \begin{subfigure}[b]{0.5\textwidth}
                \includegraphics[width=\textwidth]{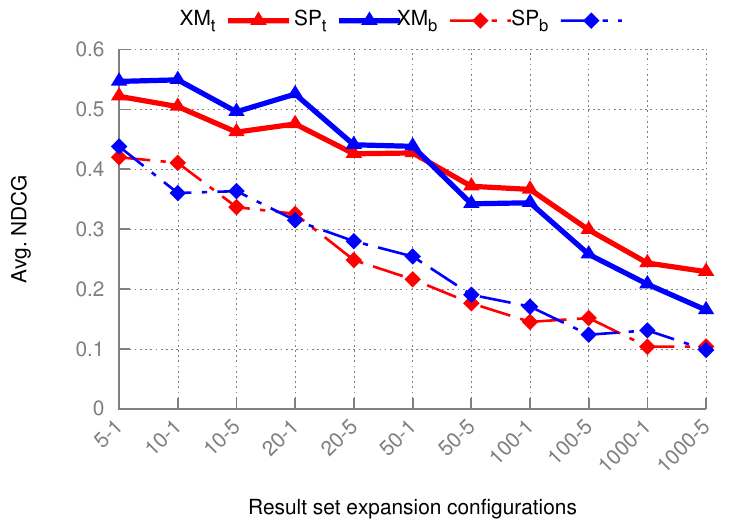}
                \caption{}
                \label{fig:result_expansion}
        \end{subfigure}
\caption{(a) The aggregated MAP for different query types and for the different retrieval approaches (note, we show the results for field \emph{body} where baseline performs best). (b) The various configurations for the number of expanded entities for $SP$ and $XM$.}
\end{figure}

Next, we present observations concerning the different \emph{query types} and the entity result set expansion  parameters. In Figure~\ref{fig:query_improvement} we show the improvement we gain in terms of MAP for the different \emph{query types}. We observe that there is quite a variance for the different query types, however, in nearly all cases, the biggest improvement is achieved through the $SP$ approach. Interestingly for the query type \emph{`Creative Work'} the state of the art is nearly as good as the $XM$ approach, whereas in the case of \emph{`Weapon'} the baseline performs best. One possible explanation for this is that in the case of \emph{`Creative Work'} the explicit entity similarity statements are abundant.

Addressing the case of optimizing our retrieval approaches, $SP$ and $XM$, we experimentally show the impact that the expansion of the result set has on the measured performance metrics. Here, we show the impact on the \textit{average NDCG} score. Figure~\ref{fig:result_expansion} shows the performance at average NDCG for the varying \emph{cluster size} and \emph{number of entities} added (result set expansion) for every entity in $E_b$. The best performance is achieved for a rather smaller \emph{cluster size} ranging between 5 and 10 entities per cluster. Regarding the number of entities with which the result set is expanded for every $e_b$, the best performance is achieved by expanding with one entity per cluster. The increase in cluster size and number of entities attributes to a decrease in performance.

\section{Conclusions}\label{sec:discussion}
In this work, we presented an approach to improve the performance of entity retrieval on structured data. Building on existing state of the art methods, we follow an approach consisting of offline preprocessing clustering, and online retrieval, results expansion and reranking. Preprocessing exploits \emph{x--means} and \emph{spectral} clustering algorithms using lexical as well as structural features. The clustering process was carried out on a large set of entities (over 450 million). The evaluation of the clustering process shows that over 80\% of clusters have an accuracy of more than 80\%. As part of the online entity retrieval, for a given a starting result set of entities as retrieved by the baseline approach BM25F we further expand the result set with relevant entities. Additionally, we propose an entity ranking model that takes into account the query type affinity. Finally, we carry out an extensive evaluation of the retrieval process using the SemSearch and the BTC12 datasets. The results show that our methods outperform the baseline and state of the art approaches. In terms of standard IR metrics, our method in combination with one of the clustering approaches, e.g. $SP_t$ improves over $S1_t$ with $\Delta P@10 = +0.19$, $\Delta MAP= +0.273$ and $\Delta R@10 = +0.1$.


\clearemptydoublepage
\chapter{Conclusion and Future Work}\label{sec:conclusion}

\section{Conclusion and Contributions}
The wide popularity of Wikipedia, with an ever growing number of real-world entities and events, has resulted in Wikipedia establishing itself as the reference point for a large number of Web users, and currently is ranked in the top--5 of the most visited internet sites~\cite{alexa_internet}. Furthermore, due to its wide coverage it has found use in a large range of applications such as Web search~\cite{DBLP:conf/kdd/0001GHHLMSSZ14,DBLP:conf/www/PoundMZ10}, Apple's Siri, etc., and as such indirectly impacts a larger audience of users. One argumentation for its popularity, is that Wikipedia, represents an open and collaboratively edited encyclopedia, consisting of information and facts for real-world entities. For popular entities and events, it was found that emerging facts are reflected with a minimal delay in Wikipedia~\cite{keegan_hot_2011}.

To better understand such dynamics, on how emerging facts and information about entities are reflected in Wikipedia, we conducted a study in Chapter~\ref{ch2:foundations}. We analyzed the amount of time it takes when something is reported in online news media, until it is reflected in Wikipedia. Additionally, we analyzed one of the key policies of providing citations to external sources as a means of evidence for added information by the Wikipedia editors. Not surprisingly, we found out that there is a long-tail of entities for which there is large gap between the time something is reported in news and until it is added in Wikipedia. Furthermore, we found that, the second most cited source in Wikipedia refers to online news media.

In this thesis we presented a holistic approach which has impact on several aspects of Wikipedia. The main outcomes and conclusions can be summarized as following:

\begin{itemize}
	\item In Chapter~\ref{ch1:citation_recommendation}, we enforce the \emph{verifiability} principle of Wikipedia, where we provide evidence for statements that require citations in Wikipedia articles, specifically we focus on \emph{news citations}. Hence, through the \emph{verifiability} principle we improve the overall quality of Wikipedia articles. The main conclusions in this Chapter~\ref{ch1:citation_recommendation} are the following:
\begin{enumerate}
	\item In order to find news citations for statements in Wikipedia articles, we first categorize the statements into the different citation categories (e.g. \emph{web, news, journal} etc.). Through this step we can focus on specific categories for which we are in hold of such specific-collections (i.e. news collections). Furthermore this ensures that we suggest most appropriate citations. The accuracy of this task varies for the different entity types, and we achieve results with up to 80\% on determining the appropriate citation type for Wikipedia statements.
	\item After having determined the required citation type for a statement, next, we focused on only statements that require news citations. For such statements, it is important that the suggested news articles, entail the statement and preferably are central to the news article. We proposed a combined approach, which first retrieves a top--$k$ set of news article candidates, and is followed by a binary classification, which tags the news article candidates as appropriate citations or not for the given statement. For this task, we achieve highly accurate results with more than 86\% accuracy. Finally, we show that in 19\% of the cases, we are able to suggest news citations of higher quality than those already existing in Wikipedia.
\end{enumerate}

	\item In Chapter~\ref{ch5:cite_span} we dealt with the problem of determining the span of citations in Wikipedia. This is an important aspect as citations in Wikipedia are assigned at various statement granularities (e.g. \emph{sub-sentence, sentence} or \emph{paragraph}), and as such there are no explicit means on knowing the exact span. This has direct consequences on enforcing the \emph{verifiability} principle, where on the presence of a citation in a Wikipedia paragraph, it is not possible to differentiate the textual fragments that are covered or uncovered by a citation.

We proposed an approach which relies on a \emph{linear-chain} CRF for determining the citation span for \emph{web} and \emph{news} references. We achieve an overall accuracy of 90\% across the different granularities of citation span.

\item The work in Chapter~\ref{ch3:news_suggestion} deals with the problem of Wikipedia's \emph{coverage} of novel and important facts, and provides means on keeping up-to-date \emph{long} and \emph{trunk} Wikipedia articles. Since the nature of Wikipedia articles is constantly evolving, new information emerges and thus such changes need to be reflected into the structure of the Wikipedia articles themselves. We proposed a two-staged approach, which focuses on news sources for providing important and novel facts for Wikipedia articles.

\begin{enumerate}
	\item In the first stage, for a given news article, we analyze its content and the entities that are mentioned therein. We first, compute entity salience measures, determining that the mentioned entities are central concept in the given article. In this way, we suggest to Wikipedia articles, only those news articles where the corresponding entity mention (i.e. thus referring to a specific Wikipedia page) is  important. We proposed novel salience measures which show significant improvement over existing state of the art approaches.
	\item After determining for a Wikipedia article a set of news articles which provide important and novel information, next, based on a section template which we construct for specific types of Wikipedia articles, we determine the exact section for which we suggest such news articles. In case, a section is missing, we add the appropriate section automatically from the constructed templates.
\end{enumerate}
\end{itemize}

Lastly, in Chapter~\ref{ch:entity_search}, we showed an application use-case of Wikipedia as the main source for generating structured datasets, and proposed means on improving existing \emph{entity search} approaches over structured datasets.

 \section{Future Work Directions}
 
 While we address several major issues with enriching and improving Wikipedia articles, there are several issues that still need to be addressed. In this thesis, through our contributions we enforce the \emph{verifiability} principle by providing citations for Wikipedia text, and correspondingly determining explicitly the span of such citations, however, we are limited to only citation markers that are already present in Wikipedia. Furthermore, an already well known problem of Wikipedia edit wars leads to content on controversial being biased~\cite{conf/cikm/DasLM13}. 
 
 Therefore, as future directions we foresee work on automatically placing citation markers, in case such markers do not exist already and are necessary. We plan to use the existing markers as ground-truth for automated approaches on placing citation markers for long-tail entities. This problem is in particular challenging, as we need to determine the type of the citation marker, that is, to what citation category the reference should point at. Alternatively, one could employ the proposed approach in Chapter~\ref{ch1:citation_recommendation} to determine for a piece of text that we want to place the citation marker which citation category is more suitable.

 Next, for controversial or polarizing topics, collaborative editing or discussion lead to a phenomena known as \emph{``echo chambers''}. As such communities that tend to agree on a specific matter are isolated from other communities who have a different stance~\cite{DBLP:journals/corr/VicarioBZPSCSQ15}. Such controversial behavior tends to be diffused in Wikipedia too, therefore, leading to Wikipedia edit wars, an in other cases to biased content towards a particular group on a controversial topic~\cite{conf/cikm/DasLM13}. To avoid biased content, we plan on analyzing the edits in the revision history of a Wikipedia article and the editor behavior to flag biased content, which may violate the neutrality of an article, respectively the \emph{neutral point of view} policy~\cite{DBLP:conf/www/Hube17}.


\clearemptydoublepage

\appendix

 \cleardoublepage

\bibliographystyle{alpha}

\newcommand{\etalchar}[1]{$^{#1}$}

 \cleardoublepage
 \singlespace
 \phantomsection
 \addcontentsline{toc}{chapter}{Index}
 \printindex

\end{document}